\documentclass[sigconf]{acmart}

\usepackage{amsmath}
\usepackage{amsthm}
\usepackage{booktabs}
\usepackage{framed} 
\usepackage{multirow}
\usepackage{enumitem}
\usepackage{caption}
\usepackage{graphicx}
\usepackage{float} 
\usepackage{etoolbox}
\usepackage{xspace}
\usepackage{accents}
\usepackage{subfig}
\usepackage{makecell}
\usepackage{bbding}
\usepackage{multicol}

\AtBeginDocument{%
  }

\setcopyright{acmlicensed}
\copyrightyear{2018}
\acmYear{2018}
\acmDOI{XXXXXXX.XXXXXXX}
\acmConference[Conference acronym 'XX]{Make sure to enter the correct
  conference title from your rights confirmation email}{June 03--05,
  2018}{Woodstock, NY}
\acmISBN{978-1-4503-XXXX-X/2018/06}

\newtheorem{myDef}{\textbf{Definition}}
\newtheorem{prob}{\textbf{Problem}}

\newcommand{\eg}{\emph{e.g.},\xspace}

\newcommand{\ie}{\emph{i.e.},\xspace}

\newcommand{\model}{UniExtreme}

\newcommand\figref[1]{Figure~\ref{#1}}

\newcommand\tabref[1]{Table~\ref{#1}}
\newcommand\secref[1]{Section~\ref{#1}}
\newcommand\equref[1]{Equation~(\ref{#1})}
\newcommand\appref[1]{Appendix~\ref{#1}}

\newcommand{\eat}[1]{}

\newcommand{\jia}[1]{{\color{purple}{#1}}}
\newcommand{\TODO}[1]{{\color{red}TODO:{#1}}}

\newcommand\beftext[1]{{\color[rgb]{0.5,0.5,0.5}{BEFORE:#1}}}

\begin{document}

\title{UniExtreme: A Universal Foundation Model for Extreme Weather Forecasting}

\author{Hang Ni}
\affiliation{%
  \institution{The Hong Kong University of Science and Technology (Guangzhou)}
  \city{Guangzhou}
  \country{China}
  }
\email{hni017@connect.hkust-gz.edu.cn}

\author{Weijia Zhang}
\affiliation{%
  \institution{The Hong Kong University of Science and Technology (Guangzhou)}
  \city{Guangzhou}
  \country{China}
  }
\email{wzhang411@connect.hkust-gz.edu.cn}

\author{Hao Liu}
\authornote{Corresponding author}
\affiliation{%
  \institution{The Hong Kong University of Science and Technology (Guangzhou)}
  \city{Guangzhou}
  \country{China}
  }
\email{liuh@ust.hk}

\renewcommand{\shortauthors}{Ni et al.}

\begin{abstract}
Recent advancements in deep learning have led to the development of Foundation Models (FMs) for weather forecasting, yet their ability to predict extreme weather events remains limited. 
Existing approaches either focus on general weather conditions or specialize in specific-type extremes, neglecting the real-world atmospheric patterns of diversified extreme events. 
In this work, we identify two key characteristics of extreme events: (1) the spectral disparity against normal weather regimes, and (2) the hierarchical drivers and geographic blending of diverse extremes. 
Along this line, we propose \textbf{UniExtreme}, a \underline{\textbf{uni}}versal \underline{\textbf{extreme}} weather forecasting foundation model that integrates (1) an Adaptive Frequency Modulation (AFM) module that captures region-wise spectral differences between normal and extreme weather, through learnable Beta-distribution filters and multi-granularity spectral aggregation, and (2) an Event Prior Augmentation (EPA) module which incorporates region-specific extreme event priors to resolve hierarchical extreme diversity and composite extreme schema, via a dual-level memory fusion network.  Extensive experiments demonstrate that UniExtreme outperforms state-of-the-art baselines in both extreme and general weather forecasting, showcasing superior adaptability across diverse extreme scenarios.
\end{abstract}

\ccsdesc[500]{Computing methodologies~Neural networks}
\ccsdesc[500]{Information systems~Data mining}
\ccsdesc[500]{Applied computing~Earth and atmospheric sciences}

\keywords{Extreme Weather Forecasting, Weather Forecasting, Foundation Models, Frequency Domain}

\received{20 February 2007}
\received[revised]{12 March 2009}
\received[accepted]{5 June 2009}

\maketitle


\begin{figure}[t]
    \centering
    \subfloat[Frequency distributions of normal, extreme, and random regions in 2024's U.S., quantified by high frequency area.]{\includegraphics[width=0.8\linewidth]{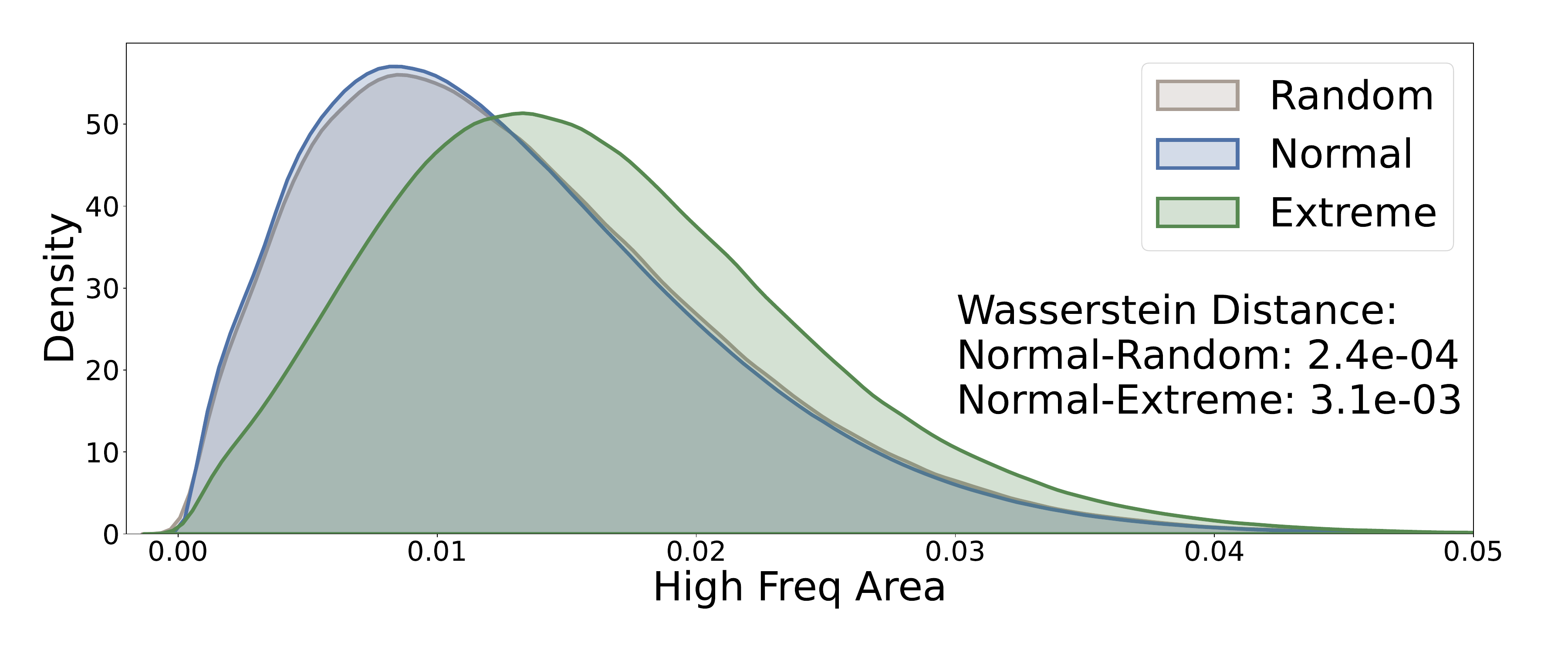}\label{figure:hfa}}\\
    \vspace{-5pt}
    \subfloat[Spatial distributions of diverse yet coexisting extreme events at Jan 9, 2024, 12 PM, exhibiting 78.30\% overlap rate.]{\includegraphics[width=0.8\linewidth]{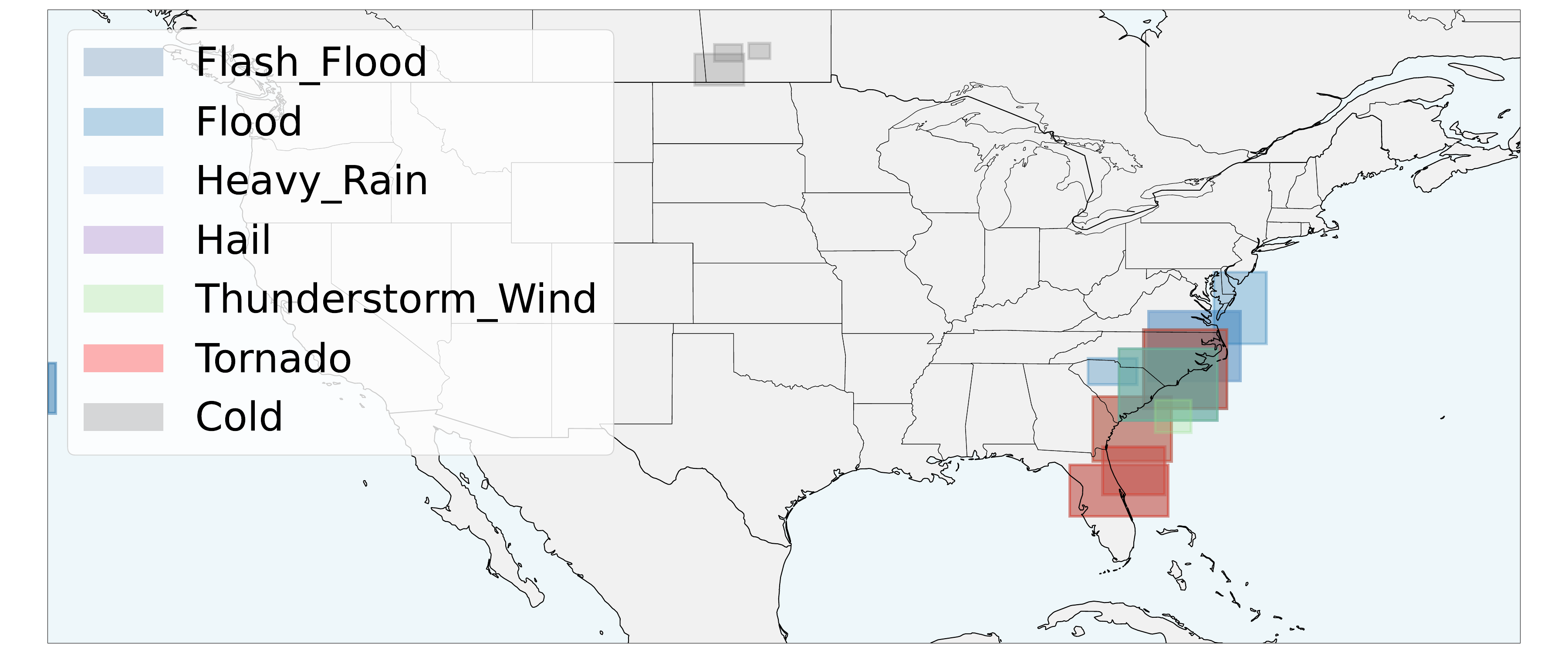}\label{figure:bbox}}
    \vspace{-8pt}
    \caption{Empirical analysis of extreme weather.}
    \vspace{-10pt}
    \label{figure:empirical}
\end{figure}

\section{Introduction} 
\label{section:intro}

In recent years, the frequency of extreme weather events has risen markedly in response to accelerated global climate change~\cite{camps2025artificial}. Accurate prediction of such events is essential for mitigating their severe socioeconomic consequences, safeguarding lives, and guiding timely disaster preparedness and response~\cite{lubchenco2012predicting}. 
Concurrently, deep learning-driven Foundation Models (FMs) have significantly reshaped the landscape of weather forecasting~\cite{chen2023foundation}, \eg~Pangu-Weather~\cite{bi2023accurate} and GraphCast~\cite{lam2023learning}, both trained on extensive global weather reanalysis datasets~\cite{hersbach2020era5} and outperform state-of-the-art numerical weather prediction (NWP) models. 
However, these FMs have predominantly concentrated on general weather conditions, whereas forecasting extreme weather poses a substantially greater challenge due to the rarity, nonlinear dynamics, and sensitivity to fine-scale processes inherent in such phenomena~\cite{olivetti2024data,materia2024artificial}. As a result, a significant research gap remains in advancing FMs to further enhance their extreme weather forecasting capabilities.

Previous studies for extreme weather forecasting primarily focus on specific event types, such as extreme precipitation~\cite{zhang2023skilful}, heatwaves~\cite{lopez2023global}, thunderstorms~\cite{guastavino2022prediction}, tornadoes~\cite{lagerquist2020deep}, and floods~\cite{sankaranarayanan2020flood}, failing to generalize across diverse extreme weather phenomena.
More recent works seek to learn generalized extreme-weather features.
For example, ExtremeCast~\cite{xu2024extremecast} mitigates the over-smoothing issue in extreme predictions via reformulating the loss function guided by the Extreme Value Theory (EVT). 
OneForecast~\cite{gao2025oneforecast} captures the extreme patterns through a high-pass Graph Neural Network~(GNN). 
However, both approaches lack supervision signals from varied real-world extreme weather events, which significantly limits their ability to model the diverse and complex nature of actual extreme weather patterns.

In this study, we aim to develop extreme weather FMs that enable the forecasting of a wide range of real-world extreme phenomena.
However, achieving this goal poses two critical challenges:
\textbf{(1)~How to discern the underlying extreme weather patterns?} 
A key to modeling extreme weather lies in distinguishing its underlying patterns from those of normal weather.
We conduct spectral analysis of ground-truth extreme events by comparing the High-Frequency Area (HFA)~\cite{tang2022rethinking} on around 36.4 million normal weather regions and 882 thousand extreme weather regions in the U.S. area of 2024. 
To avoid bias from the scarcity of extremes, at each timestep, we randomly sample regions matching the quantity of extreme regions.
As shown in \figref{figure:hfa}, it reveals that the HFA distribution of extreme regions largely overlaps with that of normal and random regions, but exhibits a noticeable "right-shift" phenomenon, \ie the Wasserstein distance between normal-extreme HFA distributions is much larger than that between normal-random ones. This indicates that extreme weather regions tend to concentrate more on high frequencies compared to general conditions.
While prior work~\cite{gao2025oneforecast} attempts to isolate high-frequency signals in meteorological data using a high-pass GNN, it suppresses valuable information at other frequencies, leading to the loss of universal and fundamental weather patterns;
\textbf{(2)~How to accommodate diverse yet coexisting extreme phenomena?} 
Extreme weather events encompass a variety of phenomena (\eg storms, floods, tornadoes), which exhibit a hierarchical structure in their formation: distinct physical drivers govern inter-type variations across event categories, while diverse atmospheric patterns lead to intra-type variations within the same event type~\cite{raymond2020understanding}. 
Moreover, through spatial analysis of real extreme weather events, we observe that extremes often occur concurrently across distributed or overlapping areas, \ie in 2024's U.S., approximately 86\% of timesteps feature composite extremes, with an average point-level overlap rate of 69\%. For example, \figref{figure:bbox} shows that the U.S. East Coast experiences compound extremes, including simultaneous heavy precipitation, flooding, and strong wind events.
Thus, prior general-purpose extreme weather models~\cite{xu2024extremecast,gao2025oneforecast} struggle to capture such fine-grained distinctions and the geographic mixture of various extreme phenomena.

\begin{table}[t]
\centering
\caption{Comparisons of foundation models in extreme event scenarios, analyzing "extreme event supervision" (emphasizing "real" and "diverse" extremes), "extreme-tailored method design", and "extreme-focused evaluation".}
\vspace{-5pt}
\resizebox{\linewidth}{!}{ 
\begin{tabular}{c|cc|c|c}
\toprule
\multirow{2}{*}{\textbf{Method}} & \multicolumn{2}{c|}{\textbf{Supervision}} & \multirow{2}{*}{\textbf{Extreme-tailored}} & \multirow{2}{*}{\textbf{Evaluation}} \\
\cmidrule{2-3}
& \textbf{Real} & \textbf{Diverse} & & \\
\midrule
GraphCast & \XSolidBrush & \XSolidBrush & \XSolidBrush & 3 Types of Records\\
PanguWhether & \XSolidBrush & \XSolidBrush & \XSolidBrush & Single-Type Records\\
ExtremeCast & \XSolidBrush & \XSolidBrush & \CheckmarkBold & Percentile Events\\
OneForecast & \XSolidBrush & \XSolidBrush & \CheckmarkBold & Single-Type Records\\
UniExtreme & \CheckmarkBold & \CheckmarkBold & \CheckmarkBold & 18 Types of Records\\
\bottomrule
\end{tabular}
}
\vspace{-15pt}
\label{table:fm}
\end{table}

To this end, we propose \textbf{\model}, a \underline{\textbf{uni}}versal \underline{\textbf{extreme}} weather foundation model that enables forecasting of 18 types of extreme weather events without the need for additional fine-tuning.
The distinction between our study and prior works is highlighted in \tabref{table:fm}.
Specifically, \model~pioneers in leveraging \textbf{labeled data from diverse real-world extreme weather events} in conjunction with general meteorological data to extend the forecasting capabilities of FMs for actual and diversified extreme weather.
The proposed UniExtreme model highlights two key innovations: 
(1) \textit{Adaptive Frequency Modulation (AFM)} module models spectral differences between normal and extreme weather regions through multiple learnable Beta-distribution spectral filters and multi-granularity frequency band aggregation; 
\textit{(2) Event Prior Augmentation (EPA)} module incorporates diverse labeled extreme atmospheric patterns as authentic prior memories customized for distinct regions, and employs a region-wise memory fusion block to resolve the hierarchical pattern structure as well as compound effects of coexisting and intertwined hazards.
These components synergize with a Swin-Transformer backbone~\cite{liu2022swin} to achieve skillful and generalizable extreme weather forecasting.

Our main contributions are summarized as follows: 
(1) To the best of our knowledge, \model~is the first extreme weather foundation model built upon both labeled data from diverse real-world extreme weather events and general meteorological data, enabling skillful forecasting on 18 types of extreme weather events through a unified model without requiring additional fine-tuning;
(2) \model~is powered by an Adaptive Frequency Modulation module to discern normal and extreme weather patterns from a spectral perspective, and an Event Prior Augmentation module to cope with diverse yet coexisting extreme phenomena; 
(3) Extensive experiments demonstrate \model's superior effectiveness on both extreme and general weather forecasting tasks compared to state-of-the-art baselines, highlighting its universality in tackling 18 types of extreme events.
\vspace{-5pt}
\section{Preliminary}
\label{section:preliminary}

\subsection{Problem Statement}
\label{section:problem}
In this paper, we study the problem of hourly weather forecasting (\ie nowcasting), as extreme weather events often emerge and dissipate rapidly, making short-term prediction critical.
The general and extreme weather forecasting problems are formulated as follows:

\begin{prob}
\textbf{General Weather Forecasting.}
Given a grid region $\mathcal{G} \subset \mathbb{R}^{H \times W \times 2}$ representing latitude-longitude coordinates over a spatial domain of height $H$ and width $W$, let $\mathbf{X}^t \in \mathbb{R}^{H \times W \times C}$ denote the observed weather state at time $t$, where $C$ represents the number of atmospheric variables. The goal is to learn a forecasting model $f_\theta: \mathbb{R}^{H \times W \times C} \rightarrow \mathbb{R}^{H \times W \times C}$ that predicts the next-hour weather state $\mathbf{X}^{t+1}$ conditioned on $\mathbf{X}^t$, such that $\mathbf{X}^{t+1} = f_\theta(\mathbf{X}^t)$.
\label{problem:general}
\end{prob}

\begin{myDef}
\textbf{Extreme Weather Event.}
An extreme weather event $e^t_j \in \mathcal{E}^t = \{e^t_j\}_{j=1}^{|\mathcal{E}^t|}$ at time $t$ is defined as a polygonal region and the meteorological observations within it.
The extreme region is represented by its bounding box vertices $\{(e^t_{j,h}, e^t_{j,y})\}_{j=1}^4$ located in $\mathcal{G}$, along with a set of event type names $e^t_{j,\mathcal{T}}$, where $e^t_{j,h}\in\mathbb{R}$ and $e^t_{j,y}\in\mathbb{R}$ denote latitude and longitude coordinates of the $j$-th corner. 
\label{definition:event}
\end{myDef}

\begin{prob}
\textbf{Extreme Weather Forecasting.}
Given the extreme event records $\mathcal{E}^t$ and $\mathbf{X}^t\in \mathbb{R}^{H \times W \times C}$, the task is to learn a model $f_\theta$ that improves forecasting accuracy specifically over extreme regions. 
The model operates on the full input $\mathbf{X}^t$ and predicts the complete output $\mathbf{X}^{t+1}$, but focuses only on extreme regions $\mathbf{X}^{t+1}|_{e^t},e^t \in \mathcal{E}^t$.
\label{problem:extreme}
\end{prob}

\begin{figure*}[ht]
    \centering
    \includegraphics[width=0.85\linewidth]{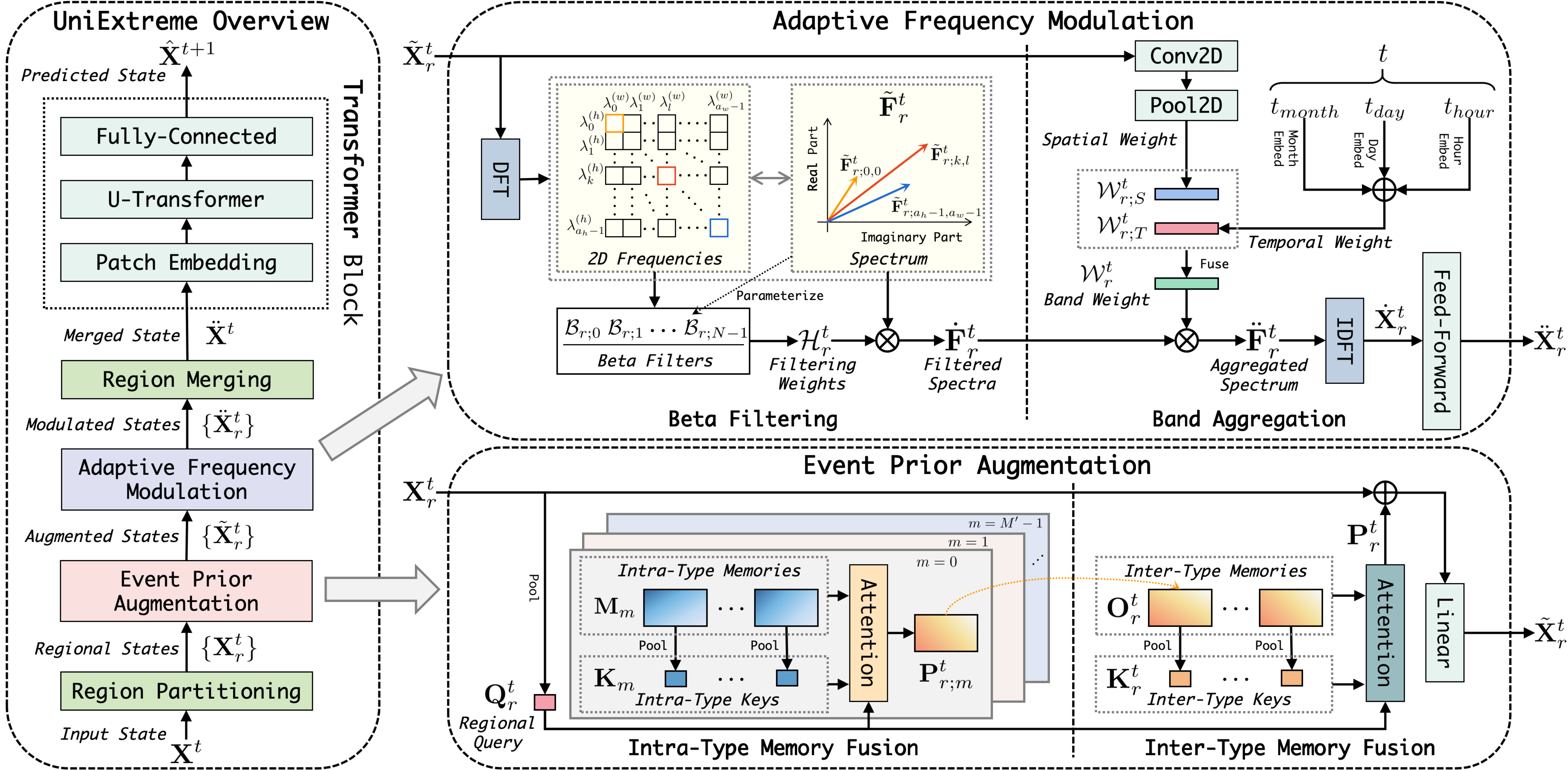}
    \vspace{-5pt}
    \caption{The overall architecture of UniExtreme.}
    \vspace{-10pt}
    \label{figure:method}
\end{figure*}

\vspace{-5pt}
\subsection{Fourier Transform and Inversion}
\label{section:fourier}
Given a weather state grid $\mathbf{X} \in \mathbb{R}^{H \times W \times C} $, its 2D Discrete Fourier Transform (DFT) is denoted by $\mathbf{F} = \mathcal{F}(\mathbf{X}) \in \mathbb{C}^{H \times W \times C}$, which decomposes the spatial signal into its frequency components. Specifically, the DFT at latitude-longitude frequencies $(\lambda_k^{(h)}, \lambda_l^{(w)})$ and variable $c$ is given by:
\begin{equation}
\mathbf{F}_{k,l,c} = \sum_{h=0}^{H-1} \sum_{w=0}^{W-1} \mathbf{X}_{h,w,c} \, e^{-2\pi i \left( \frac{k h}{H} + \frac{l w}{W} \right)} = \mathbf{R}_{k,l,c} + i\mathbf{I}_{k,l,c},
\end{equation}
where $0\leq k\leq H-1,0\leq l\leq W-1$, and $\mathbf{R}_{k,l,c}, \mathbf{I}_{k,l,c}\in\mathbb{R} $ represent the real and imaginary parts, respectively.  
The 2D inverse DFT (IDFT) reconstructs the original spatial signal from the frequency spectrum $ \mathbf{F} $ through $\mathbf{X} = \mathcal{F}^{-1}(\mathbf{F})$. For each location $ (h, w) $ and channel $ c $, the inversion is defined as
\begin{equation}
\mathbf{X}_{h,w,c} = \frac{1}{HW} \sum_{k=0}^{H-1} \sum_{l=0}^{W-1} \mathbf{F}_{k,l,c} \, e^{2\pi i \left( \frac{k h}{H} + \frac{l w}{W} \right)}.
\end{equation} 

To facilitate subsequent frequency-domain analysis, we further formulate the following definitions:

\begin{myDef}
\textbf{Spectral Energy.}
The spectral energy at frequencies $(\lambda_k^{(h)},\lambda_l^{(w)})$ for channel $c$ is defined as $E_{k,l,c} = |\mathbf{F}_{k,l,c}|^2 = \mathbf{R}_{k,l,c}^2 + \mathbf{I}_{k,l,c}^2$.
\label{definition:energy}
\end{myDef}

\begin{myDef}
\textbf{Radial Frequency.}
The radial frequency $\lambda_j$ at the flattened frequency index $j$ is defined as:
\begin{equation}
\lambda_j = \sqrt{{\lambda^{(h)}_{[j/W]}}^2 + {\lambda^{(w)}_{j\%W}}^2},\ j=0, ..., HW-1,
\end{equation}
where $({\lambda^{(h)}_{[j/W]}},{\lambda^{(w)}_{j\%W}})$ represent the corresponding 2D frequencies of $\lambda_j$ along the spatial dimensions.
\label{definition:radius}
\end{myDef}

\begin{myDef}
\textbf{Sorted Radial Frequency.}
The sorted radial frequencies $\{\lambda^{(s)}_j\}_{j=0}^{HW-1}$ are obtained by sorting the radial frequencies in ascending order, such that:
$\lambda^{(s)}_0 < \lambda^{(s)}_1 < \cdots < \lambda^{(s)}_{HW-1}$.
\label{definition:sorted_radius}
\end{myDef}

\begin{myDef}
\textbf{Normalized Sorted Radial Frequency.}
The normalized sorted radial frequency $\dot{\lambda}^{(s)}_K$ is defined as:
$\dot{\lambda}^{(s)}_K = (\lambda^{(s)}_K - \lambda_{\text{min}})/(\lambda_{\text{max}} - \lambda_{\text{min}})\in [0,1],$
where $\lambda_{\text{min}} = \lambda^{(s)}_0$ and $\lambda_{\text{max}} = \lambda^{(s)}_{HW-1}$.
\label{definition:norm_sorted_radius}
\end{myDef}
\vspace{-5pt}
\section{UniExtreme}
\label{section:method}

\figref{figure:method} illustrates the overall architecture of UniExtreme.
First, the Region Partitioning module divides input weather states into uniform spatial regions to enable specialized modeling of extreme events. 
Next, the Event Prior Augmentation module constructs a diverse memory pool of ground-truth extreme patterns and captures hierarchical and compound extreme features via intra- and inter-type memory fusion. 
The Adaptive Frequency Modulation module then utilizes region-adaptive Beta filtering and spectral band aggregation to differentiate normal and extreme weather spectra. 
Then, the Region Merging module reconstructs the full-shape weather grid. 
A Transformer backbone finally integrates these enhanced representations for prediction. 
This model collectively addresses spectral distinctions, hierarchical diversity, and geographic blending of extreme weather, with each module detailed as follows.

\vspace{-5pt}
\subsection{Region Partitioning and Merging}
\label{section:partition}

Extreme weather events occur in the form of spatially contiguous regions, requiring region-specific processing. Therefore, we uniformly partition the weather inputs $\mathbf{X}^t \in \mathbb{R}^{H \times W \times C}$ into regional states, formulated as:
\begin{equation}
\mathbf{X}^t_{r} = \mathbf{X}^t[r_h\cdot a_h:(r_h+1)\cdot a_h, r_w\cdot a_w:(r_w+1)\cdot a_w, :],
\end{equation}
where $\mathbf{X}^t_{r} \in \mathbb{R}^{a_h \times a_w \times C}$ denotes the $r$-th region's weather state, $0\leq r\leq H^\prime W^\prime-1$.
Parameters $(a_h,a_w)$ denote the region size, and $0\leq r_h=[r/W^\prime] \leq H^\prime-1$, $0\leq r_w=r\%W^\prime\leq W^\prime-1$, where $H^\prime=H/a_h$, and $W^\prime=W/a_w$. 
After being processed via posterior region-wise modules, AFM and EPA, we merge the regions to recover the original shape. Unlike patch embedding (detailed in \secref{section:transformer}), this strategy preserves all original grid observations.

\subsection{Adaptive Frequency Modulation}

Prior studies like~\cite{gao2025oneforecast} have noted a link between extreme weather and high-frequency signals, but compelling evidence is still lacking.
We investigate the HFA distributions of ground-truth extreme and normal weather regions in 2024's U.S. area.
\figref{figure:hfa} reveals the real-world spectral distinction between extreme and normal weather regimes: extreme events exhibit significantly stronger high-frequency components (\ie right-shift).
Empirical details and more analysis previous to 2024 are provided in \appref{appendix:empirical_freq}, which further reveals the broader universality of such frequency property and deficiencies of indiscriminate weather modeling. 
Thus, we propose an Adaptive Frequency Modulation (AFM) module, which aims to adjust frequencies for distinct regions in a customized manner, so as to capture the normal-extreme differences.

Although frequency filtering has shown promise~\cite{yi2024filternet}, existing methods face three kinds of limitations which impede region-wise normal-extreme discrimination: they either (1) pass only certain bands of frequencies which disregards both the global weather spectrum and fine-grained atmospheric spectral details~\cite{du2023frequency,wang2024empowering,li2025climatellm}, or (2) use rigid filter parameters lacking adaptability to handle the distinct frequency concentration for normal and extreme weather regions~\cite{tang2022rethinking}, or (3) apply fully-learnable filters that are unable to smoothly curate the continuous spectral distribution of normal and extreme weather~\cite{yi2024filternet,zhang2025frequency}.
In this study, inspired by spectral modeling capabilities of Beta kernels~\cite{tang2022rethinking}, we design multiple region-specific Beta filters to modulate fine-grained spectral signals in diverging local frequency bands.
The parameters of Beta distributions are learnable specific to each region, which achieve the balance between kernel smoothness and region-wise adaptability.
In addition, we design a band aggregation network to synthesize the nuanced modulation results of different local filters to consolidate the global spectral information. 

\noindent\textbf{Beta Filtering.} 
Applying 2D Fourier transform to EPA-generated $\tilde{\mathbf{X}}^t_{r}$ yields $\tilde{\mathbf{F}}^t_{r}=\mathcal{F}(\tilde{\mathbf{X}}^t_{r})=\mathbf{R}^t_{r}+i\mathbf{I}^t_{r}\in\mathbb{C}^{a_h\times a_w\times C}$ with associated region-agnostic radial frequency parameters $\{\lambda_j,\lambda^{(s)}_j,\dot\lambda^{(s)}_j\}_{j=0}^{a_h a_w-1}$. 
Then, we define the functions of $N$ region-adaptive Beta filters to modulate local signals of distinct frequency ranges:
\begin{equation}
\mathcal{B}_{r;n}(x)=\frac{\dot{x}^{\alpha_{r;n}-1}(1-\dot{x})^{\beta_{r;n}-1}}{\tilde{\lambda}_{r;n}^{\alpha_{r;n}-1}(1-\tilde{\lambda}_{r;n})^{\beta_{r;n}-1}},x\in[\lambda_{\text{min}},\lambda_{\text{max}}],
\label{equation:beta}
\end{equation}
where $n=0,...,N-1$, and $\dot{x}=(\dot{x}-\lambda_{\text{min}})/(\lambda_{\text{max}}-\lambda_{\text{min}})\in[0,1]$ denotes the normalized variable of sorted radial frequency. 
The denominator in \equref{equation:beta} is applied to normalize $\mathcal{B}_{r;n}(x) \in [0,1]$, ensuring the fairness of local spectral modulation. 
The Beta parameters satisfy $
\alpha_{r;n}=\tilde{\lambda}_{r;n}(\kappa_{r;n}-2)+1\geq 1$, $\beta_{r;n}=(1-\tilde{\lambda}_{r;n})(\kappa_{r;n}-2)+1 \geq 1$, where $0\leq\tilde{\lambda}_{r;n}=\frac{\alpha_{r;n}-1}{\alpha_{r;n}+\beta_{r;n}-2}\leq1$ and $\kappa_{r;n}=\alpha_{r;n}+\beta_{r;n}\geq2$. 
Intuitively, two parameters, $\tilde{\lambda}_{r;n}$ and $\kappa_{r;n}$, determine the effects of frequency manipulation of a filter. Specifically, $\tilde{\lambda}_{r;n}$, \ie \textit{mode}, controls the radial frequency matching the distribution peak; and $\kappa_{r;n}$, \ie \textit{spread}, controls the concentration of the distribution around the mode.
Once $\tilde{\lambda}$ and $\kappa$ are confirmed, we can infer $\alpha,\beta$, thereby determining all parameters of the filter $\mathcal{B}_{r;n}$ in \equref{equation:beta}.

Thus, our goal is to confirm $\tilde{\lambda}$ and $\kappa$ for the filters.
First, to enforce coverage of diverse frequency bands, we manually specify the mode parameters $\tilde{\lambda}_{r;n}$, making them agnostic to regions. From \figref{figure:hfa}, we find that the weather states cater primarily to low frequencies (\ie high frequency area close to 0), inspiring us to employ a logarithmic-scale strategy for mode determination. In particular, we split the normalized sorted radial frequency $\{\dot{\lambda}^{(s)}_j\}_{j=0}^{a_h a_w-1}$ into $N$ bands, with band width increased by radial frequency at a growth rate $\gamma$, which ensures finer-grained processing for lower frequency components.
Then we define the modes as each band's medium:
\begin{align}
&\tilde{\lambda}_{r;n}=\text{Medium}(\{\dot\lambda^{(s)}_{B_{n,0}},...,\dot\lambda^{(s)}_{B_{n,|B_n|-1}}\}),\\
&|B_{n+1}|=\gamma\cdot|B_n|,n=0,...,N-1,
\end{align}
where $B_n$ denotes the indices of sorted radial frequency in the $n$-th band, satisfying $B_{n+1,0}=B_{n,|B_n|-1}+1,B_{n,j+1}=B_{n,j}+1$ and boundary conditions: $|B_0|=1,\sum_{n=0}^{N-1} |B_n|=a_ha_w$.
As for the spread $\kappa$ of Beta filter, we make them learnable by applying linear transformation to facilitate the region-adaptive modeling of diverse local spectral properties, detailed as:
\begin{equation}
\kappa_r=2+\text{MAX}_\kappa\sigma([\mathbf{R}^t_{r}:\mathbf{I}^t_{r}]\mathbf{W}_\kappa+\mathbf{b}_\kappa),
\end{equation}
where $\mathbf{W}_\kappa\in\mathbb{R}^{2C\times N}, \mathbf{b}_\kappa\in\mathbb{R}^N$, $\sigma(x)=1/(1+e^{-x})$ is the sigmoid function, $[:]$ denotes the concatenation function. Thus, the parameters $\kappa_r\in\mathbb{R}^N$ are settled for $N$ Beta filters, wherein we ensure $2\leq\kappa_{r;n}\leq \text{MAX}_\kappa+2$ to avoid numerical errors. 

Based on the learned Beta filters, we modulate the spectrum at diverging frequency scopes by 
\begin{equation}
\dot{\mathbf{F}}^t_{r}=\mathcal{H}^t_{r} \cdot \tilde{\mathbf{F}}^t_{r} \in\mathbb{R}^{N\times a_h\times a_w\times C}.
\end{equation}
We broadcast the spectrum to the same shape of the filtering weight $\mathcal{H}^t_{r}\in \mathbb{R}^{N\times a_h\times a_w}$, which is defined as: 
\begin{equation}
\begin{aligned}
&\mathcal{H}^t_{r;n,k,l}=\mathcal{B}_{r;n}(\lambda_{k\cdot a_w+l}), \ n=0,...,N-1,\\ 
&\lambda_{k\cdot a_w+l}=\sqrt{{\lambda_k^{(h)}}^2+{\lambda_l^{(w)}}^2}, \ k=0,...,a_h-1;l=0,...,a_w-1.\\
\end{aligned}
\end{equation}

\noindent\textbf{Band Aggregation.}
To combine modulated signals across multi-granularity frequencies, we aggregate $N$ local filtered spectra through a learnable band aggregation weight $\mathcal{W}^t_{r}\in\mathbb{R}^{N\times C}$, formulated as:
\begin{equation}
\ddot{\mathbf{F}}^t_{r}=\text{Sum}(\mathcal{W}^t_{r}\cdot \dot{\mathbf{F}}^t_{r}), \ \ddot{\mathbf{F}}^t_{r} \in\mathbb{R}^{a_h\times a_w\times C}.
\end{equation} 
To determine the weight, we utilize a Convolution Neural Network (CNN) followed by mean pooling to exploit spatial information. In addition, to identify time-evolving frequency differences, we also integrate time embedding into the weights. Spatial and temporal weights are fused into a spatiotemporal weight by a linear layer:
\begin{align}
&\mathcal{W}^t_{r;S} = \text{MeanPool2D}(\text{Conv2D}(\mathbf{X}^t_{r})),\\
&\mathcal{W}^t_{r;T} = (\mathbf{E}_m(t_{\text{month}}) + \mathbf{E}_d(t_{\text{day}}) + \mathbf{E}_h(t_{\text{hour}}))\mathbf{W}_t+\mathbf{b}_t,\\
&\mathcal{W}^t_{r} = \text{Softmax}([\mathcal{W}^t_{r;S}:\mathcal{W}^t_{r;T}]\mathbf{W}_w+\mathbf{b}_w), \label{equation:band_weight}
\end{align}
where $\mathcal{W}^t_{r;S}\in\mathbb{R}^{N\times C},\mathcal{W}^t_{r;T}\in\mathbb{R}^{N}$ are spatial and temporal weights, respectively. 
The temporal weight is merged by three learnable embeddings $\mathbf{E}_m\in\mathbb{R}^{12\times D_T},\mathbf{E}_d\in\mathbb{R}^{31\times D_T},\mathbf{E}_h\in\mathbb{R}^{24\times D_T}$, representing month of year, day of month, and hour of day, respectively, for a given time scalar $t$, where $D_T$ denotes the embedding dimension. 
The linear parameters $\mathbf{W}_t\in\mathbb{R}^{D_T\times N},\mathbf{b}_t\in\mathbb{R}^N,\mathbf{W}_w\in\mathbb{R}^{2N\times N},\mathbf{b}_w\in\mathbb{R}^N$.

Finally, we apply IDFT on the aggregated spectrum to derive $\dot{\mathbf{X}}^t_{r}=\mathcal{F}^{-1}(\ddot{\mathbf{F}}^t_{r})$. Moreover, we transform the modulated grids by a Feed-Forward Network (FFN) with residual connection and layer normalization~\cite{vaswani2017attention}, which are formulated as:
\begin{equation}
\ddot{\mathbf{X}}_r^t=\dot{\mathbf{X}}_r^t+\text{FFN}(\text{LayerNorm}(\dot{\mathbf{X}}_r^t)), 
\end{equation}
where $\ddot{\mathbf{X}}^t_{r}\in\mathbb{R}^{a_h\times a_w\times C}$. The modulated states $\{\ddot{\mathbf{X}}^t_{r}\}_{r=0}^{H'W'-1}$ are merged to reconstruct the full-resolution format $\ddot{\mathbf{X}}^t \in \mathbb{R}^{H \times W \times C}$.

\vspace{-5pt}
\subsection{Event Prior Augmentation}
\label{section:epa}

To address the hierarchical heterogeneity and intricate co-occurrence structures of extreme events illustrated in \figref{figure:bbox} (more examples are provided in \appref{appendix:empirical_space}), we propose an Event Prior Augmentation (EPA) module which incorporates diversified extreme priors as event memories to enrich different geographic areas in a specialized way, and utilizes an attention-based memory fusion network to model event pattern hierarchy as well as cumulative effects of compound disasters. 

Memory-related designs have emerged in recent studies such as memory networks~\cite{sukhbaatar2015end,yuan2024unist} and prompt learning~\cite{jia2022visual,liutoken}, but they encounter several issues hindering adaptation to our extreme-focused task: 
(1) the memories/prompts are either fully-learnable~\cite{liutoken,yuan2024unist} or extracted from model-generated features~\cite{liu2023cross}, which struggle to automatically capture authentic extreme priors; (2) the memories/prompts are either added to the pixel-level raw inputs~\cite{wang2024lion,tsaoautovp} which fail to accommodate region-scale extreme events, or integrated with patch-level feature sequences~\cite{jia2022visual,liutoken} which are incapable of sensing the raw extreme details; (3) the flat memory/prompt architectures lack hierarchical modeling capacity to capture both inter-type distinctions and intra-type pattern variations in extreme weather events.
To this end, we first construct a categorical memory pool of multifarious region-wise events under the indirect guidance of ground-truth extreme weather records, which operate as event priors.
Then, an attention-based dual-level memory fusion network is applied for each partitioned regional input $\{\mathbf{X}^t_{r}\}_{r=0}^{H^\prime W^\prime-1}$ in a specialized way, which aims to capture the hierarchical intra- and inter-type discrepancies, along with the coupling impacts of intertwined extremes.

\noindent\textbf{Memory Construction.}
To leverage real extreme event priors, the event prior memory pool is constructed directly from a subset of training data containing extreme event records. To accommodate disparate spatial scales of extreme events, we implement the uniform region partitioning strategy described in \secref{section:partition}. Each selected region grid $\mathbf{X}_{r}$ is annotated with a multi-hot type vector $\mathbf{c}_r\in\{0,1\}^{M^\prime}$, where $M^\prime=M+1$ represents the total number of event types (including $M$ extreme types and one "normal" type). For quantity balance, we sample normal regions matching the quantity of extreme regions during memory construction. 
This process yields a categorized memory pool $\mathcal{M}=\{\mathcal{M}_m\}_{m=0}^{M^\prime-1}$, where each $\mathcal{M}_m=\{\mathbf{X}^{t^m_j}_{r^m_j}\}_{j=0}^{|\mathcal{M}_m|-1}$ contains relevant weather regions of spatiotemporal indices $(t^m_j,r^m_j)$. To address memory quantity disparities across types that might hinder parallel processing, we apply KMeans clustering to standardize each $\mathcal{M}_m$ into $\dot{\mathcal{M}}_m$ with a fixed capacity of $U$ memories. The final memory pool $\dot{\mathcal{M}}=\{\dot{\mathcal{M}}_m\}_{m=0}^{M^\prime-1}$ incorporates zero-padding for types with insufficient samples, complemented by a masking mechanism to circumvent the influence of padded entries during memory fusion.

\noindent\textbf{Memory Fusion.}
To integrate hierarchical patterns and superimposed extremes,
the memory fusion process employs the attention mechanism~\cite{vaswani2017attention} with two levels: (1) intra-type fusion aggregates $\dot{\mathcal{M}}_m$ into type-specific fused memory $\mathbf{P}^t_{r;m}$, and (2) inter-type fusion integrates $\{\mathbf{P}^t_{r;m}\}_{m=0}^{M^\prime-1}$ into a hybrid-type memory $\mathbf{P}^t_{r}$.

First, each $\mathbf{X}^t_{r}$ independently queries memories within each type, through concatenated representations $\mathbf{M}_m=\text{Concat}(\dot{\mathcal{M}}_m)\in\mathbb{R}^{U\times a_h\times a_w\times C}$. 
To handle the spatial dimensions of meteorological inputs and memories, we implement the querying process via the learnable pooling layer:
\begin{align}
&\mathbf{Q}^t_{r}=\text{MeanPool2D}(\text{Conv2D}(\mathbf{X}^t_{r})),\\
&\mathbf{K}_m=\text{MeanPool2D}(\text{Conv2D}(\mathbf{M}_m)),\\
&\mathbf{P}^t_{r;m}=\text{Attention}(\mathbf{Q}^t_{r},\mathbf{K}_m,\mathbf{M}_m),
\end{align}
where $\mathbf{Q}^t_{r}\in\mathbb{R}^{C},\mathbf{K}_m\in\mathbb{R}^{U\times C},\mathbf{P}^t_{r;m}\in\mathbb{R}^{a_h\times a_w\times C}$.
Secondly, the inter-type fusion similarly unites multiple type-specific fused memories $\{\mathbf{P}^t_{r;m}\}_{m=0}^{M^\prime-1}$, with $\mathbf{Q}^t_{r}$ as the query:
\begin{align}
&\mathbf{O}^t_{r}=\text{Concat}(\{\mathbf{P}^t_{r;m}\}_{m=0}^{M^\prime-1},\\
&\mathbf{K}^t_{r}=\text{MeanPool2D}(\text{Conv2D}(\mathbf{O}^t_{r})),\\
&\mathbf{P}^t_{r}=\text{Attention}(\mathbf{Q}^t_{r},\mathbf{K}^t_{r},\mathbf{O}^t_{r}), \label{equation:epa_attn}
\end{align}
where $\mathbf{O}^t_{r}\in\mathbb{R}^{M^\prime\times a_h\times a_w\times C},\mathbf{K}^t_{r}\in\mathbb{R}^{M^\prime\times C},\mathbf{P}^t_{r}\in\mathbb{R}^{a_h\times a_w\times C}$.
Lastly, we enhance the raw inputs $\mathbf{X}^t_{r}$ with the hybrid-type memory $\mathbf{P}^t_{r}$ to obtain augmented weather states, through residual connection:
\begin{equation}
\tilde{\mathbf{X}}^t_{r} = (\mathbf{X}^t_{r}+\mathbf{P}^t_{r})\mathbf{W}_m +\mathbf{b}_m,
\end{equation}
where $\mathbf{W}_m\in\mathbb{R}^{C\times C},\mathbf{b}\in\mathbb{R}^C$.

\begin{table*}[t]
\centering
\caption{Normalized weather forecasting results. "Gen." and "Ext." indicate "General" and "Extreme", respectively.}
\begin{tabular}{c|ccc|ccc|ccc}
\toprule
\multirow{2}{*}{\textbf{Method}} & \multicolumn{3}{c|}{\textbf{MAE ($1\times e^{-2}$)}} & \multicolumn{3}{c|}{\textbf{RMSE ($1\times e^{-2}$)}} & \multicolumn{3}{c}{\textbf{ACC ($1\times e^{-1}$)}}\\
\cmidrule(lr){2-4} \cmidrule(lr){5-7} \cmidrule(lr){8-10} 
&  \textbf{Ext.$\downarrow$} & \textbf{Gen.$\downarrow$} & \textbf{Gap$\downarrow$} &  \textbf{Ext.$\downarrow$} & \textbf{Gen.$\downarrow$} & \textbf{Gap$\downarrow$} &  \textbf{Ext.$\uparrow$} & \textbf{Gen.$\uparrow$} & \textbf{Gap$\downarrow$}\\
\hline\textbf{NWP} & 12.4471 & 8.1758 & 4.2714 & 19.7998 & 12.7407 & 7.0591 & 9.4632 & 9.8157 & 0.3525\\ 
\hline
\textbf{GraphCast} & \underline{\textit{9.0245}} & \underline{\textit{6.0371}} & \underline{\textit{2.9874}} & \underline{\textit{13.9938}} & \underline{\textit{9.1884}} & \underline{\textit{4.8054}} & 9.6823 & 9.9002 & 0.2179\\ 
\hline
\textbf{Pangu} & 10.5085 & 6.8582 & 3.6503 & 16.5176 & 10.4857 & 6.0320 & 9.5570 & 9.8680 & 0.3110\\ 
\hline
\textbf{FengWu} & 9.9437 & 6.5556 & 3.3880 & 15.4971 & 9.9643 & 5.5328 & 9.6241 & 9.8824 & 0.2583\\ 
\hline
\textbf{FuXi} & 9.3014 & 6.1365 & 3.1649 & 14.3104 & 9.2007 & 5.1097 & \underline{\textit{9.6914}} & \underline{\textit{9.9040}} & \underline{\textit{0.2126}}\\ 
\hline
\textbf{OneForecast} & 9.2762 & 6.2198 & 3.0564 & 14.3574 & 9.4508 & 4.9066 & 9.6681 & 9.8933 & 0.2252\\ 
\hline
\textbf{UniExtreme} & \textbf{8.0402} & \textbf{5.1602} & \textbf{2.8799} & \textbf{12.6157} & \textbf{7.9845} & \textbf{4.6312} & \textbf{9.7359} & \textbf{9.9214} & \textbf{0.1855}\\ 
\bottomrule
\end{tabular}
\vspace{-5pt}
\label{table:main_norm}
\end{table*}

\begin{figure*}[h!]
    \centering
    \includegraphics[width=0.88\linewidth]{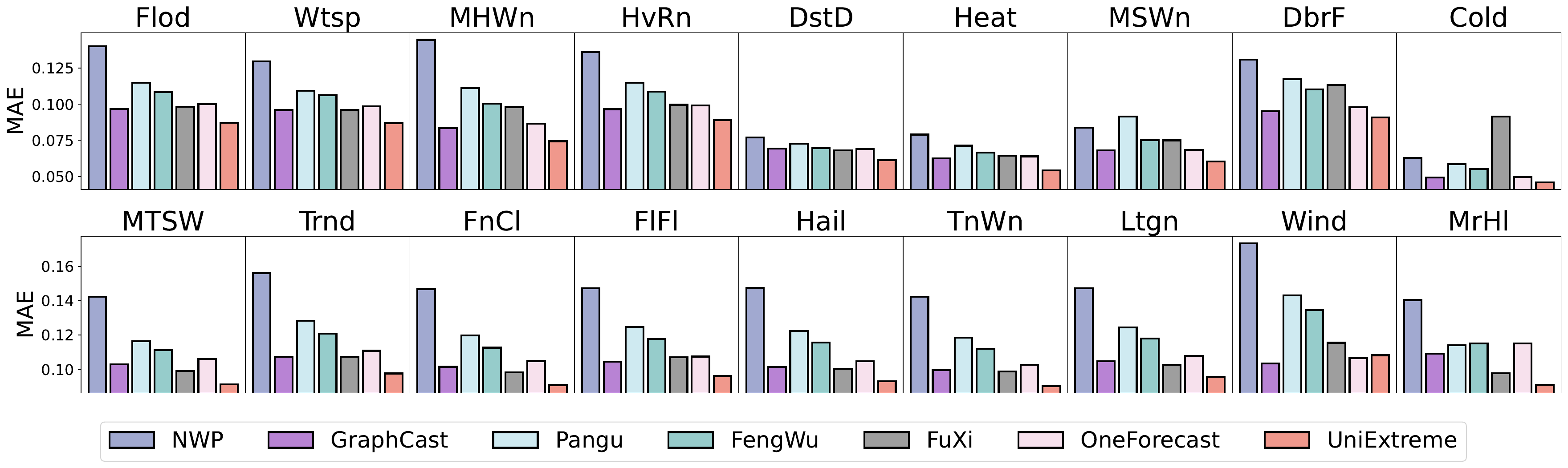}
    \vspace{-8pt}
    \caption{Comparison of different methods in forecasting 18 types of extreme events.}
    \vspace{-12pt}
    \label{figure:main_type}
\end{figure*}

\subsection{Transformer Backbone}
\label{section:transformer}
We instantiate our Transformer backbone with FuXi~\cite{chen2023fuxi}, adapted for single-time-step processing through three key components:

\noindent\textbf{Patch Embedding Layer.} 
Unlike the original FuXi architecture, our implementation processes weather states at individual time steps. The patch embedding layer transforms AFM-processed states $\ddot{\mathbf{X}}^t$ via $\mathbf{Z}^t_0 = \text{LayerNorm}(\text{Conv2D}(\ddot{\mathbf{X}}^t))$,
yielding patch embeddings $\mathbf{Z}^t_0 \in \mathbb{R}^{\overline{H} \times \overline{W} \times D}$, where $\overline{H}=H/p_h$, $\overline{W}=W/p_w$ denote the spatial dimensions in patch units, $(p_h,p_w)$ represents the patch size, and $D$ is the embedding dimension.

\noindent\textbf{U-Transformer.} The core processing comprises $L$ Swin Transformer~\cite{liu2021swin,liu2022swin} blocks employing window-based attention and shifted-window strategies for computational efficiency:
\begin{equation}
\mathbf{Z}^t_j = \text{SwinTransformer}^l(\mathbf{Z}^t_{j-1}), \quad j=1,...,L.
\end{equation}

\noindent\textbf{Fully-Connected Layer.} The final transformation recovers the original input dimensions by $\mathbf{Z}^t_{L+1} = \mathbf{Z}^t_L \mathbf{W}_z + \mathbf{b}_z$,
where $\mathbf{W}_z\in\mathbb{R}^{D\times(p_h \cdot p_w \cdot C)}, \mathbf{b}_z\in\mathbb{R}^{(p_h \cdot p_w \cdot C)}$. Finally, we permute and reshape the representation $\mathbf{Z}^t_{L+1}\in\mathbb{R}^{\overline{H} \times \overline{W} \times (p_h \cdot p_w \cdot C)}$ into $\hat{X}^{t+1}\in\mathbb{R}^{H \times W \times C}$ that serves as the predicted weather state at time $t+1$.

\subsection{Optimization}

Following established practices~\cite{chen2023fuxi,bi2023accurate}, we optimize UniExtreme using the L1 loss function, without using the extreme records $\mathcal{E}^t$:
\begin{equation}
\mathcal{L}^t = \frac{1}{HWC}\sum_{h=0}^{H-1}\sum_{w=0}^{W-1}\sum_{c=0}^{C-1} |\hat{\mathbf{X}}^{t+1}_{h,w,c}-\mathbf{X}^{t+1}_{h,w,c}|,
\end{equation}
where $0\leq t\leq T-1$, and $T$ denotes the number of timesteps in training. All important notations are summarized in~\appref{appendix:notation}.

\vspace{-5pt}
\section{Experiment}

\subsection{Experimental Setups}

\noindent\textbf{Dataset.}
In this study, we follow HR-Extreme dataset~\cite{ran2025hr}, which provides high-resolution (3km$\times$3km, 1-hour interval) weather states across the contiguous U.S., spanning latitudes 21.1°N to 52.6°N and longitudes 225.9°E to 299.1°E. The original dataset contains weather states from 2019-2020 across 69 variables, along with extreme event annotations for 2020.
To enable more comprehensive evaluation, we construct \textit{HR-Extreme-V2}, an enhanced 26TB dataset covering atmospheric states from 2019-2024 and corresponding annotated extreme events, spanning 18 event types.
For computational efficiency, we convert the spatial resolution to approximately 6km$\times$6km (530$\times$900 grid points). 
The training set comprises 2019-2022 data, with 2023 and 2024 reserved for validation and testing respectively. 
Additional dataset details are provided in~\appref{appendix:dataset}, while the ethical use of data is discussed in \appref{appendix:ethical}.

\noindent\textbf{Baseline.}
We compare our model against one NWP method, WRF-ARWv3.9+~\cite{WRF-ARW}; and five DL-based method: GraphCast~\cite{lam2023learning}, PanguWeather~\cite{bi2023accurate}, FengWu~\cite{chen2023fengwu}, FuXi~\cite{chen2023fuxi}, OneForecast~\cite{gao2025oneforecast}.

\noindent\textbf{Metric.}
Following existing works~\cite{liucirt,gao2025oneforecast}, we adopt three evaluation metrics: Mean Absolute Error (MAE), Root Mean Squared Error (RMSE), and ACC (Anomaly Correlation Coefficient), which are computed across three dimensions: \textit{General} denotes the overall prediction performance; \textit{Extreme} indicates the results in extreme weather regions, \textit{Gap} is the discrepancy between general and extreme forecasting capability. The \textit{Gap} metric focuses on improvements in extreme event prediction independent of general forecasting gains. Detailed metric formulations appear in~\appref{appendix:metric}.
Note that the reported ACC results are generally high (\ie $>0.9$), as we focus on predicting solely the 1-hour future state, \ie nowcasting, as described in \secref{section:problem}.

To address weather data non-stationarity, we apply instance normalization using 2019-2022 temporal statistics. More implementation details are provided in~\appref{appendix:setup}.

\begin{table*}[t]
\centering
\caption{Raw-scale weather forecasting MAE results. "Gen." and "Ext." indicate "General" and "Extreme", respectively.}
\resizebox{\linewidth}{!}{ 
\begin{tabular}{c|ccc|ccc|ccc|ccc}
\toprule
\multirow{2}{*}{\textbf{Method}} & \multicolumn{3}{c|}{\textbf{MSL}} & \multicolumn{3}{c|}{\textbf{V150}} & \multicolumn{3}{c|}{\textbf{Z500}} & \multicolumn{3}{c}{\textbf{T850}} \\
\cmidrule(lr){2-4} \cmidrule(lr){5-7} \cmidrule(lr){8-10} \cmidrule(lr){11-13}
& \textbf{Ext.$\downarrow$} & \textbf{Gen.$\downarrow$} & \textbf{Gap$\downarrow$} & \textbf{Ext.$\downarrow$} & \textbf{Gen.$\downarrow$} & \textbf{Gap$\downarrow$} & \textbf{Ext.$\downarrow$} & \textbf{Gen.$\downarrow$} & \textbf{Gap$\downarrow$} & \textbf{Ext.$\downarrow$} & \textbf{Gen.$\downarrow$} & \textbf{Gap$\downarrow$} \\
\hline\textbf{NWP} & 57.6042 & 45.9517 & 11.6525 & 1.1245 & 0.9072 & 0.2173 & 4.6947 & 4.1438 & 0.5509 & 0.6089 & 0.4117 & 0.1972\\ 
\hline
\textbf{GraphCast} & \underline{\textit{31.0413}} & \underline{\textit{28.1268}} & 2.9144 & 0.9938 & 0.7844 & 0.2095 & \underline{\textit{3.5127}} & \underline{\textit{2.9135}} & 0.5993 & \underline{\textit{0.3903}} & \underline{\textit{0.2727}} & \textbf{0.1175}\\ 
\hline
\textbf{Pangu} & 50.5370 & 41.8219 & 8.7150 & 0.9591 & 0.7567 & 0.2025 & 4.2033 & 3.6609 & 0.5425 & 0.5296 & 0.3656 & 0.1640\\ 
\hline
\textbf{FengWu} & 41.4503 & 35.6901 & 5.7602 & 0.9695 & 0.7679 & \underline{\textit{0.2017}} & 3.7781 & 3.2339 & 0.5442 & 0.4517 & 0.3146 & 0.1371\\ 
\hline
\textbf{FuXi} & 38.3128 & 33.0047 & 5.3081 & \underline{\textit{0.8182}} & \underline{\textit{0.6054}} & 0.2128 & 6.0101 & 4.5552 & 1.4549 & 0.5590 & 0.3502 & 0.2088\\ 
\hline
\textbf{OneForecast} & 32.3066 & 29.4170 & \underline{\textit{2.8896}} & 0.9948 & 0.7887 & 0.2061 & 3.5650 & 3.0467 & \underline{\textit{0.5183}} & 0.4004 & 0.2820 & \underline{\textit{0.1184}}\\ 
\hline
\textbf{UniExtreme} & \textbf{29.4688} & \textbf{27.6502} & \textbf{1.8187} & \textbf{0.6540} & \textbf{0.4888} & \textbf{0.1652} & \textbf{3.1841} & \textbf{2.7648} & \textbf{0.4193} & \textbf{0.3709} & \textbf{0.2517} & 0.1192\\ 
\bottomrule
\end{tabular}
}
\vspace{-5pt}
\label{table:main_raw}
\end{table*}

\begin{figure*}[h!]
    \centering
    \includegraphics[width=0.9\linewidth]{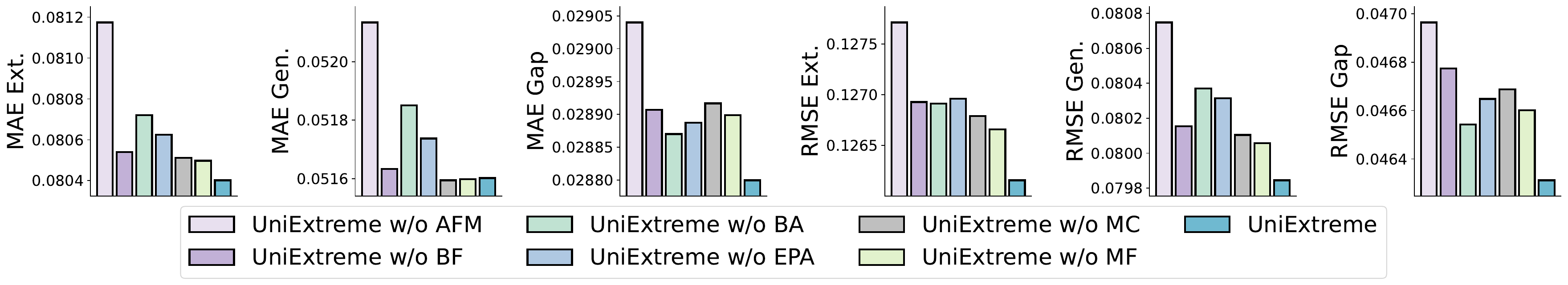}
    \vspace{-8pt}
    \caption{Ablation study of UniExtreme.}
    \vspace{-12pt}
    \label{figure:ablation}
\end{figure*}

\vspace{-5pt}
\subsection{Comparison Results}
\label{section:compare}

In this section, we present comprehensive comparisons across normalized states, raw-scale predictions, and extreme event-specific categorical performance.

\noindent\textbf{Normalized Forecasting Performance.}
\tabref{table:main_norm} presents normalized results averaged across all variables, yielding several key insights:
(1) \textit{UniExtreme demonstrates superior general and extreme forecasting capability.} Especially, it achieves approximately 11$\%$ MAE improvement and 10$\%$ RMSE gain, over the best baseline in extreme weather forecasting.
(2) \textit{DL-based approaches exhibit advantages over NWP models}, across all metrics. The data-driven methods significantly outperform the physical-based model, highlighting the inherent limitations of relying solely on numerical solutions to atmospheric equations, particularly in weather forecasting scenarios.
(3) \textit{Explicit pressure-level modeling provides limited benefits}. While PanguWeather and FengWu employ 3D Swin-Transformers to process pressure levels, they underperform across all metrics compared to methods using spatial modeling only, \eg GraphCast and OneForecast apply the GNN, while FuXi and UniExtreme utilize the 2D Swin-Transformer.

\noindent\textbf{Raw-Scale Forecasting Performance.}
\tabref{table:main_raw} presents MAE results at original scales, for four representative variables (\ie MSL, V150, Z600, T850) from distinct pressure levels. Key findings include:
(1) \textit{UniExtreme achieves significant improvements in extreme event prediction}, particularly for MSL where it reduces the normal-extreme performance gap by around $37\%$ compared to best baseline.
(2) \textit{Frequency modulation shows promise in enhancing extreme event forecasting}. OneForecast and UniExtreme are frequency-aware models which achieve strong performance in the Gap metric, underscoring the importance of frequency-domain manipulation for improving extreme event predictions.
Complete results (see~\appref{appendix:raw_var}) corroborate these conclusions across all measured variables.

\noindent\textbf{Categorical Forecasting Performance.}
\figref{figure:main_type} showcases the results of distinct methods in forecasting various extreme events.
We observe that UniExtreme \textit{consistently outperforms both NWP and DL-based baselines across all event categories}, demonstrating strong adaptability to diverse extreme weather patterns.
The event-wise comparisons in terms of other metrics (see~\appref{appendix:categorical}) further validate the universality and effectiveness of our model.

\vspace{-5pt}
\subsection{Ablation Study}

The ablation study investigates the impact of various model components on forecasting performance by comparing UniExtreme with several variants. Specifically, "w/o AFM" and "w/o EPA" denote the removal of the AFM and EPA modules, respectively. Additionally, four further degraded variants are evaluated: "w/o BF" replaces the Beta filters with fully learnable filters, "w/o BA" employs simple summation instead of weighted band aggregation, "w/o MC" substitutes the extreme event memories with randomly initialized memories, and "w/o MF" removes the intra-type fusion in the hierarchical memory fusion block.

\figref{figure:ablation} demonstrates the normalized forecasting performance in terms of MAE and RMSE metrics, revealing several critical insights:
(1) Both adaptive frequency modulation (w/o AFM) and extreme event prior integration (w/o EPA) are critical to UniExtreme's superior performance in extreme weather forecasting.
(2) Learnable Beta filters outperform purely learnable filters (w/o BF) in fine-grained spectral manipulation, while spatiotemporal weighting further improves local frequency aggregation compared to uniform summation (w/o BA).
(3) Authentic extreme patterns enhance extreme event comprehension over random patterns (w/o MC), and modeling intra-type variations improves hierarchical representation compared to flat type-level memory fusion (w/o MF).
These findings underscore the essential role of each proposed component in achieving robust and accurate extreme weather predictions.
\begin{figure*}[t]
    \centering
    \includegraphics[width=0.975\linewidth]{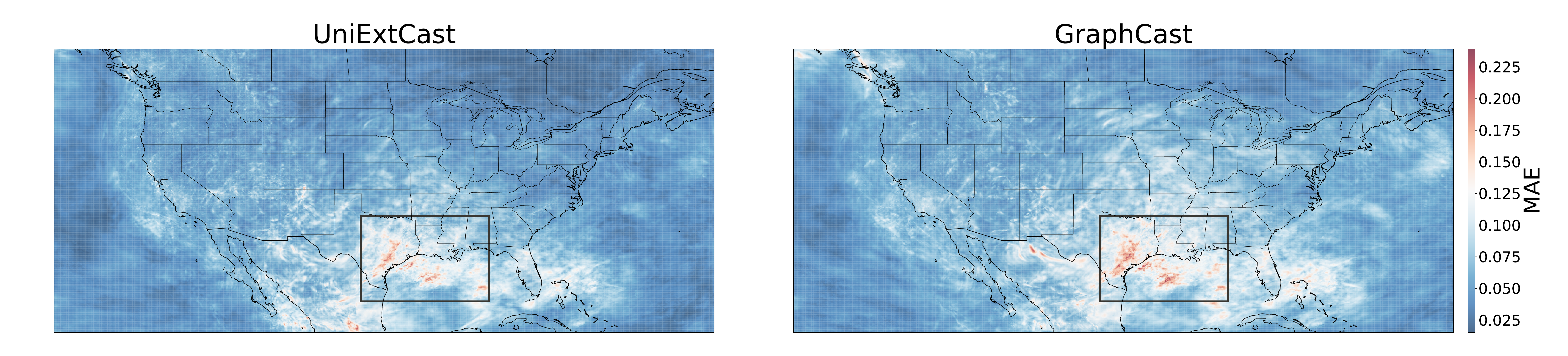}
    \vspace{-15pt}
    \caption{The MAE distribution of UniExtreme and GraphCast. The gray box marks the extreme weather region.}
    \vspace{-10pt}
    \label{figure:case_ext}
\end{figure*}

\vspace{-5pt}
\subsection{Visualized Analysis}
\label{section:case}

In this section, we conduct a comprehensive evaluation of UniExtreme's ability to address the two key challenges identified in \secref{section:intro}, as well as a particular case study of its efficacy in extreme weather forecasting.

\noindent\textbf{Band Aggregation Analysis.} 
To examine whether our AFM module effectively captures the "right-shift" property in frequency modulation, we analyze the learned band aggregation weights from \equref{equation:band_weight} as a proxy. 
Following the formulas in \secref{section:preliminary}, we compute the High-Frequency Area (HFA) metric for band weights similarly (detailed in~\appref{appendix:baa}). \figref{figure:case_hfa} presents the KDE plot of HFA measures for both normal and extreme weather regions across all timestamps (Year 2024) and variables. The results also reveal a "right-shift" pattern in the band aggregation weights, consistent with that in \figref{figure:hfa}, which confirms the module's capability in adaptive frequency modulation.

\noindent\textbf{Memory Fusion Analysis.} 
To assess how our EPA module integrates diverse event priors into regional weather states, we visualize the attention weights derived from the inter-type memory fusion in \equref{equation:epa_attn}. As a case study, we examine the weather state on Sep 29, 2024, 12:00 AM, which includes multiple flood and tornado events along the U.S. East Coast. The ground-truth type indices of regions are highlighted in red boxes in \figref{figure:case_attn}. The visualization demonstrates that the learned attention weights align well with the actual event types, indicating that our framework can leverage event prior knowledge to enhance extreme weather prediction.

\begin{figure}[h]
    \centering
    \vspace{-10pt}
    \subfloat[HFA distributions of band weights.]{\includegraphics[width=0.499\linewidth]{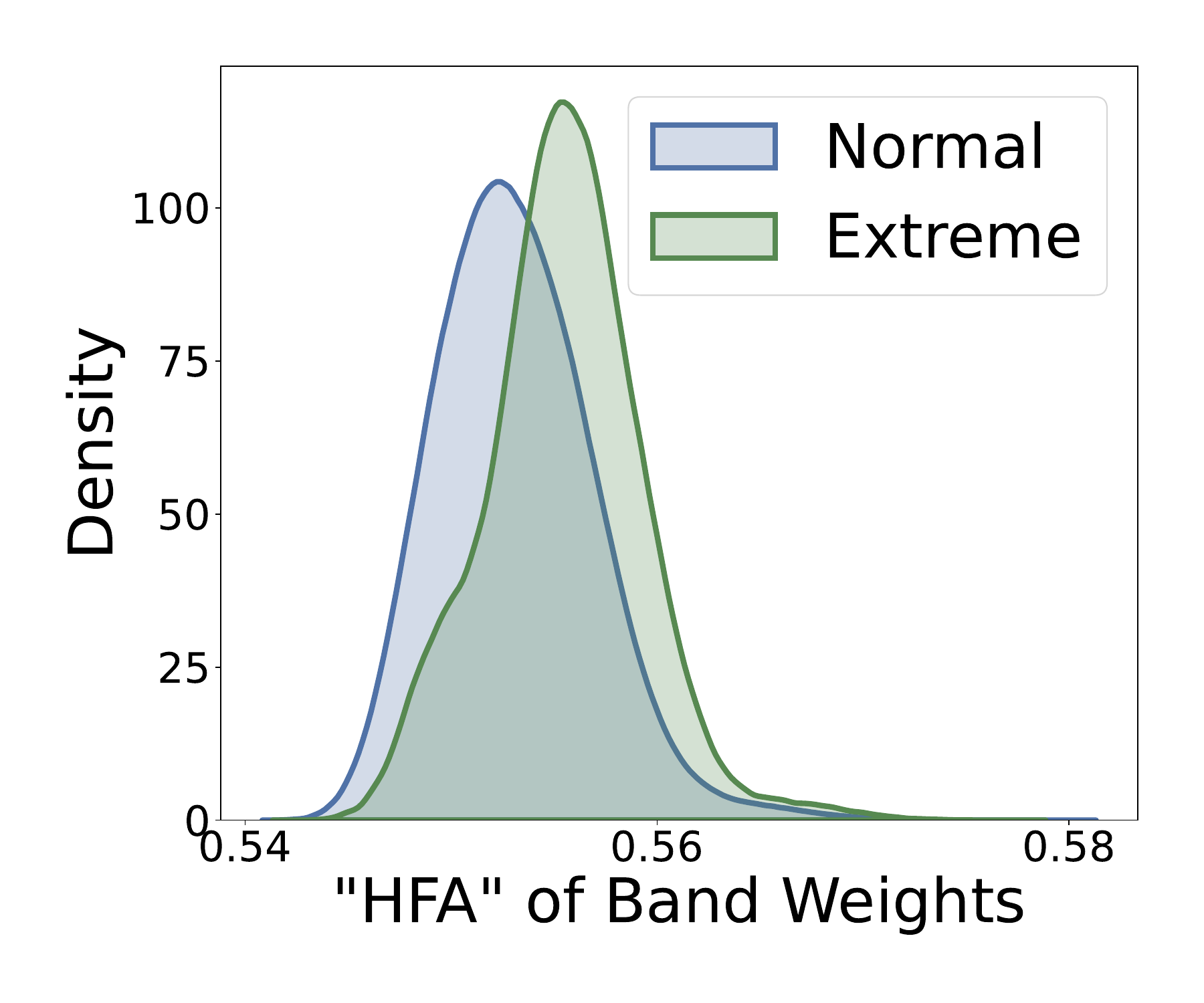}\label{figure:case_hfa}}
    \subfloat[Visualization of attention weights.]{\includegraphics[width=0.499\linewidth]{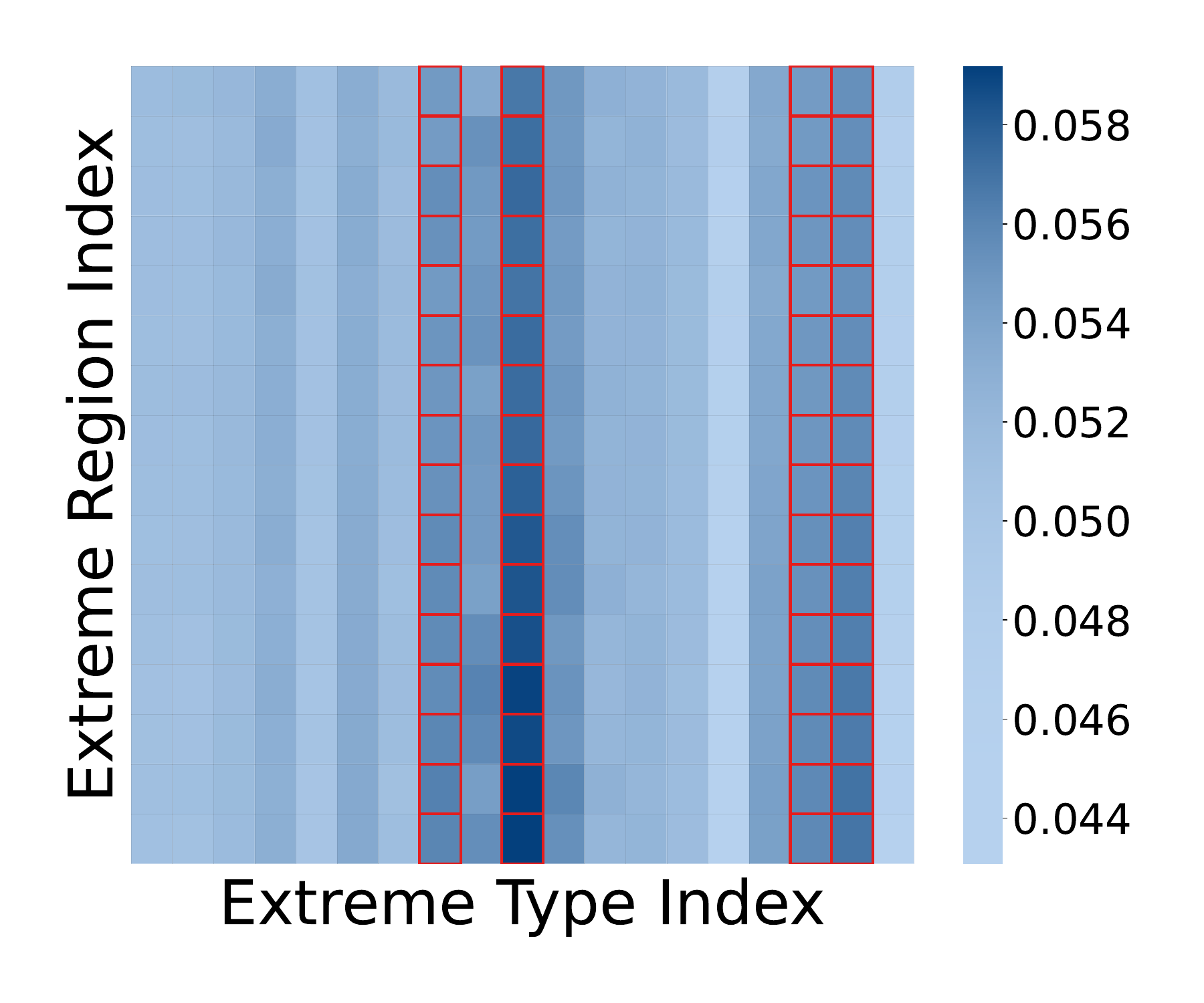}\label{figure:case_attn}}\\
    \vspace{-8pt}
    \caption{Empirical analysis of extreme weather.}
    \vspace{-8pt}
    \label{figure:case}
\end{figure}

\noindent\textbf{Extreme Case Analysis.}
To further demonstrate the superiority of our method in extreme weather scenarios, we compare the absolute errors of UniExtreme and GraphCast across the entire grid map, averaged over all variables. As illustrated in \figref{figure:case_ext}, we focus on 6:00 PM, January 8, 2024, a timestep marked by severe thunderstorms, tornadoes, and floods along the U.S. South Coast. The results show that UniExtreme achieves significantly lower prediction errors in extreme regions compared to GraphCast, the best-performing baseline in terms of MAE, underscoring its robustness in forecasting extreme events.
\vspace{-5pt}
\section{Related Work}


In this section, we review recent studies in general and extreme weather forecasting.

\noindent\textbf{Weather Forecasting.}
As an alternative to traditional umerical weather prediction (NWP) methods~\cite{bauer2015quiet,skamarock2019description}, recent years have witnessed growing interests in DL-based weather forecasting, which can be broadly categorized into deterministic and probabilistic methods~\cite{shi2025deep}.
Deterministic methods~\cite{nguyen2024scaling,lyu2025physics}, to which our work belongs, predict only future weather states, and many foundation models (FMs) have emerged in recent years. 
For instance, GraphCast~\cite{lam2023learning} introduces a graph-based framework, constructing a mesh grid over weather states and performing message passing via Graph Neural Networks (GNNs)~\cite{gcn}.
PanguWeather~\cite{bi2023accurate} employs a 3D Earth-specific Swin-Transformer~\cite{liu2021swin} architecture that explicitly incorporates atmospheric pressure levels into forecasts.
Building upon this, FengWu~\cite{chen2023fengwu} treats different atmospheric variables as separate modalities, processing their features independently.
FuXi~\cite{chen2023fuxi} proposes a cube embedding strategy for spatiotemporal fusion, and adopts 2D Swin-Transformer V2~\cite{liu2022swin} blocks as its backbone.
In contrast, probabilistic methods emphasize uncertainty quantification, often leveraging generative models~\cite{gao2023prediff,oskarsson2024probabilistic,price2025probabilistic}.
For example, GenCast~\cite{price2025probabilistic} builds a probabilistic FM that through diffusion models~\cite{ho2020denoising} conditioned on the previous two observations.
Despite these advances, accurate prediction of extreme weather remains an open challenge, with existing methods primarily optimized for general weather conditions.

\noindent\textbf{Extreme Weather Forecasting.}
Preliminary research efforts have been dedicated to improving forecasts for specific extreme weather phenomena, including extreme precipitation~\cite{zhang2023skilful}, heatwaves~\cite{lopez2023global}, thunderstorms~\cite{guastavino2022prediction}, tornadoes~\cite{lagerquist2020deep}, floods~\cite{sankaranarayanan2020flood}, etc.
For example, NowcastNet~\cite{zhang2023skilful} enhances extreme precipitation nowcasting by integrating an evolution network guided by atmospheric physics principles.
López-Gómez et al.~\cite{lopez2023global} address heatwave prediction by reweighting extreme cases using an exponential loss function. 
However, these approaches are narrowly tailored to specific event types and lack generalizability across diverse extreme weather patterns.
Recent studies~\cite{ni2023kunyu,zhong2024fuxi,xu2024extremecast,gao2025oneforecast} have sought to develop global forecasting models, even FMs, with enhancements in extreme prediction, primarily using the ERA5 dataset~\cite{hersbach2020era5}.
For instance, ExtremeCast~\cite{xu2024extremecast} mitigates the over-smoothing of extreme predictions by reformulating the loss function using insights from Extreme Value Theory (EVT). 
Meanwhile, OneForecast~\cite{gao2025oneforecast} argues that extreme events are linked to high-frequency signals and proposes a high-pass GNN to capture such features. 
Nevertheless, current methods for assessing extreme weather prediction expertise often rely on percentile thresholds or limited extreme types, lacking both reliable validation across diverse real-world cases and sufficient supervision signals from varied extreme events, which hinders their ability to capture the complex meteorological patterns and inter-categorical distinctions of actual extreme weather.
\vspace{-5pt}
\section{Conclusion}

In this work, we present UniExtreme, a foundation model designed for universal extreme weather forecasting. 
By addressing the spectral and spatial complexities of extreme events through Adaptive Frequency Modulation (AFM) and Event Prior Augmentation (EPA), our framework achieves significant improvements over existing methods. Specifically, AFM adaptively distinguishes between normal and extreme weather regimes using learnable spectral filters and a frequency aggregation network, while EPA incorporates region-wise extreme event memories through a hierarchical memory fusion block, to handle diverse and coexisting hazards. 
Extensive evaluations confirm UniExtreme's effectiveness in forecasting both general and extreme weather conditions, as well as its versatility across multiple types of extreme events. 
This work advances the field of data-driven weather forecasting by bridging the gap between general and extreme prediction, paving the way for more reliable and comprehensive weather modeling.

\bibliographystyle{ACM-Reference-Format}
\bibliography{sample-base}

\appendix
\begin{table*}[h]
\centering
\caption{Summary of Notations.}
\begin{tabular}{ll}
\toprule
\textbf{Notation} & \textbf{Description} \\
\midrule
$\mathbf{X}^t$ & Weather state at time $t$ \\
$\mathcal{G}$ & Latitude-longitude grid region \\
$f_\theta$ & Forecasting model with parameters $\theta$ \\
$\mathcal{E}^t$ & Set of extreme weather event records at time $t$ \\
$e^t_j$ & Bounding box coordinates of the $j$-th extreme event at time $t$ \\
\midrule
$\mathbf{F}$ & 2D Discrete Fourier Transform of $\mathbf{X}$ \\
$\mathbf{R}, \mathbf{I}$ & Real and imaginary parts of Fourier coefficients \\
$\lambda^{(h)}_j$ & $j$-th frequency along the height dimension \\
$\lambda^{(w)}_j$ & $j$-th frequency along the width dimension \\
$\lambda_j$ & Radial frequency of $j$-th frequency component \\
$\lambda_j^{(s)}$ & $j$-th smallest radial frequency \\
$\dot\lambda_j^{(s)}$ & Normalization of the $j$-th smallest radial frequency \\
$\eta_{K,c}$ & Energy ratio for channel $c$ up to $K$-th frequency \\
$S_{\text{high}}(c)$ & High frequency area for channel $c$ \\
\midrule
$\mathbf{X}^t_{r}$ & Partitioned regional data at position $r$ \\
$\tilde{\mathbf{X}}^t_{r}$ & Regional augmented weather data transformed by the EPA module \\
$\ddot{\mathbf{X}}^t_{r}$ & Regional modulated weather data transformed by the AFM module \\
$a_h, a_w$ & Height and width of each partitioned region \\
$H^\prime, W^\prime$ & Number of regions along height and width ($H^\prime=H/a_h$, $W^\prime=W/a_w$) \\
$\mathcal{B}_{r;n}(x)$ & $n$-th Beta filter for region $r$ \\
$\tilde{\lambda}_{r;n}, \kappa_{r;n}$ & Mode and spread parameters for Beta filter $n$ and region $r$ \\
$\alpha_{r;n}, \beta_{r;n}$ & Original distribution parameters for Beta filter $n$ at region $r$ \\
$\dot{\mathbf{F}}^t_{r}$ & Filtered spectra for region $r$ \\
$\mathcal{W}^t_{r}$ & Learnable band aggregation weight for region $r$ \\
$\ddot{\mathbf{F}}^t_{r}$ & Aggregated spectrum for region $r$ \\
$\mathcal{M}, \dot{\mathcal{M}}$ & Original and clustered event memory pools \\
$M^\prime$ & Number of event types (including normal) \\
$U$ & Fixed capacity of each memory type \\
$\mathbf{P}^t_{r}$ & Hybrid-type memory for region $r$ \\
$\mathbf{Z}^t_j$ & Transformer hidden states at layer $j$ \\
$p_h, p_w$ & Patch size for patch embedding \\
$D$ & Embedding dimension \\
$\mathbf{W},\mathbf{b}$ & Parameters of liner layers \\
\bottomrule
\end{tabular}
\label{table:notation}
\end{table*}

\section{Notation}
\label{appendix:notation}
The key notations used in this paper are summarized in~\tabref{table:notation}.

\section{Empirical Analysis}
\label{appendix:empirical}
This section presents details of the empirical study from both frequency and spatial perspectives.

\begin{figure}[ht]
    \centering
    \subfloat[HFA of 2019.]{\includegraphics[width=0.499\linewidth]{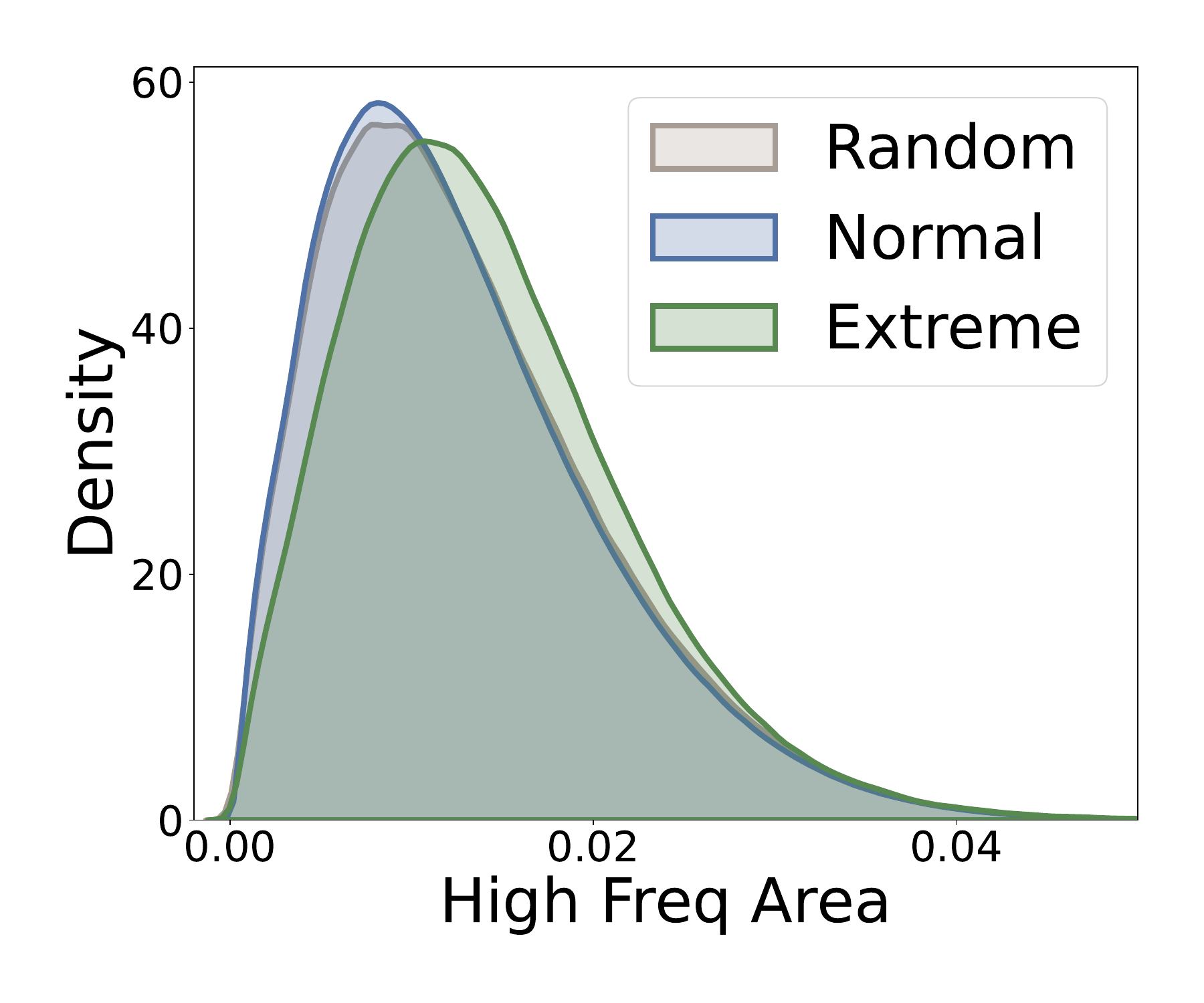}}
    \subfloat[HFA of 2020.]{\includegraphics[width=0.499\linewidth]{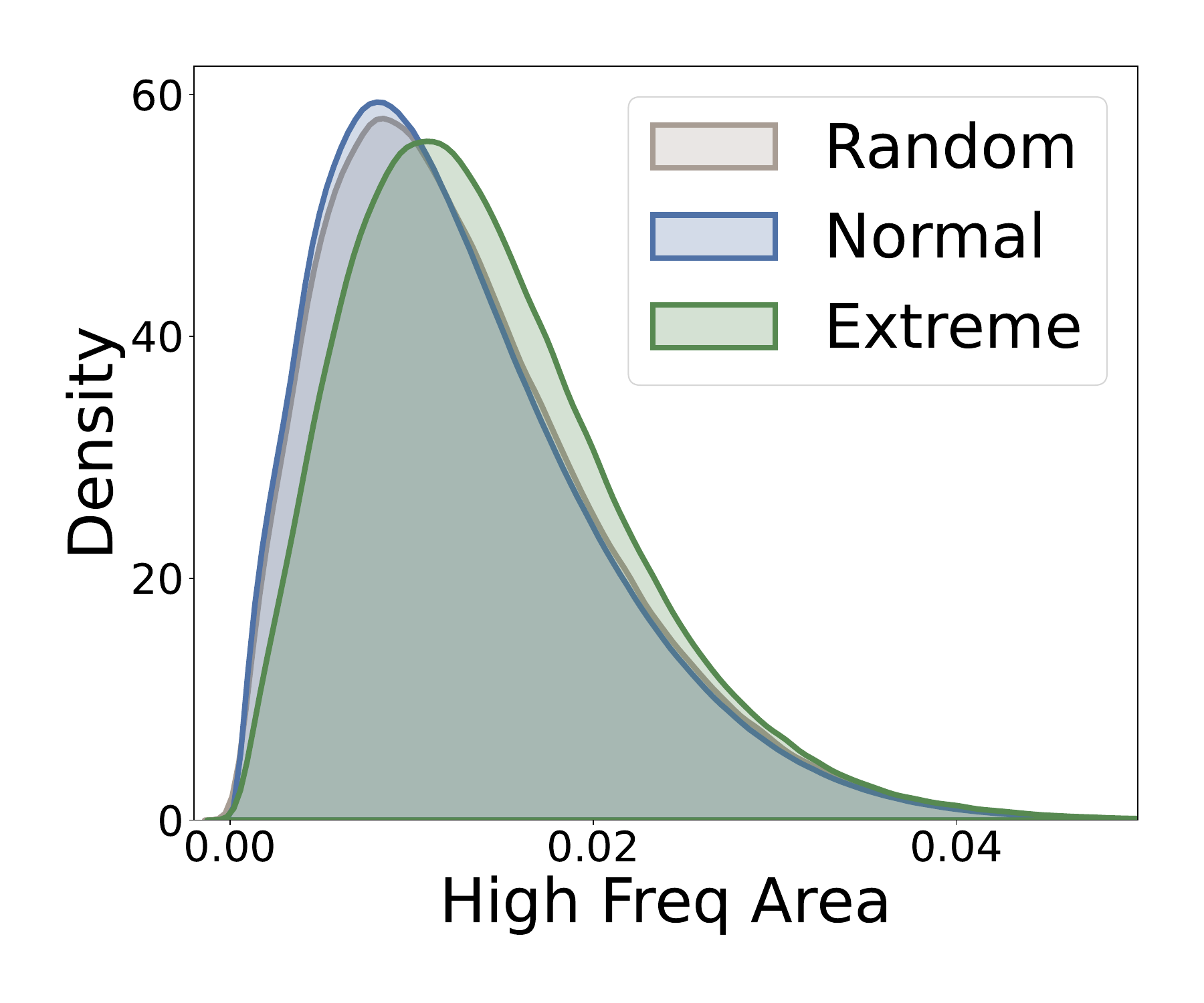}}\\
    \subfloat[HFA of 2021.]{\includegraphics[width=0.499\linewidth]{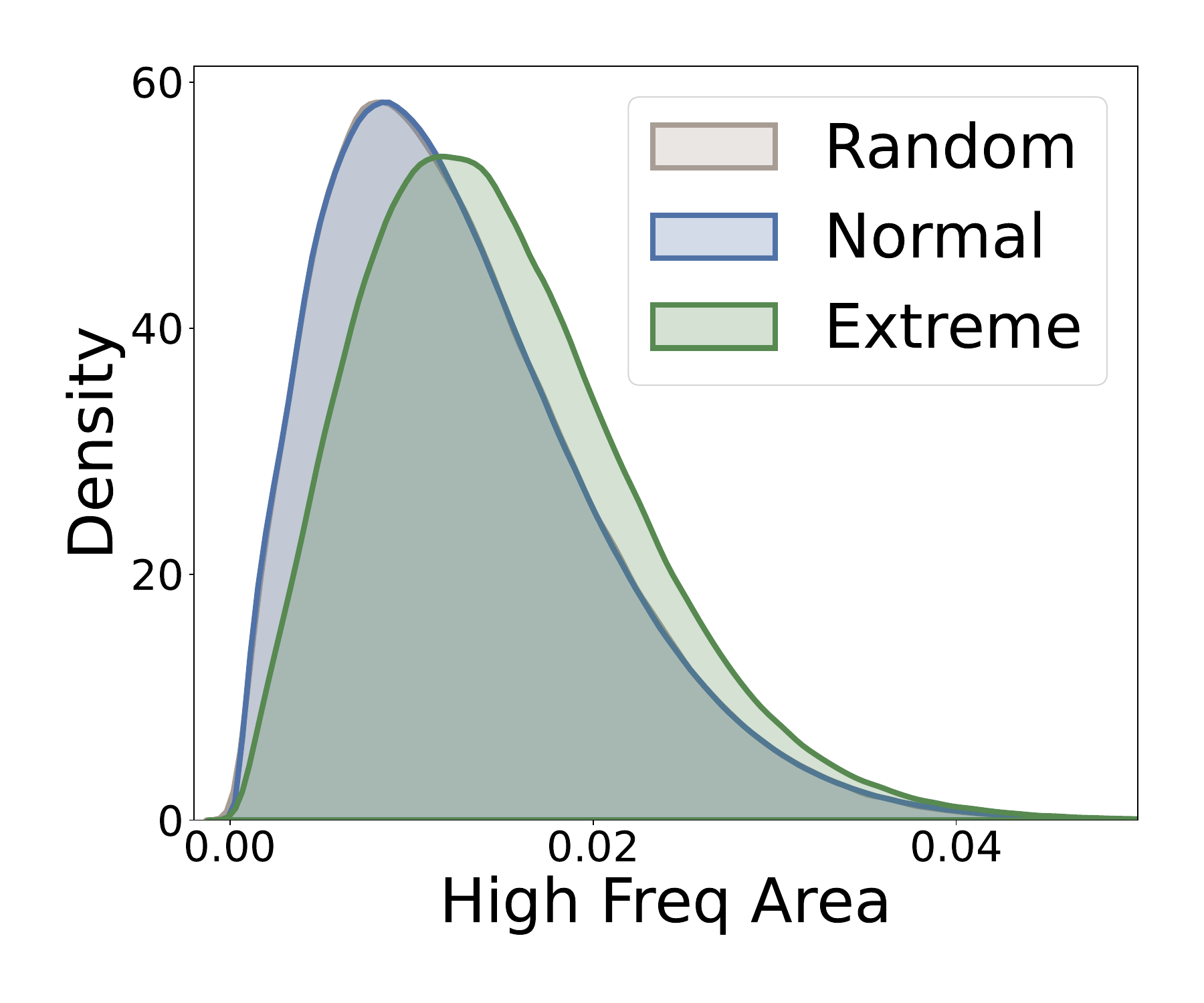}}
    \subfloat[HFA of 2022.]{\includegraphics[width=0.499\linewidth]{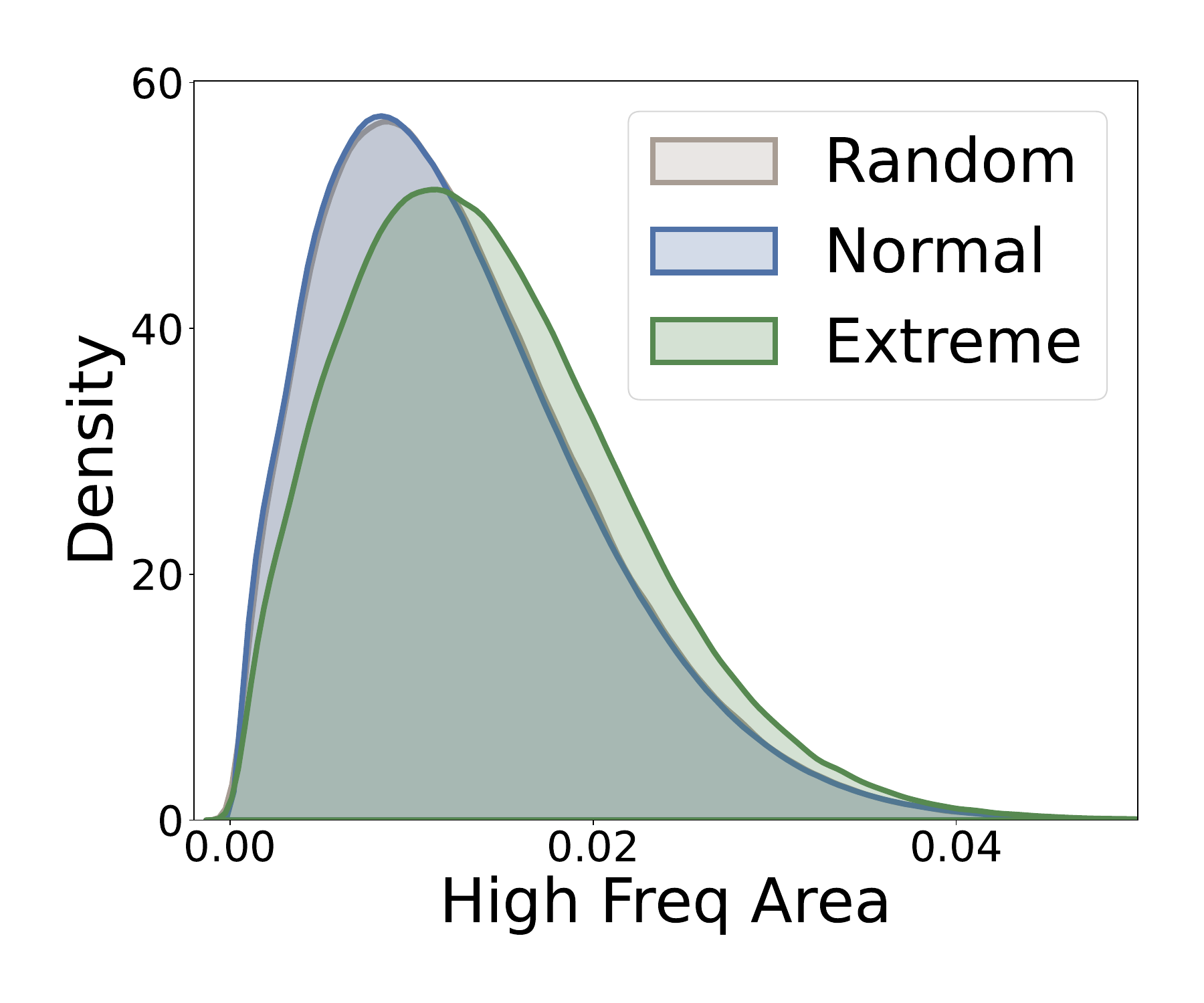}}\\
    \subfloat[HFA of 2023.]{\includegraphics[width=0.499\linewidth]{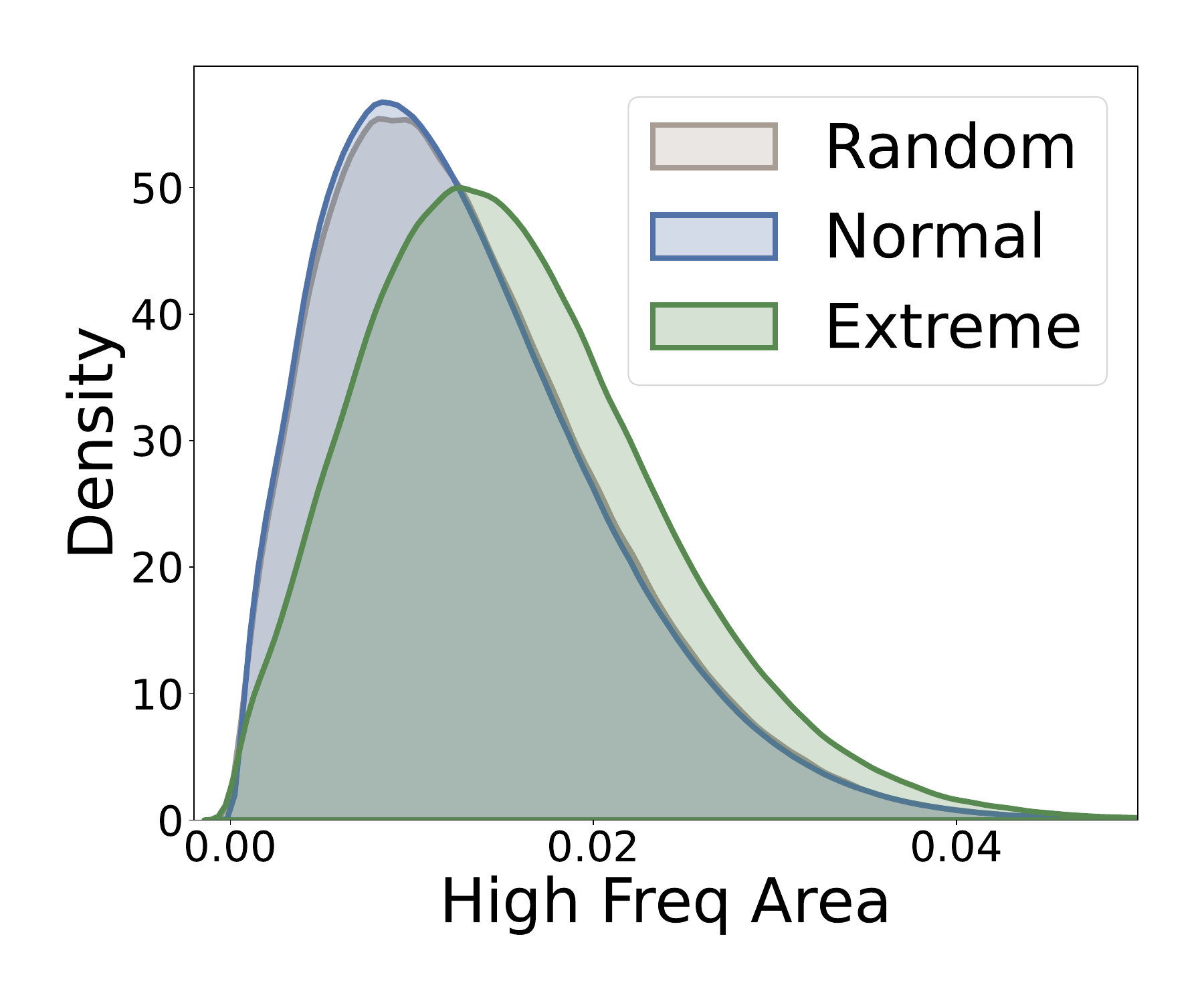}}
    \subfloat[HFA of 2024.]{\includegraphics[width=0.499\linewidth]{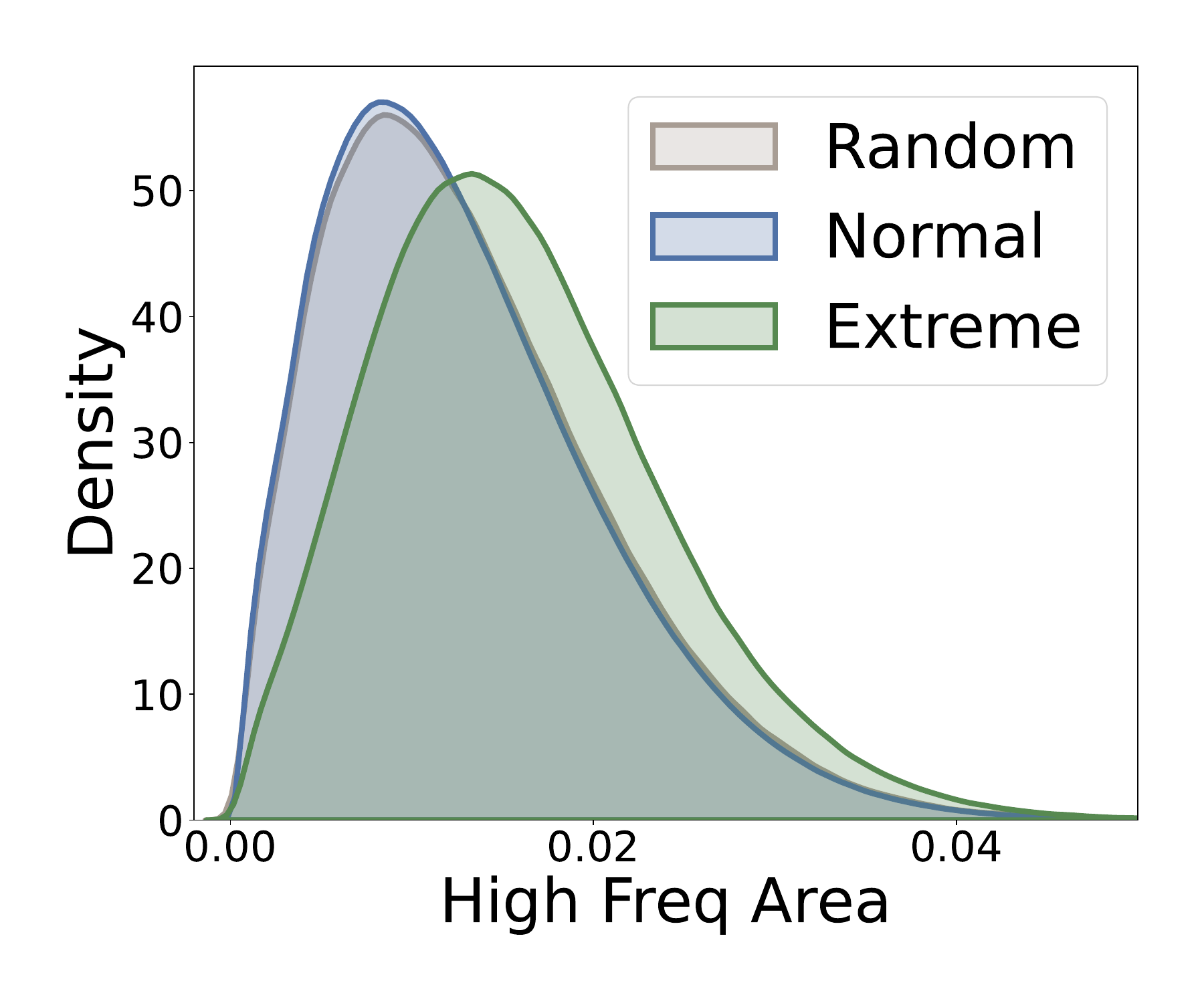}}\\
    \caption{KDE plot of high frequency area (HFA) from 2019-2024.}
    \label{figure:empirical_freq_all}
\end{figure}

\subsection{Frequency-Domain Analysis}
\label{appendix:empirical_freq}

While prior work has suggested a correlation between extreme weather and high-frequency disturbances~\cite{gao2025oneforecast}, systematic empirical validation remains limited. Building on~\cite{tang2022rethinking}, we quantitatively assess the spectral properties of normal and extreme weather using \textit{High Frequency Area (HFA)}, which are formulated as follows:

\begin{myDef}
\textbf{Energy Ratio.}
The $K$-th energy ratio for channel $c$ is defined as the cumulative energy proportion of the first $K$ sorted raidal frequencies:  
\begin{align*}
\eta_{K,c} = \frac{\sum_{j=1}^K E^{(s)}_{j,c}}{\sum_{j=1}^{HW} E^{(s)}_{j,c}},
\end{align*}
where $E^{(s)}_{j,c}$ denotes the energy for each channel $c$, corresponding to the $j$-th sorted requency radius $\lambda^{(s)}_j$.
Intuitively, a higher $ \eta_K(c) $ indicates greater energy concentration in frequency components below $\lambda_K$ for the channel $c$.
\end{myDef}

\begin{myDef}
\textbf{High Frequency Area.}
High frequency area, measuring the frequency concentration for channel $c$, is defined as 
\begin{align*}
S_{\text{high}}(c) = \int_{\dot\lambda_0}^{\dot\lambda_{HW-1}} \left(1 - f_c(t)\right) dt \in[0,1],
\end{align*}
where $f_c(t)$ is a piecewise function of energy ratio curve defined as:
\begin{align*}
f_c(t) = \eta_K(c), \quad t \in \left[\dot\lambda^{(s)}_K, \dot\lambda^{(s)}_{K+1}\right), \quad K=0,...,HW-2.
\end{align*} 
A larger $S_{\text{high}}(c)$ implies a higher concentration of energy in high-frequency components for the channel $c$. 
\end{myDef}

To enable fine-grained localized analysis, we partition the weather grid $\mathbf{X}^t \in \mathbb{R}^{H \times W \times C}$ at time $t$ into uniform regions $\{\mathbf{X}^t_r\}_{r=1}^{H^\prime W^\prime-1} \subset \mathbb{R}^{a_h \times a_w \times C}$ (see \secref{section:partition}), labeling each as normal or extreme based on extreme event records $\mathcal{E}$. For each region, we compute the 2D Fourier transform across all timesteps and atmospheric variables, then derive HFA statistics.
To mitigate bias from extreme region scarcity, we compare three region types: \textit{normal}, \textit{extreme} and \textit{random}, where the random regions are the ones randomly sampled from the partitioned regions, matching to extreme regions in quantity.
As complements to \figref{figure:hfa}, \figref{figure:empirical_freq_all} visualizes the kernel density estimate (KDE) distribution of HFA $S_{\text{high}}$ averaging over all variables and across our complete dataset spanning 2019-2024. 
We can observe a \textit{Right-Shift Phenomenon}: Extreme regions exhibit significantly higher HFA values compared to both normal and randomly sampled regions, confirming their spectral concentration in high frequencies.
Such spectral disparities persist consistently throughout all years in our study period, and verify the insights of prior work~\cite{gao2025oneforecast,tang2022rethinking}. 
In addition, this finding underscores the necessity and challenge of adaptively discriminating between extreme and normal weather regimes based on their spectral signatures.

\begin{figure*}[ht]
    \centering
    \subfloat[9:00 PM, January 23, 2024.]{\includegraphics[width=0.4\linewidth]{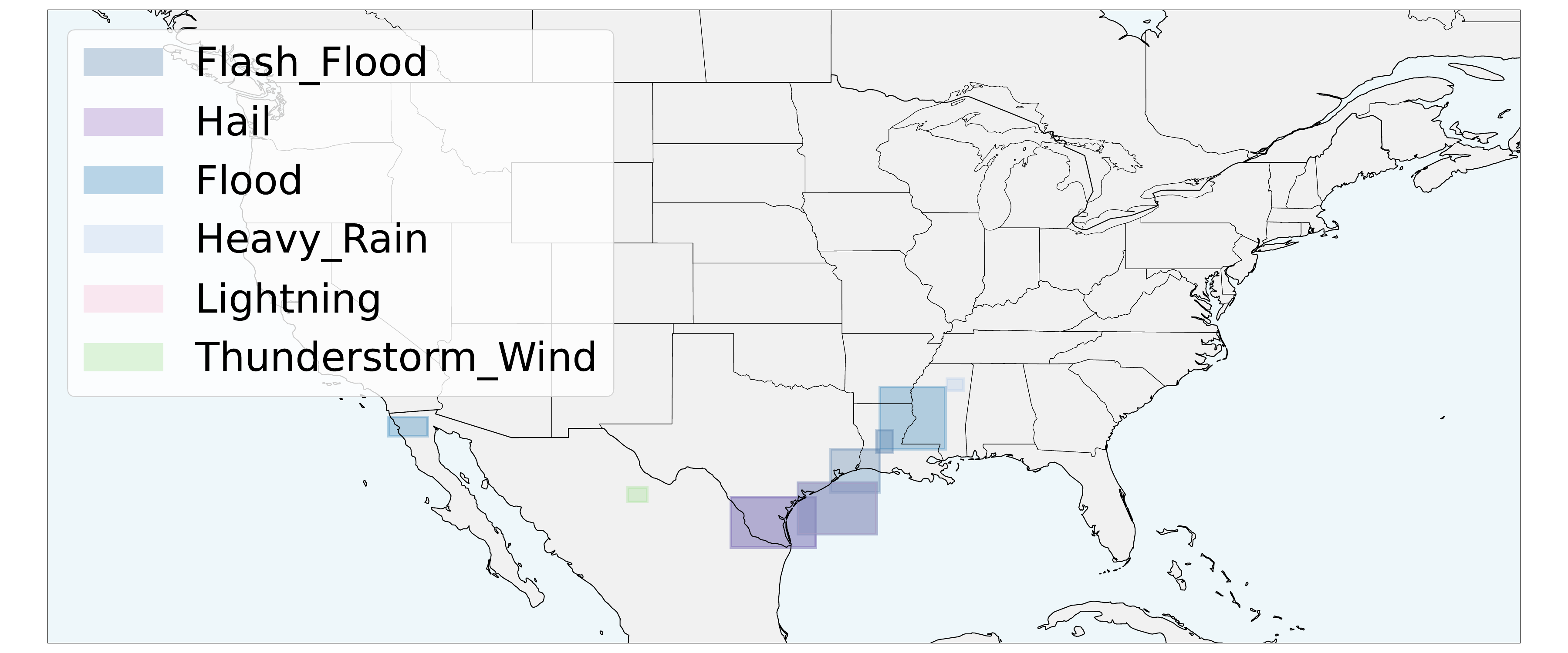}}
    \subfloat[9:00 AM, February 20, 2024.]{\includegraphics[width=0.4\linewidth]{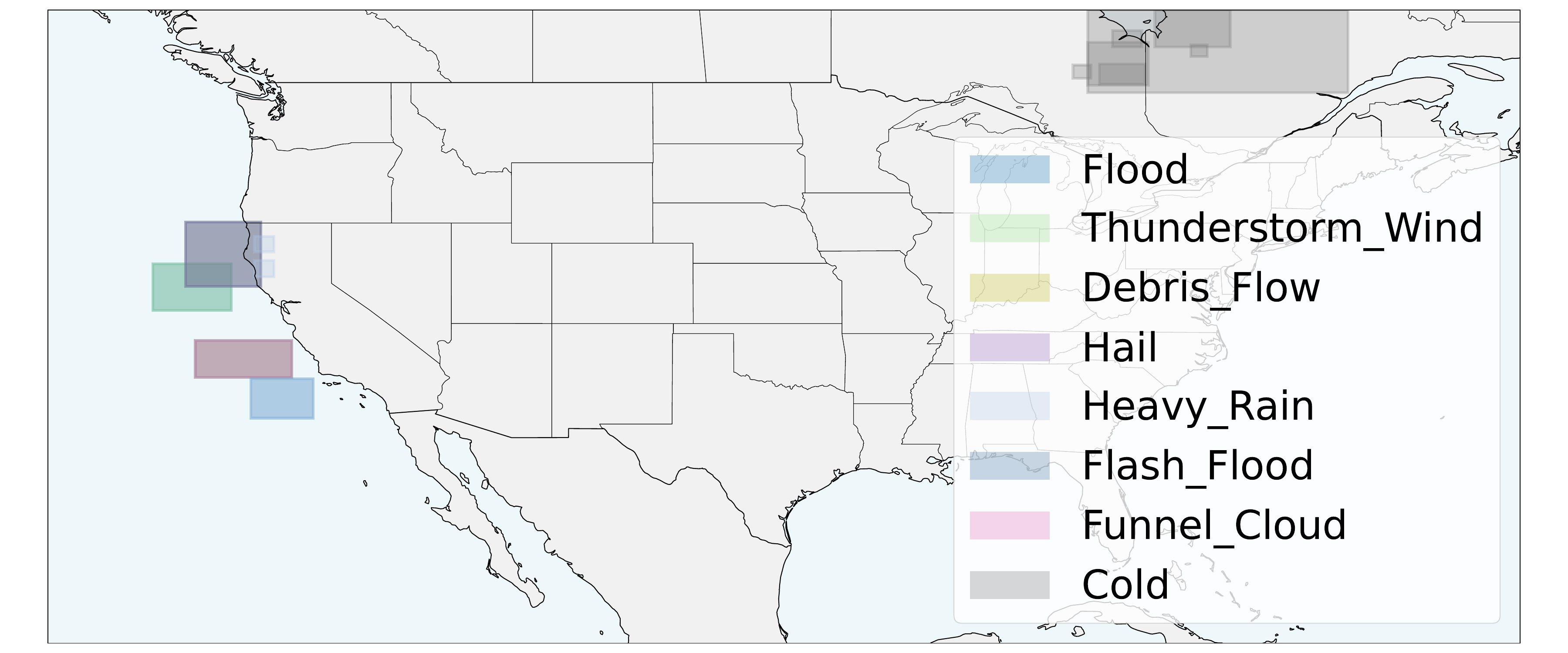}}\\
    \subfloat[3:00 PM, March 14, 2024.]{\includegraphics[width=0.4\linewidth]{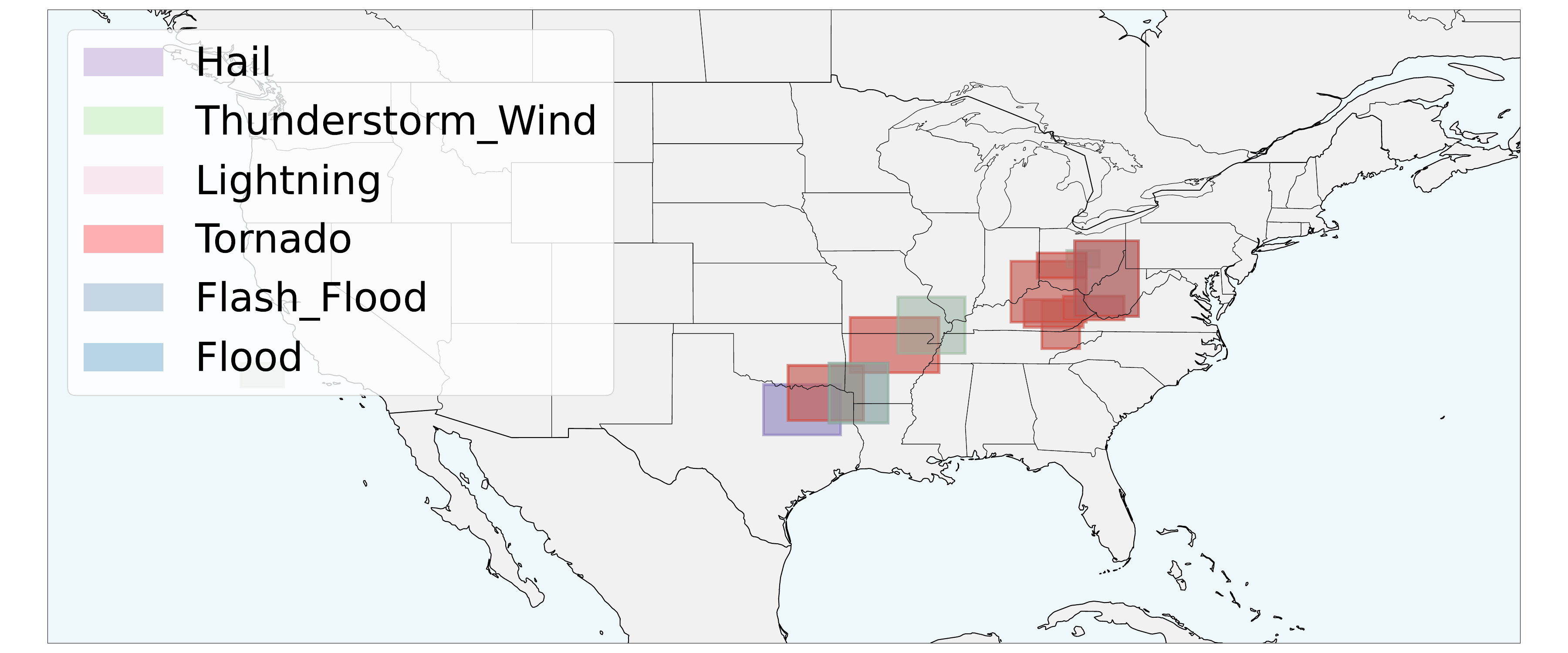}}
    \subfloat[7:00 PM, April 8, 2024.]{\includegraphics[width=0.4\linewidth]{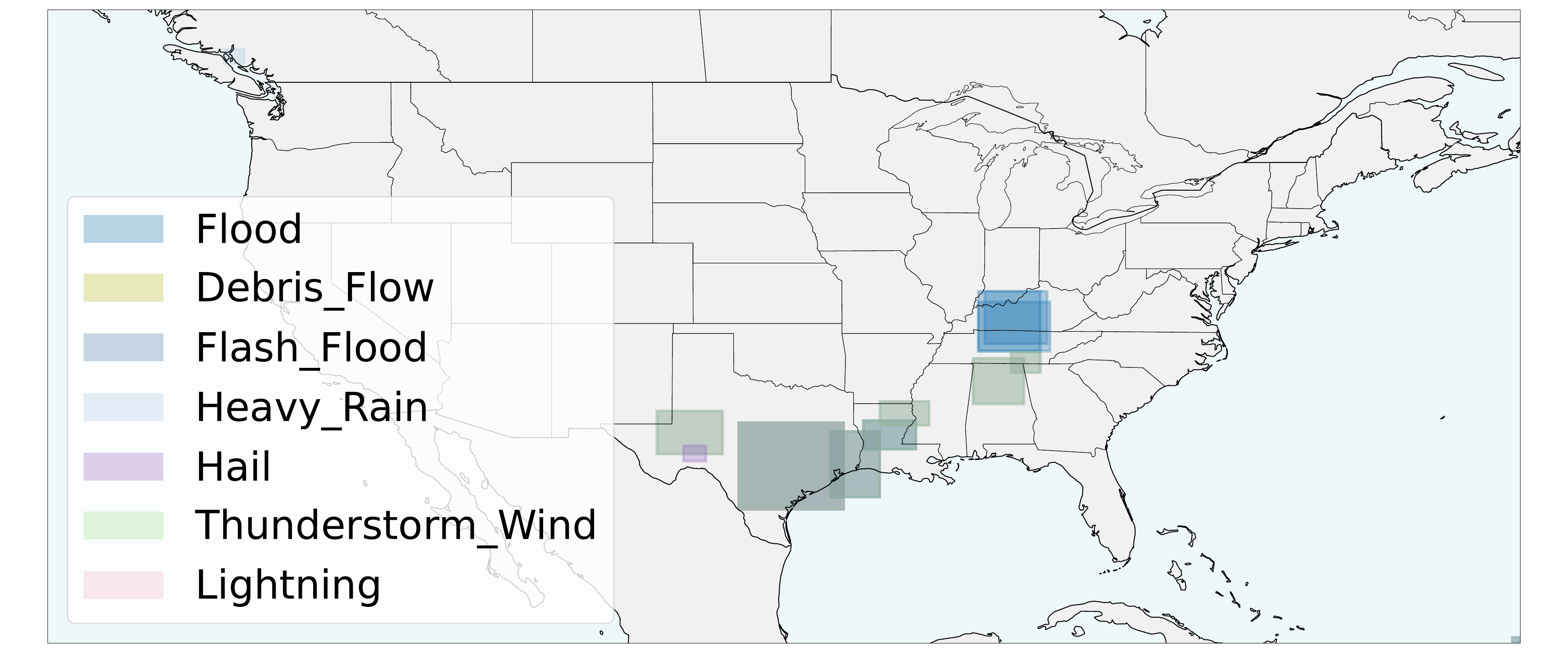}}\\
    \subfloat[8:00 AM, May 3, 2024.]
    {\includegraphics[width=0.4\linewidth]{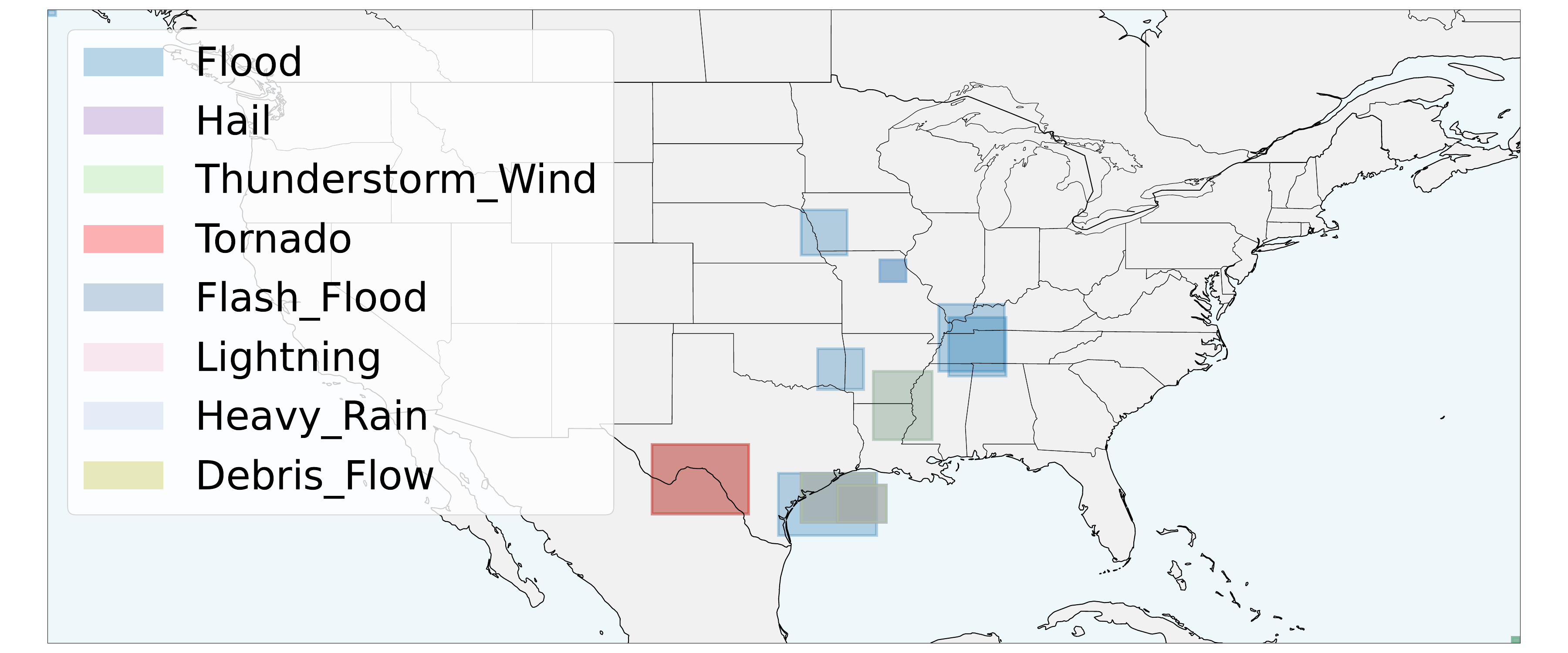}}
    \subfloat[12:00 PM, June 4, 2024.]
    {\includegraphics[width=0.4\linewidth]{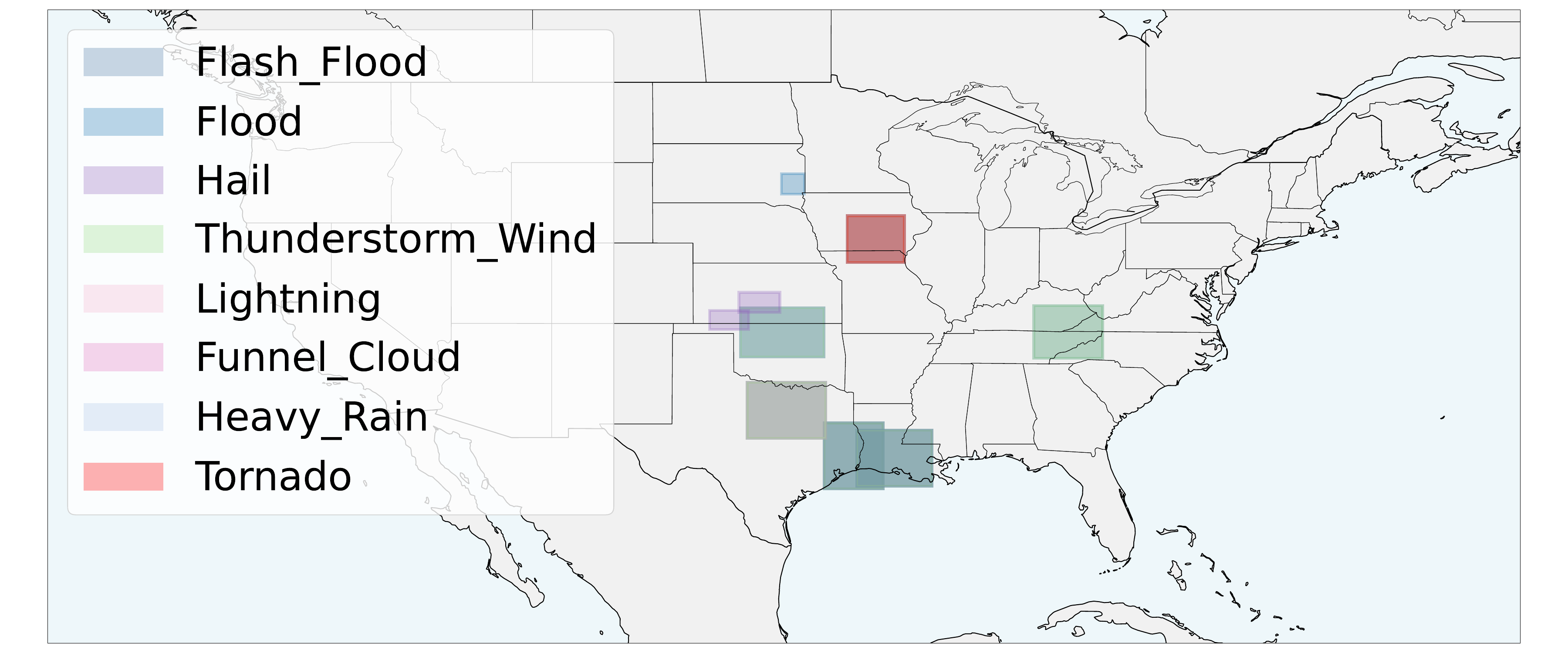}}\\
    \caption{Bounding boxes of extreme events in from January to June in 2024.}
    \label{figure:empirical_space_all_1}
\end{figure*}

\begin{figure*}[ht]
    \centering
    \subfloat[12:00 AM, July 4, 2024.]{\includegraphics[width=0.4\linewidth]{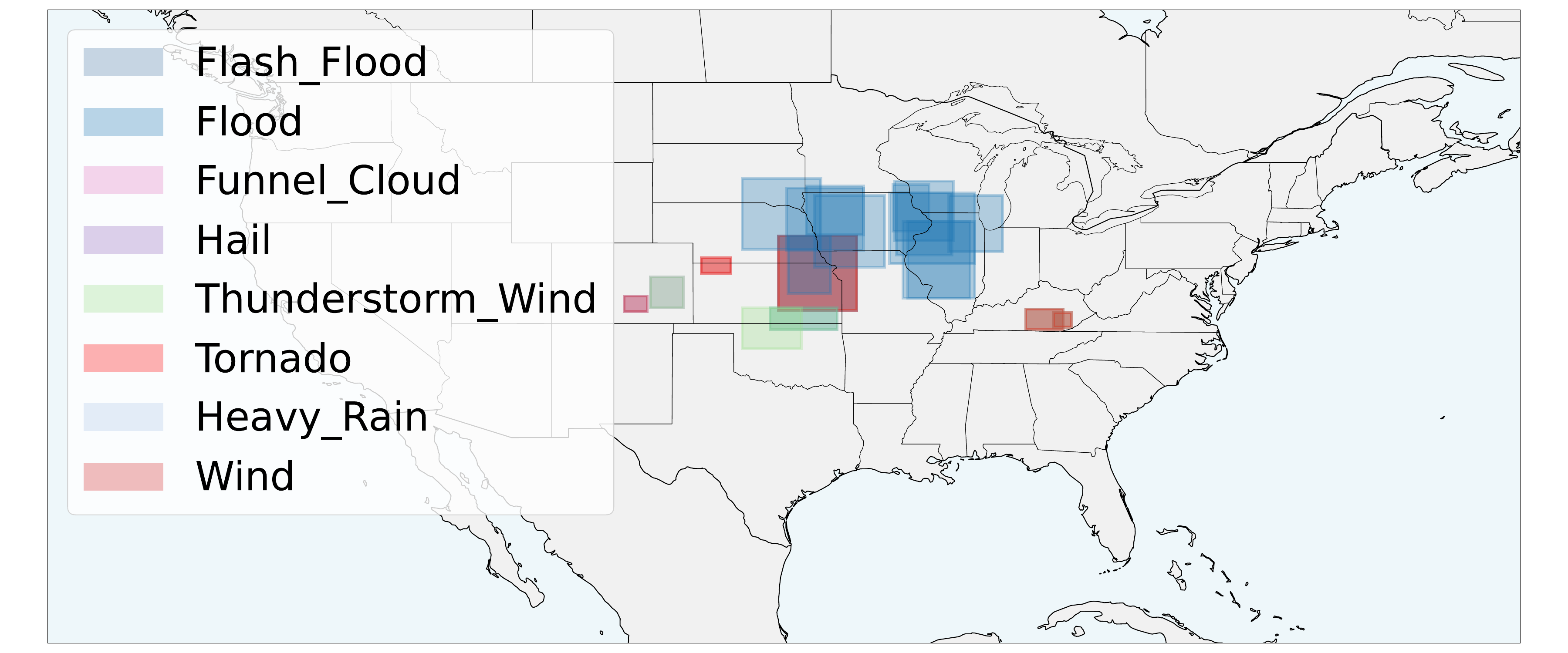}}
    \subfloat[6:00 AM, August 17, 2024.]{\includegraphics[width=0.4\linewidth]{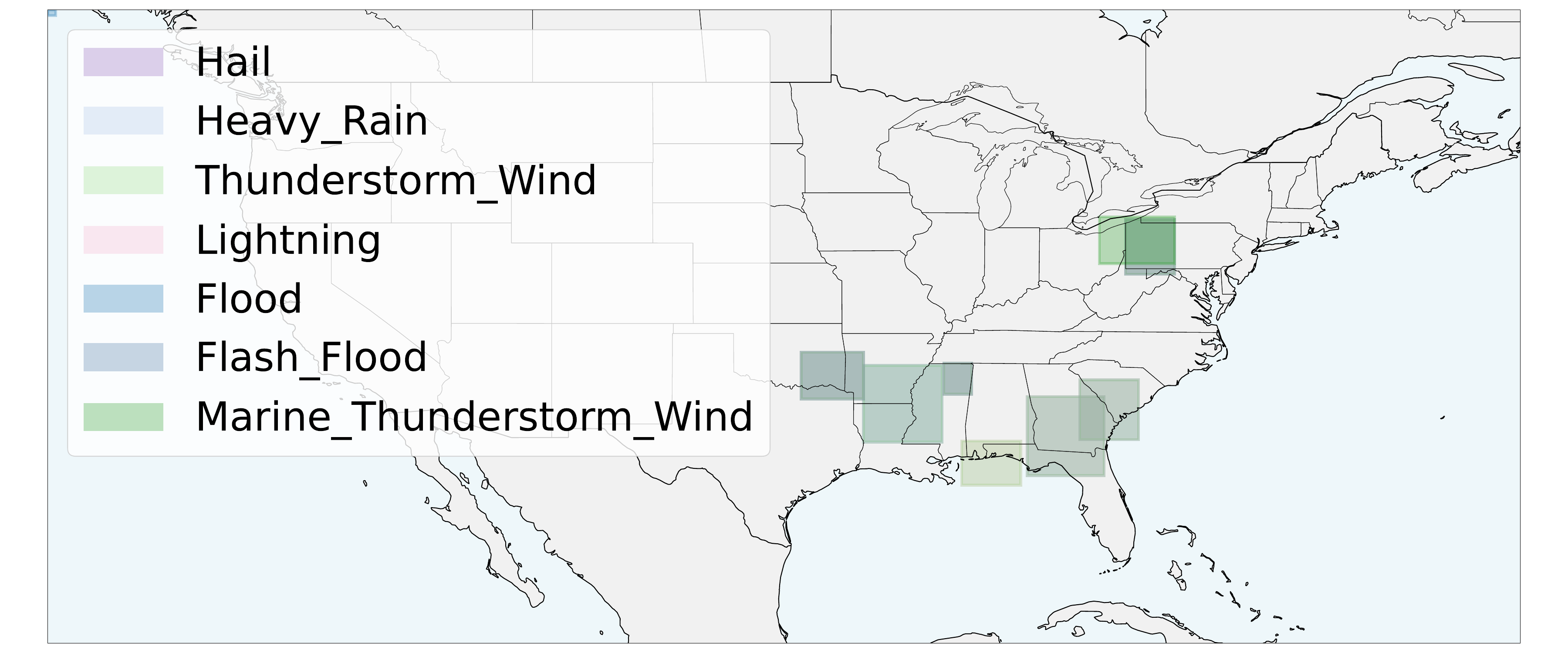}}\\
    \subfloat[2:00 PM, September 14, 2024.]{\includegraphics[width=0.4\linewidth]{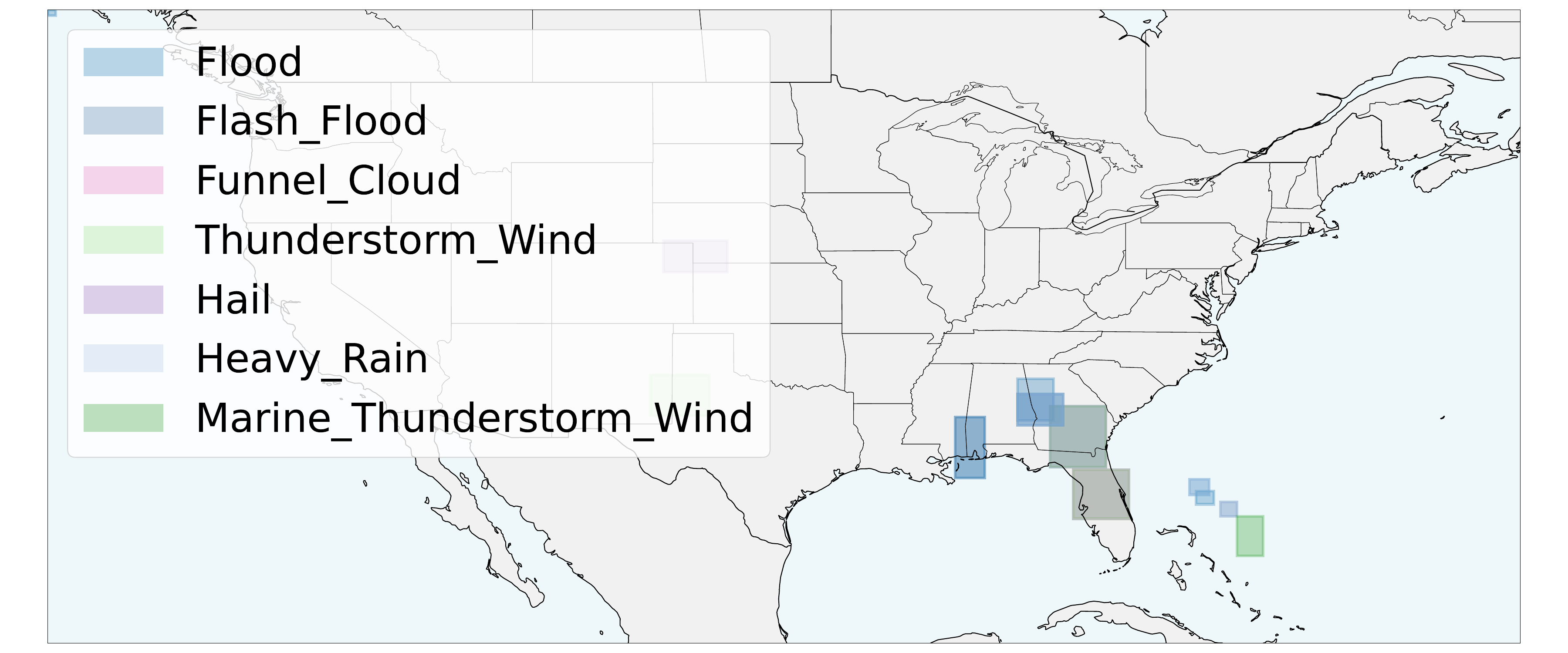}}
    \subfloat[8:00 PM, October 30, 2024.]{\includegraphics[width=0.4\linewidth]{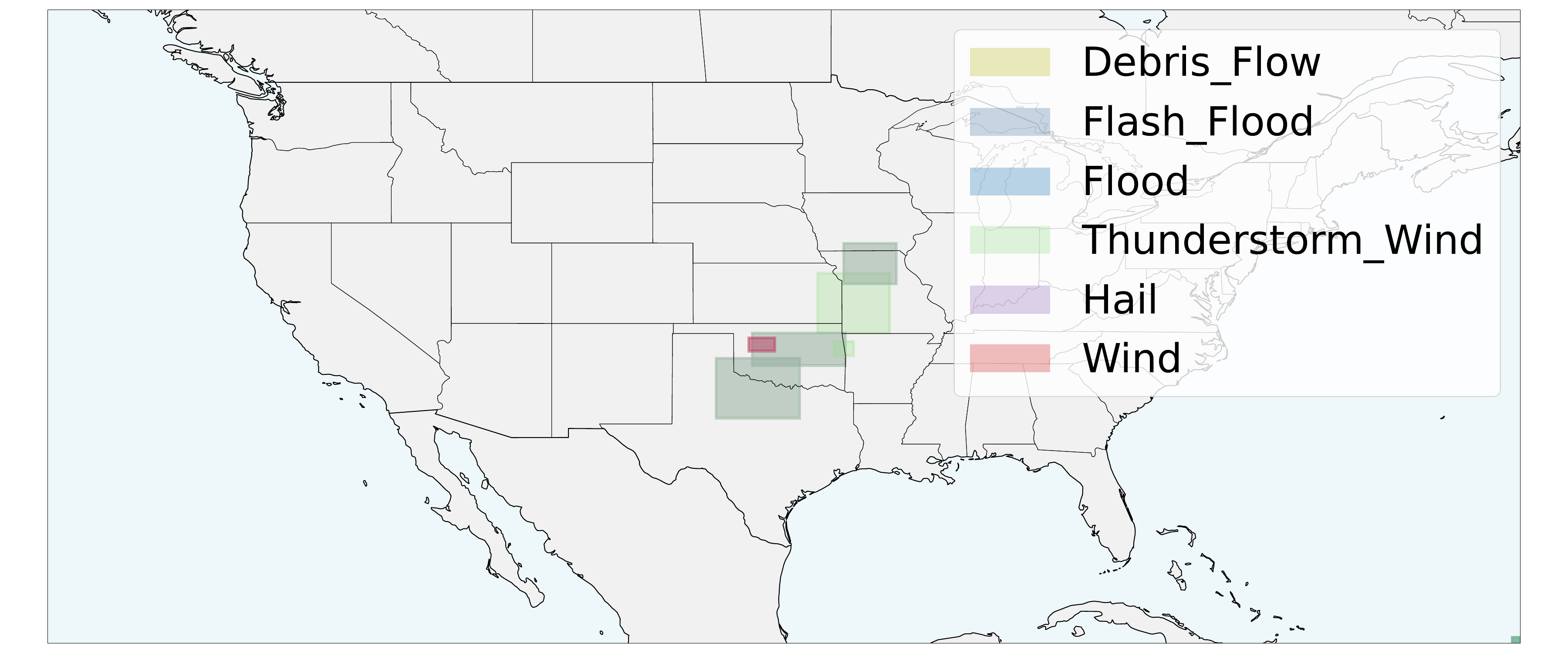}}\\
    \subfloat[10:00 PM, November 4, 2024.]
    {\includegraphics[width=0.4\linewidth]{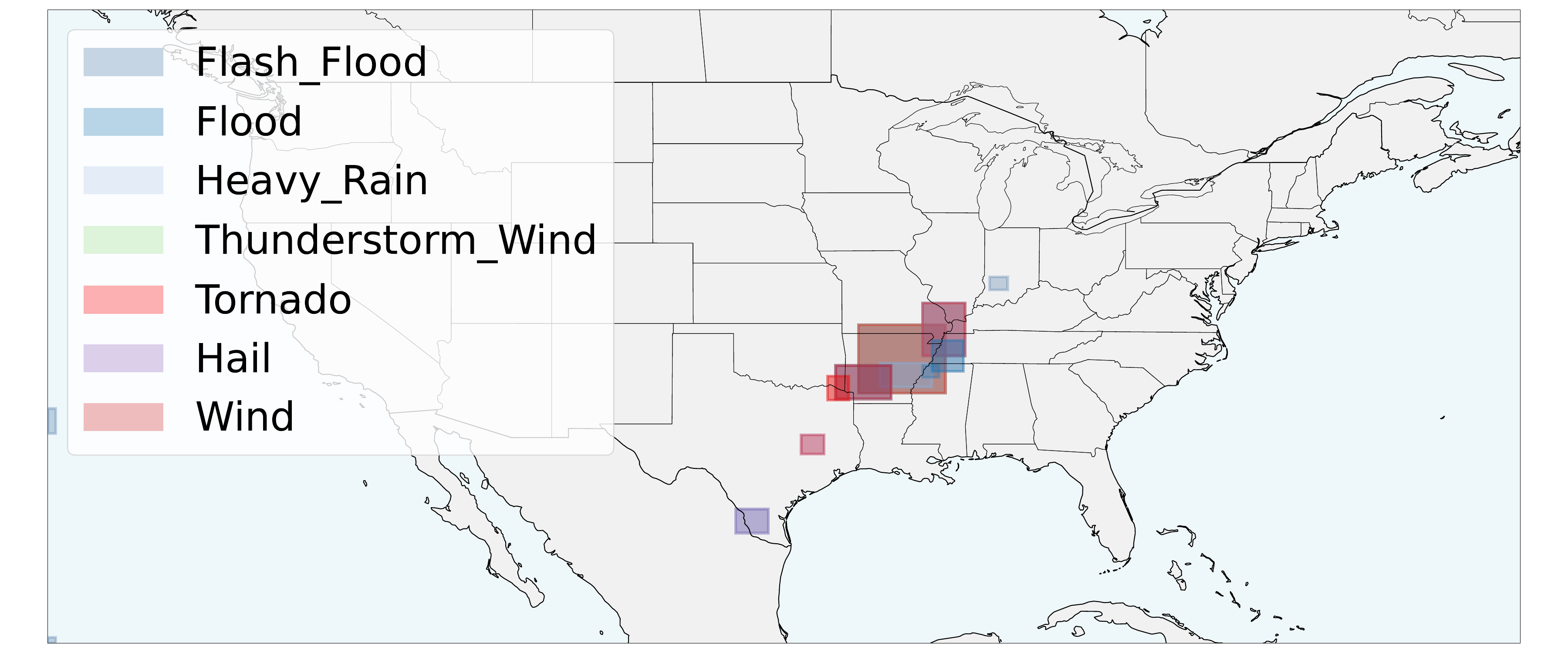}}
    \subfloat[11:00 AM, December 28, 2024.]
    {\includegraphics[width=0.4\linewidth]{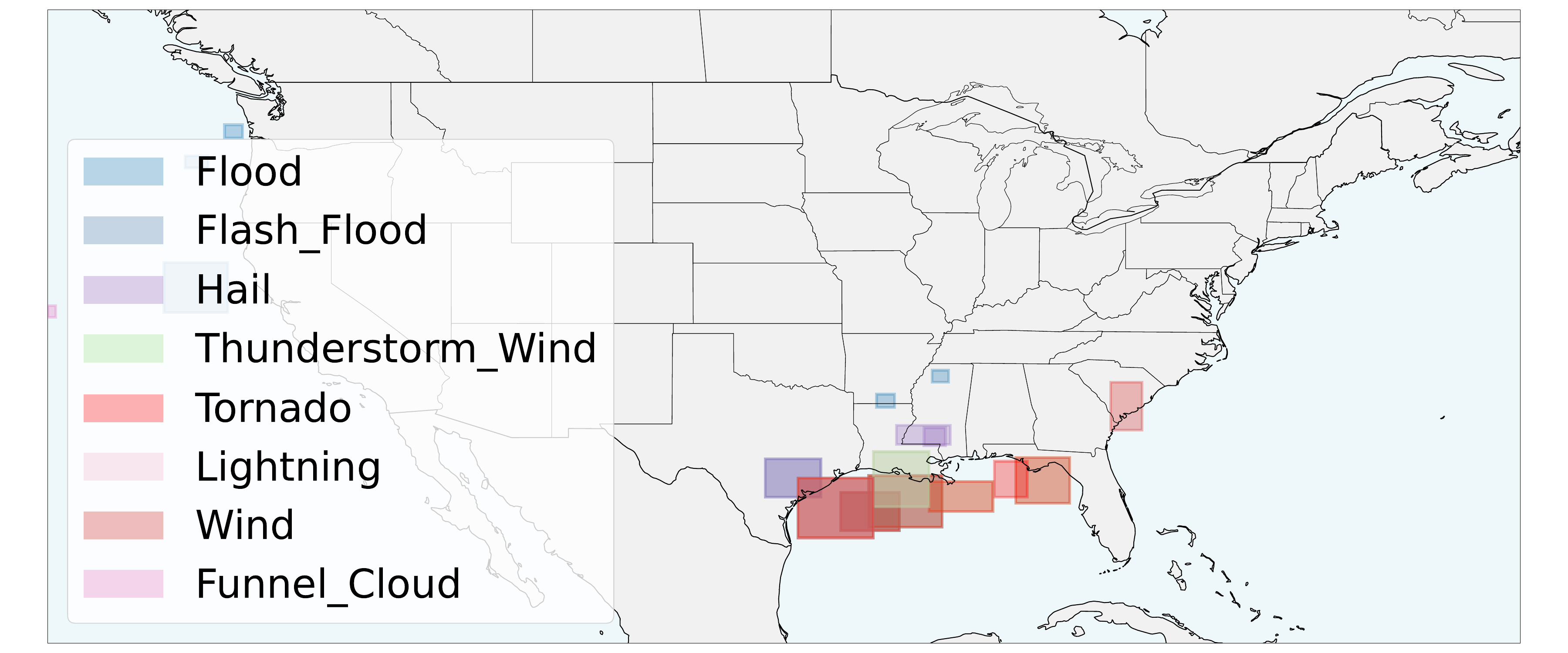}}\\
    \caption{Bounding boxes of extreme events in from July to December in 2024.}
    \label{figure:empirical_space_all_2}
\end{figure*}

\subsection{Spatial-Domain Analysis}
\label{appendix:empirical_space}

In addition to \figref{figure:bbox}, \figref{figure:empirical_space_all_1} and \figref{figure:empirical_space_all_2} display the spatial distributions of extreme events in different periods of 2024. The visualization reveals both the widespread occurrence of distributed disasters across geographical regions, and the prevalence of compound disaster events, which further consolidate our insights of event-aware extreme modeling.

\begin{table*}[h]
\centering
\caption{Summary of the 69 atmospheric variables.}
\begin{tabular}{cccc}
\toprule
\textbf{Variable} & \textbf{Definition} & \textbf{Unit} & \textbf{Range} \\
\midrule
MSL & Mean Sea Level Pressure & Pa & - \\T2M & Temperature 2 m above ground & K & - \\
U10 & U-component Wind Speed 10 m above ground & m/s & - \\
V10 & V-component Wind Speed 10 m above ground & m/s & - \\
\midrule
Z & Geopotential Height & gpm & \multirow{5}{*}{\makecell[l]{At 50, 100, 150, 200, 250, 300, \\ 400, 500, 600, 700, 850, 925, \\ 1000 millibars, totally 13 levels}} \\
U & U-component Wind Speed & m/s & \\
V & V-component Wind Speed & m/s & \\
T & Temperature & K & \\
Q & Specific Humidity & kg/kg & \\
\bottomrule
\end{tabular}
\label{table:variable}
\end{table*}

\section{Dataset}
\label{appendix:dataset}
The HR-Extreme dataset~\cite{ran2025hr} is constructed from the High-Resolution Rapid Refresh (HRRR) data~\cite{noaaHRRR}, provided by the National Oceanic and Atmospheric Administration (NOAA). 
This dataset spans a spatial resolution of 3 km, with dimensions of 1059$\times$1799 latitude-longitude grid points, covering the contiguous United States (latitudes 21.1°–52.6°, longitudes 225.9°–299.1°). 
It features hourly temporal resolution and includes 69 atmospheric variables, the details of which are provided in~\tabref{table:variable}.

In addition, HR-Extreme incorporates an associated extreme event dataset derived from three primary sources:
(1) \textit{NOAA Storm Events Database}~\cite{noaaSED}. This database contains comprehensive records of storms and severe weather events from 1950 to 2024. 
To ensure data quality, HR-Extreme filters out events lacking precise spatiotemporal information or those involving non-meteorological hazards (\eg avalanches), retaining only well-documented meteorological extremes.
(2) \textit{NOAA Storm Prediction Center (SPC)}~\cite{noaaSPC}. While SPC provides real-time reports of hail, tornadoes, and wind events, its records lack a structured classification. 
To address this, HR-Extreme applies DBSCAN clustering to aggregate spatially proximate reports and remove noise, ensuring coherent event boundaries for robust model evaluation.
(3) \textit{Manually Filtered Extreme Temperatures}. Extreme temperature events are identified by thresholding HRRR’s 2-meter temperature data at the 5th and 95th percentiles. DBSCAN clustering is also employed to isolate large-scale heatwaves and cold spells while filtering out localized anomalies.

Given that HR-Extreme is limited to 2019–2020, we extend its methodology to construct \textit{HR-Extreme-V2}, covering 2019–2024 with updated extreme event records (2022–2024). This expanded dataset includes 18 distinct types of extreme events, detailed in~\tabref{table:event}. To improve computational efficiency, we reduce the original spatial resolution to 6 km, resulting in a grid of 530$\times$900 points while preserving key meteorological features.
The discussion of ethical use is introduced in \appref{appendix:ethical}.

\section{Ethical Use of Data}
\label{appendix:ethical}

The HR-Extreme-V2 dataset is constructed exclusively from publicly available meteorological data (\ie HRRR~\cite{noaaHRRR}) and extreme event records (\ie SED~\cite{noaaSED}, SPC~\cite{noaaSPC}) provided by the National Oceanic and Atmospheric Administration (NOAA) under open-access licenses. 
All original data sources are devoid of any personal or sensitive human subject information, as they solely document atmospheric variables (\eg temperature, wind speed) and aggregated extreme weather events (\eg storms, floods). 

The dataset is intended solely for scientific research in weather modeling and extreme weather prediction. Our data construction process follows the methodology of HR-Extreme~\cite{ran2025hr}, utilizing the Herbie~\cite{herbie} Python library for downloading general meteorological data. All code resources are publicly available on GitHub\footnote{HR-Extreme: \url{https://github.com/HuskyNian/HR-Extreme}; Herbie: \url{https://github.com/herbie/herbie}}.

\section{Metric}
\label{appendix:metric}
This section provides the formulations of our evaluation metrics: MAE, RMSE, and ACC, from 3 aspects: \textit{General}, \textit{Extreme}, and \textit{Gap}. 
For time $t$, given the prediction results $\hat{\mathbf{X}}^{t+1}$ and ground-truth $\mathbf{X}^{t+1}$, the general metrics for the variable $c$ are defined as:
\begin{align*}
&MAE_{Gen}(c) = \frac{1}{THW}\sum_{t=0}^{T-1}\sum_{h=0,w=0}^{H-1,W-1}|\hat{\mathbf{X}}^{t+1}_{h,w,c}-\mathbf{X}^{t+1}_{h,w,c}|, \\
&RMSE_{Gen}(c) = \sqrt{\frac{1}{THW}\sum_{t=0}^{T-1}\sum_{h=0,w=0}^{H-1,W-1}(\hat{\mathbf{X}}^{t+1}_{h,w,c}-\mathbf{X}^{t+1}_{h,w,c})^2}, \\
ACC&_{Gen}(t,c) = \frac{\sum_{h=0,w=0}^{H-1,W-1}\hat{\mathbf{D}}^{t+1}_{h,w,c}\mathbf{D}^{t+1}_{h,w,c}}{\sqrt{\sum_{h=0,w=0}^{H-1,W-1}(\hat{\mathbf{D}}^{t+1}_{h,w,c})^2\sum_{h=1}^H\sum_{w=0}^{W-1}(\mathbf{D}^{t+1}_{h,w,c})^2}},
\end{align*}
where $\hat{\mathbf{D}}^{t+1}_{h,w,c}=\hat{\mathbf{X}}^{t+1}_{h,w,c}-\mathbf{C}^{t+1}_{h,w,c}$, $\mathbf{D}^{t+1}_{h,w,c}=\mathbf{X}^{t+1}_{h,w,c}-\mathbf{C}^{t+1}_{h,w,c}$, and $T$ denotes the number of timesteps in the validation or testing set.
$\mathbf{C}^{t+1}_{h,w,c}$ represents The \textit{climatology} for variable $c$ at time $t+1$ and location $(h,w)$, which is pre-calculated using our training samples from 2019-2022, following the methodology described in \cite{lam2023learning,bi2023accurate}.
The overall ACC is computed by $ACC_{Gen}(c)=\sum_{t=0}^{T-1}ACC_{Gen}(t,c)$.
Note that our reported ACC results are generally high, since we focus on the nowcasting task, which aims to predict only the one-hour future states.

For the calculation of \textit{Extreme} metrics, we adopt similar formulas but mask the grid values in the normal weather regions as 0. Specifically, we apply an extreme-region mask $\mathbf{S}^t\in\mathbb{R}^{H\times W}$ on the predictions and ground-truths $\hat{\mathbf{X}}^{t+1},\mathbf{X}^{t+1}$, which is defined as:
\begin{align*}
\mathbf{S}^t_{h,w} = 
\begin{cases}
1, & if\ \exists e_j \in \mathcal{E}, (h,w)\ is\ contained\ in\ e_j\\
0, & otherwise.
\end{cases}\\
\end{align*}
Then, extreme metrics are calculated by:
\begin{align*}
&MAE_{Ext}(c) = \frac{\sum_{t=0}^{T-1}\sum_{h=0,w=0}^{H-1,W-1}\mathbf{S}_{h,w}^{t+1}|\hat{\mathbf{X}}^{t+1}_{h,w,c}-\mathbf{X}^{t+1}_{h,w,c}|}{\sum_{t=0}^{T-1}\sum_{h=0,w=0}^{H-1,W-1}\mathbf{S}^t_{h,w}}, \\
&RMSE_{Ext}(c) = \sqrt{\frac{\sum_{t=0}^{T-1}\sum_{h=0,w=0}^{H-1,W-1}\mathbf{S}_{h,w}^{t+1}(\hat{\mathbf{X}}^{t+1}_{h,w,c}-\mathbf{X}^{t+1}_{h,w,c})^2}{\sum_{t=0}^{T-1}\sum_{h=0,w=0}^{H-1,W-1}\mathbf{S}^t_{h,w}}}, \\
ACC&_{Ext}(t,c) = \frac{\sum_{h=0,w=0}^{H-1,W-1}\mathbf{S}_{h,w}^{t+1}\hat{\mathbf{D}}^{t+1}_{h,w,c}\mathbf{D}^{t+1}_{h,w,c}}{\sqrt{\sum_{h=0,w=0}^{H-1,W-1}\mathbf{S}_{h,w}^{t+1}(\hat{\mathbf{D}}^{t+1}_{h,w,c})^2\sum_{h=1}^H\sum_{w=0}^{W-1}\mathbf{S}_{h,w}^{t+1}(\mathbf{D}^{t+1}_{h,w,c})^2}}.
\end{align*}

Lastly, we compute \textit{Gap} metrics as the difference between \textit{General} and \textit{Extreme} results:
\begin{align*}
&MAE_{Gap}(t,c) = MAE_{Ext}(t,c) - MAE_{Gen}(t,c), \\
&RMSE_{Gap}(t,c) = RMSE_{Ext}(t,c) - RMSE_{Gen}(t,c), \\
&ACC_{Gap}(t,c) = ACC_{Gen}(t,c) - ACC_{Ext}(t,c).
\end{align*}

\begin{table*}[h]
\centering
\caption{Descriptions of Extreme Weather Events.}
\resizebox{0.875\linewidth}{!}{
\begin{tabular}{lll}
\toprule
\textbf{Type} & \textbf{Abbr.} & \textbf{Description} \\
\midrule
Flood & Flod & \makecell[l]{Any high flow, overflow, or inundation by water causing damage, \\ typically from inundation of normally dry areas due to increased \\ water levels.} \\
\midrule
Marine Thunderstorm Wind & MTSW & \makecell[l]{Thunderstorm winds over marine areas with speeds of at least 34 \\ knots (39 mph) for up to 2 hours.} \\
\midrule
Waterspout & Wtsp & \makecell[l]{Rotating column of air extending from cloud base to water surface \\ in bays and Great Lakes areas.} \\
\midrule
Thunderstorm Wind & TnWn & \makecell[l]{Convective winds occurrin within 30 minutes of lightning, with \\ speeds $\geq$50 knots (58 mph).} \\
\midrule
Funnel Cloud & FnCl & \makecell[l]{Rotating cloud extension not reaching ground, precursor to severe \\ weather with potential aviation hazards from wind shear.} \\
\midrule
Tornado & Trnd & \makecell[l]{Violently rotating air column extending from cloud to ground, often \\ visible as condensation funnel.} \\
\midrule
Wind & Wind & \makecell[l]{Severe thunderstorm or strong wind causing damage, as recorded \\ by NOAA Storm Prediction Center.} \\
\midrule
Hail & Hail & \makecell[l]{Frozen precipitation in form of balls or irregular lumps of ice.} \\
\midrule
Flash Flood & FlFl & \makecell[l]{Life-threatening rapid water rise in normally dry areas within minutes \\ to hours following intense rainfall or other triggering events.} \\
\midrule
Lightning & Ltgn & \makecell[l]{Sudden electrical discharge from thunderstorms causing fatalities, \\ injuries, or damage.} \\
\midrule
Heavy Rain & HvRn & \makecell[l]{Unusually large rainfall amount causing damage without meeting \\ flood criteria.} \\
\midrule
Cold & Cold & \makecell[l]{Large area of excessive cold below -29°C in US during nighttime.} \\
\midrule
Marine High Wind & MHWn & \makecell[l]{Non-convective sustained winds $\geq$48 knots (55 mph) over marine \\ areas causing fatalities, injuries, or damage.} \\
\midrule
Debris Flow & DbrF & \makecell[l]{Slurry of loose materials triggered by intense rain (often post-wildfire),\\ capable of carrying large particles in suspension.} \\
\midrule
Dust Devil & DstD & \makecell[l]{Small, rotating updraft of air visible by dust or debris it picks up, \\ typically not severe.} \\
\midrule
Marine Hail & MrHl & \makecell[l]{Hail $\geq$0.75 inch diameter occurring over marine forecast zones.} \\
\midrule
Heat & Heat & \makecell[l]{Large area of excessive heat above 37°C in US during daytime.} \\
\midrule
Marine Strong Wind & MSWn & \makecell[l]{Non-convective sustained winds up to 47 knots (54 mph) over marine \\ areas causing fatalities, injuries, or damage.} \\
\bottomrule
\end{tabular}
}
\label{table:event}
\end{table*}

\section{Implementation Details}
\label{appendix:setup}
Our experimental setup utilizes weather states of 2019 to 2022 for training, with data from 2023 and 2024 reserved for validation and inference. For the Numerical Weather Prediction (NWP) baseline, we download the predictions of the WRF-ARW model using the Herbie library~\cite{herbie}.
All deep learning baselines, including our proposed UniExtreme model, employ consistent hyperparameters: a patch size of $8 \times 8$, batch size of 12 (except for two GNN~\cite{gcn,ni2025unsupervised}-based models GraphCast and OneForecast, which require batch size=1 in their original implementations), initial learning rate of $1 \times 10^{-3}$, and 50 training epochs. We optimize the models using the AdamW optimizer with weight decay $3 \times 10^{-6}$ and employ a StepLR learning rate scheduler with a decay step size of 1 and a decay factor of 0.85. Early stopping is implemented based on validation performance (measured by MAE$_{Ext}$) with a patience of 5 epochs. All experiments are conducted using PyTorch on 4 NVIDIA A800-80G GPUs.

For our UniExtreme model, we partition the spatial domain into regions of size $(a_h, a_w) = (10, 10)$. The AFM module incorporates 10 Beta filters (\ie $N=10$) with a band growth rate $\gamma = 1.3$ and a numerical limit $\text{MAX}_\kappa = 70$. Temporal features are encoded using time embeddings of dimension $D_T = 72$. To ensure computational efficiency when processing real-valued weather observations, we implement frequency analysis and modulation using the Real-Valued Fast Fourier Transform (RFFT).
The EPA module constructs its memory using only extreme weather records from 2022, with fixed memory capacity $U = 5$. 
Following the architectural choices of FuXi, we set the dimension $D$ as 512 in the Transformer backbone.


\begin{figure*}[h]
    \centering
    \includegraphics[width=0.9\linewidth]{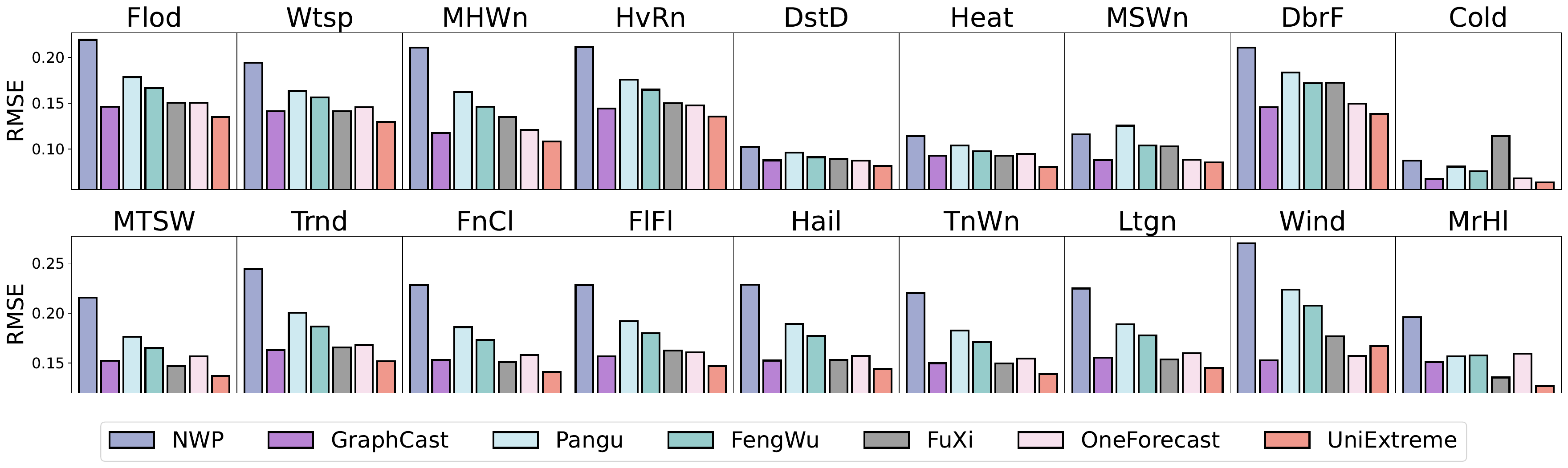}
    \caption{RMSE performance forecasting 18 types of extreme events.}
    \label{figure:type_rmse}
\end{figure*}

\begin{figure*}[h]
    \centering
    \includegraphics[width=0.9\linewidth]{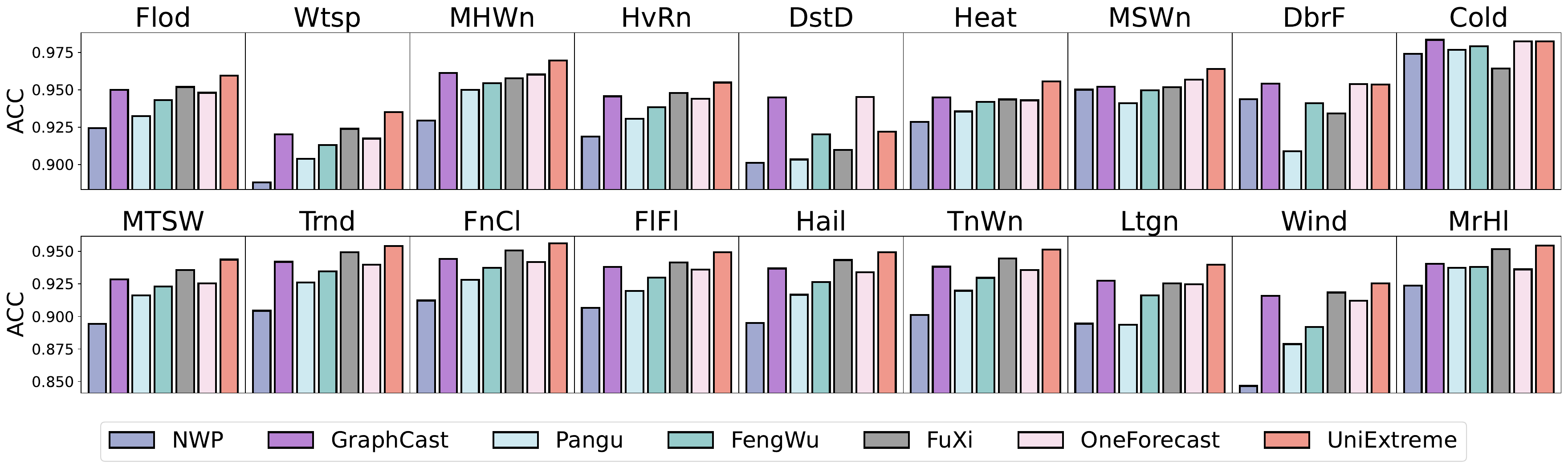}
    \caption{ACC performance forecasting 18 types of extreme events.}
    \label{figure:type_acc}
\end{figure*}

\section{Complete Raw-Scale Results}
\label{appendix:raw_var}

\begin{figure*}[!ht]
\begin{minipage}[t]{0.48\textwidth}
    \centering
    \includegraphics[width=\linewidth]{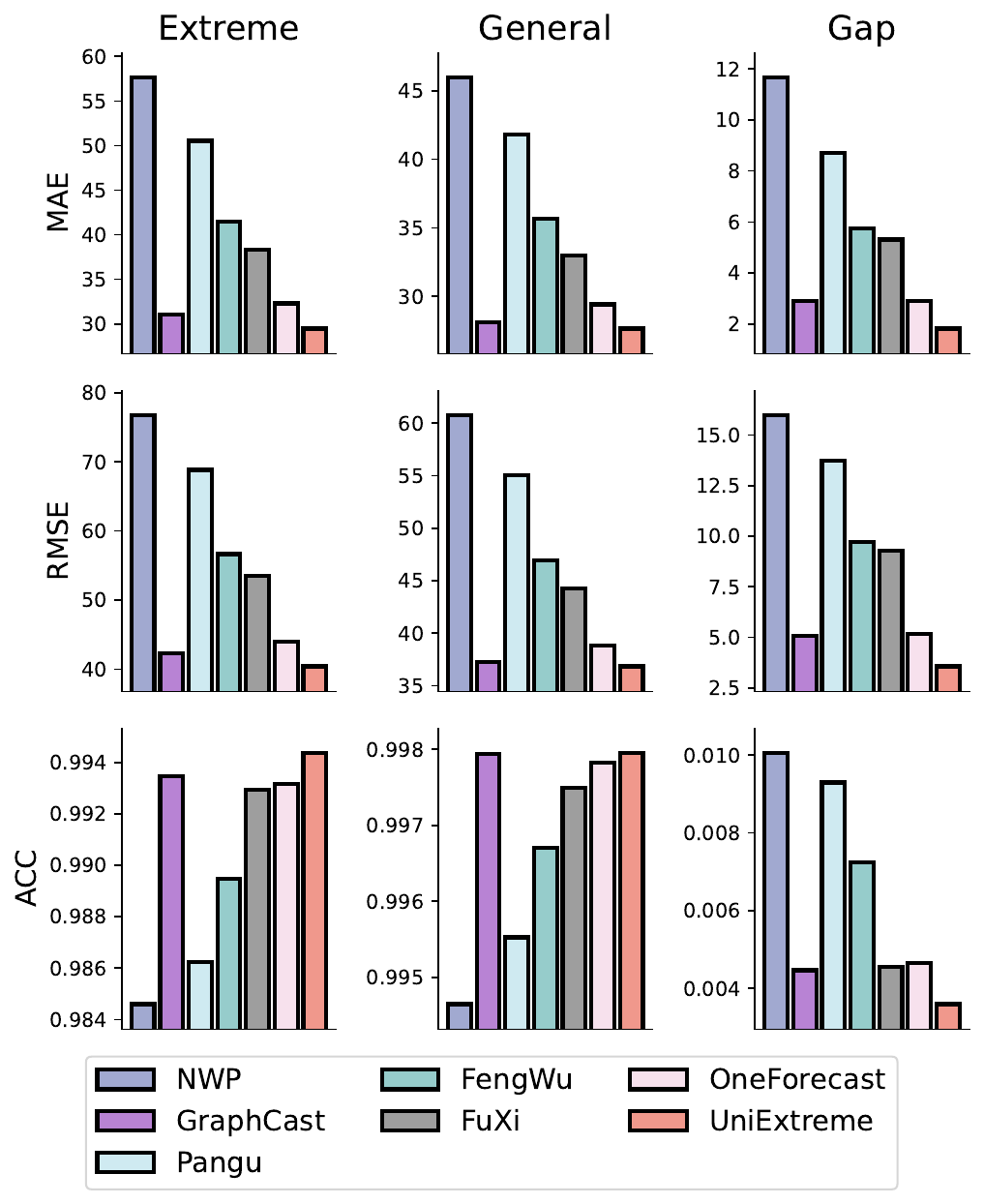}
    \vspace{-25pt}
    \caption{Raw forecasting results of variable MSL.}
    \vspace{-5pt}
    \label{fig:raw_msl_left}
\end{minipage}
\hfill
\begin{minipage}[t]{0.48\textwidth}
    \centering
    \includegraphics[width=\linewidth]{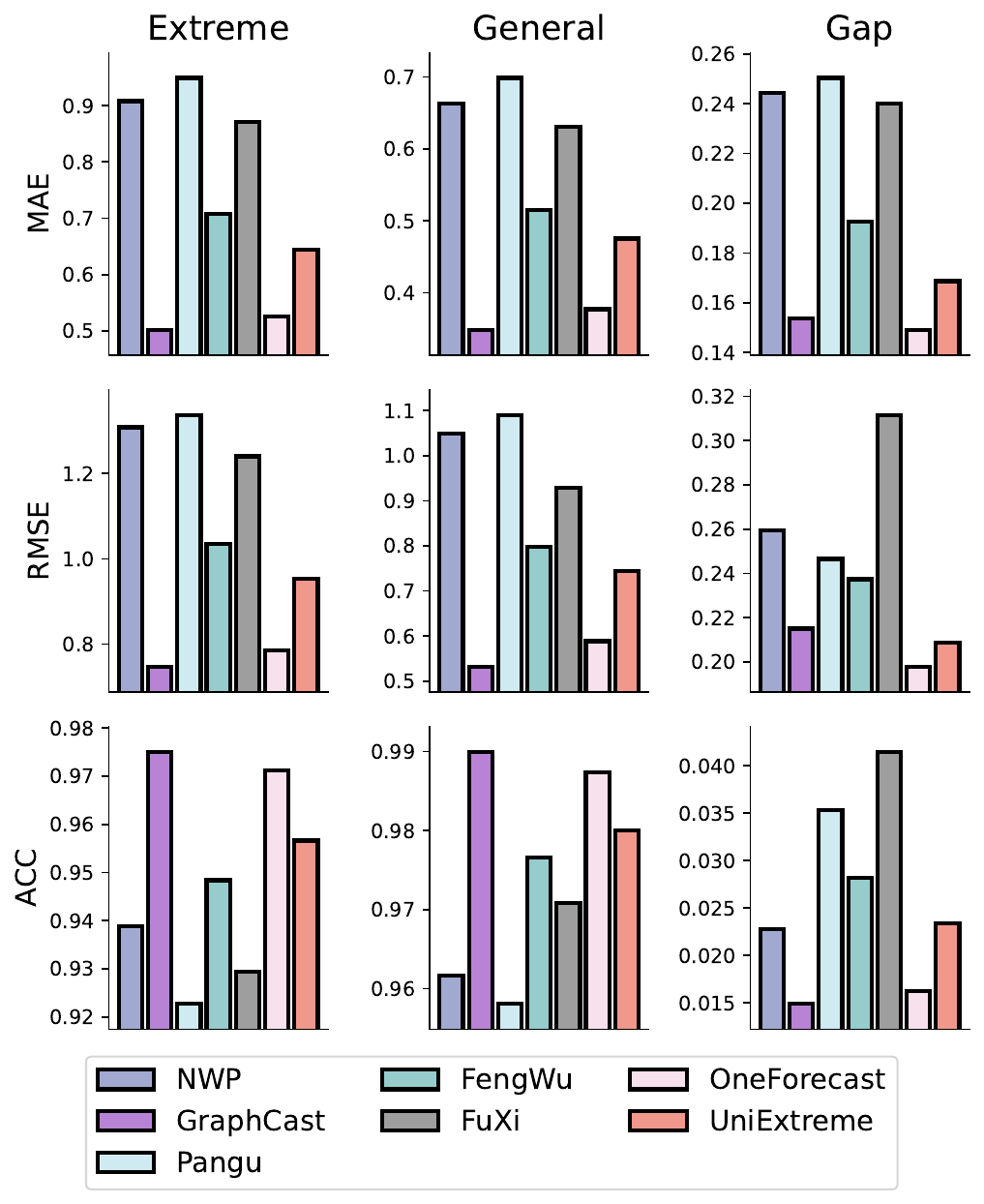}
    \vspace{-25pt}
    \caption{Raw forecasting results of variable T2M.}
    \vspace{-5pt}
    \label{fig:raw_t2m_right}
\end{minipage}
\\[10pt]
\begin{minipage}[t]{0.48\textwidth}
    \centering
    \includegraphics[width=\linewidth]{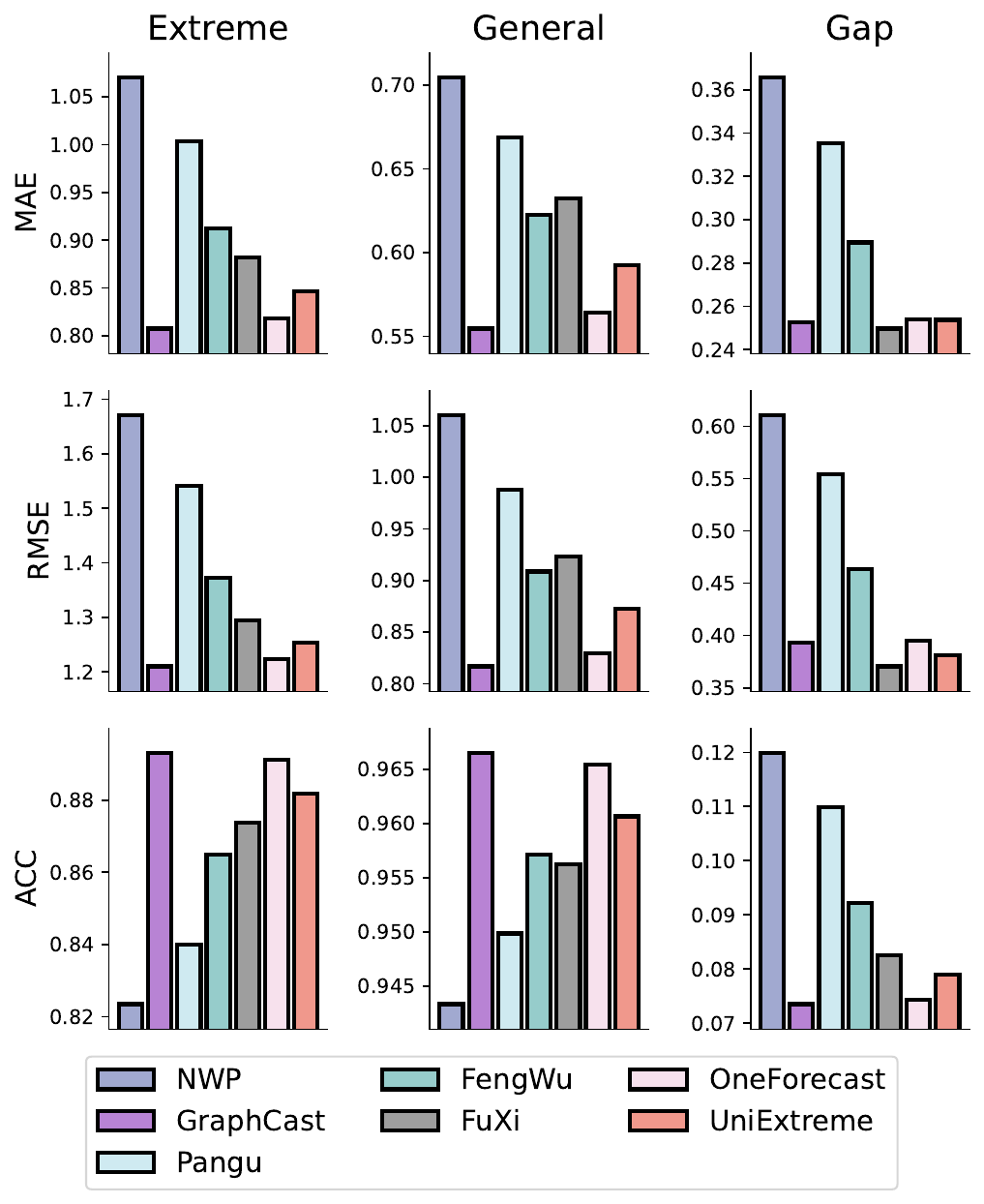}
    \vspace{-25pt}
    \caption{Raw forecasting results of variable U10.}
    \vspace{-5pt}
    \label{fig:raw_u10_left}
\end{minipage}
\hfill
\begin{minipage}[t]{0.48\textwidth}
    \centering
    \includegraphics[width=\linewidth]{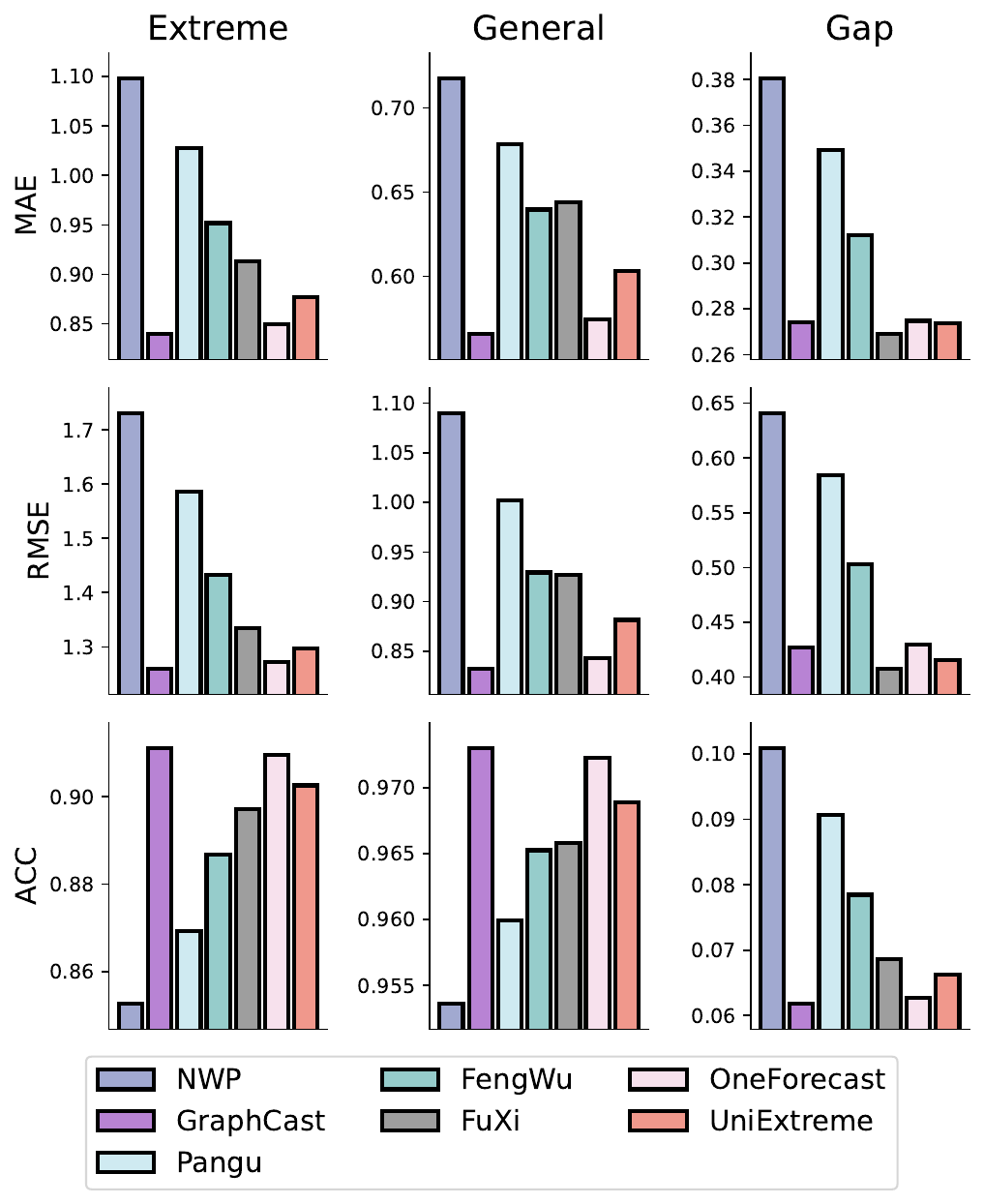}
    \vspace{-25pt}
    \caption{Raw forecasting results of variable V10.}
    \vspace{-5pt}
    \label{fig:raw_v10_right}
\end{minipage}
\end{figure*}

\begin{figure*}[!ht]
\begin{minipage}[t]{0.48\textwidth}
    \centering
    \includegraphics[width=\linewidth]{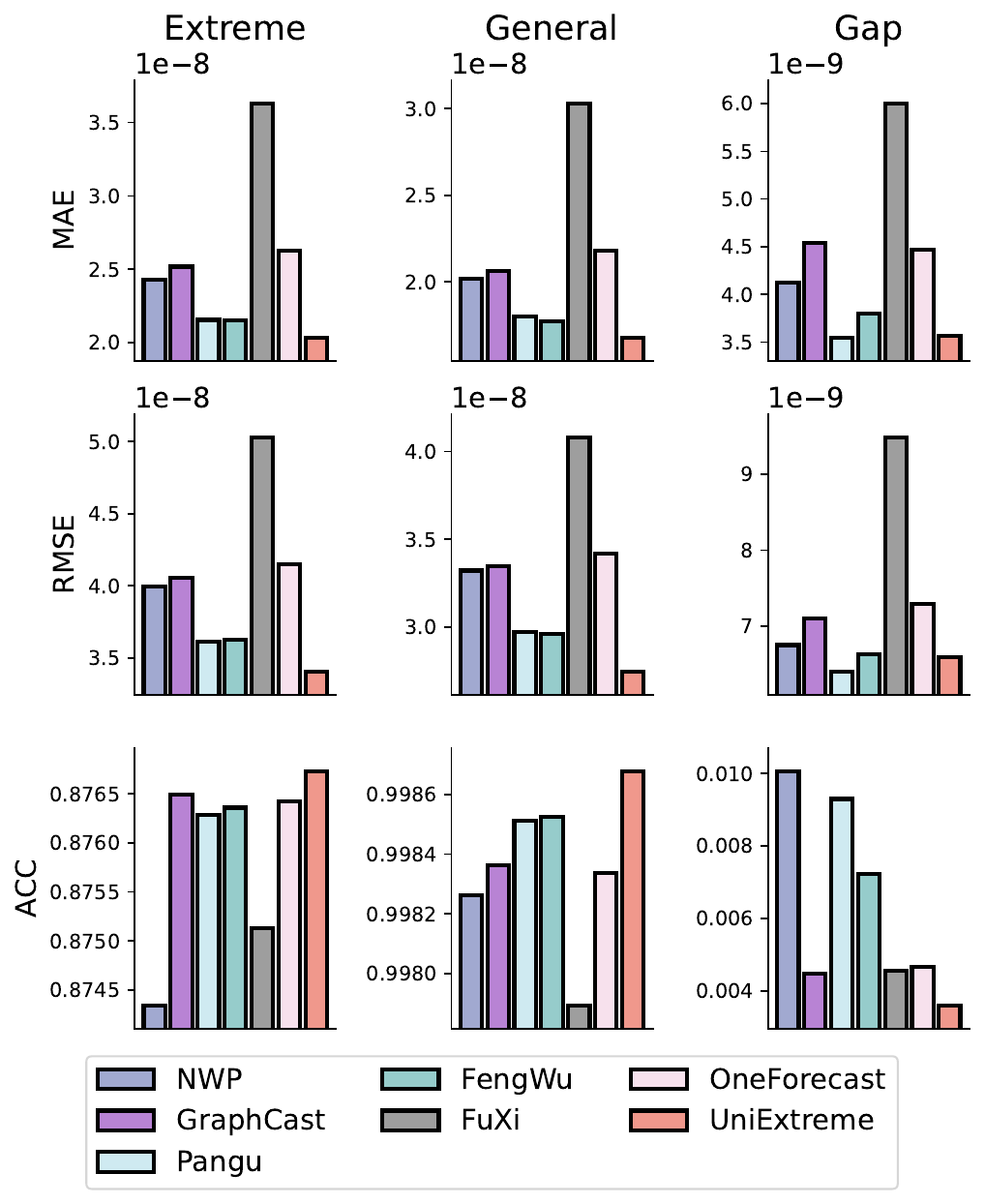}
    \vspace{-25pt}
    \caption{Raw forecasting results of variable Q50.}
    \vspace{-5pt}
    \label{fig:raw_q_50_left}
\end{minipage}
\hfill
\begin{minipage}[t]{0.48\textwidth}
    \centering
    \includegraphics[width=\linewidth]{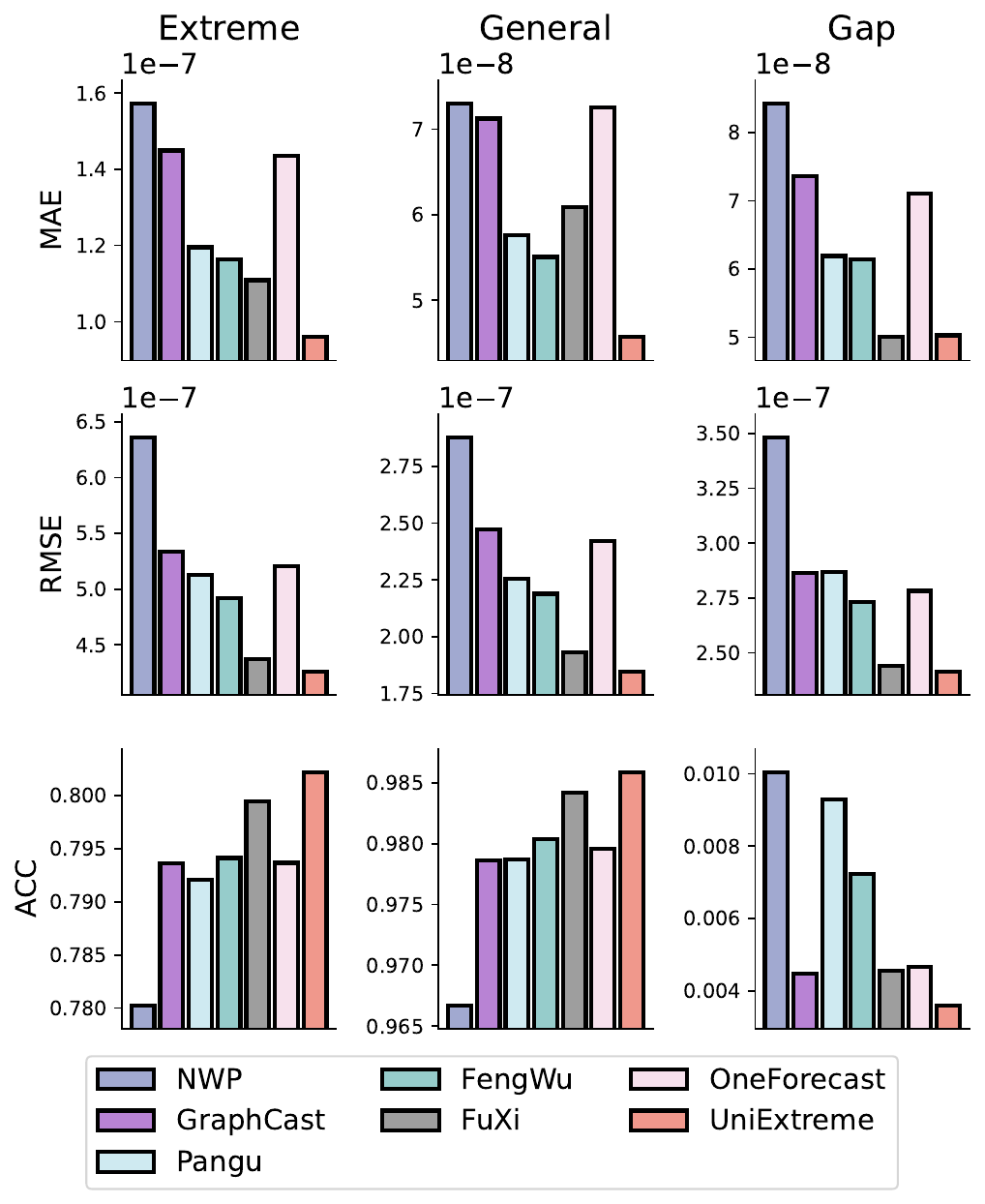}
    \vspace{-25pt}
    \caption{Raw forecasting results of variable Q100.}
    \vspace{-5pt}
    \label{fig:raw_q_100_right}
\end{minipage}
\\[10pt]
\begin{minipage}[t]{0.48\textwidth}
    \centering
    \includegraphics[width=\linewidth]{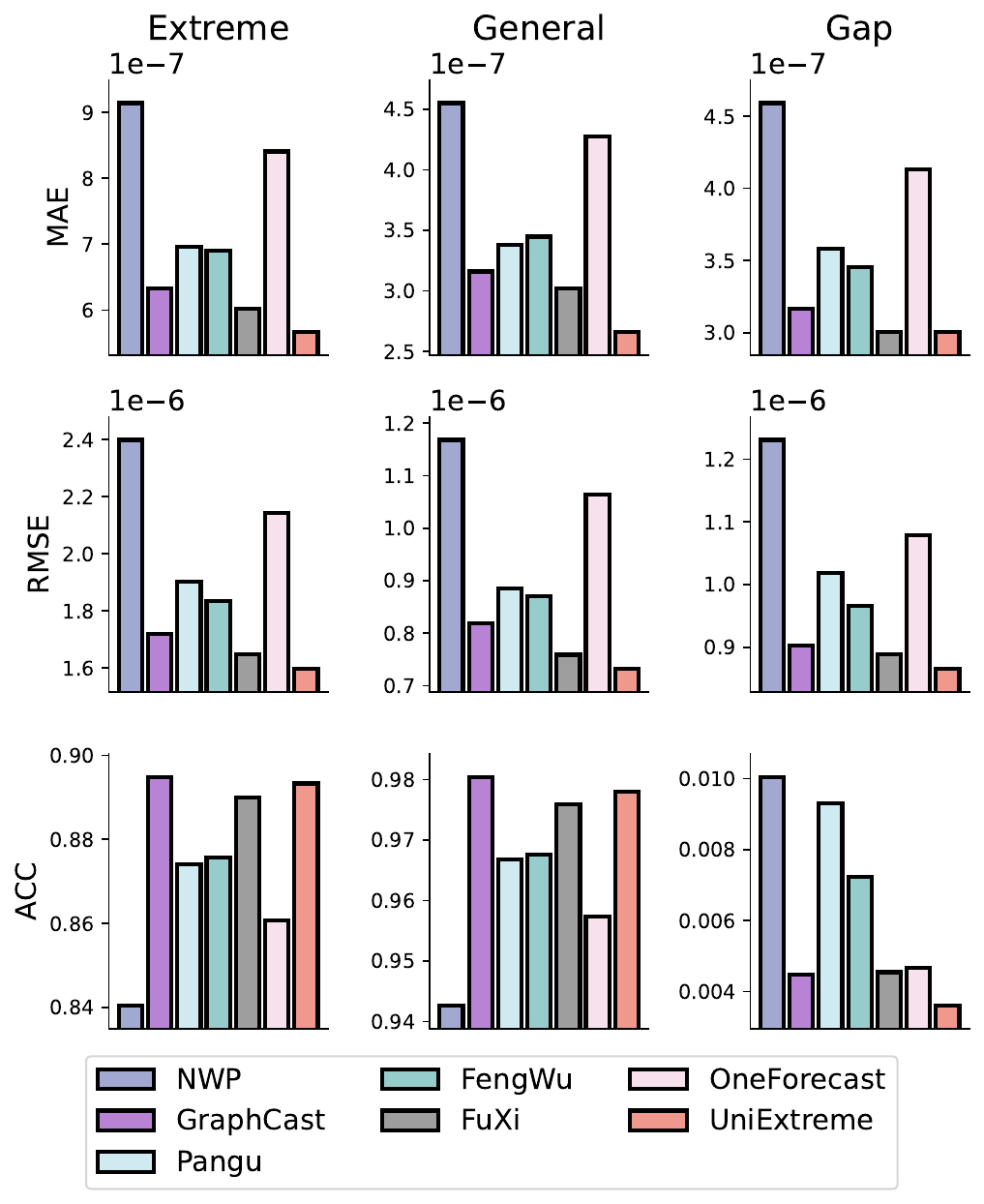}
    \vspace{-25pt}
    \caption{Raw forecasting results of variable Q150.}
    \vspace{-5pt}
    \label{fig:raw_q_150_left}
\end{minipage}
\hfill
\begin{minipage}[t]{0.48\textwidth}
    \centering
    \includegraphics[width=\linewidth]{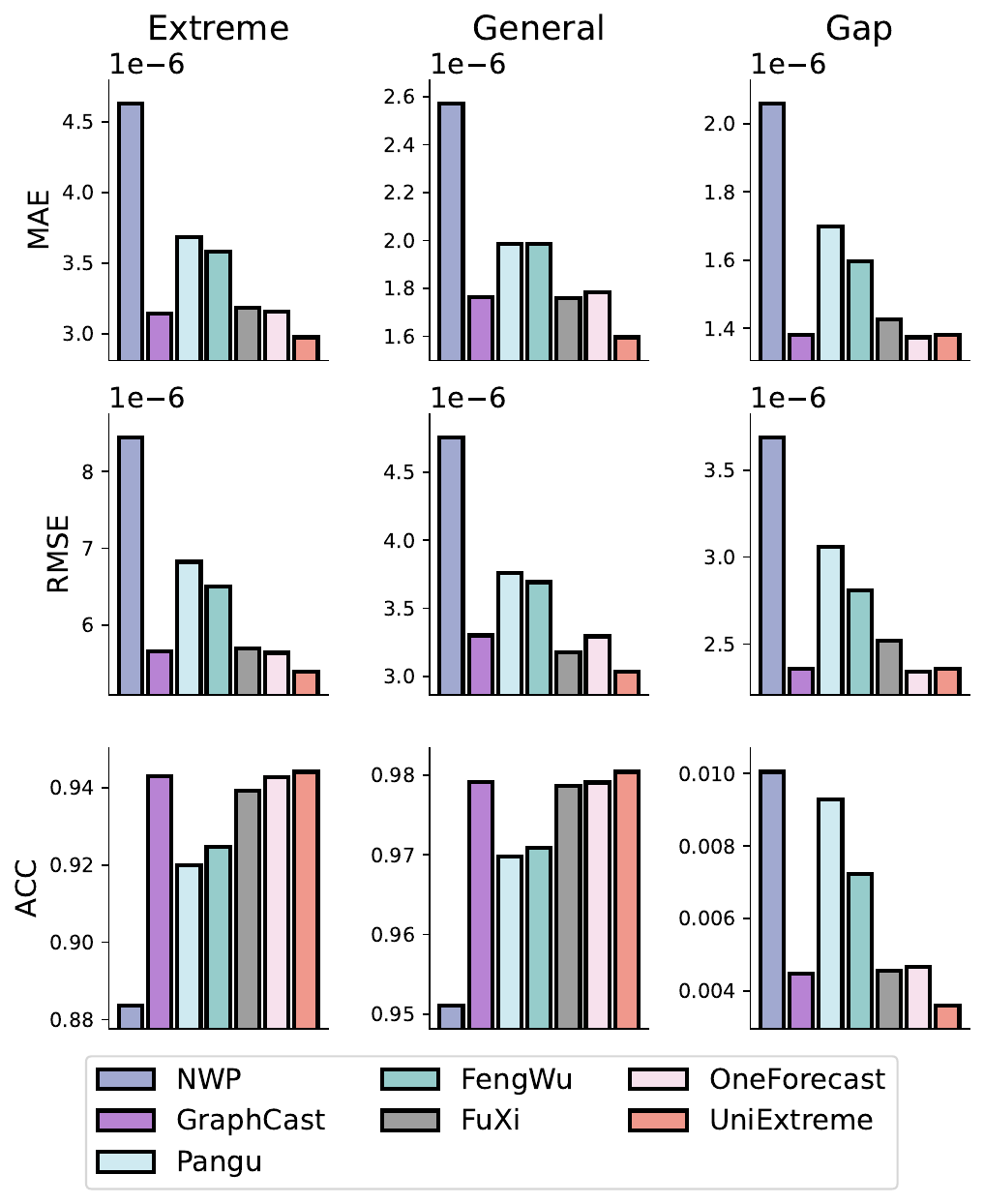}
    \vspace{-25pt}
    \caption{Raw forecasting results of variable Q200.}
    \vspace{-5pt}
    \label{fig:raw_q_200_right}
\end{minipage}
\end{figure*}

\begin{figure*}[!ht]
\begin{minipage}[t]{0.48\textwidth}
    \centering
    \includegraphics[width=\linewidth]{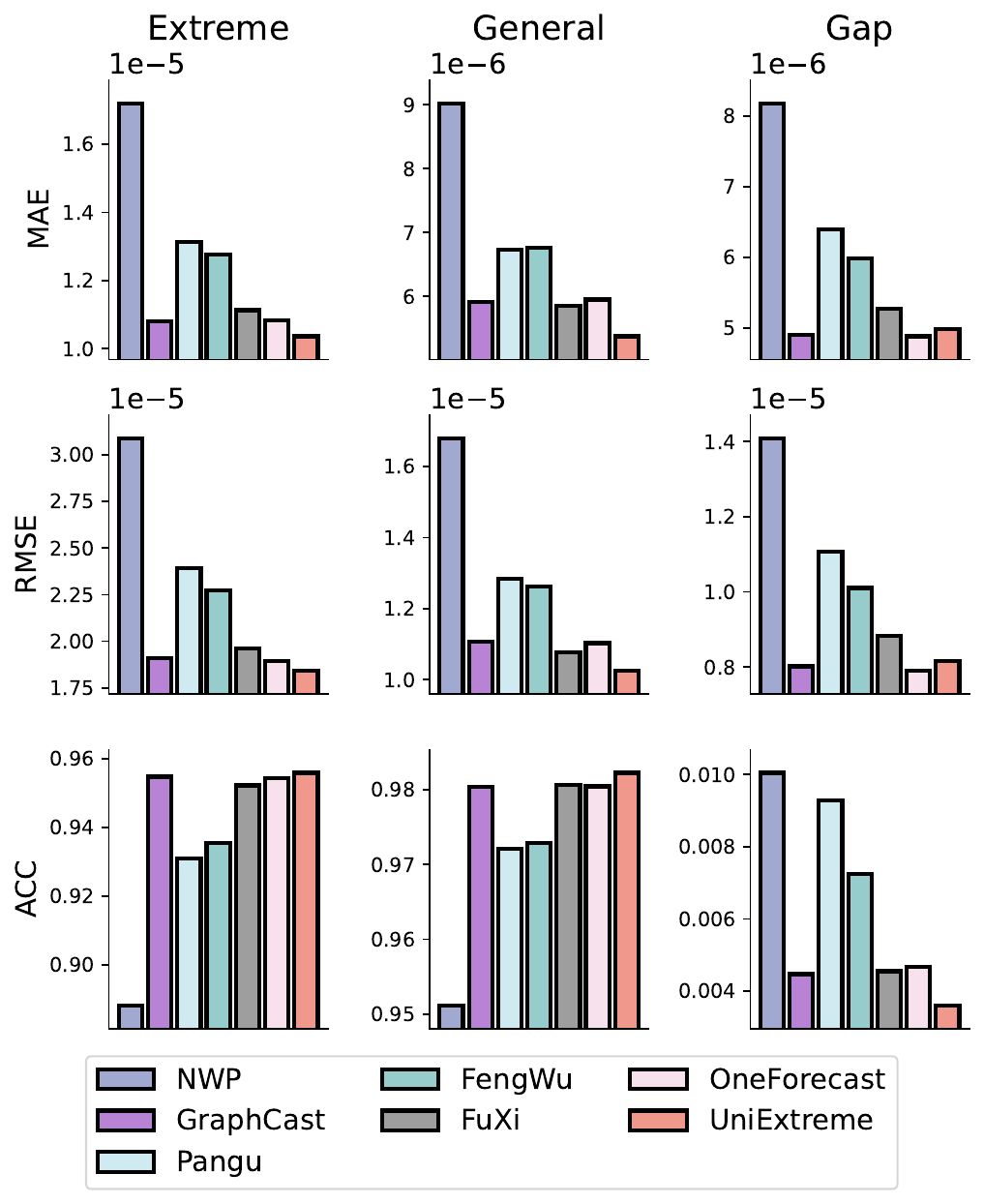}
    \vspace{-25pt}
    \caption{Raw forecasting results of variable Q250.}
    \vspace{-5pt}
    \label{fig:raw_q_250_left}
\end{minipage}
\hfill
\begin{minipage}[t]{0.48\textwidth}
    \centering
    \includegraphics[width=\linewidth]{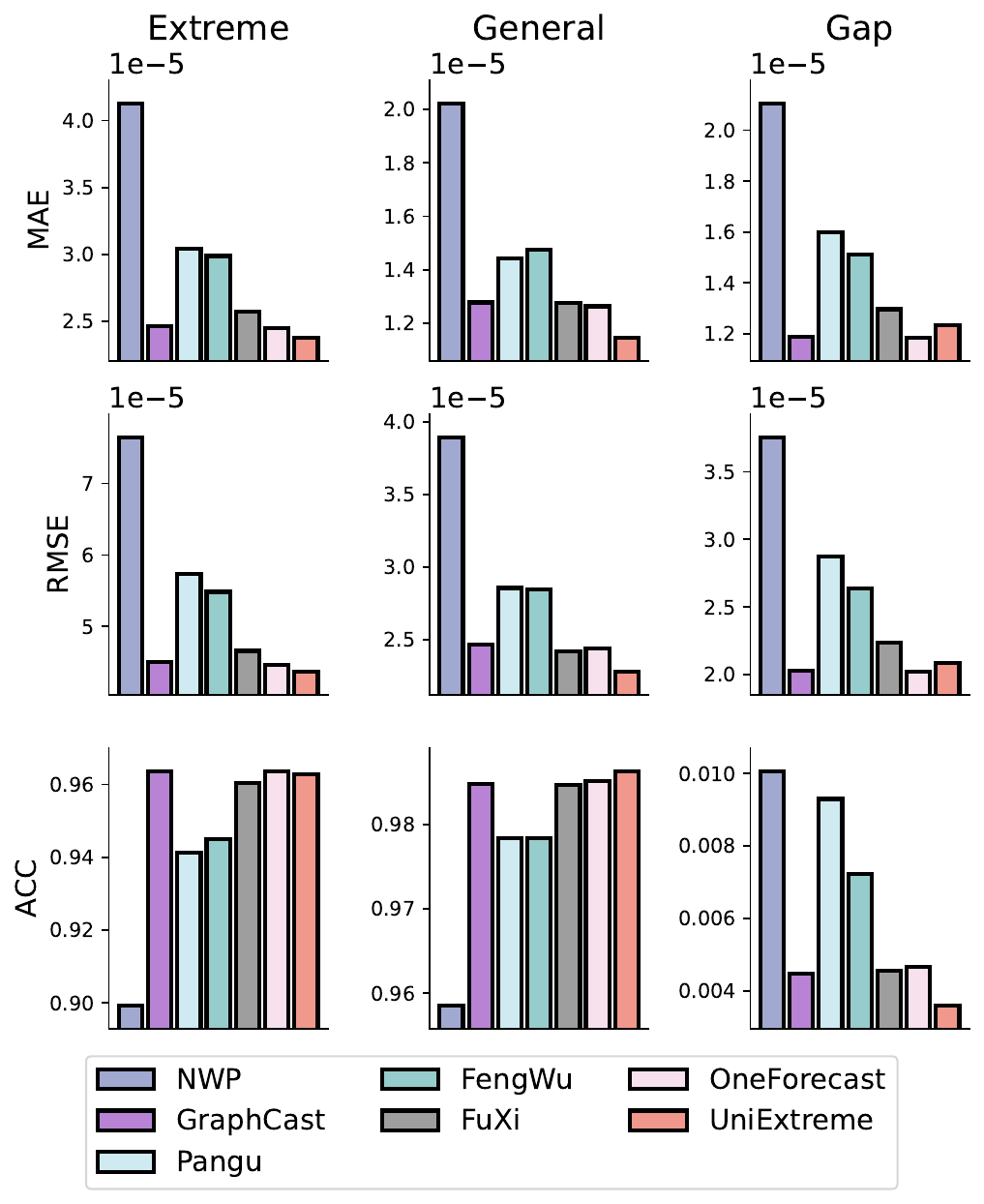}
    \vspace{-25pt}
    \caption{Raw forecasting results of variable Q300.}
    \vspace{-5pt}
    \label{fig:raw_q_300_right}
\end{minipage}
\\[10pt]
\begin{minipage}[t]{0.48\textwidth}
    \centering
    \includegraphics[width=\linewidth]{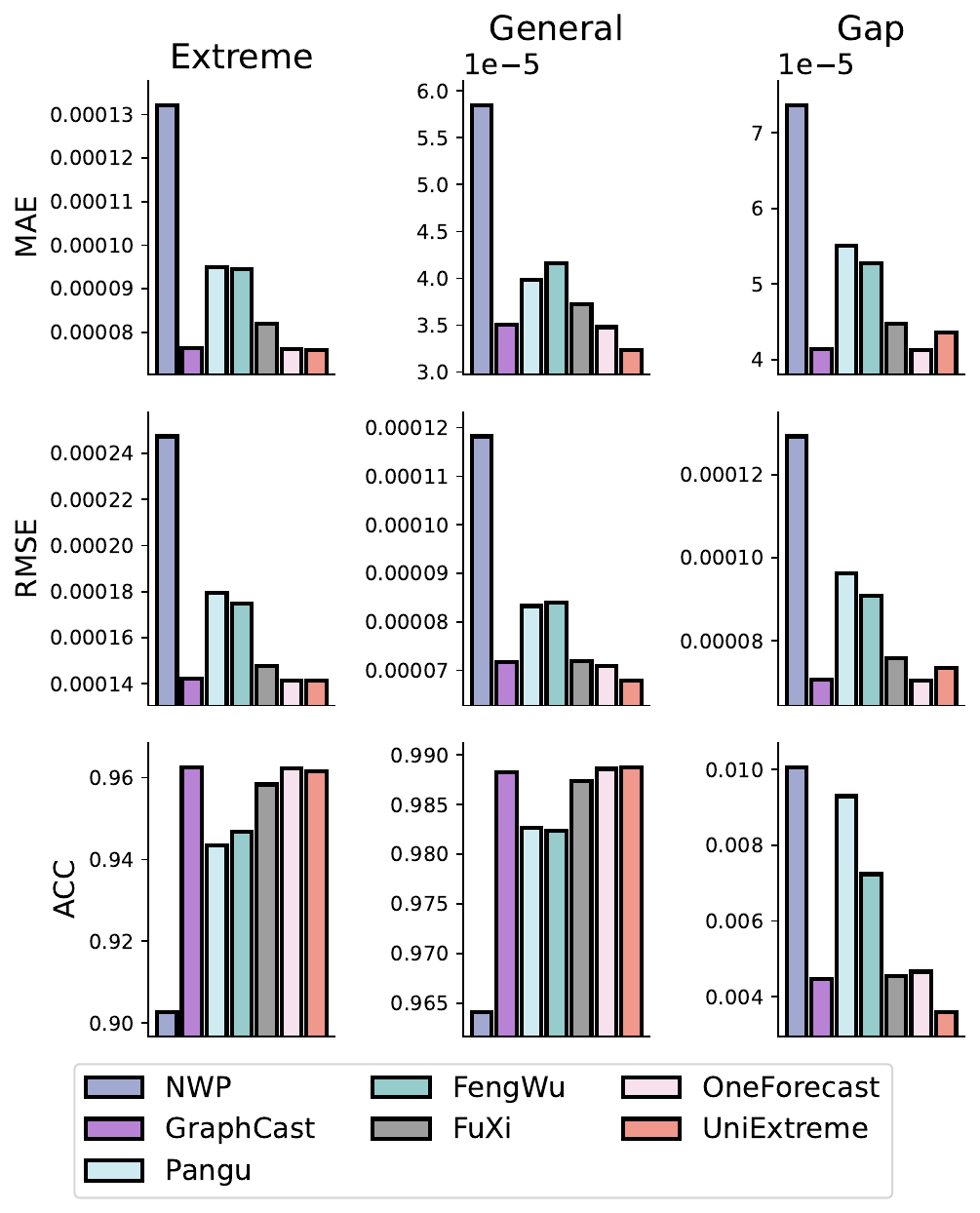}
    \vspace{-25pt}
    \caption{Raw forecasting results of variable Q400.}
    \vspace{-5pt}
    \label{fig:raw_q_400_left}
\end{minipage}
\hfill
\begin{minipage}[t]{0.48\textwidth}
    \centering
    \includegraphics[width=\linewidth]{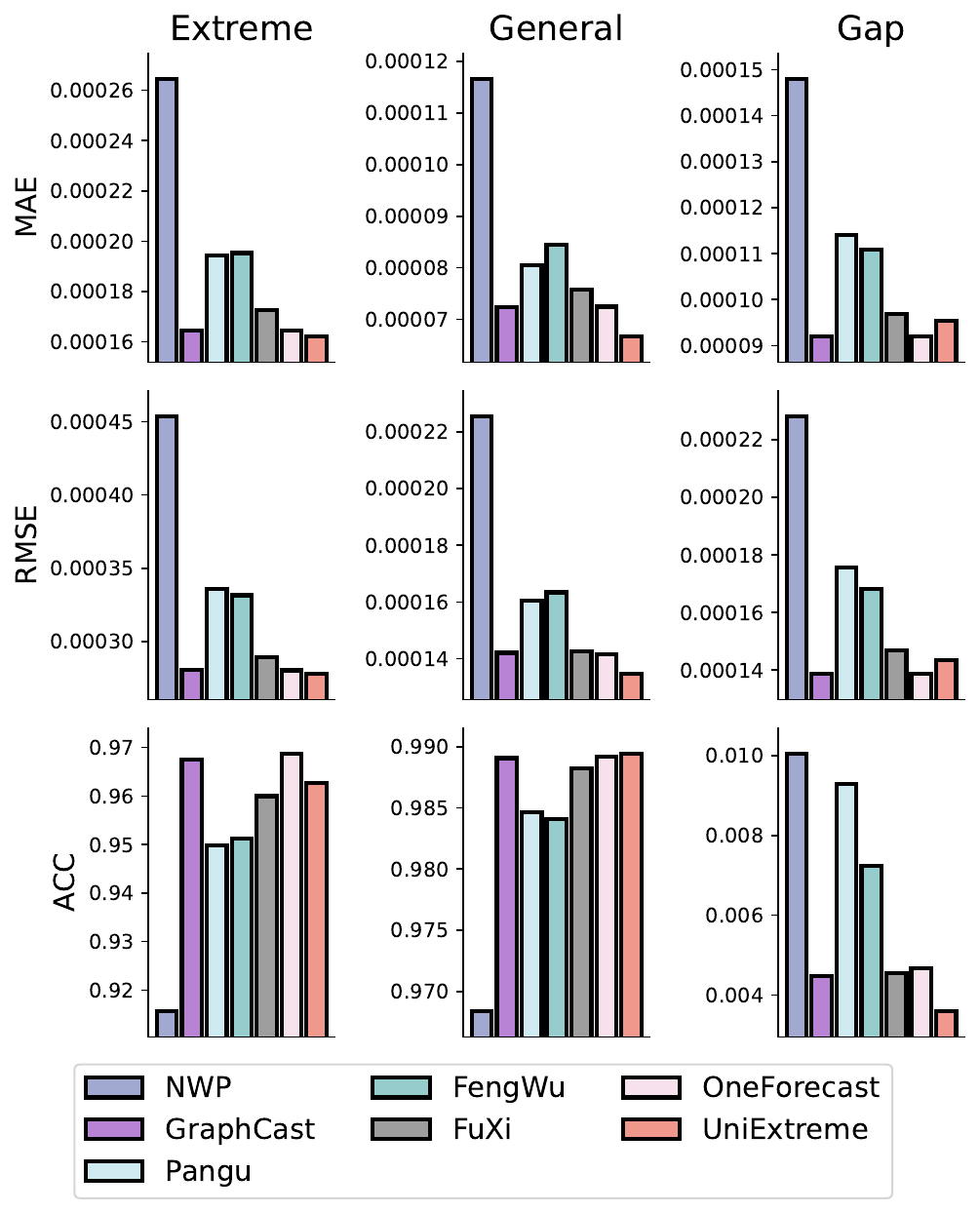}
    \vspace{-25pt}
    \caption{Raw forecasting results of variable Q500.}
    \vspace{-5pt}
    \label{fig:raw_q_500_right}
\end{minipage}
\end{figure*}

\begin{figure*}[!ht]
\begin{minipage}[t]{0.48\textwidth}
    \centering
    \includegraphics[width=\linewidth]{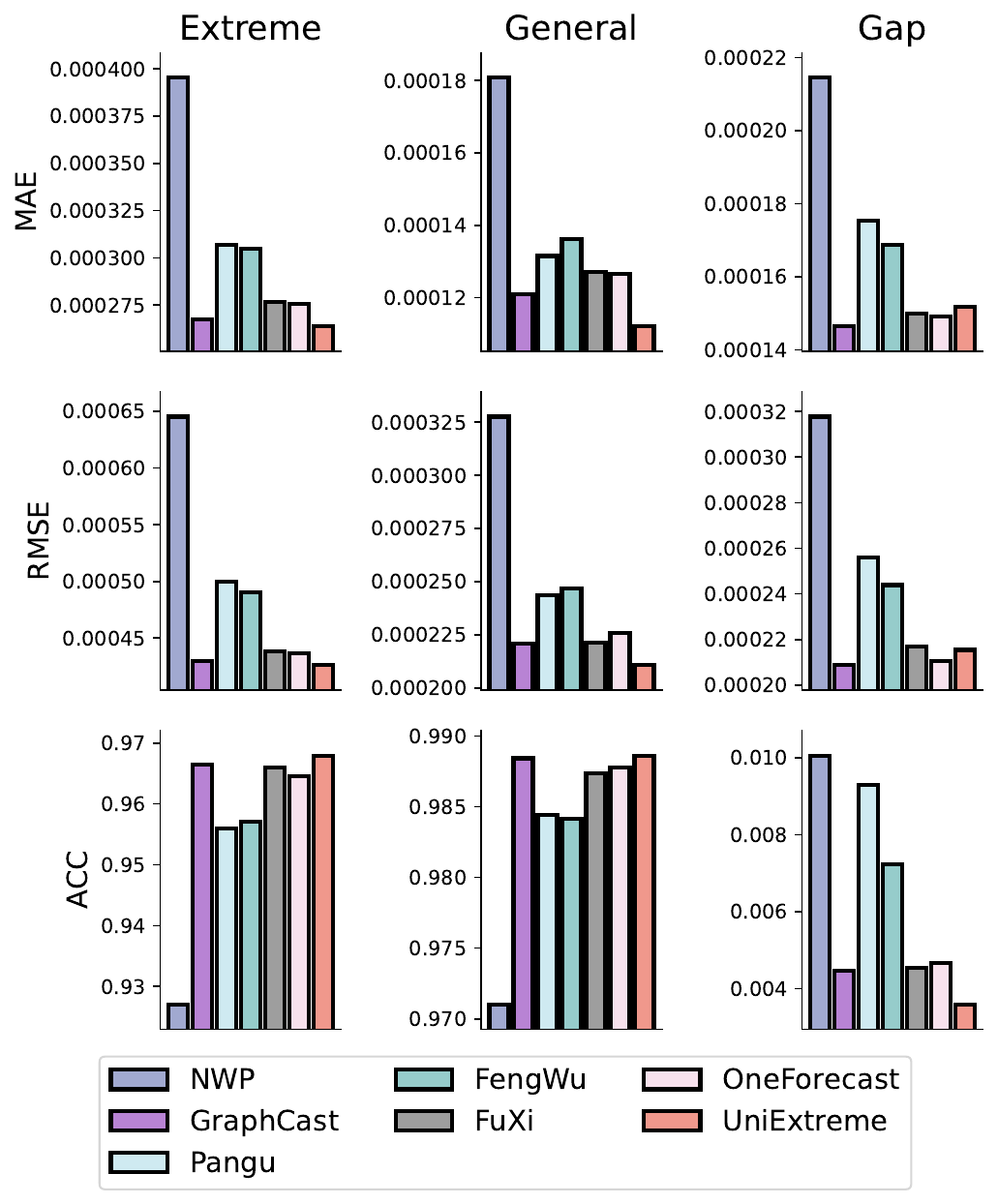}
    \vspace{-25pt}
    \caption{Raw forecasting results of variable Q600.}
    \vspace{-5pt}
    \label{fig:raw_q_600_left}
\end{minipage}
\hfill
\begin{minipage}[t]{0.48\textwidth}
    \centering
    \includegraphics[width=\linewidth]{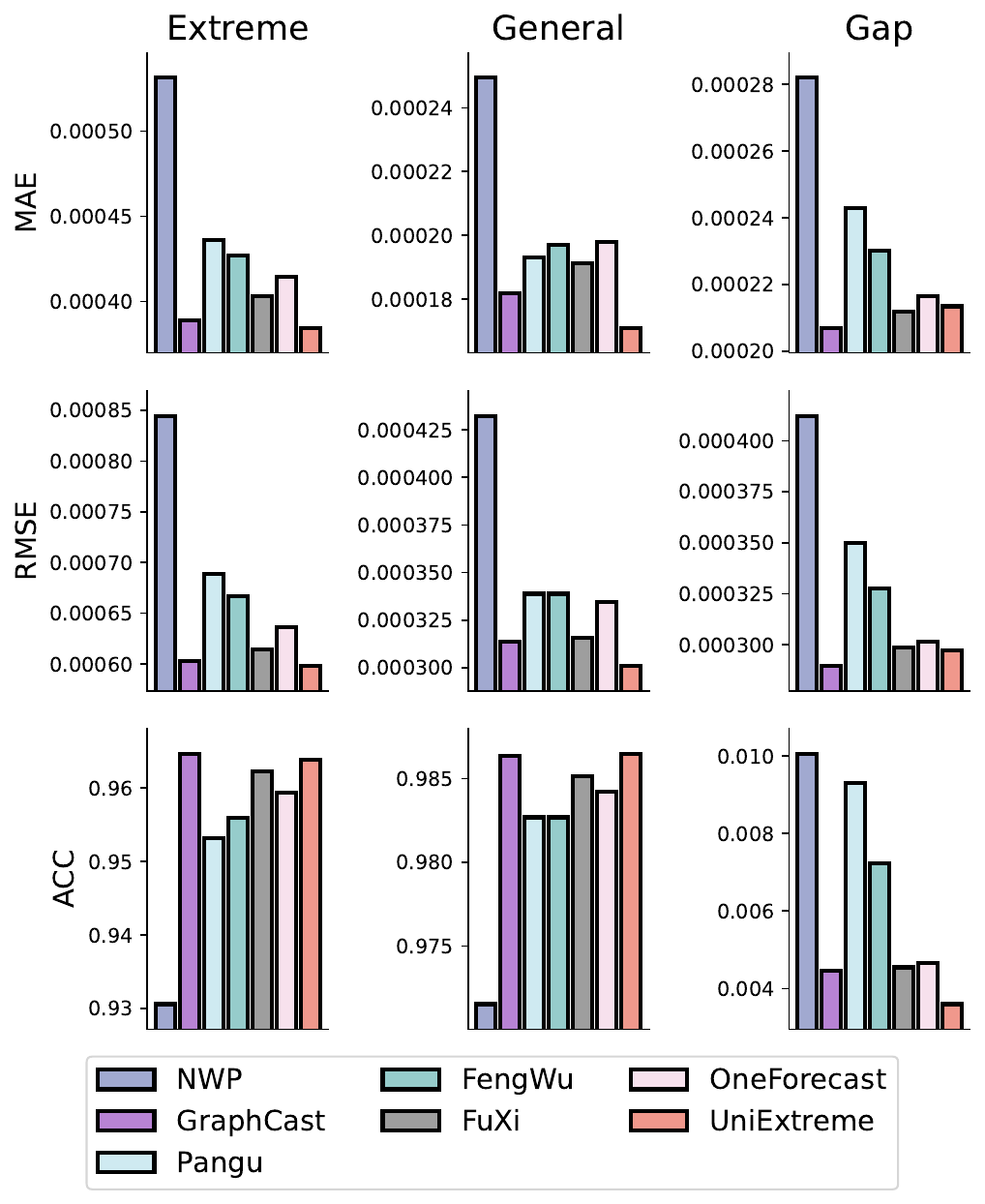}
    \vspace{-25pt}
    \caption{Raw forecasting results of variable Q700.}
    \vspace{-5pt}
    \label{fig:raw_q_700_right}
\end{minipage}
\\[10pt]
\begin{minipage}[t]{0.48\textwidth}
    \centering
    \includegraphics[width=\linewidth]{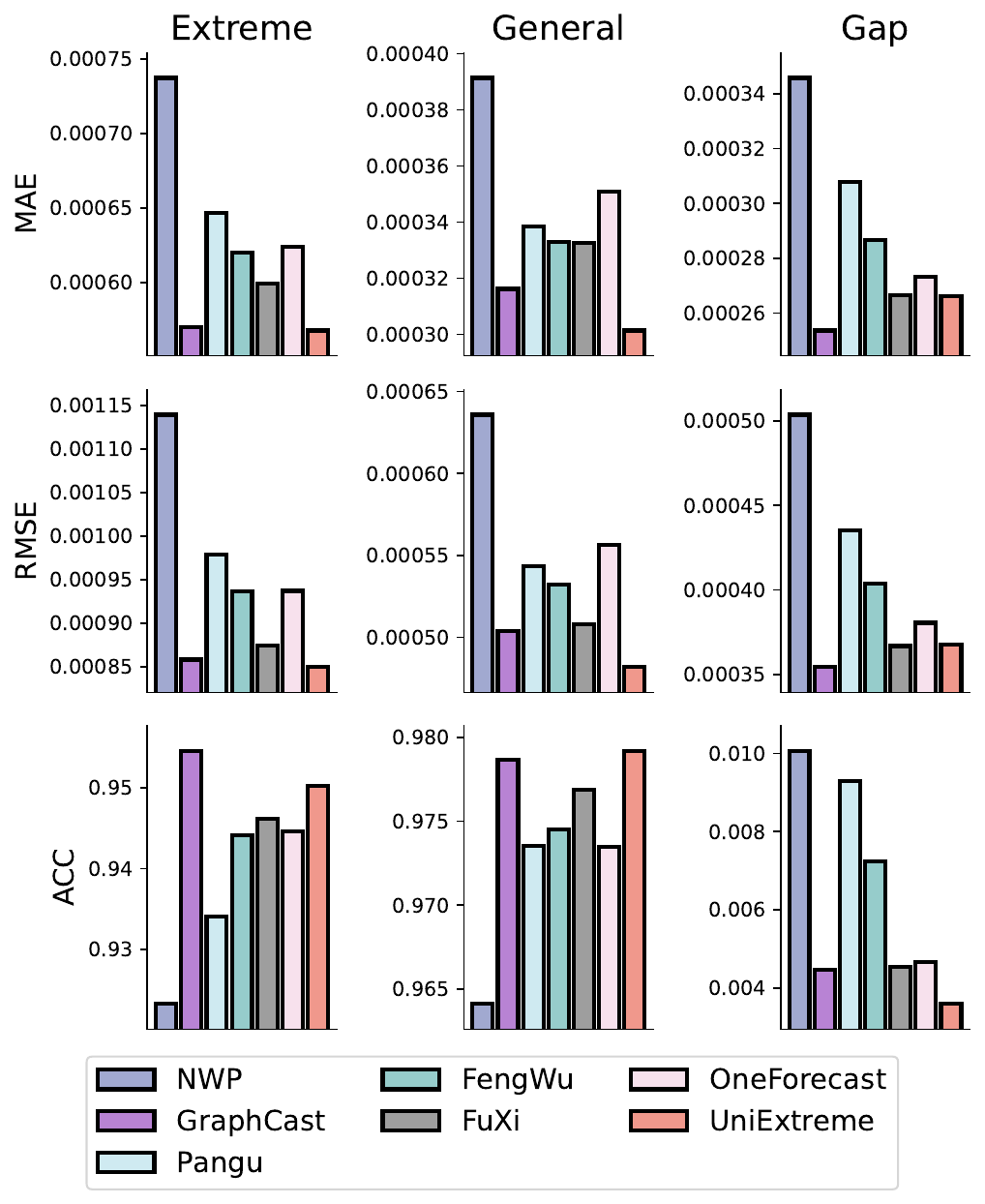}
    \vspace{-25pt}
    \caption{Raw forecasting results of variable Q850.}
    \vspace{-5pt}
    \label{fig:raw_q_850_left}
\end{minipage}
\hfill
\begin{minipage}[t]{0.48\textwidth}
    \centering
    \includegraphics[width=\linewidth]{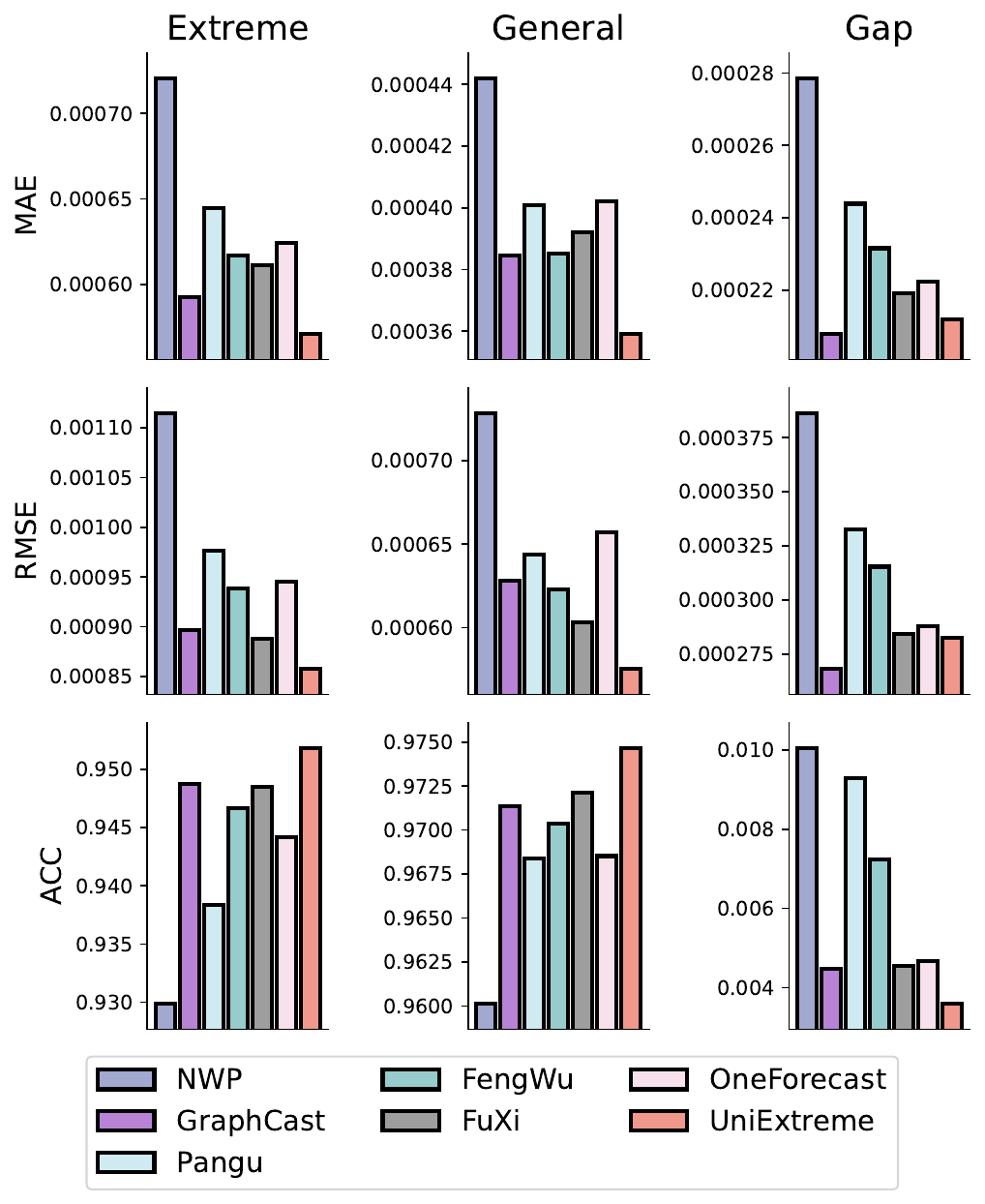}
    \vspace{-25pt}
    \caption{Raw forecasting results of variable Q925.}
    \vspace{-5pt}
    \label{fig:raw_q_925_right}
\end{minipage}
\end{figure*}

\begin{figure*}[!ht]
\begin{minipage}[t]{0.48\textwidth}
    \centering
    \includegraphics[width=\linewidth]{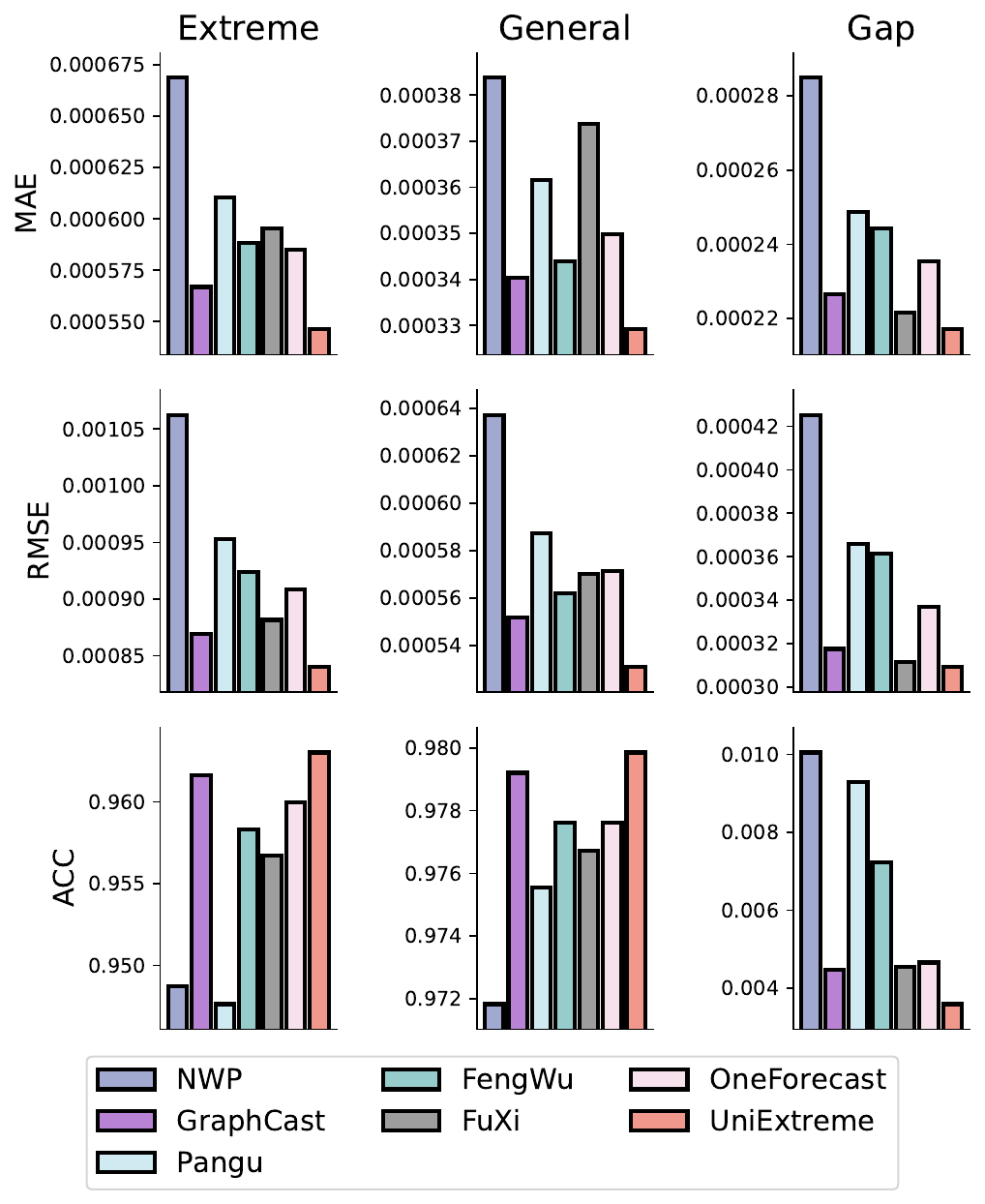}
    \vspace{-25pt}
    \caption{Raw forecasting results of variable Q1000.}
    \vspace{-5pt}
    \label{fig:raw_q_1000_left}
\end{minipage}
\hfill
\begin{minipage}[t]{0.48\textwidth}
    \centering
    \includegraphics[width=\linewidth]{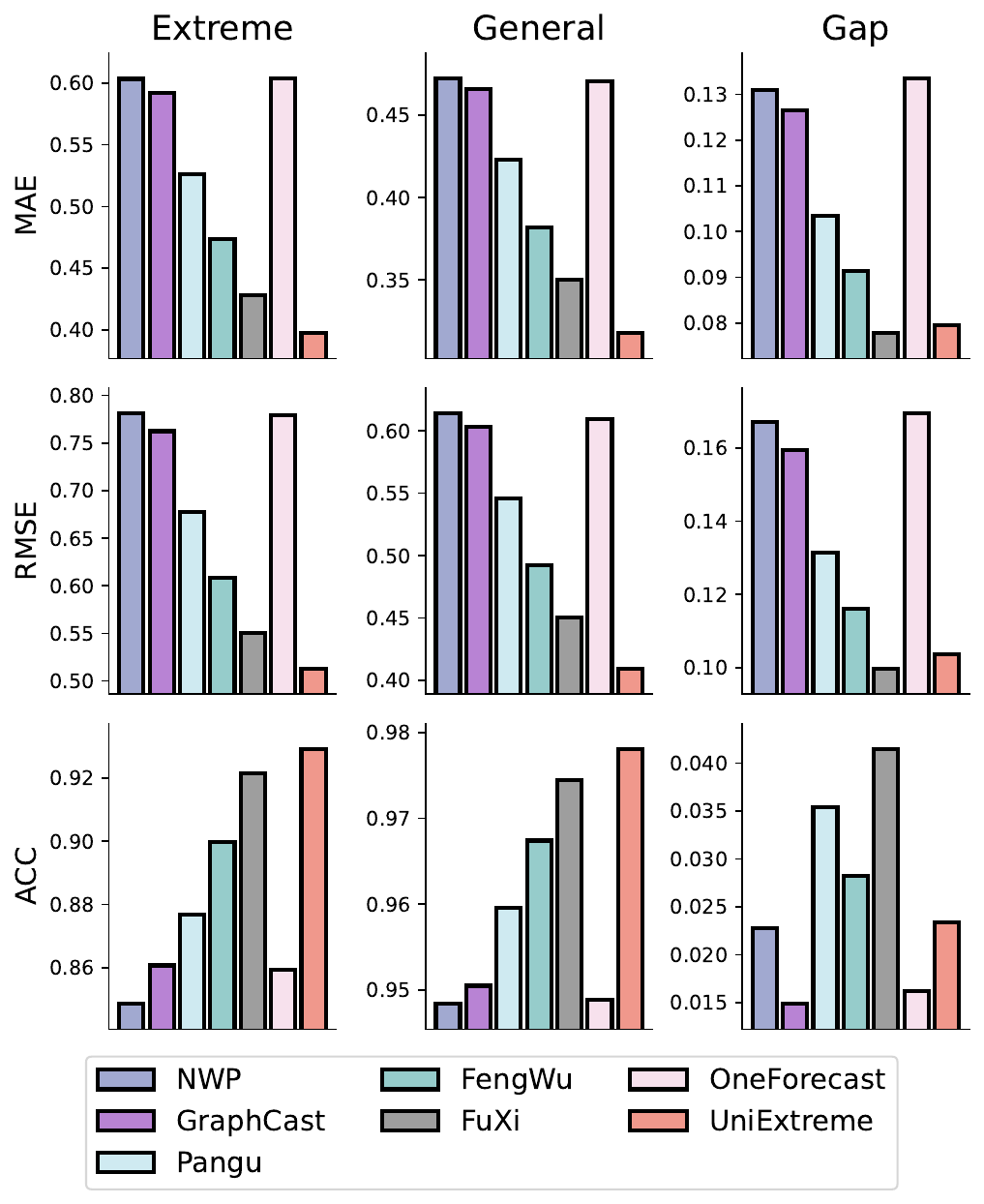}
    \vspace{-25pt}
    \caption{Raw forecasting results of variable T50.}
    \vspace{-5pt}
    \label{fig:raw_t_50_right}
\end{minipage}
\\[10pt]
\begin{minipage}[t]{0.48\textwidth}
    \centering
    \includegraphics[width=\linewidth]{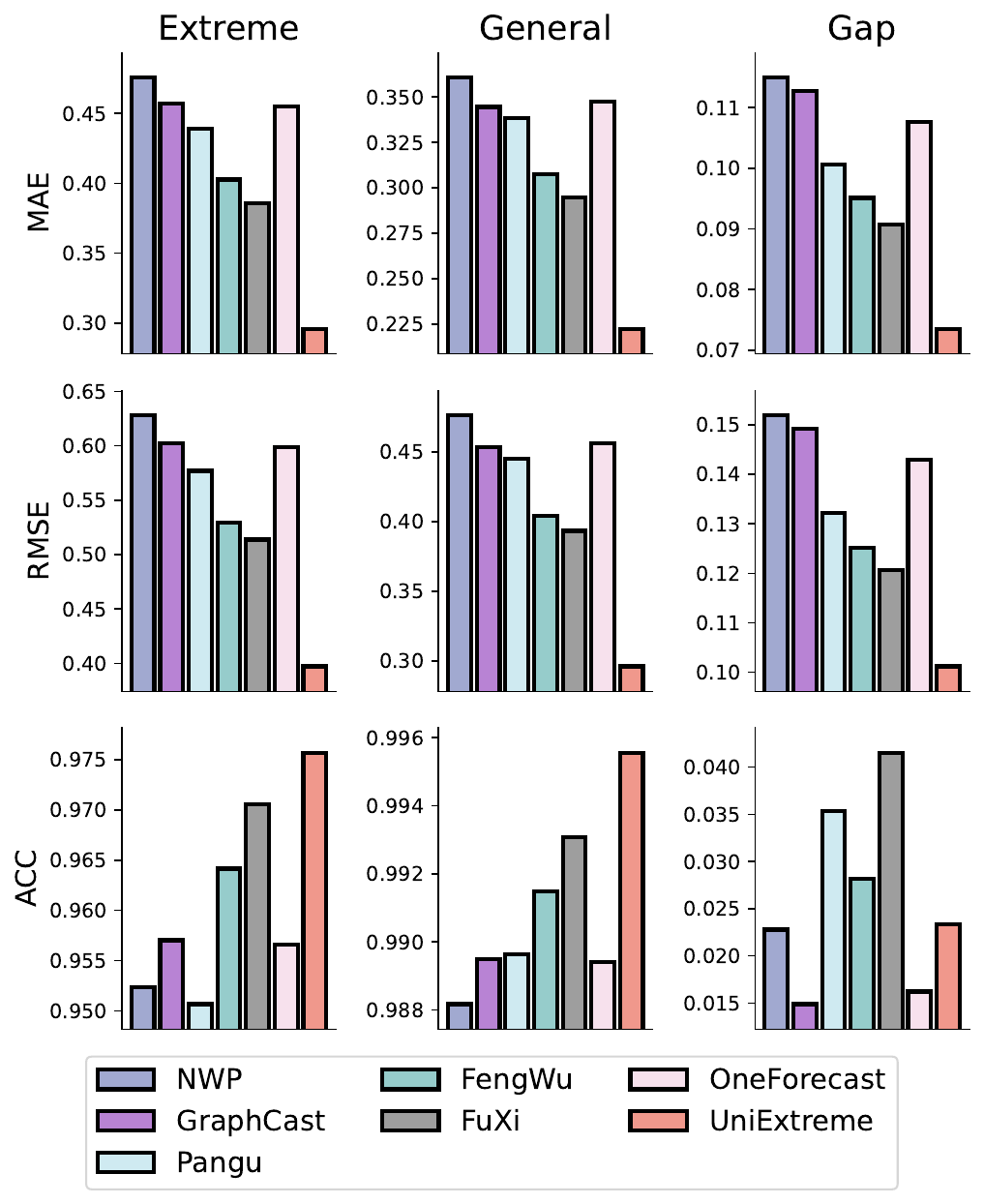}
    \vspace{-25pt}
    \caption{Raw forecasting results of variable T100.}
    \vspace{-5pt}
    \label{fig:raw_t_100_left}
\end{minipage}
\hfill
\begin{minipage}[t]{0.48\textwidth}
    \centering
    \includegraphics[width=\linewidth]{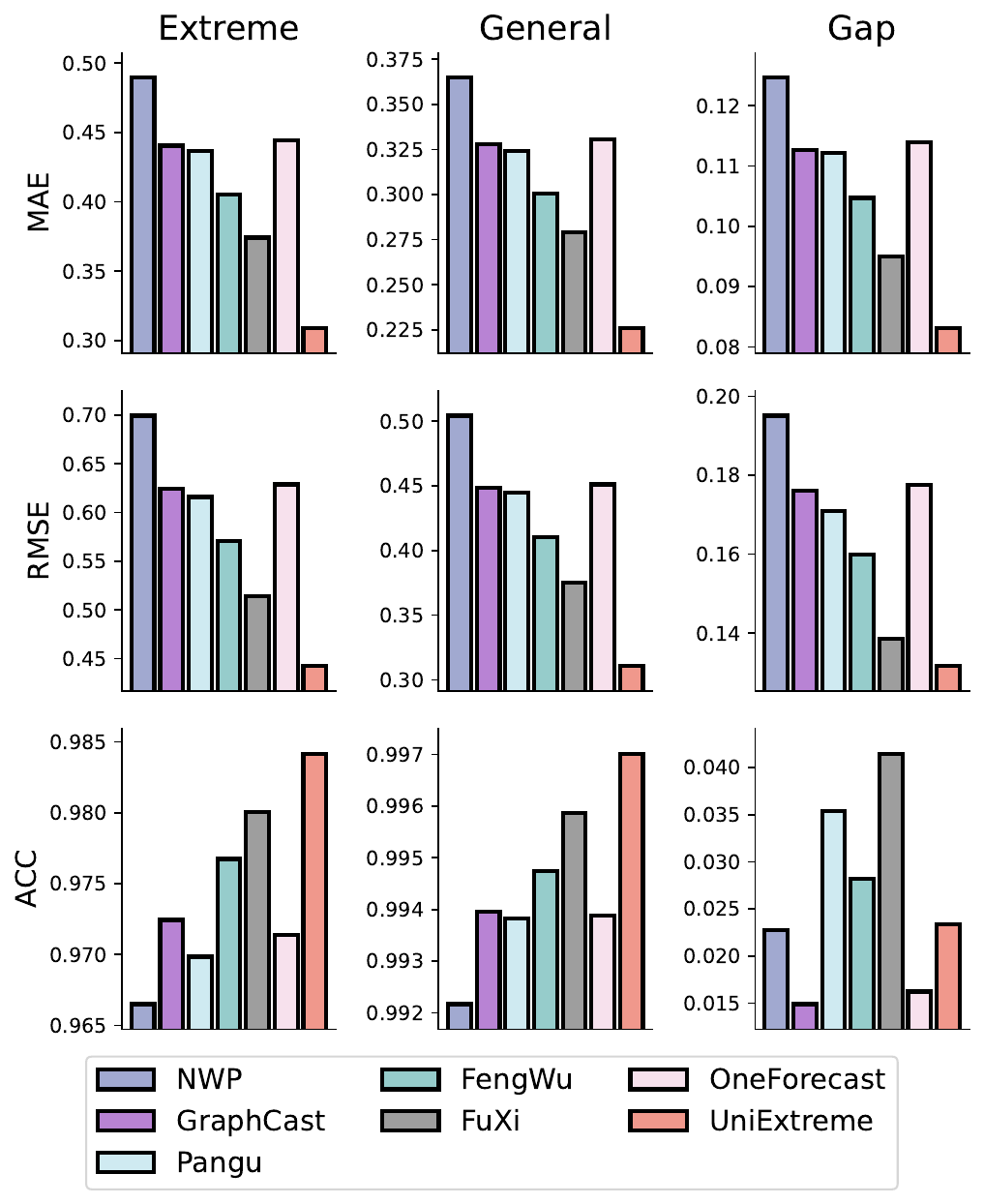}
    \vspace{-25pt}
    \caption{Raw forecasting results of variable T150.}
    \vspace{-5pt}
    \label{fig:raw_t_150_right}
\end{minipage}
\end{figure*}

\begin{figure*}[!ht]
\begin{minipage}[t]{0.48\textwidth}
    \centering
    \includegraphics[width=\linewidth]{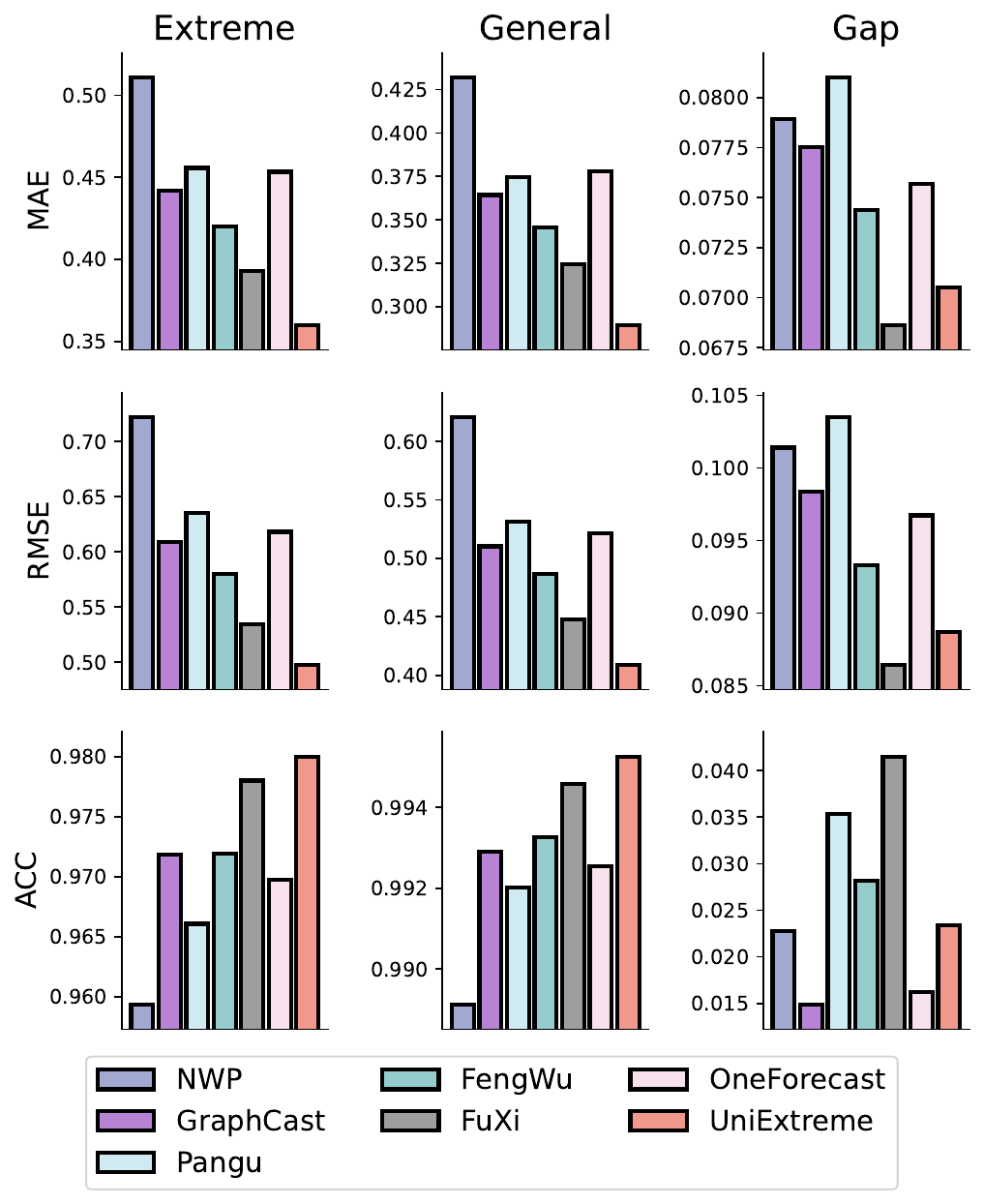}
    \vspace{-25pt}
    \caption{Raw forecasting results of variable T200.}
    \vspace{-5pt}
    \label{fig:raw_t_200_left}
\end{minipage}
\hfill
\begin{minipage}[t]{0.48\textwidth}
    \centering
    \includegraphics[width=\linewidth]{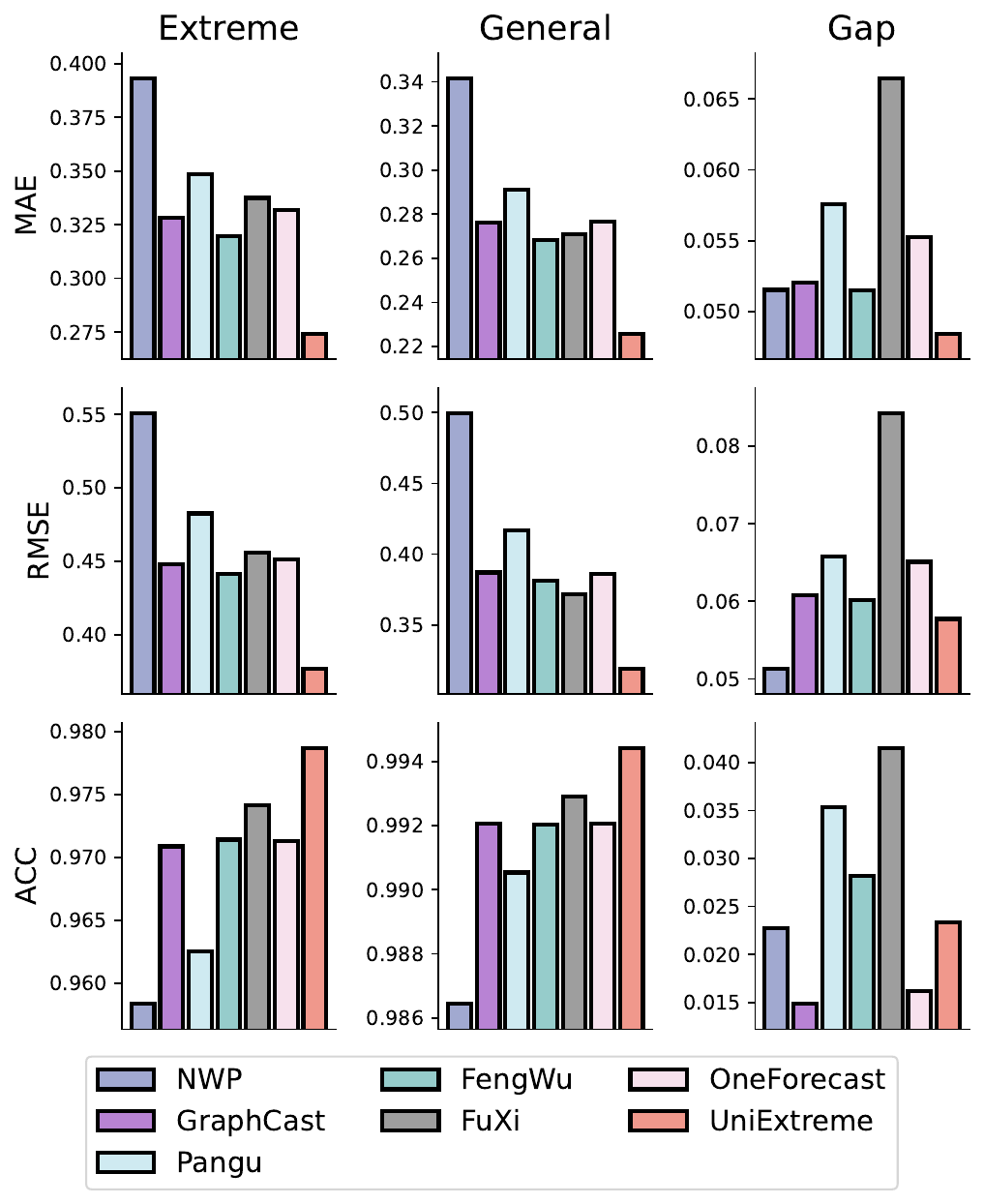}
    \vspace{-25pt}
    \caption{Raw forecasting results of variable T250.}
    \vspace{-5pt}
    \label{fig:raw_t_250_right}
\end{minipage}
\\[10pt]
\begin{minipage}[t]{0.48\textwidth}
    \centering
    \includegraphics[width=\linewidth]{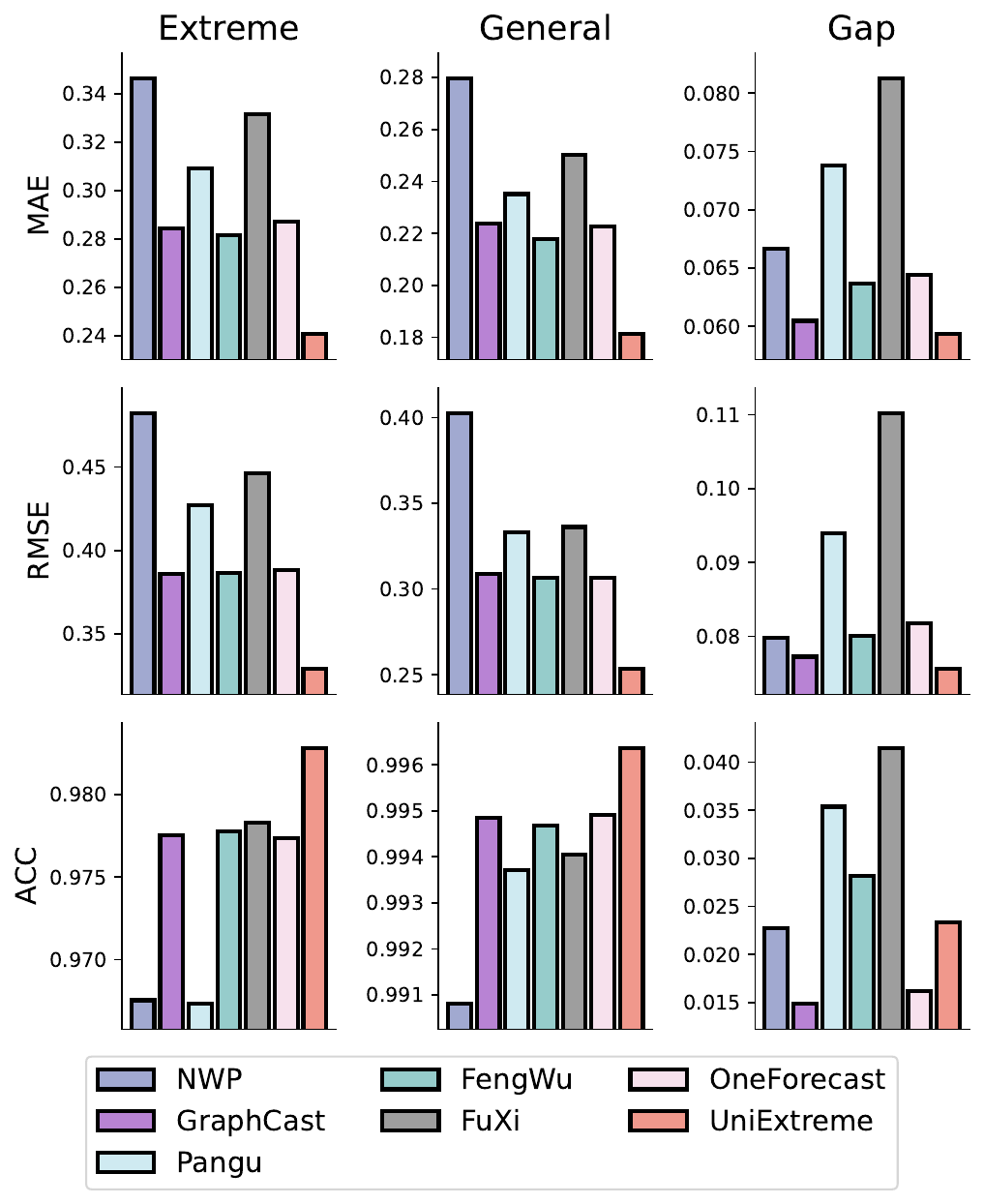}
    \vspace{-25pt}
    \caption{Raw forecasting results of variable T300.}
    \vspace{-5pt}
    \label{fig:raw_t_300_left}
\end{minipage}
\hfill
\begin{minipage}[t]{0.48\textwidth}
    \centering
    \includegraphics[width=\linewidth]{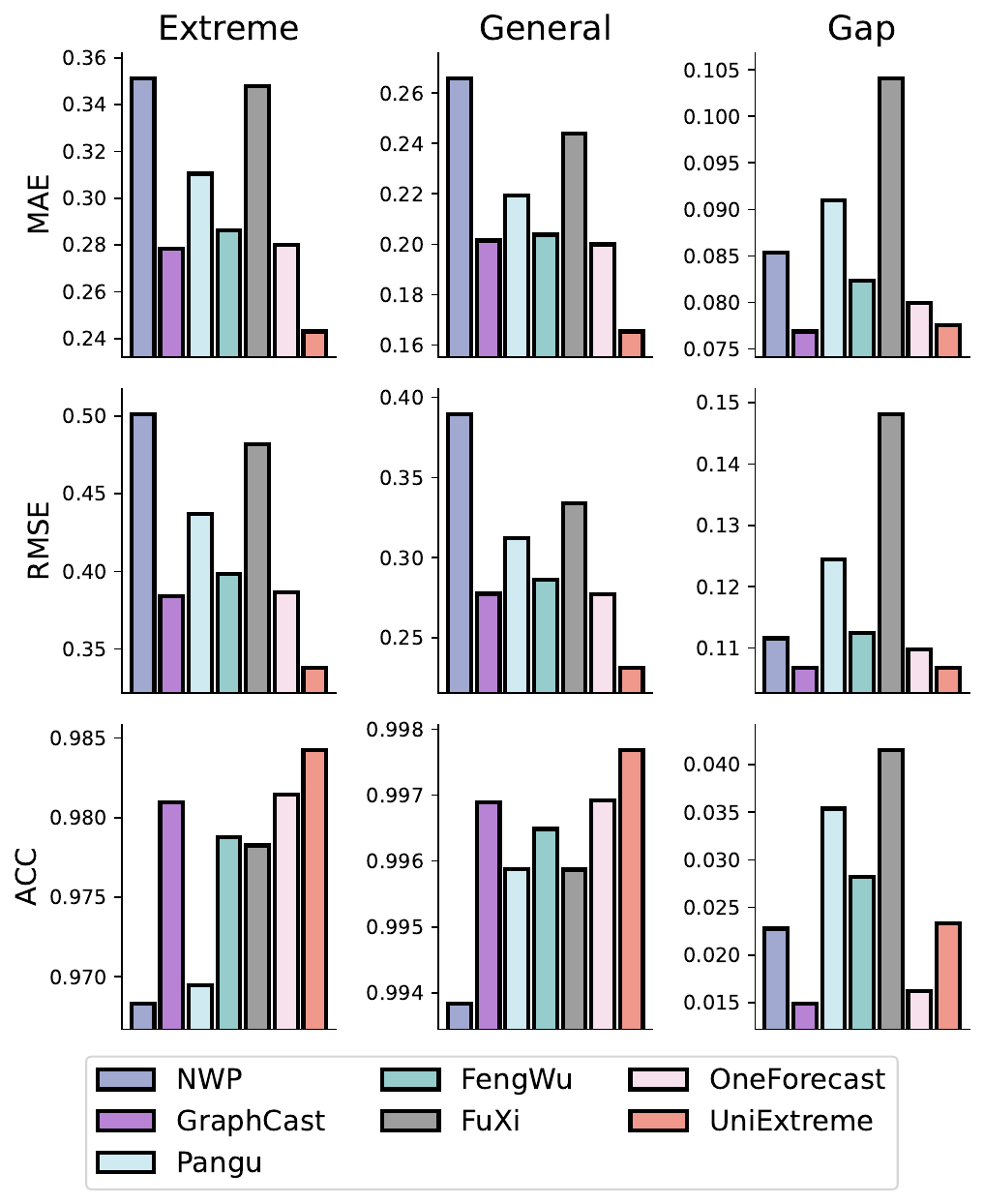}
    \vspace{-25pt}
    \caption{Raw forecasting results of variable T400.}
    \vspace{-5pt}
    \label{fig:raw_t_400_right}
\end{minipage}
\end{figure*}

\begin{figure*}[!ht]
\begin{minipage}[t]{0.48\textwidth}
    \centering
    \includegraphics[width=\linewidth]{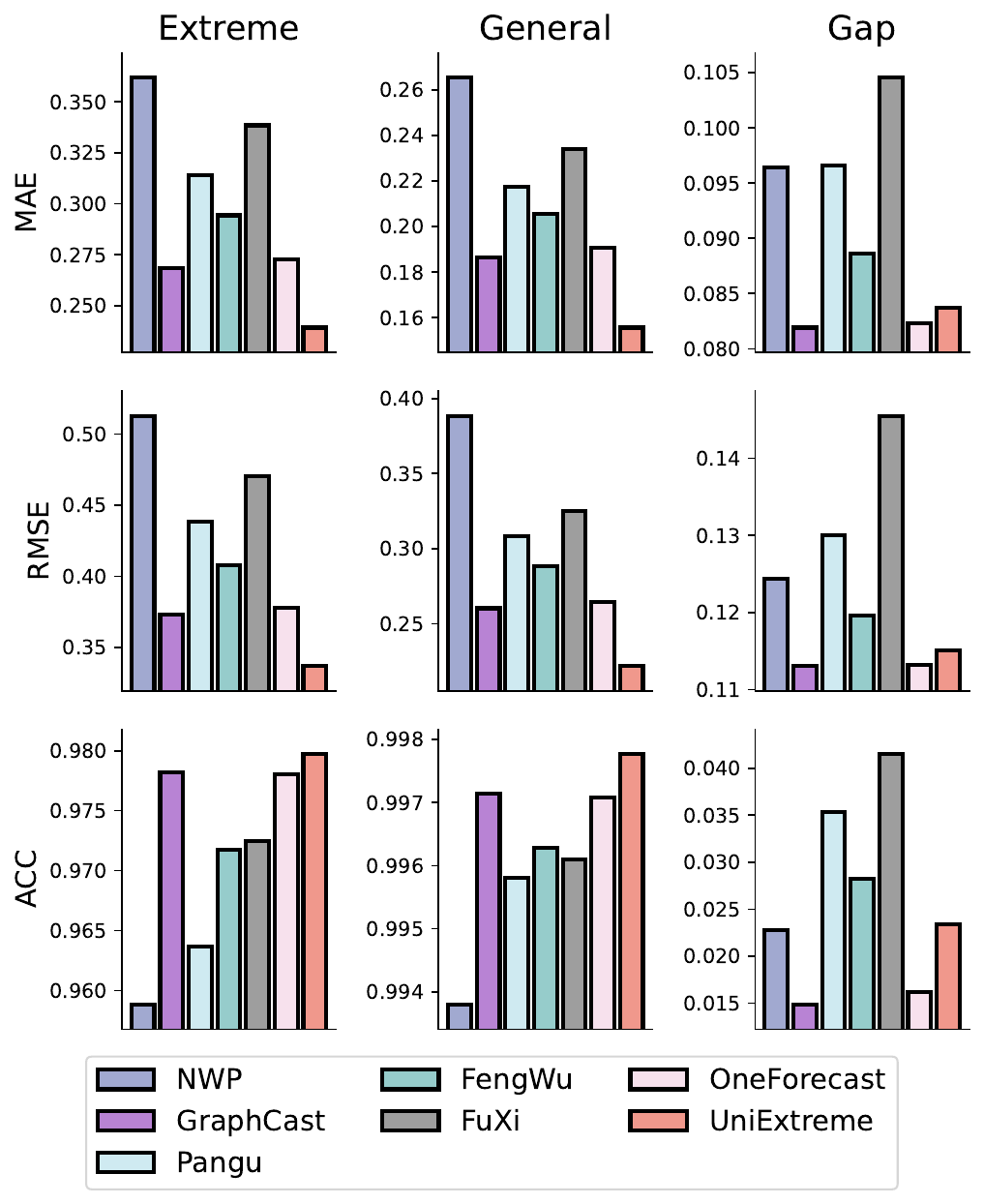}
    \vspace{-25pt}
    \caption{Raw forecasting results of variable T500.}
    \vspace{-5pt}
    \label{fig:raw_t_500_left}
\end{minipage}
\hfill
\begin{minipage}[t]{0.48\textwidth}
    \centering
    \includegraphics[width=\linewidth]{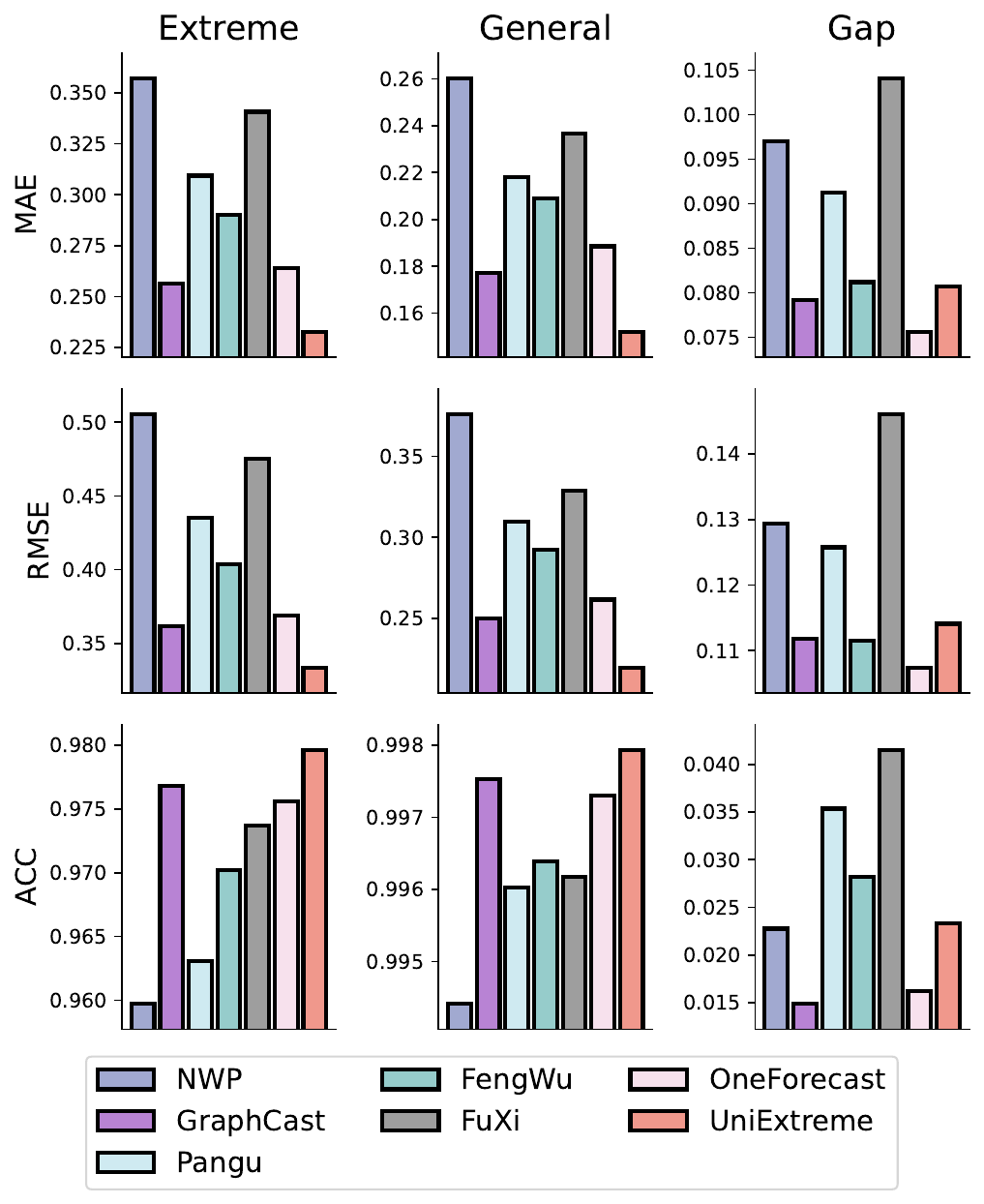}
    \vspace{-25pt}
    \caption{Raw forecasting results of variable T600.}
    \vspace{-5pt}
    \label{fig:raw_t_600_right}
\end{minipage}
\\[10pt]
\begin{minipage}[t]{0.48\textwidth}
    \centering
    \includegraphics[width=\linewidth]{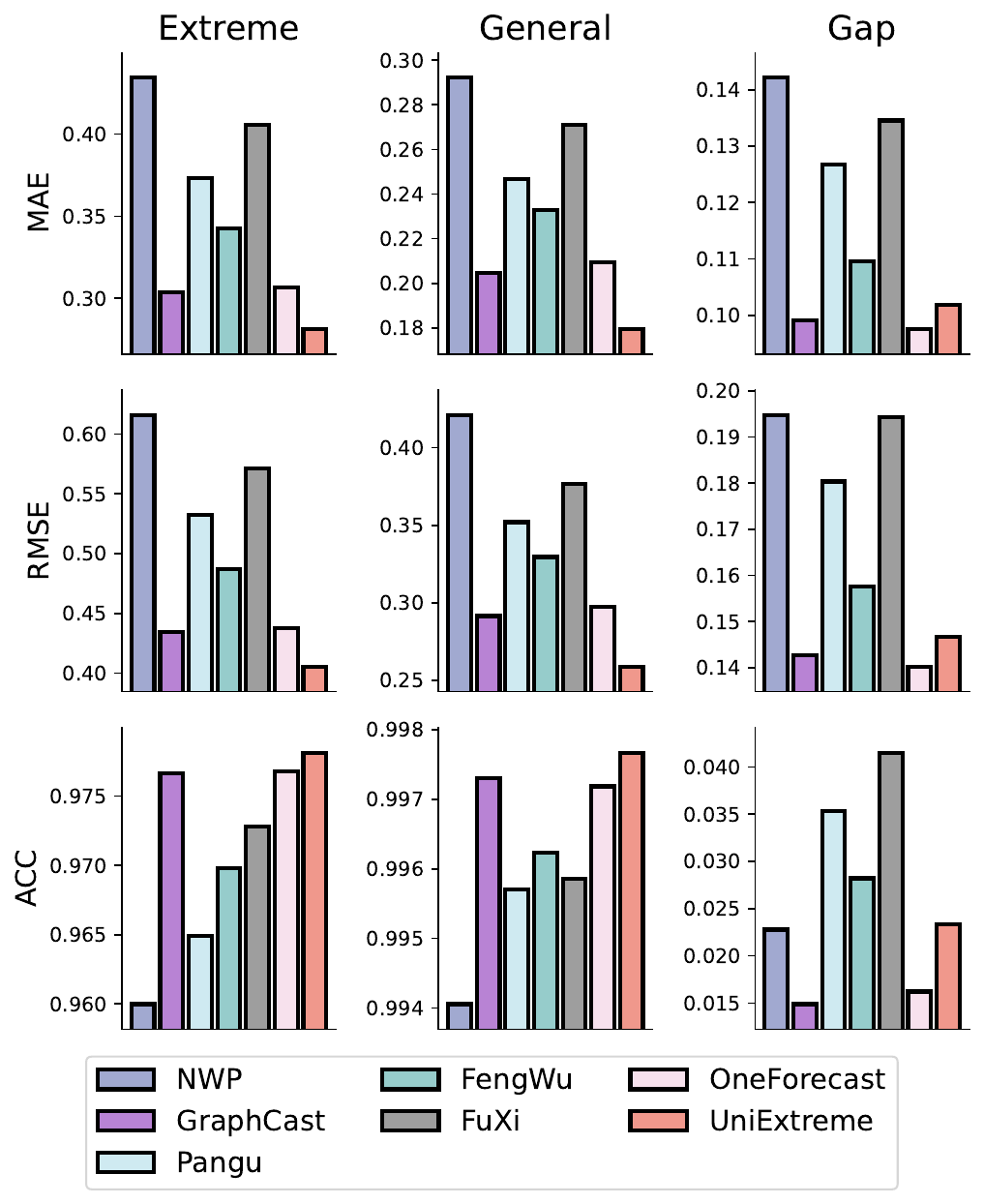}
    \vspace{-25pt}
    \caption{Raw forecasting results of variable T700.}
    \vspace{-5pt}
    \label{fig:raw_t_700_left}
\end{minipage}
\hfill
\begin{minipage}[t]{0.48\textwidth}
    \centering
    \includegraphics[width=\linewidth]{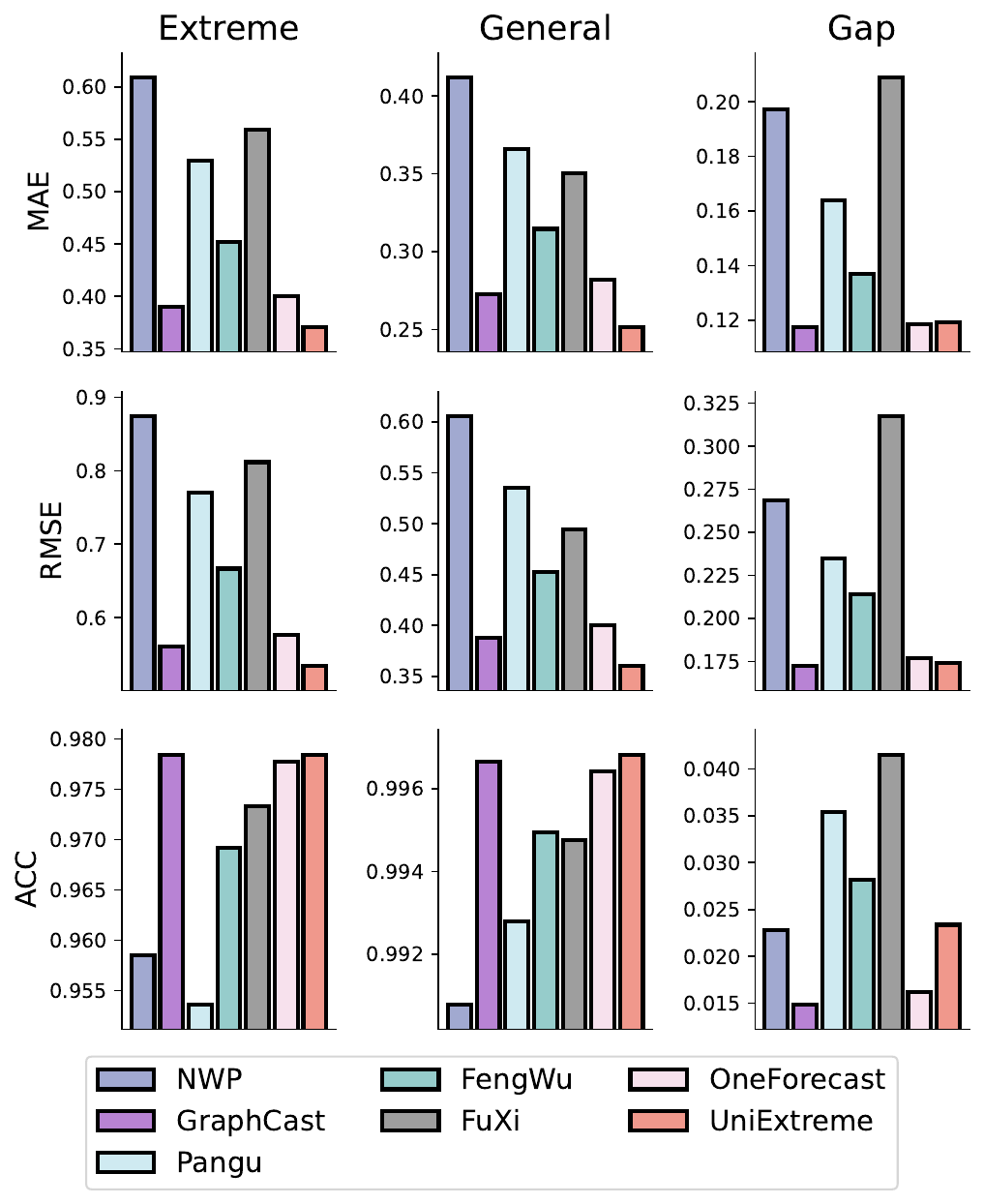}
    \vspace{-25pt}
    \caption{Raw forecasting results of variable T850.}
    \vspace{-5pt}
    \label{fig:raw_t_850_right}
\end{minipage}
\end{figure*}

\begin{figure*}[!ht]
\begin{minipage}[t]{0.48\textwidth}
    \centering
    \includegraphics[width=\linewidth]{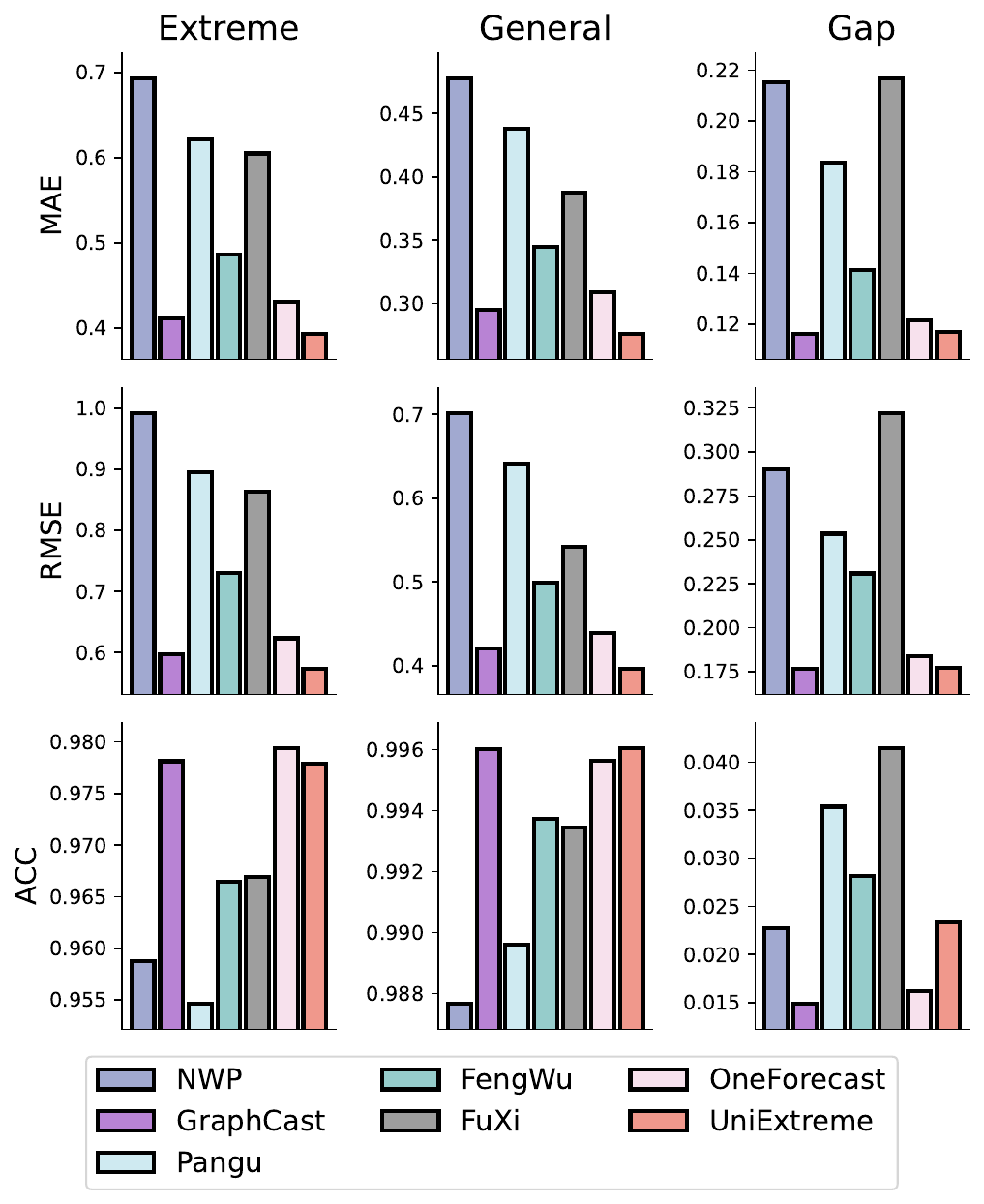}
    \vspace{-25pt}
    \caption{Raw forecasting results of variable T925.}
    \vspace{-5pt}
    \label{fig:raw_t_925_left}
\end{minipage}
\hfill
\begin{minipage}[t]{0.48\textwidth}
    \centering
    \includegraphics[width=\linewidth]{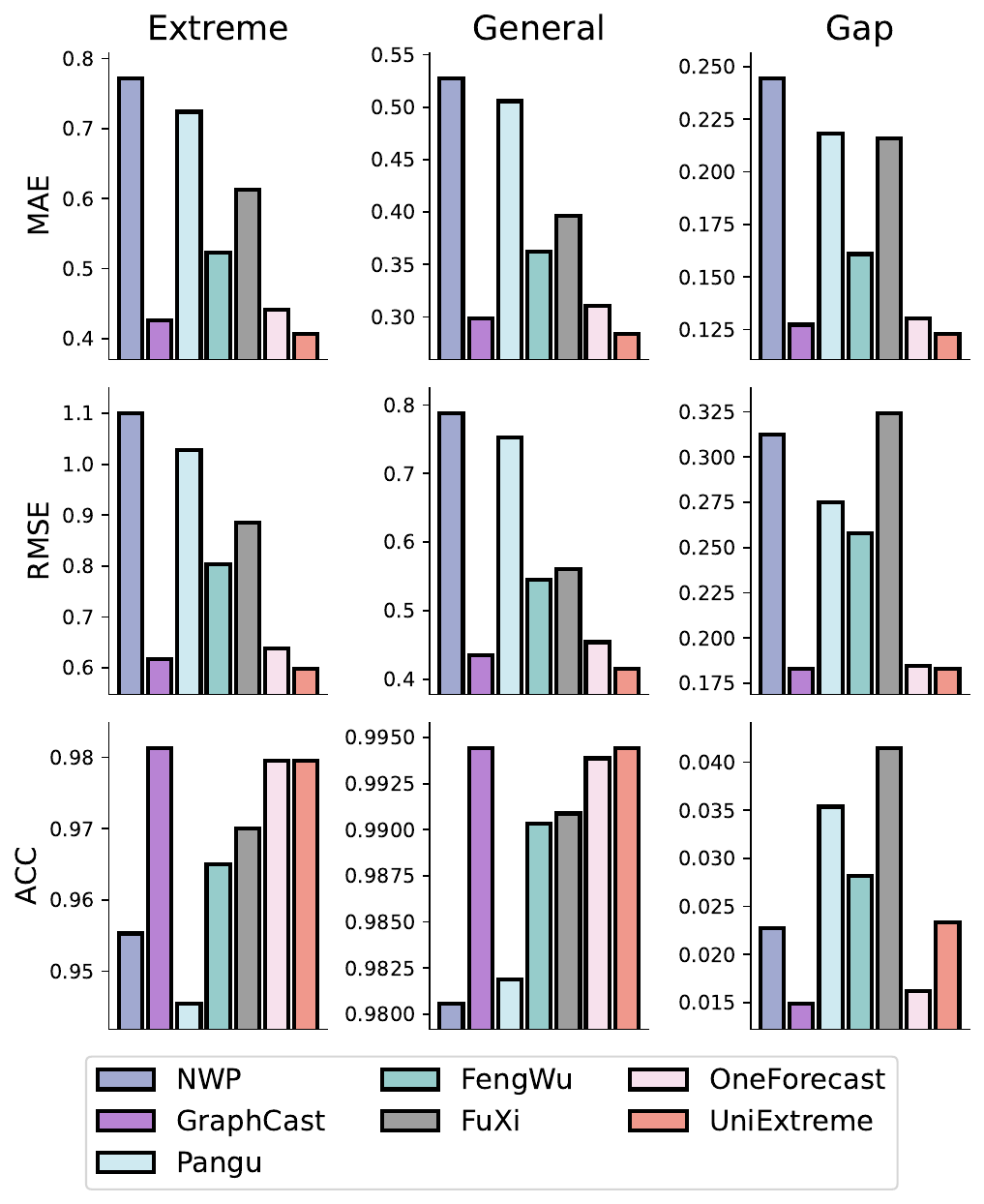}
    \vspace{-25pt}
    \caption{Raw forecasting results of variable T1000.}
    \vspace{-5pt}
    \label{fig:raw_t_1000_right}
\end{minipage}
\\[10pt]
\begin{minipage}[t]{0.48\textwidth}
    \centering
    \includegraphics[width=\linewidth]{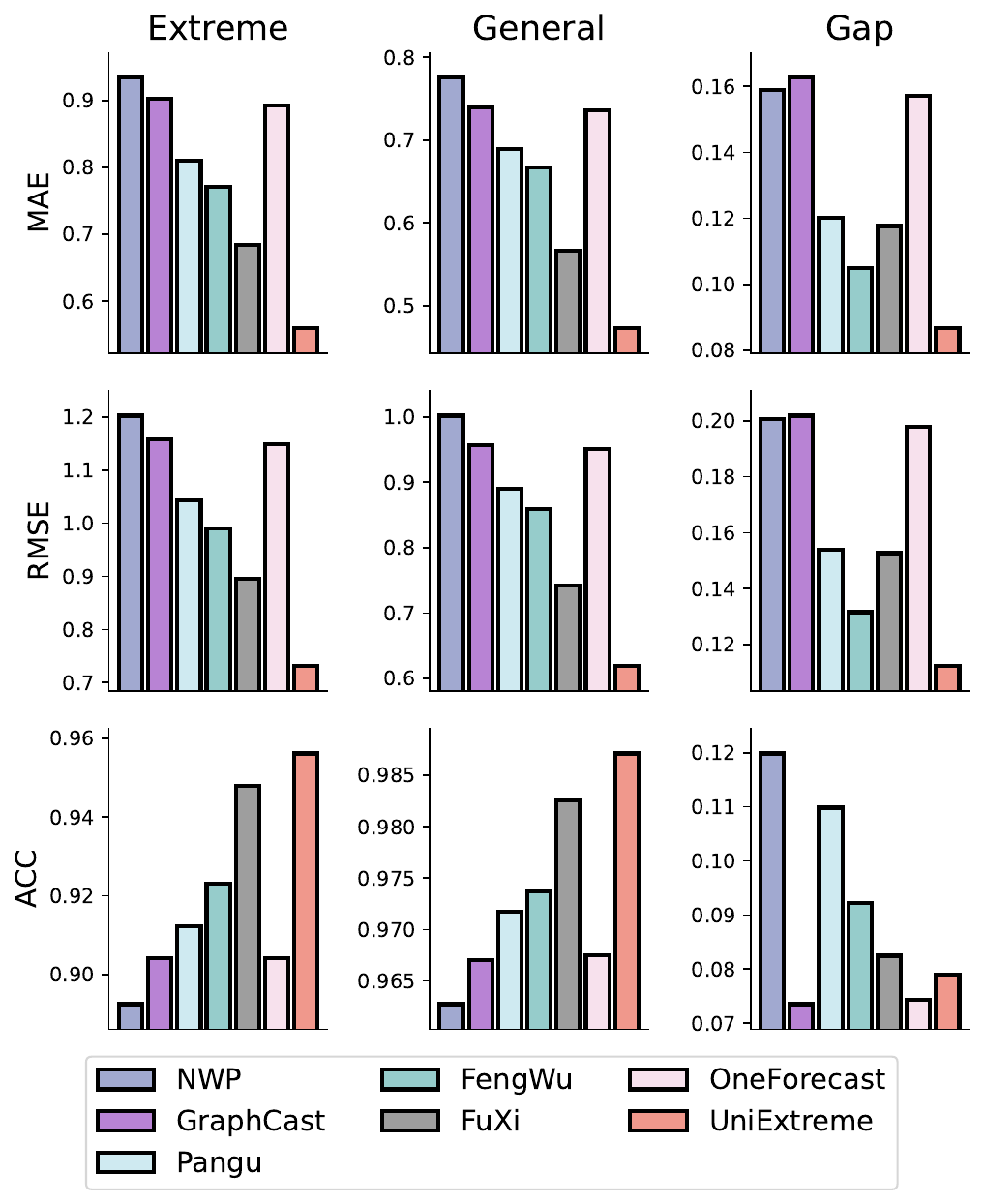}
    \vspace{-25pt}
    \caption{Raw forecasting results of variable U50.}
    \vspace{-5pt}
    \label{fig:raw_u_50_left}
\end{minipage}
\hfill
\begin{minipage}[t]{0.48\textwidth}
    \centering
    \includegraphics[width=\linewidth]{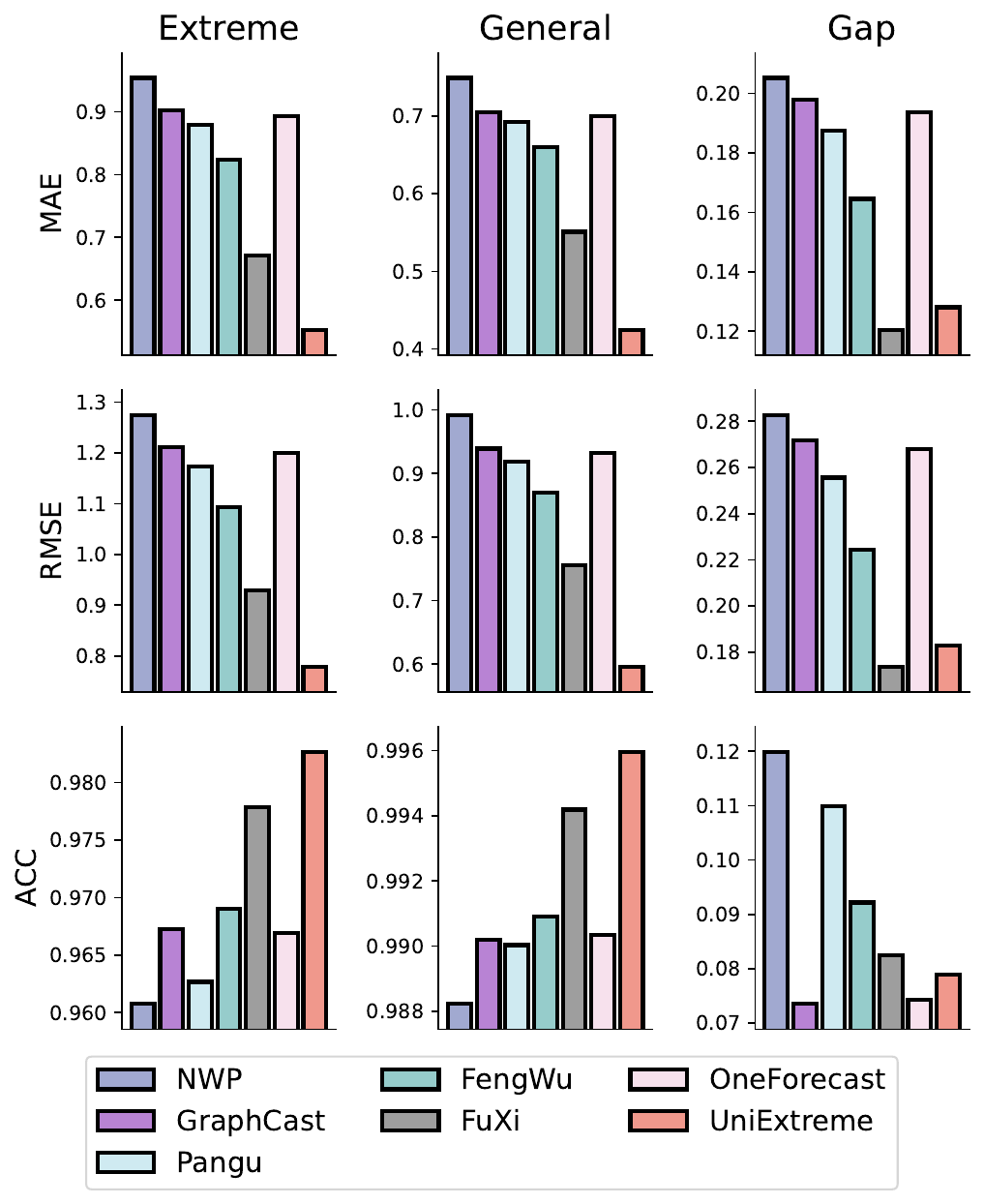}
    \vspace{-25pt}
    \caption{Raw forecasting results of variable U100.}
    \vspace{-5pt}
    \label{fig:raw_u_100_right}
\end{minipage}
\end{figure*}

\begin{figure*}[!ht]
\begin{minipage}[t]{0.48\textwidth}
    \centering
    \includegraphics[width=\linewidth]{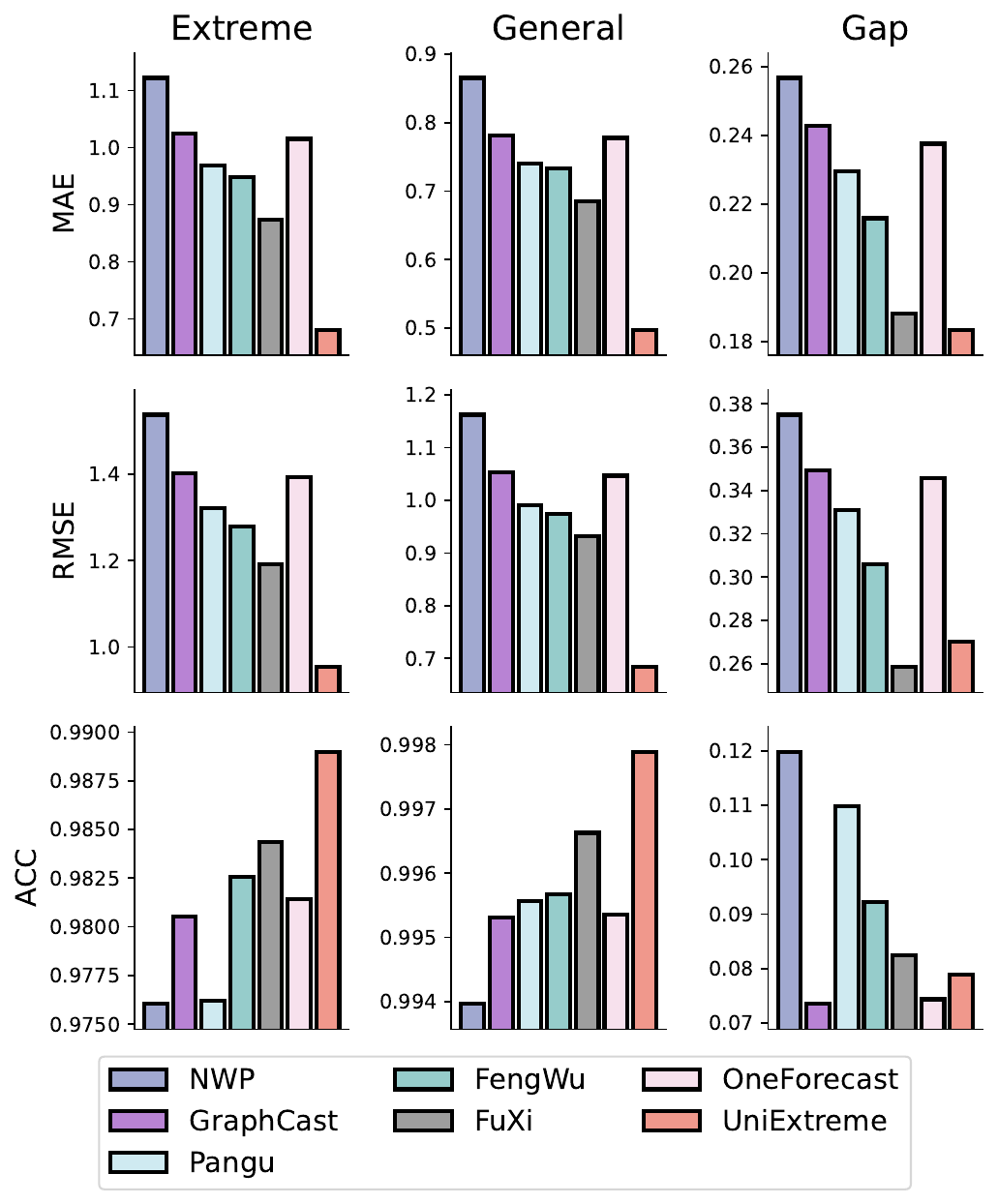}
    \vspace{-25pt}
    \caption{Raw forecasting results of variable U150.}
    \vspace{-5pt}
    \label{fig:raw_u_150_left}
\end{minipage}
\hfill
\begin{minipage}[t]{0.48\textwidth}
    \centering
    \includegraphics[width=\linewidth]{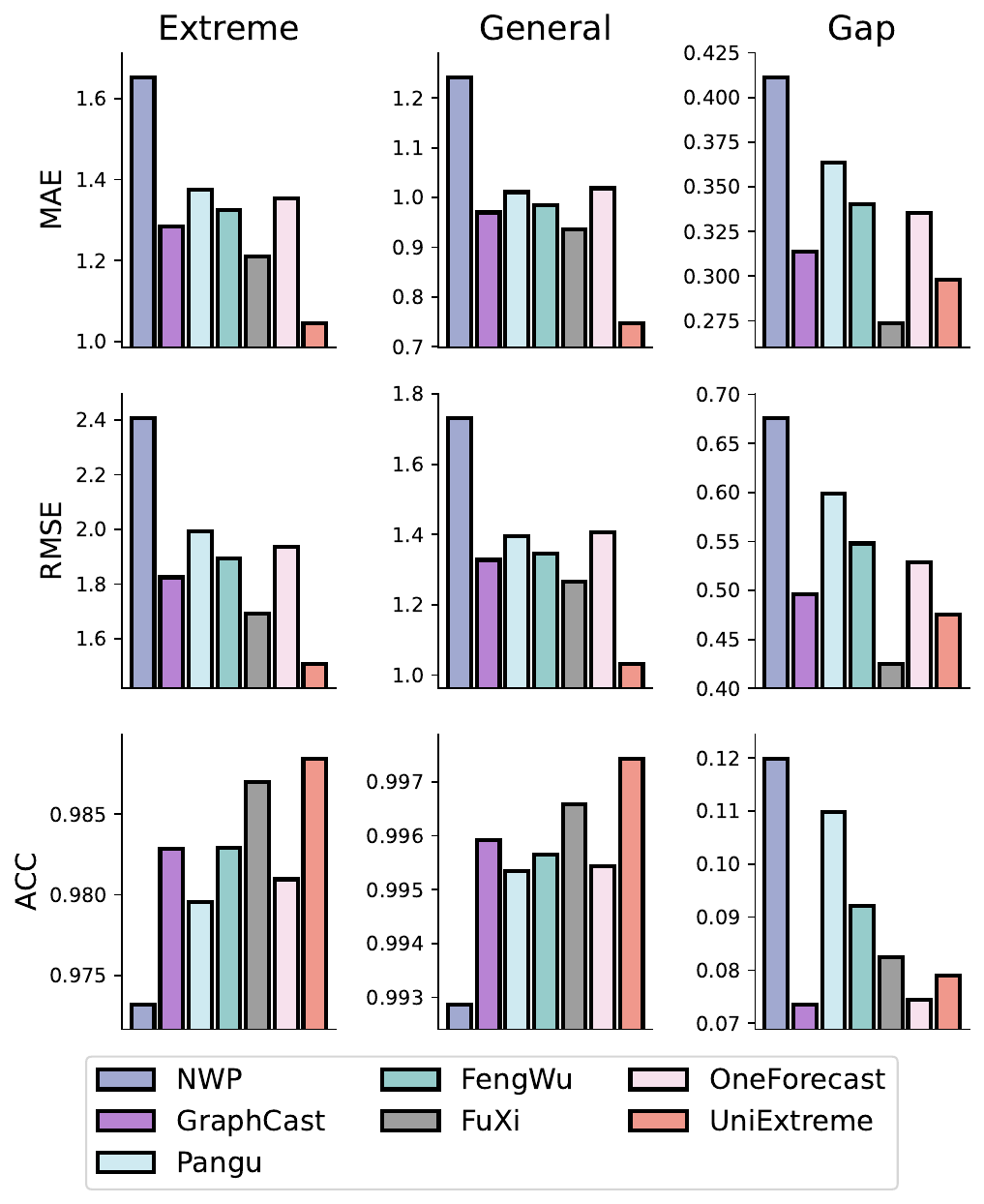}
    \vspace{-25pt}
    \caption{Raw forecasting results of variable U200.}
    \vspace{-5pt}
    \label{fig:raw_u_200_right}
\end{minipage}
\\[10pt]
\begin{minipage}[t]{0.48\textwidth}
    \centering
    \includegraphics[width=\linewidth]{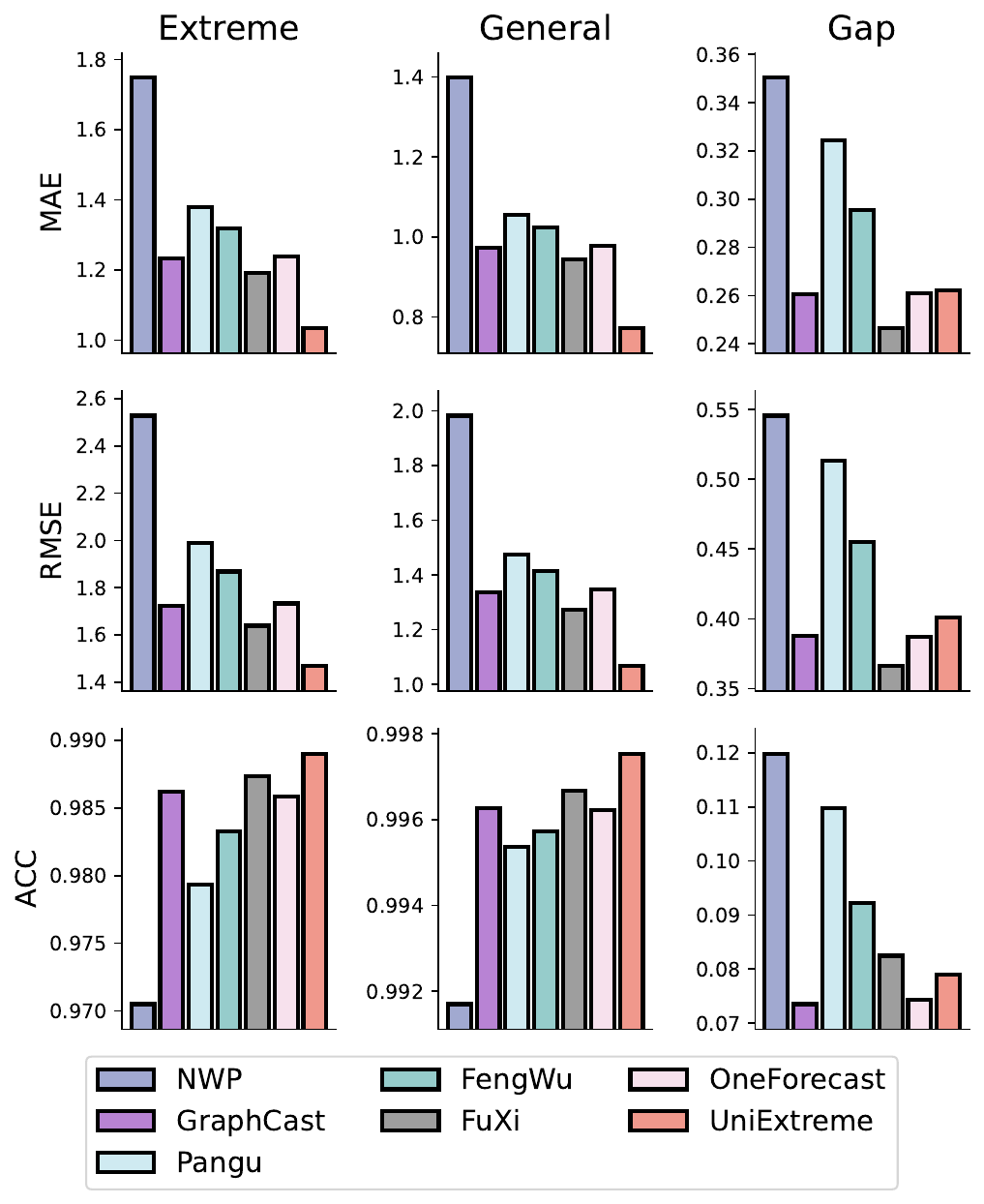}
    \vspace{-25pt}
    \caption{Raw forecasting results of variable U250.}
    \vspace{-5pt}
    \label{fig:raw_u_250_left}
\end{minipage}
\hfill
\begin{minipage}[t]{0.48\textwidth}
    \centering
    \includegraphics[width=\linewidth]{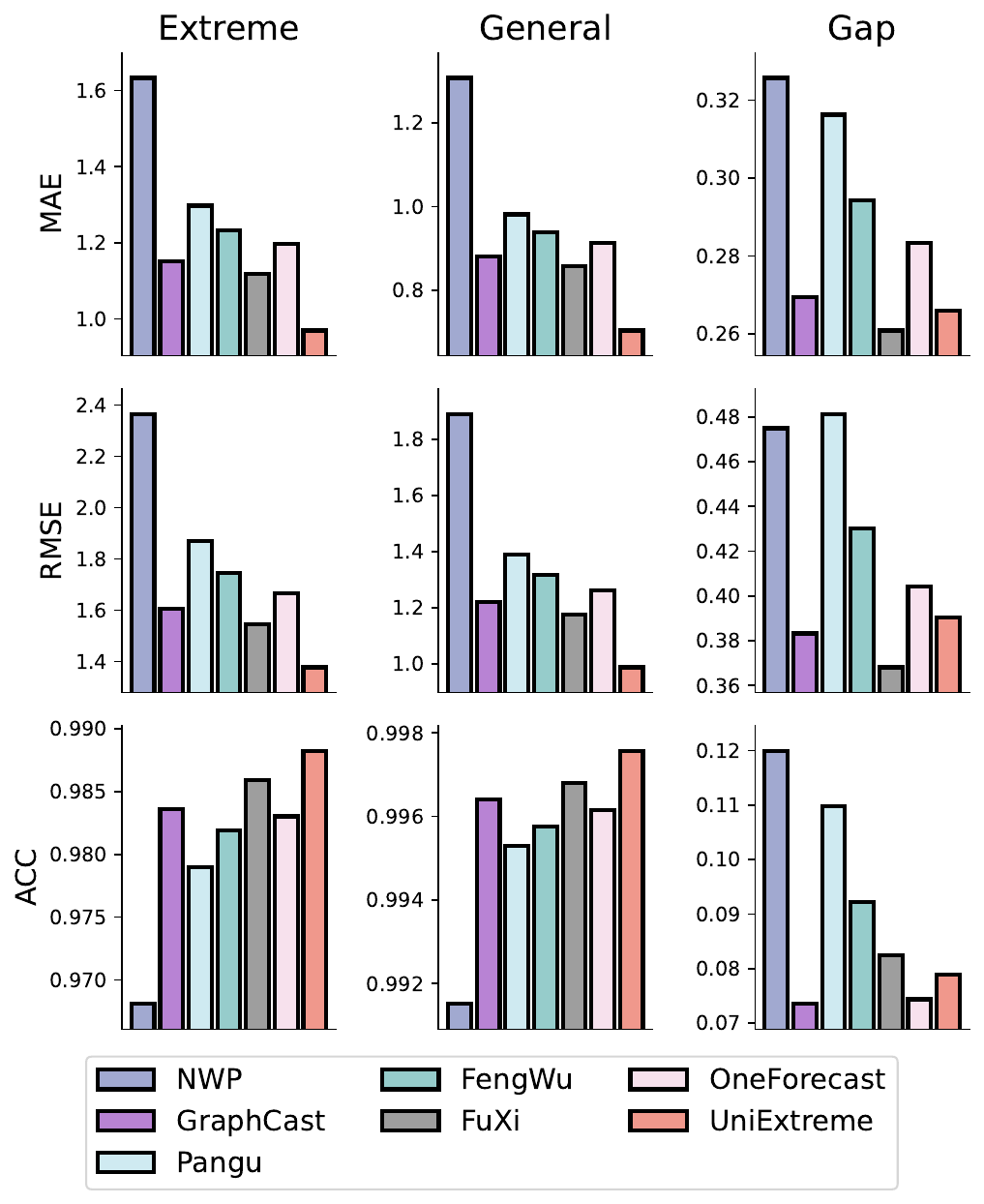}
    \vspace{-25pt}
    \caption{Raw forecasting results of variable U300.}
    \vspace{-5pt}
    \label{fig:raw_u_300_right}
\end{minipage}
\end{figure*}

\begin{figure*}[!ht]
\begin{minipage}[t]{0.48\textwidth}
    \centering
    \includegraphics[width=\linewidth]{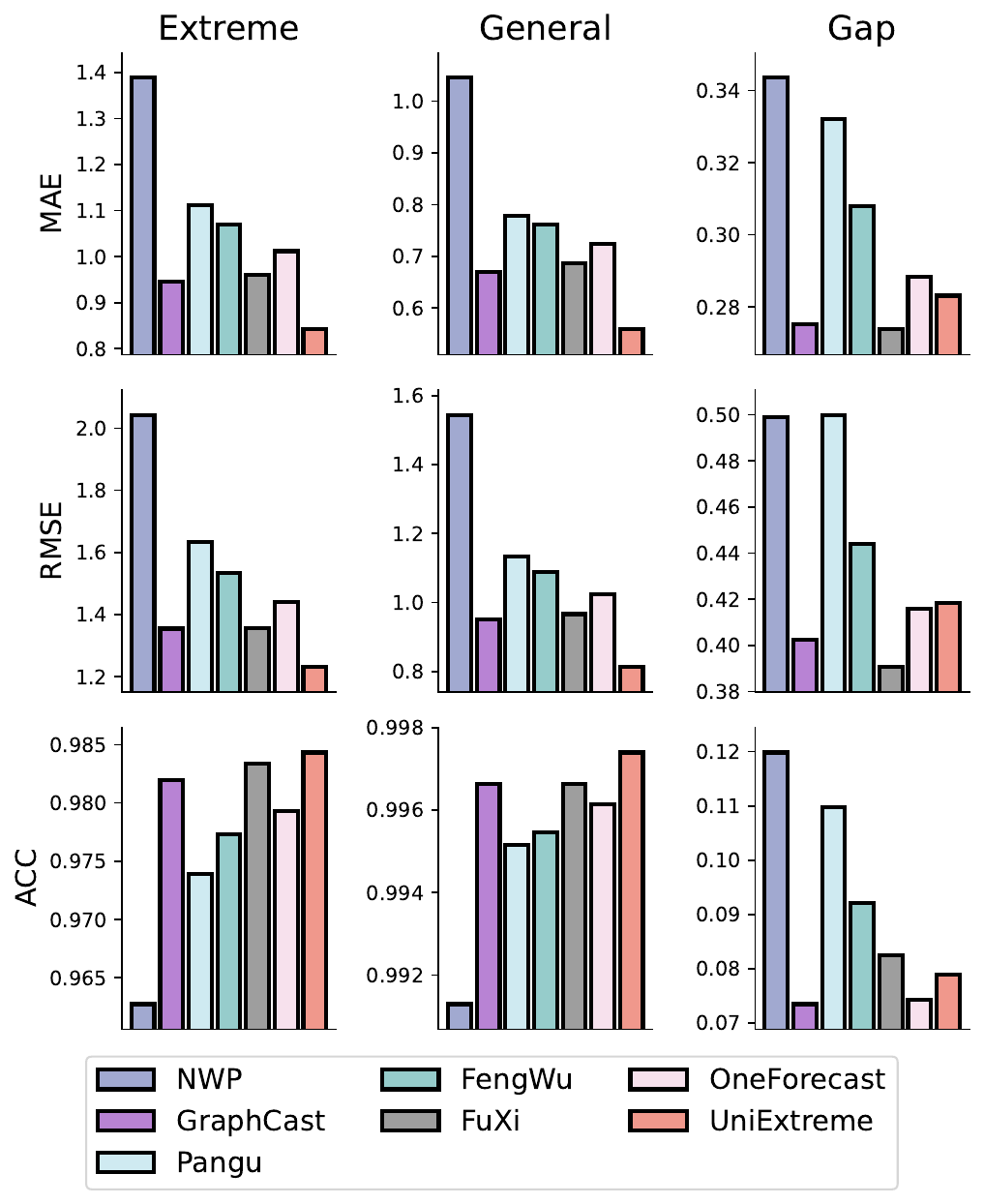}
    \vspace{-25pt}
    \caption{Raw forecasting results of variable U400.}
    \vspace{-5pt}
    \label{fig:raw_u_400_left}
\end{minipage}
\hfill
\begin{minipage}[t]{0.48\textwidth}
    \centering
    \includegraphics[width=\linewidth]{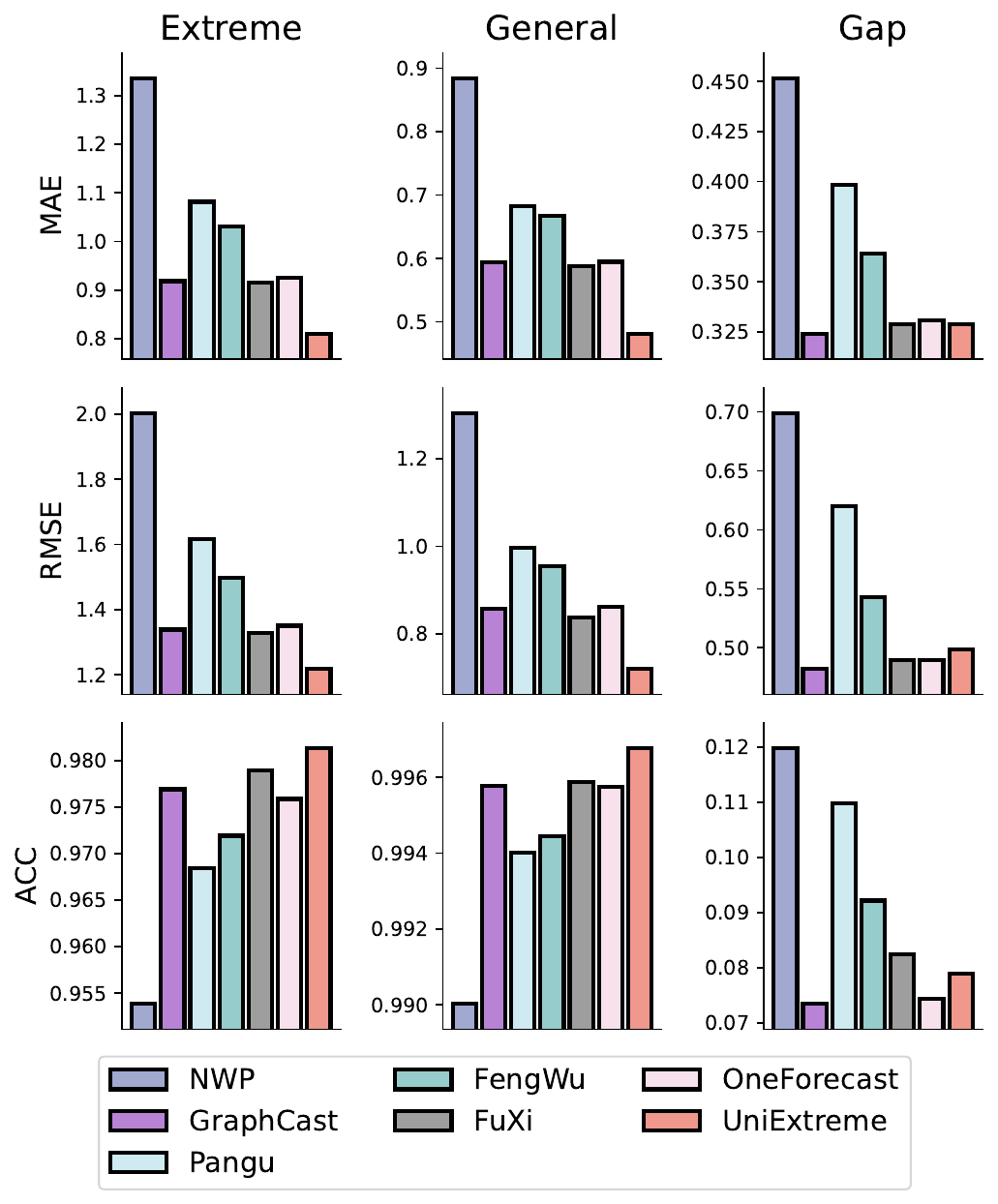}
    \vspace{-25pt}
    \caption{Raw forecasting results of variable U500.}
    \vspace{-5pt}
    \label{fig:raw_u_500_right}
\end{minipage}
\\[10pt]
\begin{minipage}[t]{0.48\textwidth}
    \centering
    \includegraphics[width=\linewidth]{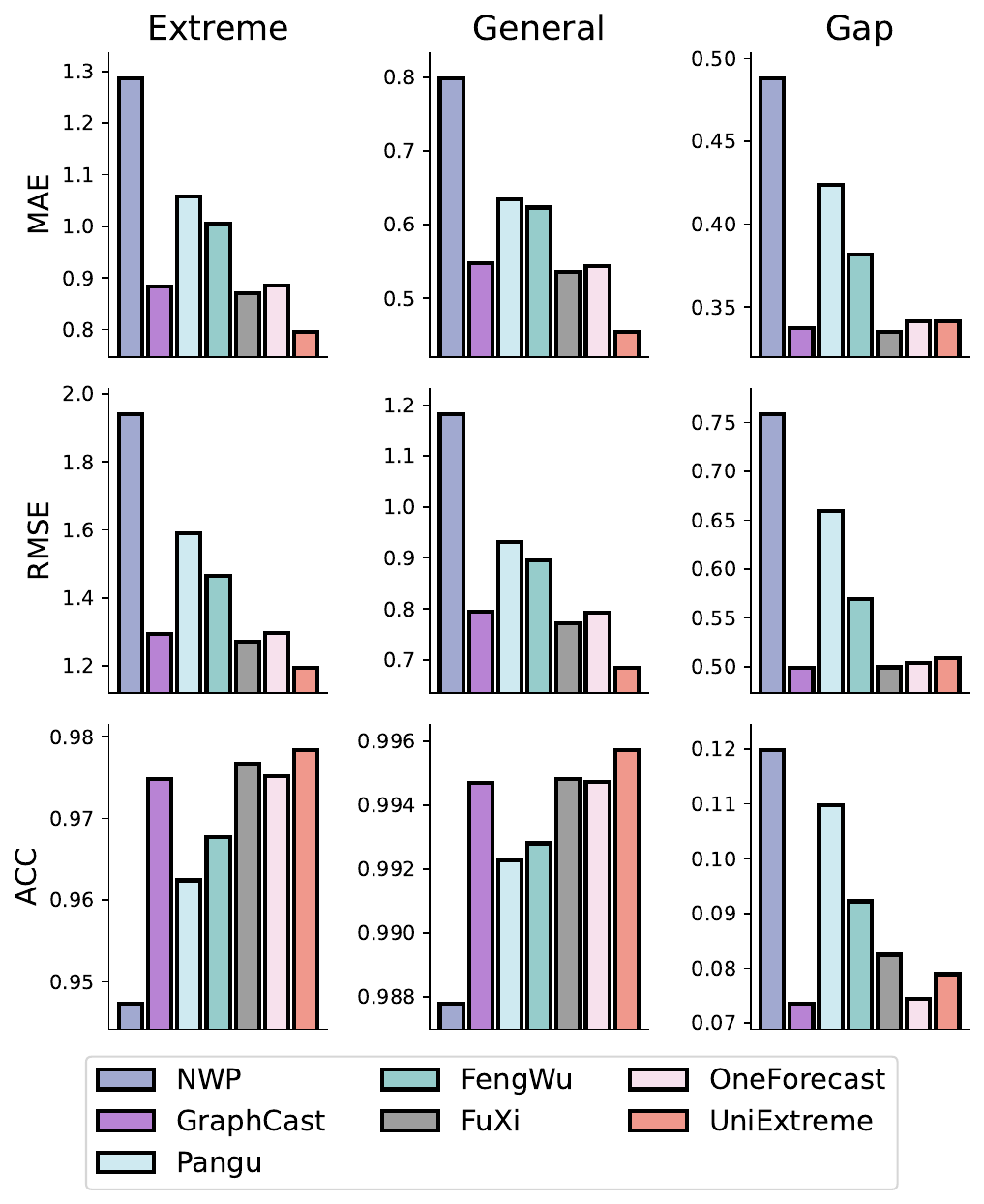}
    \vspace{-25pt}
    \caption{Raw forecasting results of variable U600.}
    \vspace{-5pt}
    \label{fig:raw_u_600_left}
\end{minipage}
\hfill
\begin{minipage}[t]{0.48\textwidth}
    \centering
    \includegraphics[width=\linewidth]{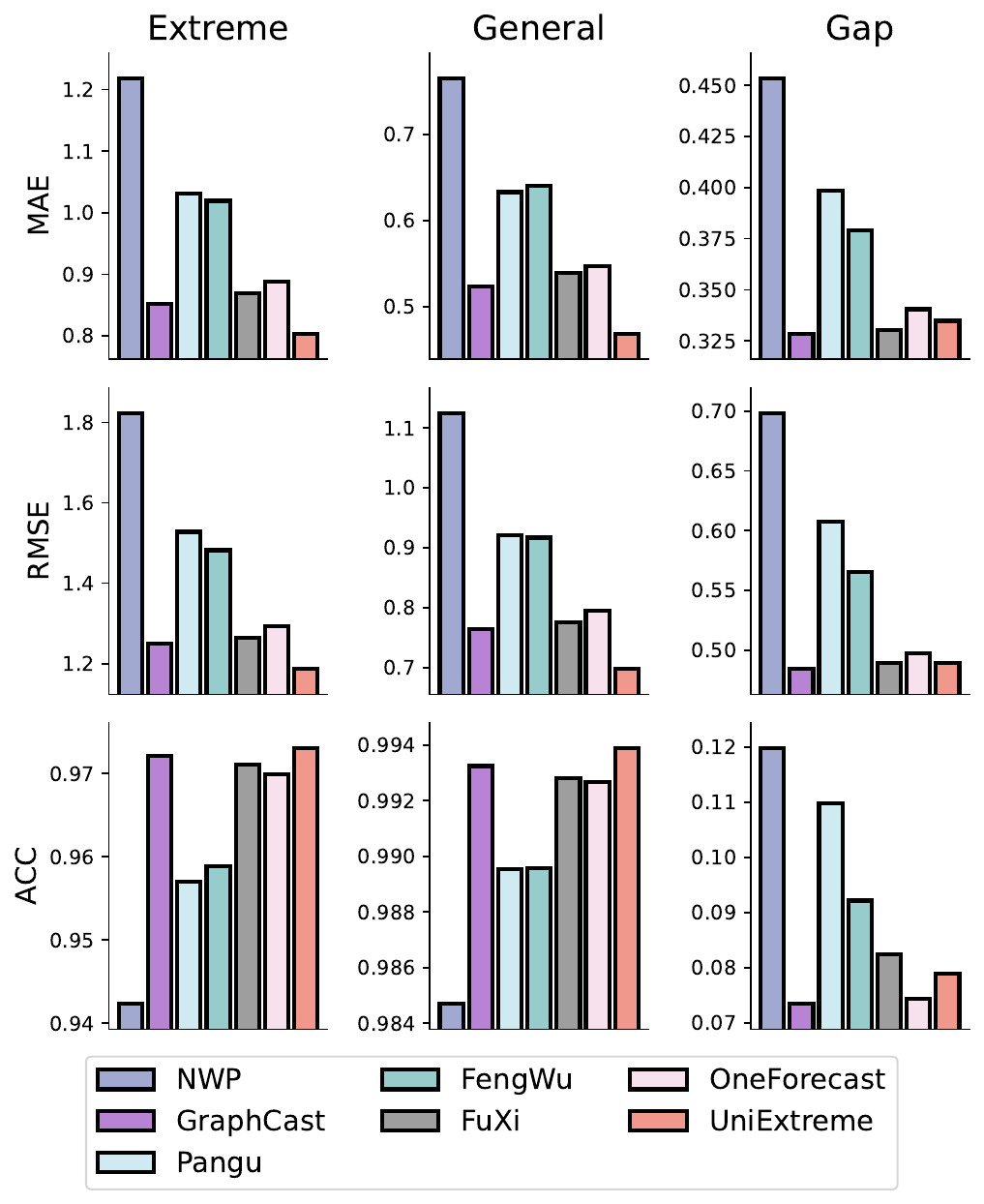}
    \vspace{-25pt}
    \caption{Raw forecasting results of variable U700.}
    \vspace{-5pt}
    \label{fig:raw_u_700_right}
\end{minipage}
\end{figure*}

\begin{figure*}[!ht]
\begin{minipage}[t]{0.48\textwidth}
    \centering
    \includegraphics[width=\linewidth]{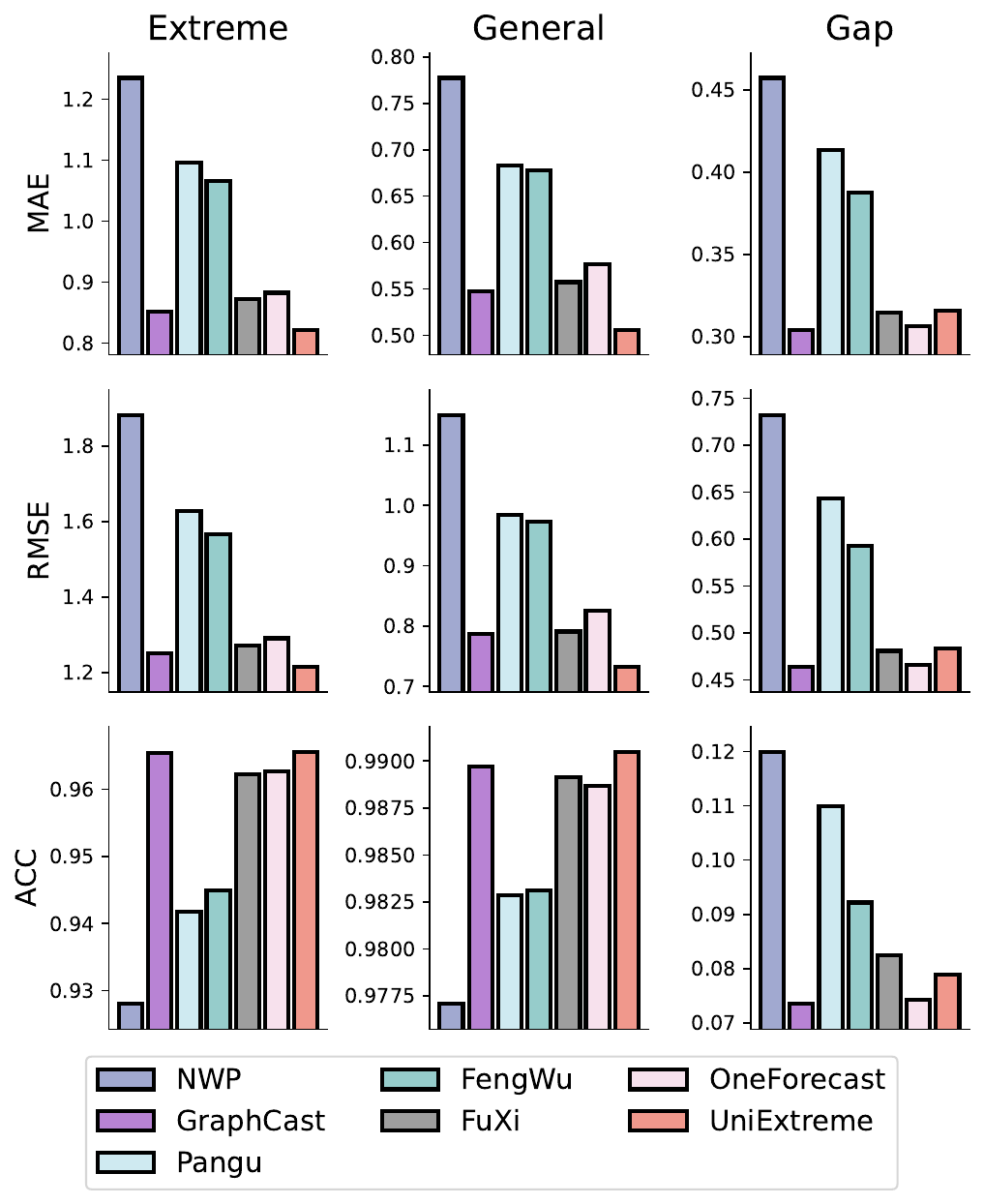}
    \vspace{-25pt}
    \caption{Raw forecasting results of variable U850.}
    \vspace{-5pt}
    \label{fig:raw_u_850_left}
\end{minipage}
\hfill
\begin{minipage}[t]{0.48\textwidth}
    \centering
    \includegraphics[width=\linewidth]{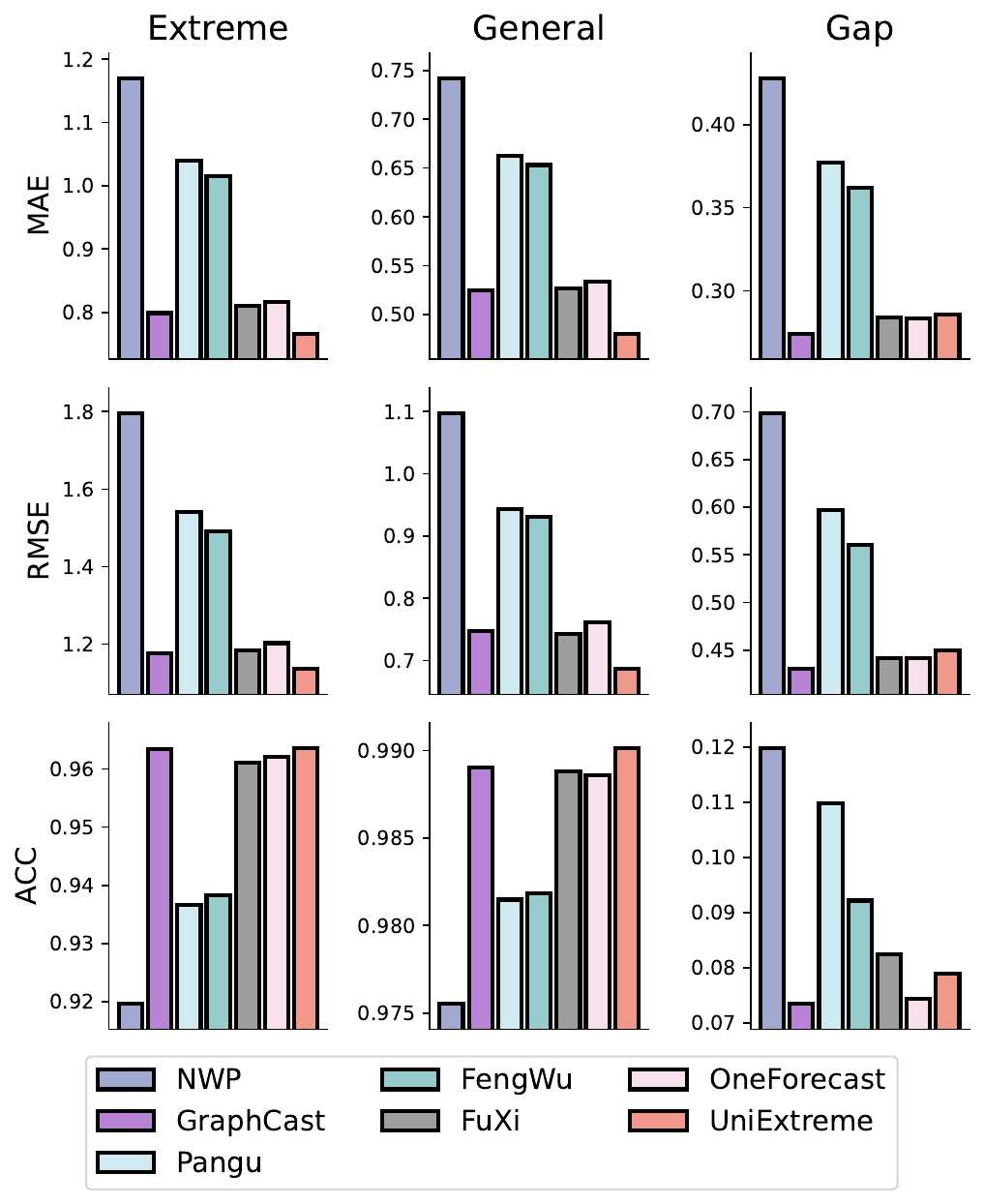}
    \vspace{-25pt}
    \caption{Raw forecasting results of variable U925.}
    \vspace{-5pt}
    \label{fig:raw_u_925_right}
\end{minipage}
\\[10pt]
\begin{minipage}[t]{0.48\textwidth}
    \centering
    \includegraphics[width=\linewidth]{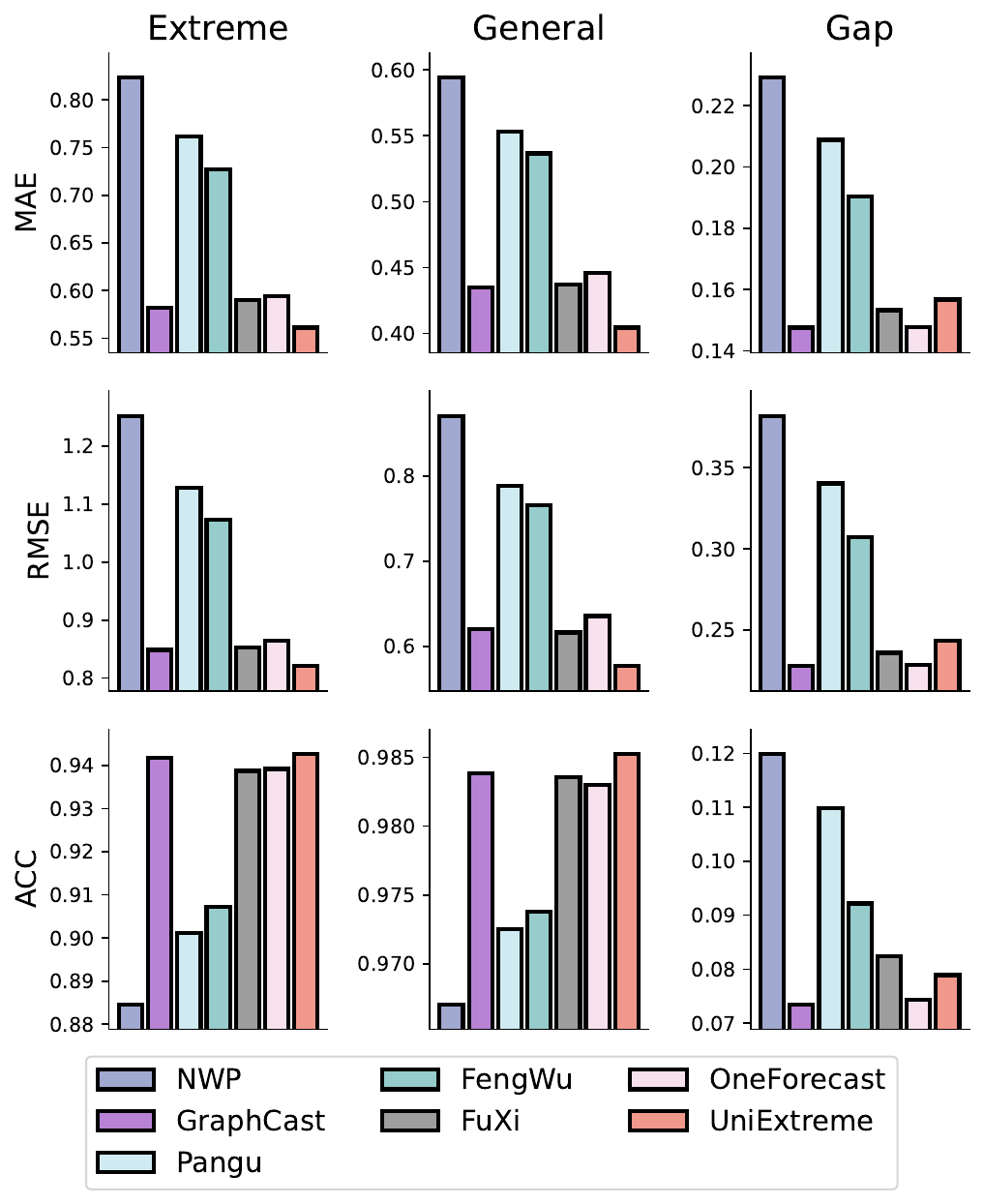}
    \vspace{-25pt}
    \caption{Raw forecasting results of variable U1000.}
    \vspace{-5pt}
    \label{fig:raw_u_1000_left}
\end{minipage}
\hfill
\begin{minipage}[t]{0.48\textwidth}
    \centering
    \includegraphics[width=\linewidth]{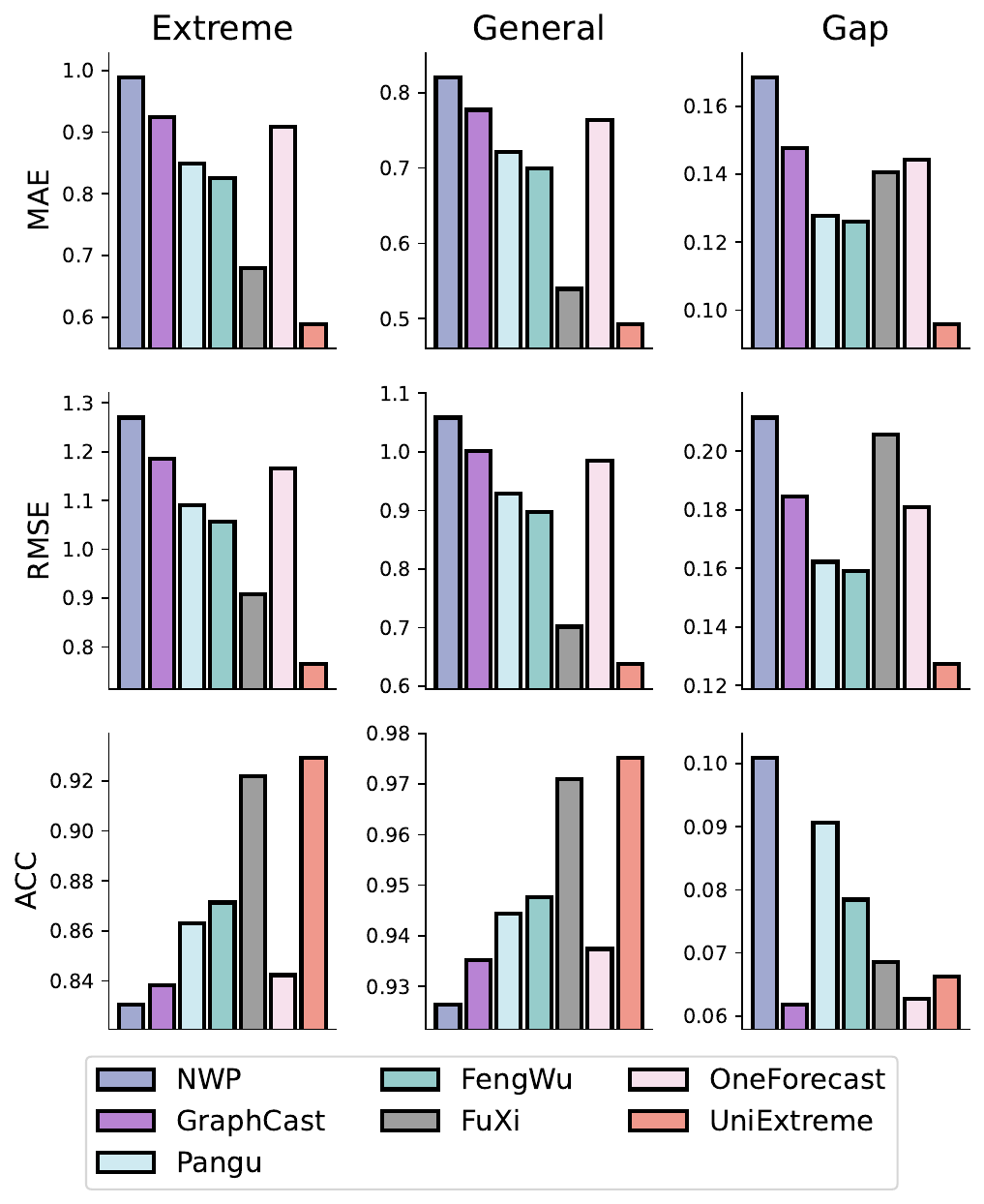}
    \vspace{-25pt}
    \caption{Raw forecasting results of variable V50.}
    \vspace{-5pt}
    \label{fig:raw_v_50_right}
\end{minipage}
\end{figure*}

\begin{figure*}[!ht]
\begin{minipage}[t]{0.48\textwidth}
    \centering
    \includegraphics[width=\linewidth]{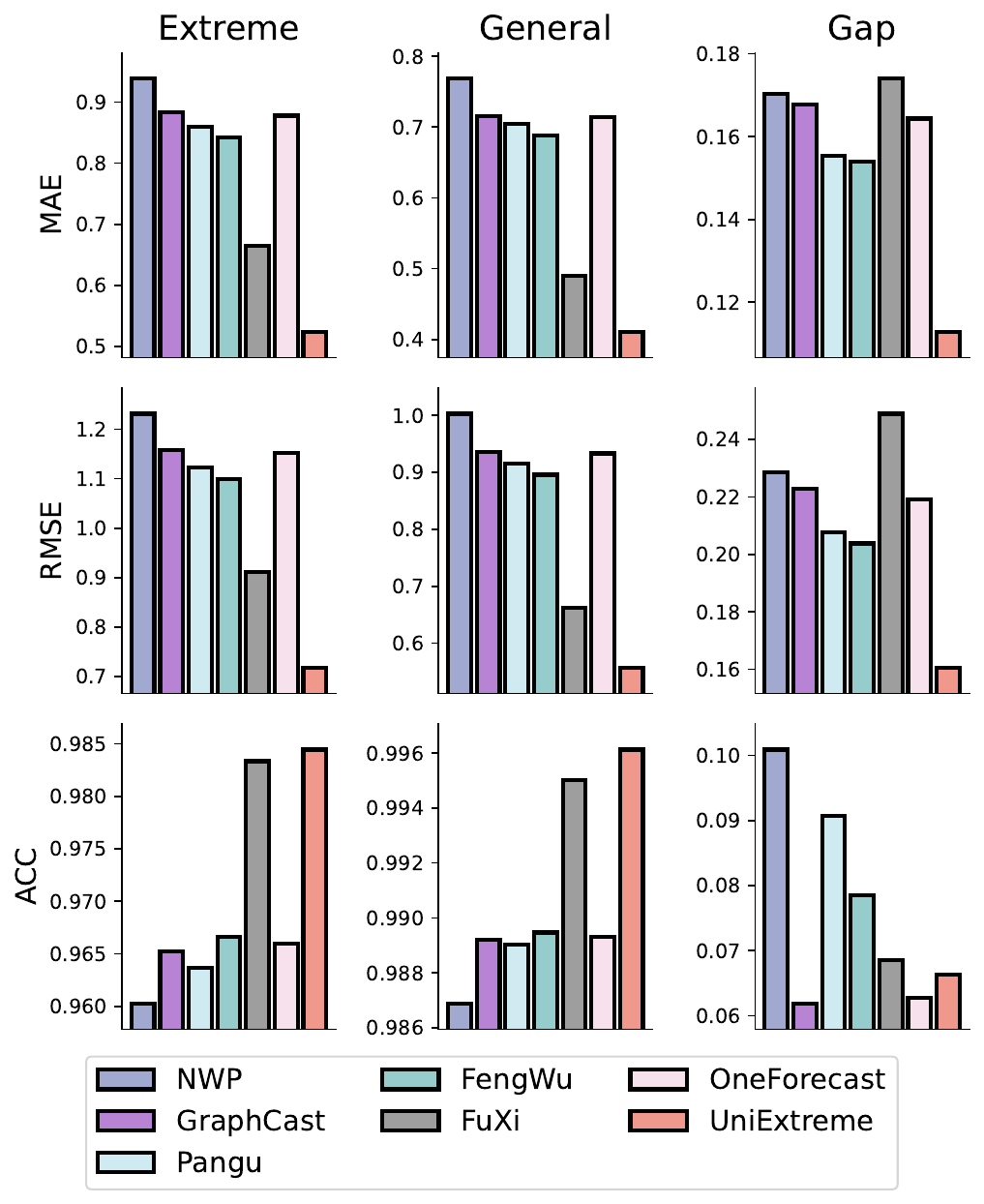}
    \vspace{-25pt}
    \caption{Raw forecasting results of variable V100.}
    \vspace{-5pt}
    \label{fig:raw_v_100_left}
\end{minipage}
\hfill
\begin{minipage}[t]{0.48\textwidth}
    \centering
    \includegraphics[width=\linewidth]{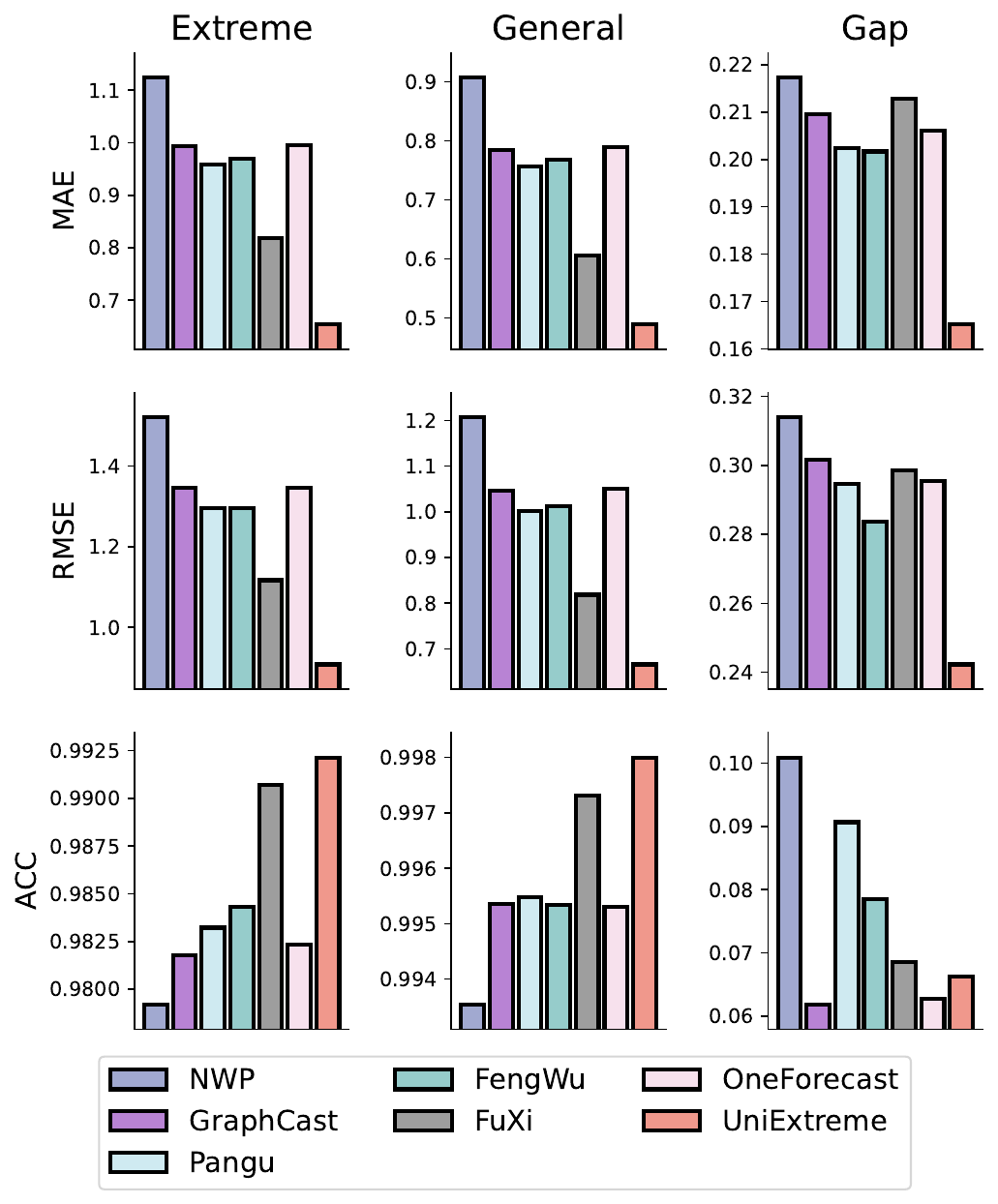}
    \vspace{-25pt}
    \caption{Raw forecasting results of variable V150.}
    \vspace{-5pt}
    \label{fig:raw_v_150_right}
\end{minipage}
\\[10pt]
\begin{minipage}[t]{0.48\textwidth}
    \centering
    \includegraphics[width=\linewidth]{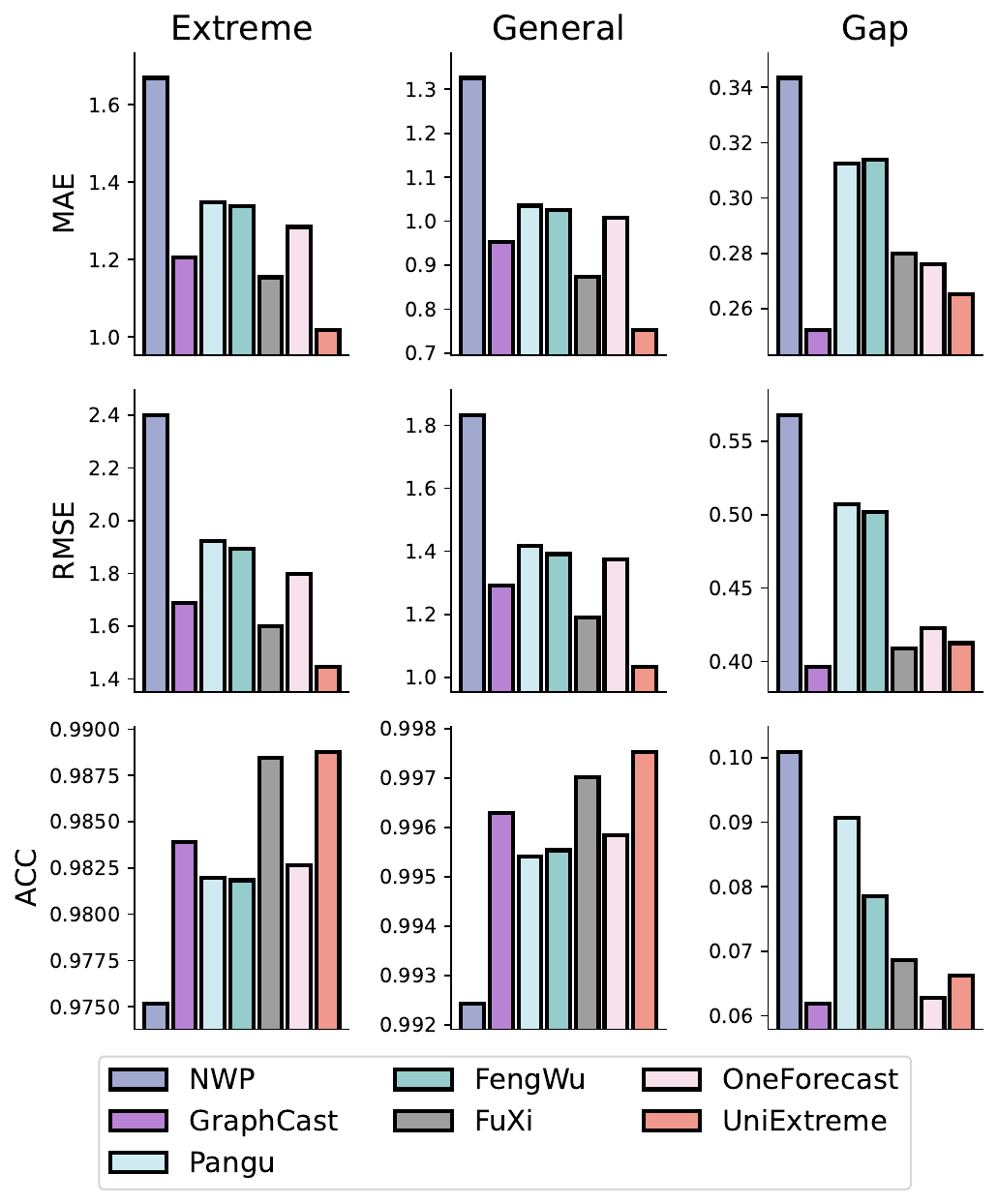}
    \vspace{-25pt}
    \caption{Raw forecasting results of variable V200.}
    \vspace{-5pt}
    \label{fig:raw_v_200_left}
\end{minipage}
\hfill
\begin{minipage}[t]{0.48\textwidth}
    \centering
    \includegraphics[width=\linewidth]{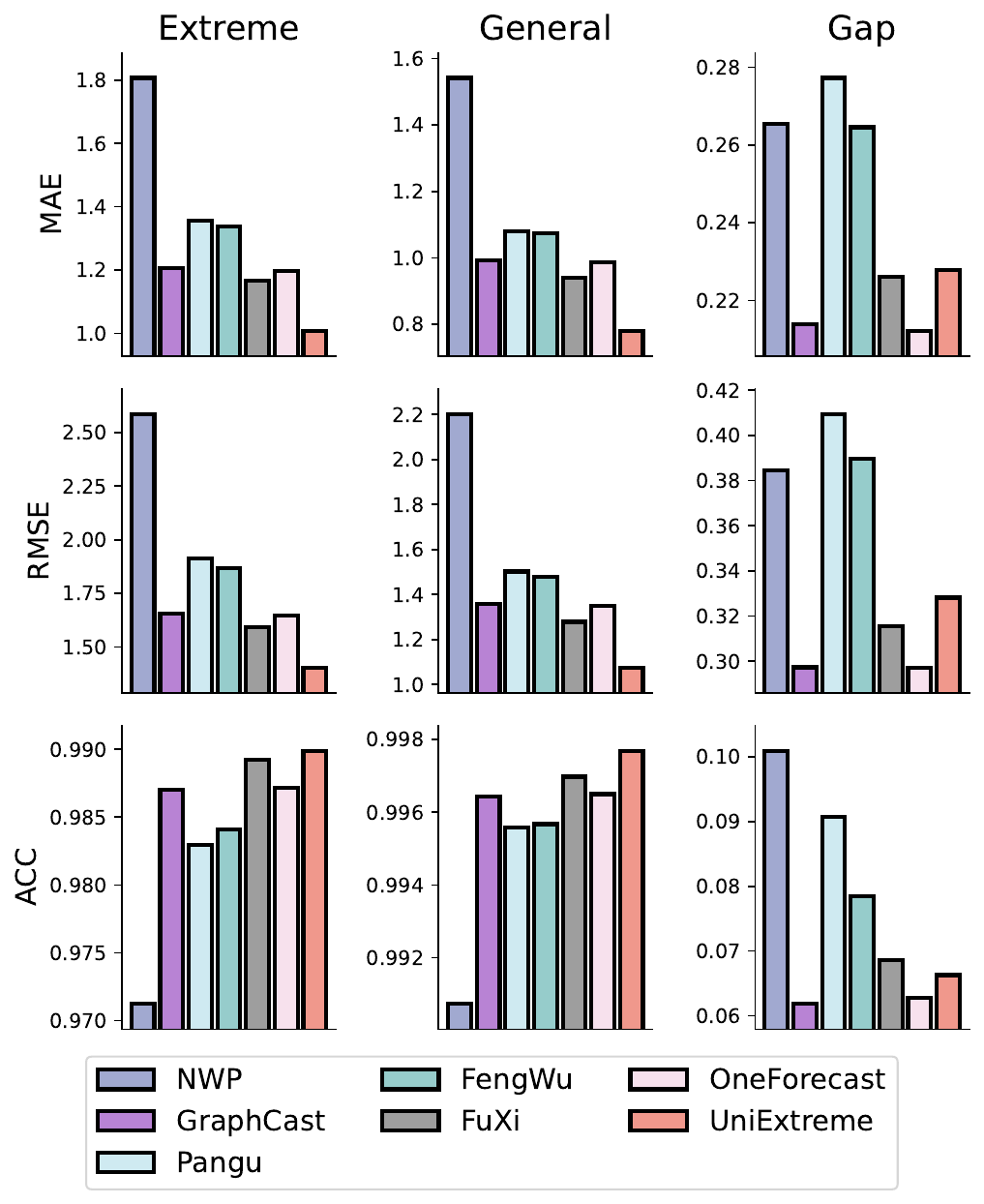}
    \vspace{-25pt}
    \caption{Raw forecasting results of variable V250.}
    \vspace{-5pt}
    \label{fig:raw_v_250_right}
\end{minipage}
\end{figure*}

\begin{figure*}[!ht]
\begin{minipage}[t]{0.48\textwidth}
    \centering
    \includegraphics[width=\linewidth]{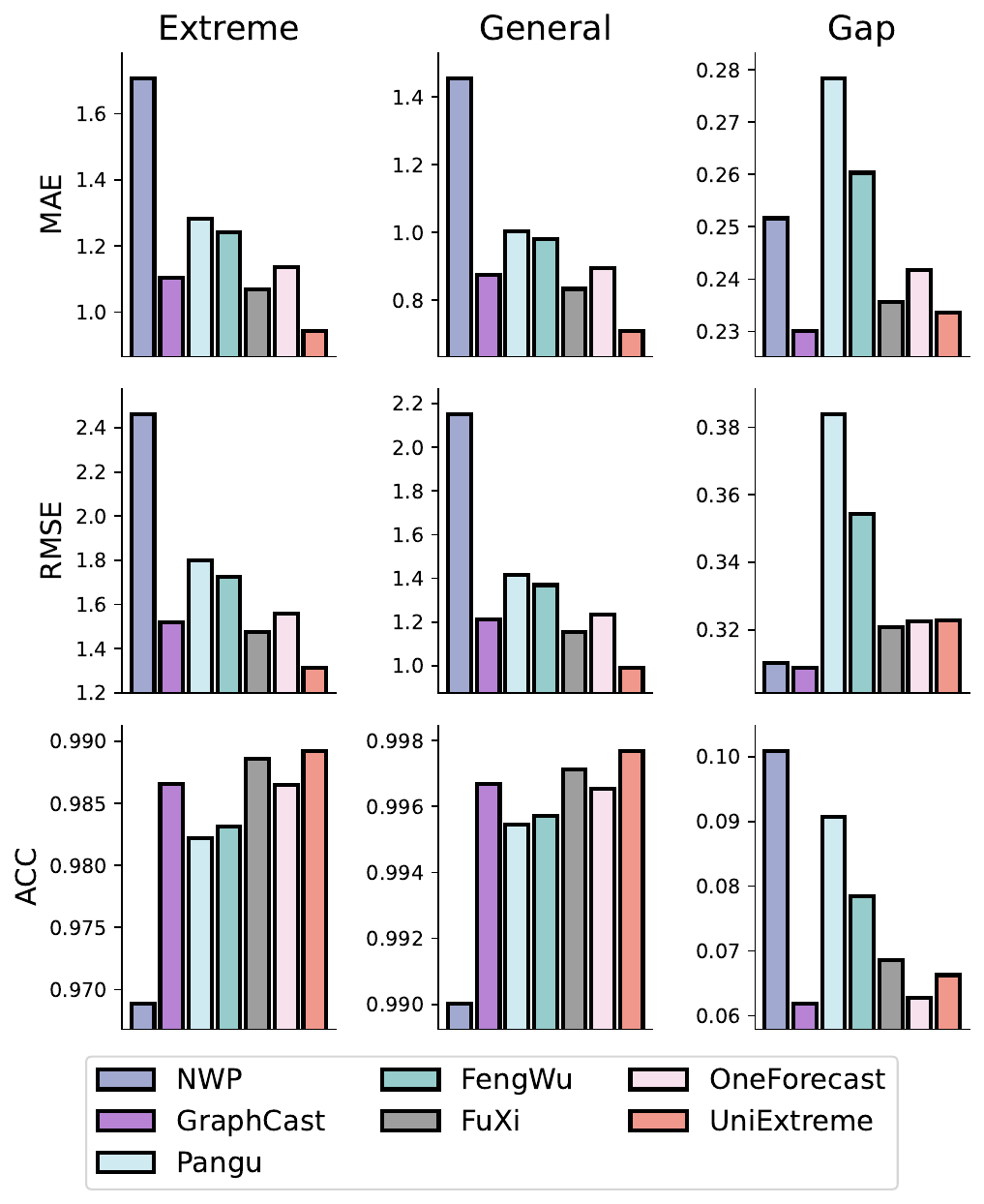}
    \vspace{-25pt}
    \caption{Raw forecasting results of variable V300.}
    \vspace{-5pt}
    \label{fig:raw_v_300_left}
\end{minipage}
\hfill
\begin{minipage}[t]{0.48\textwidth}
    \centering
    \includegraphics[width=\linewidth]{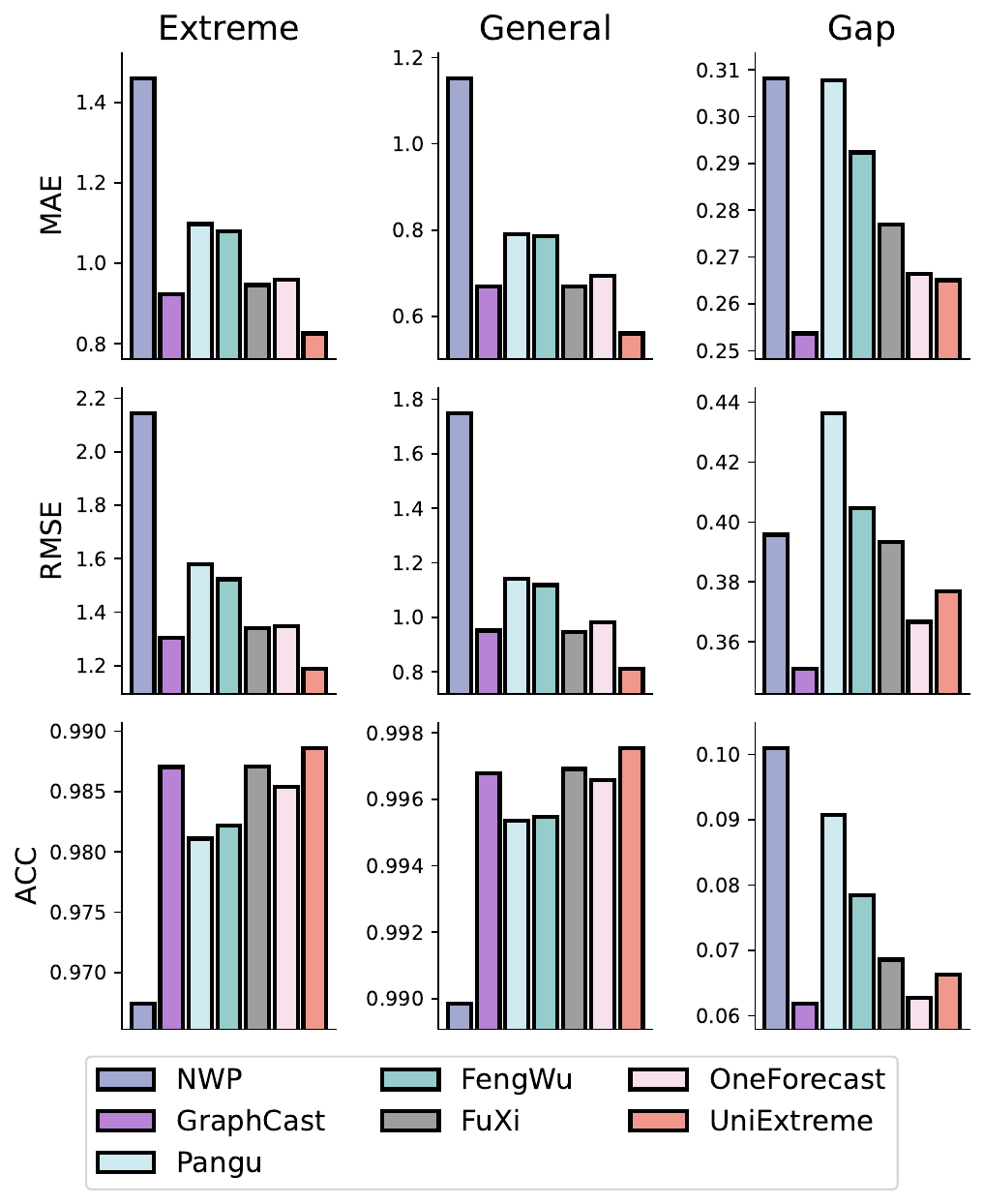}
    \vspace{-25pt}
    \caption{Raw forecasting results of variable V400.}
    \vspace{-5pt}
    \label{fig:raw_v_400_right}
\end{minipage}
\\[10pt]
\begin{minipage}[t]{0.48\textwidth}
    \centering
    \includegraphics[width=\linewidth]{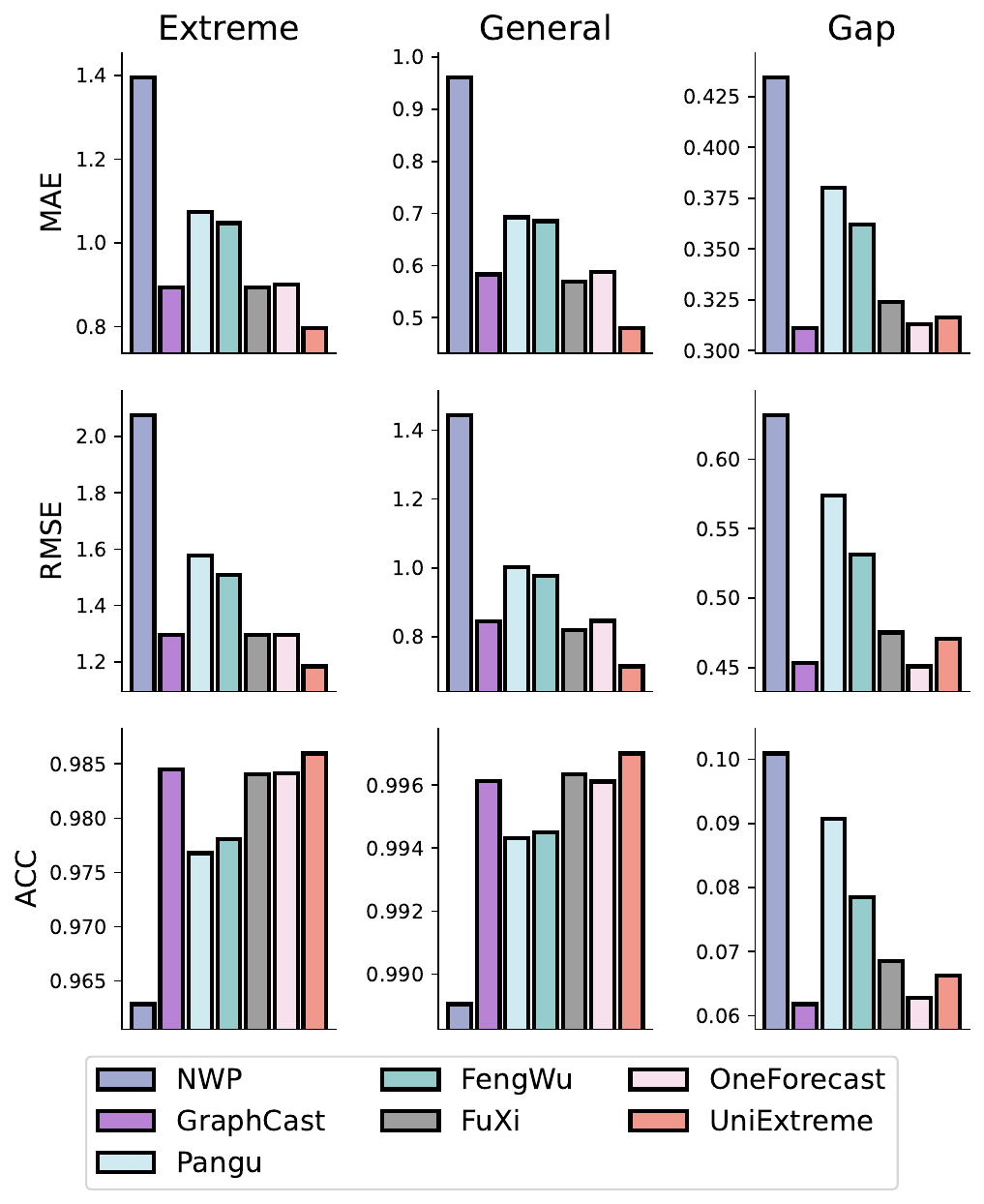}
    \vspace{-25pt}
    \caption{Raw forecasting results of variable V500.}
    \vspace{-5pt}
    \label{fig:raw_v_500_left}
\end{minipage}
\hfill
\begin{minipage}[t]{0.48\textwidth}
    \centering
    \includegraphics[width=\linewidth]{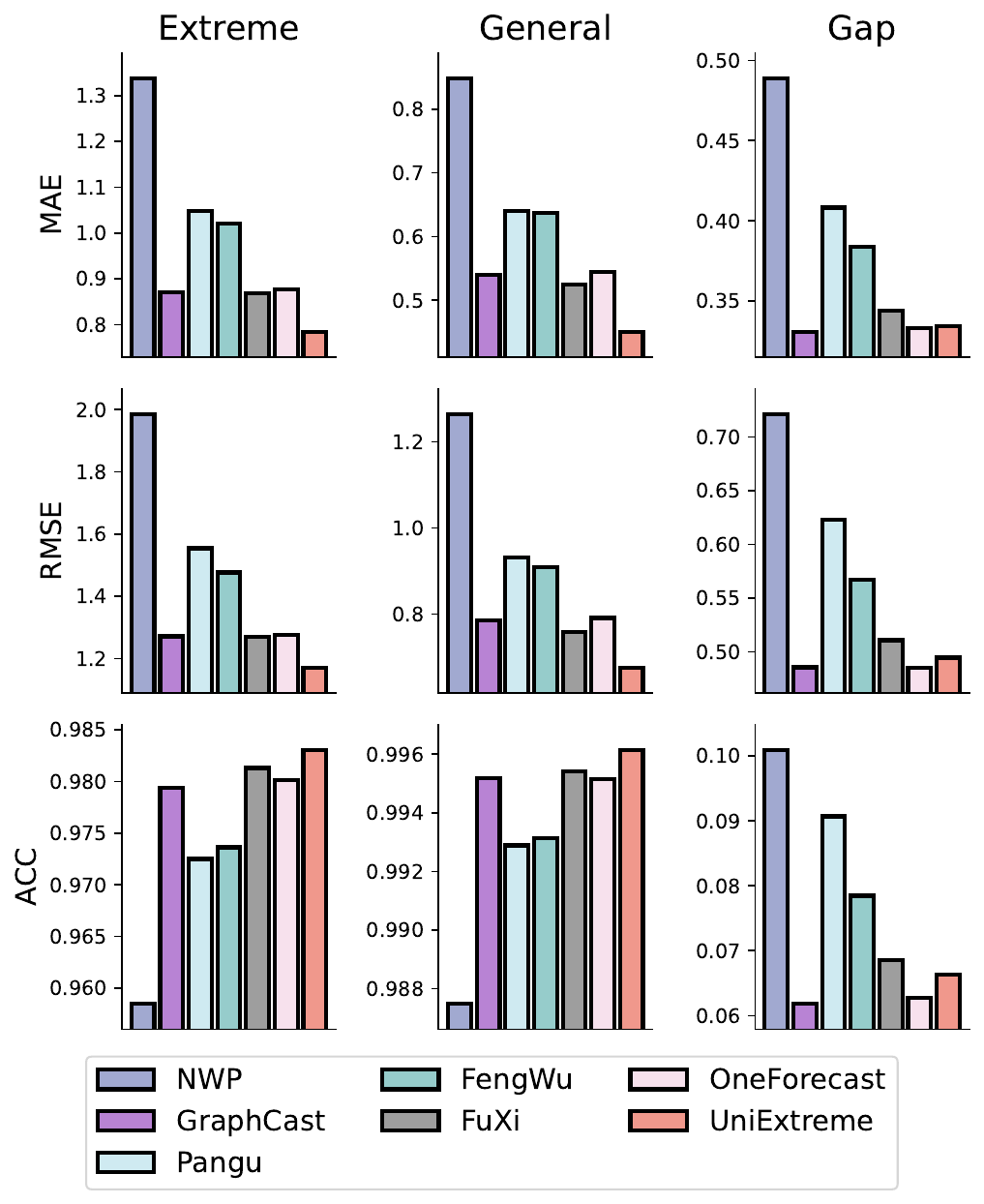}
    \vspace{-25pt}
    \caption{Raw forecasting results of variable V600.}
    \vspace{-5pt}
    \label{fig:raw_v_600_right}
\end{minipage}
\end{figure*}

\begin{figure*}[!ht]
\begin{minipage}[t]{0.48\textwidth}
    \centering
    \includegraphics[width=\linewidth]{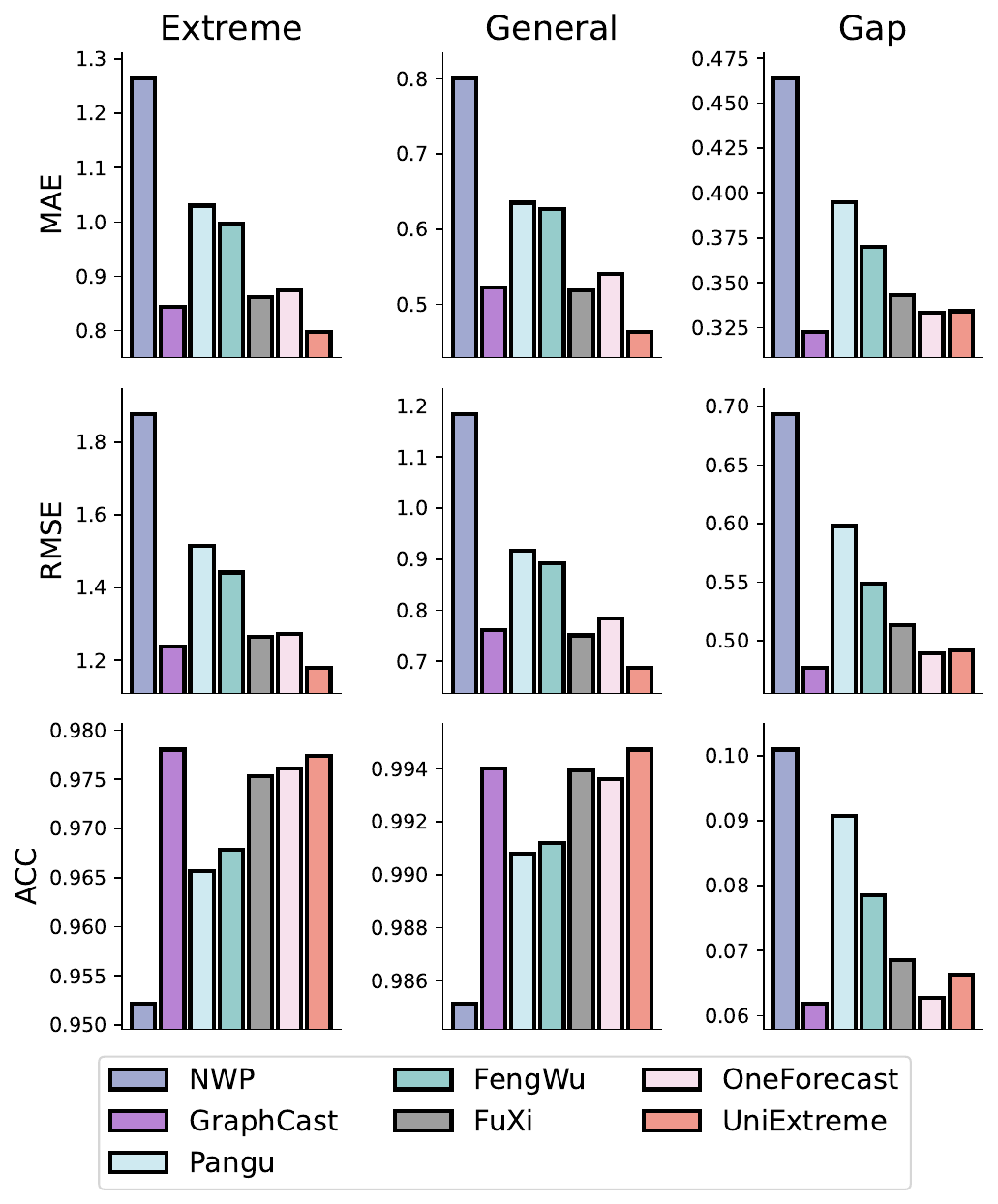}
    \vspace{-25pt}
    \caption{Raw forecasting results of variable V700.}
    \vspace{-5pt}
    \label{fig:raw_v_700_left}
\end{minipage}
\hfill
\begin{minipage}[t]{0.48\textwidth}
    \centering
    \includegraphics[width=\linewidth]{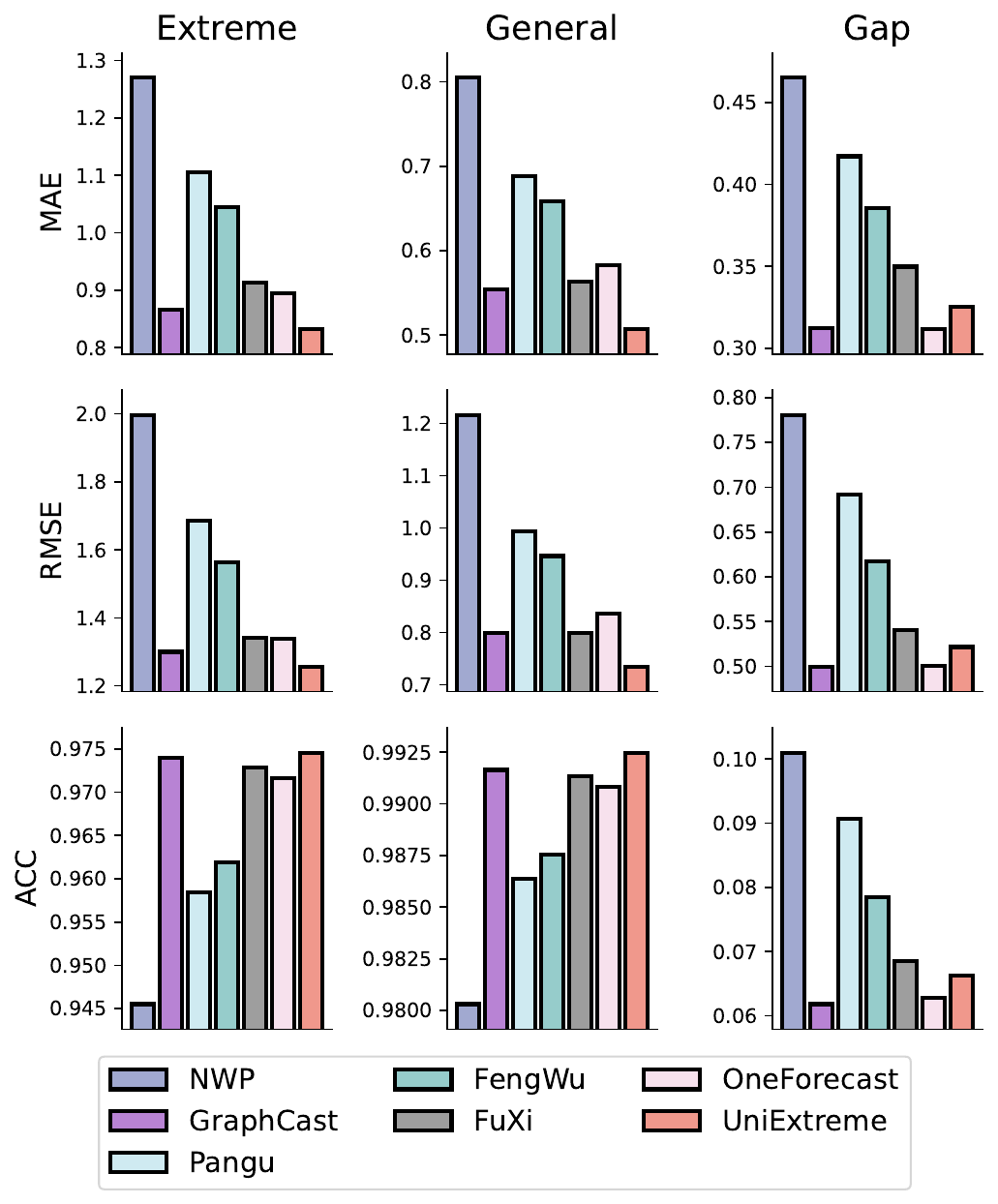}
    \vspace{-25pt}
    \caption{Raw forecasting results of variable V850.}
    \vspace{-5pt}
    \label{fig:raw_v_850_right}
\end{minipage}
\\[10pt]
\begin{minipage}[t]{0.48\textwidth}
    \centering
    \includegraphics[width=\linewidth]{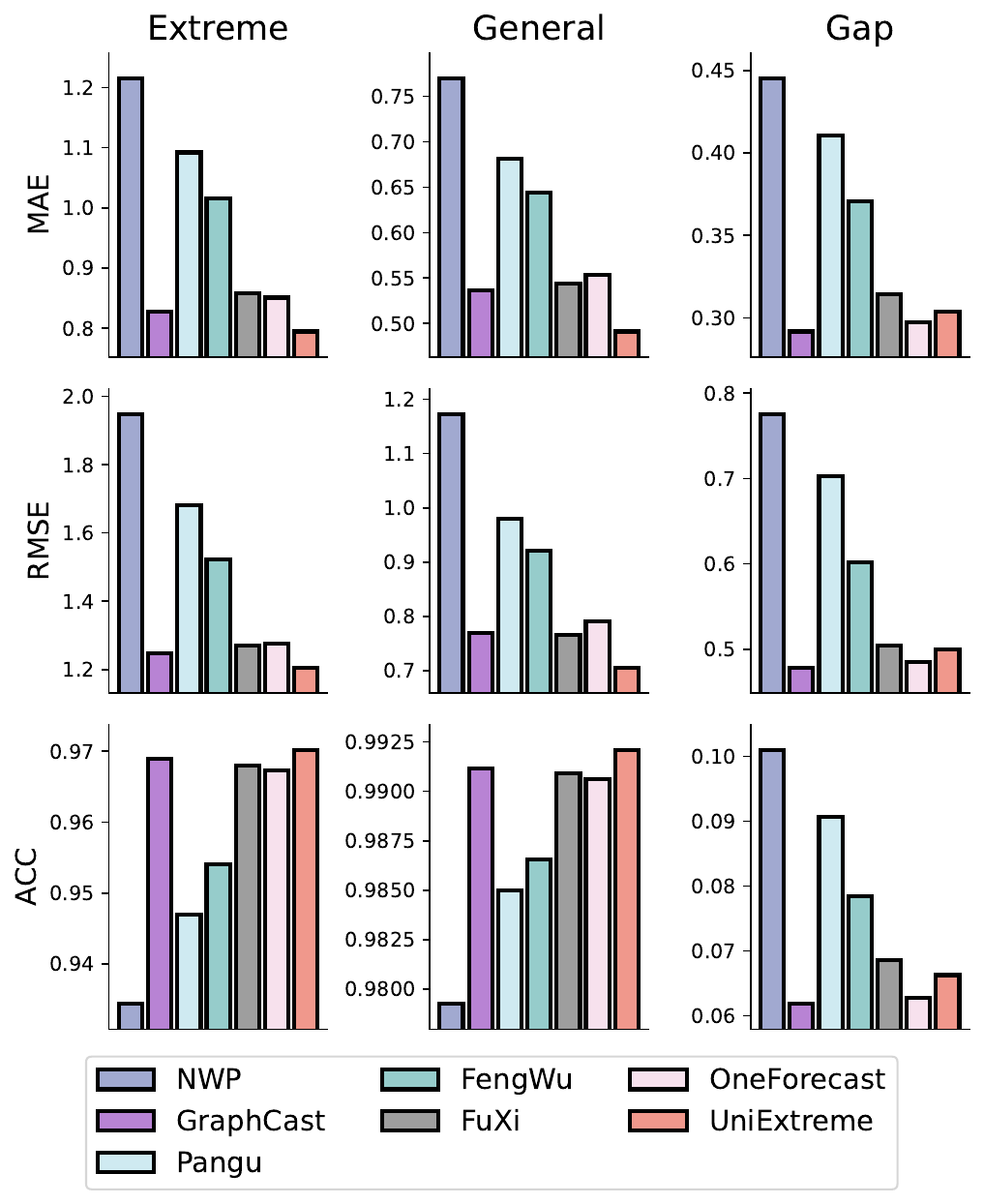}
    \vspace{-25pt}
    \caption{Raw forecasting results of variable V925.}
    \vspace{-5pt}
    \label{fig:raw_v_925_left}
\end{minipage}
\hfill
\begin{minipage}[t]{0.48\textwidth}
    \centering
    \includegraphics[width=\linewidth]{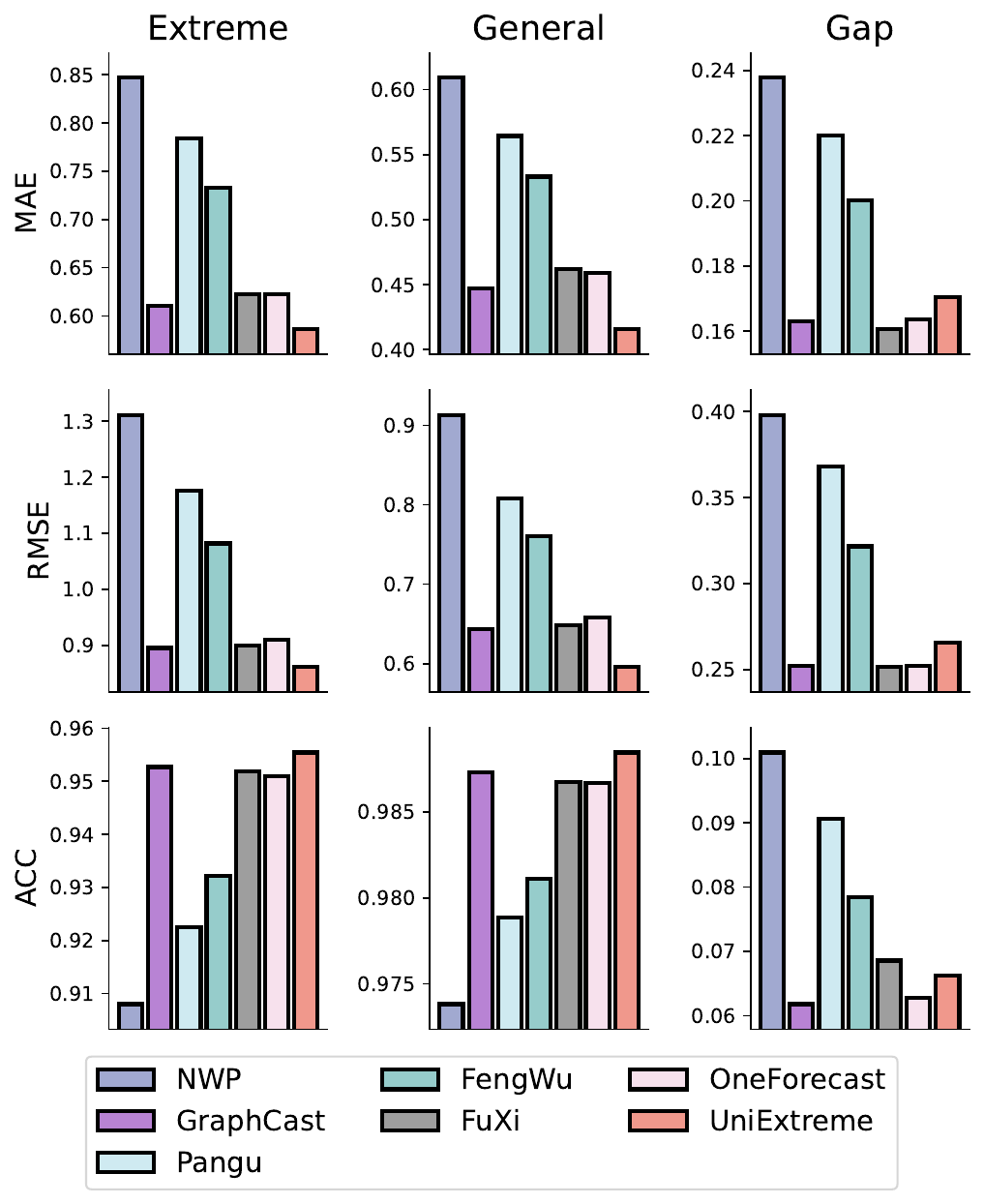}
    \vspace{-25pt}
    \caption{Raw forecasting results of variable V1000.}
    \vspace{-5pt}
    \label{fig:raw_v_1000_right}
\end{minipage}
\end{figure*}

\begin{figure*}[!ht]
\begin{minipage}[t]{0.48\textwidth}
    \centering
    \includegraphics[width=\linewidth]{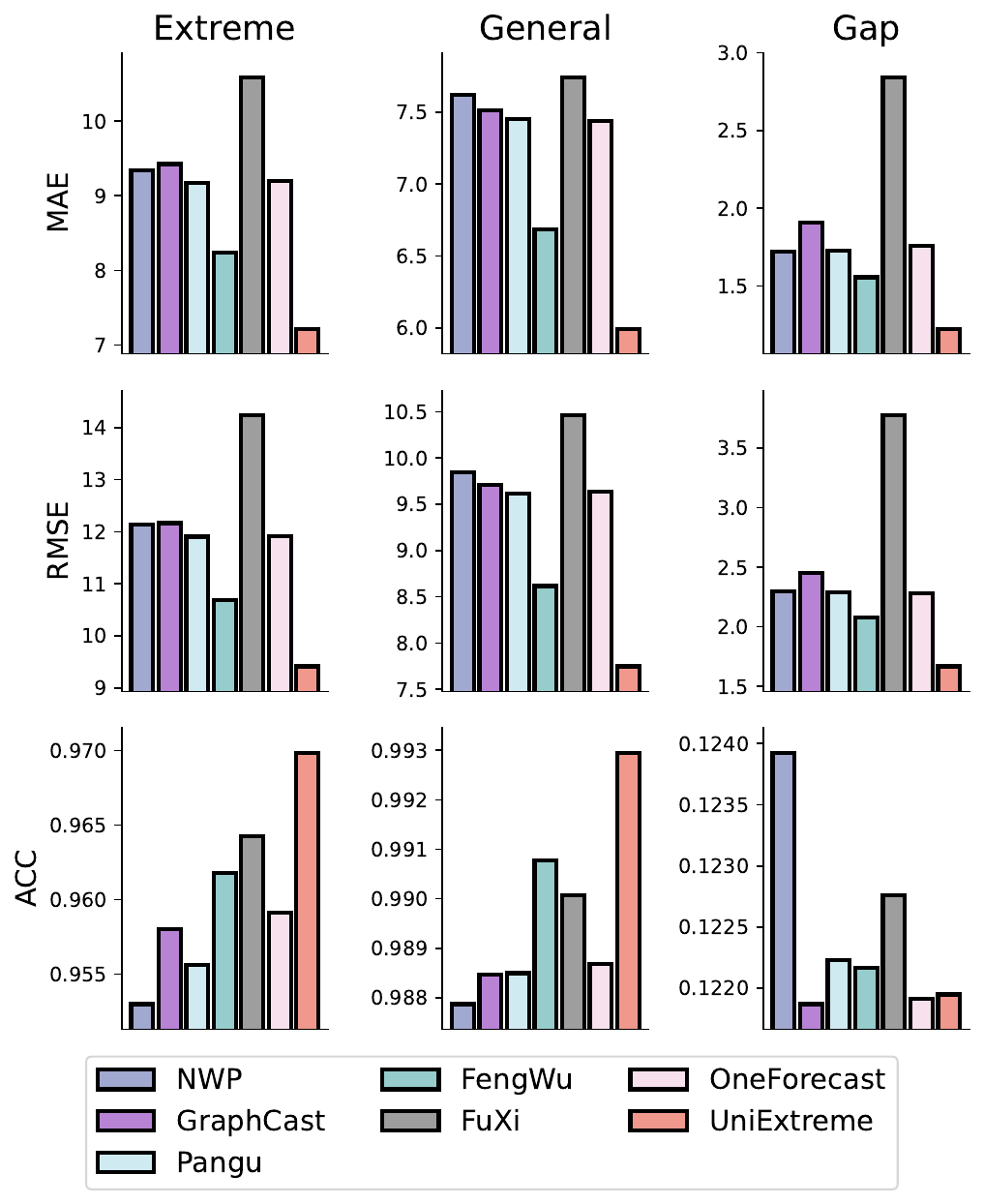}
    \vspace{-25pt}
    \caption{Raw forecasting results of variable Z50.}
    \vspace{-5pt}
    \label{fig:raw_z_50_left}
\end{minipage}
\hfill
\begin{minipage}[t]{0.48\textwidth}
    \centering
    \includegraphics[width=\linewidth]{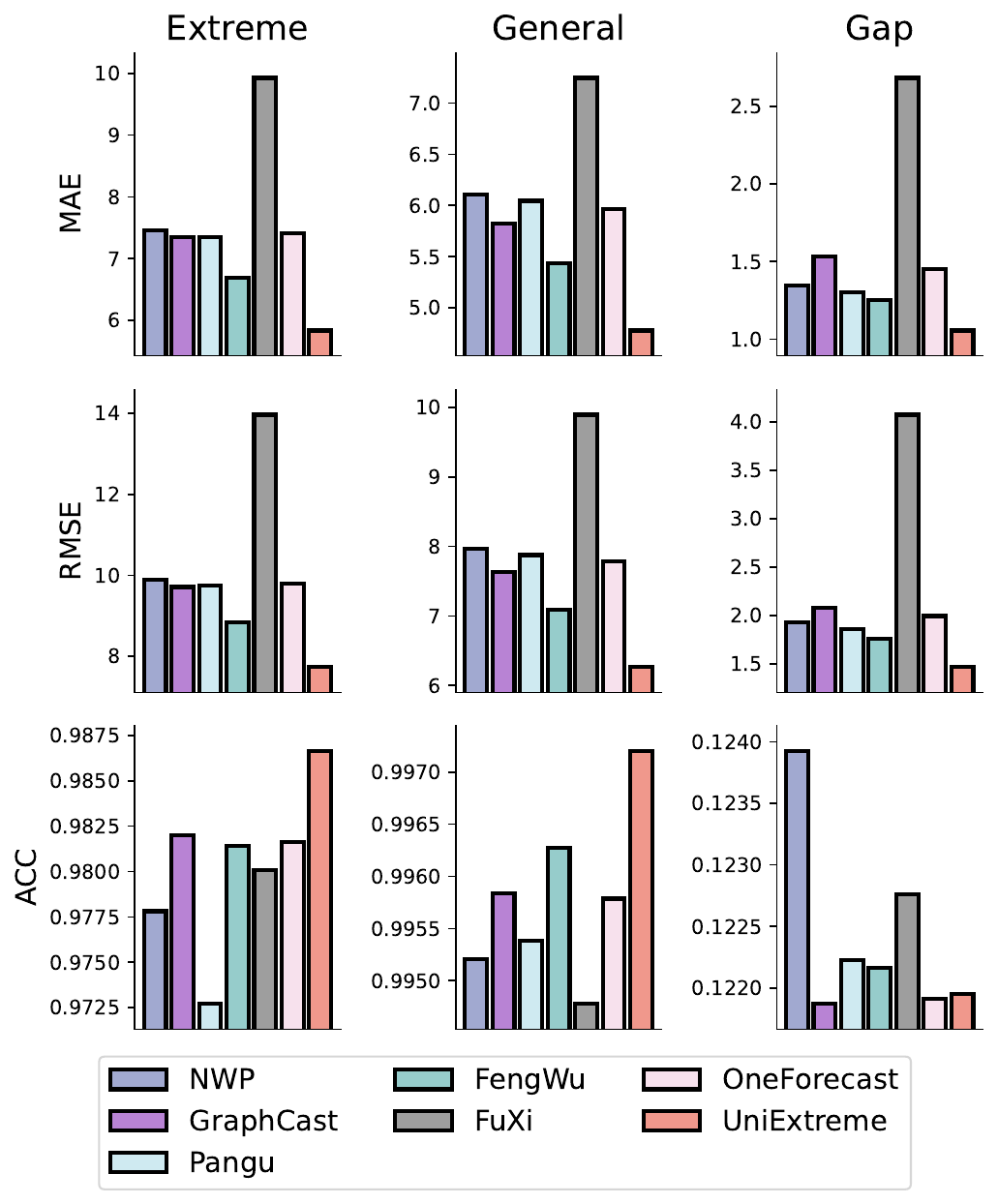}
    \vspace{-25pt}
    \caption{Raw forecasting results of variable Z100.}
    \vspace{-5pt}
    \label{fig:raw_z_100_right}
\end{minipage}
\\[10pt]
\begin{minipage}[t]{0.48\textwidth}
    \centering
    \includegraphics[width=\linewidth]{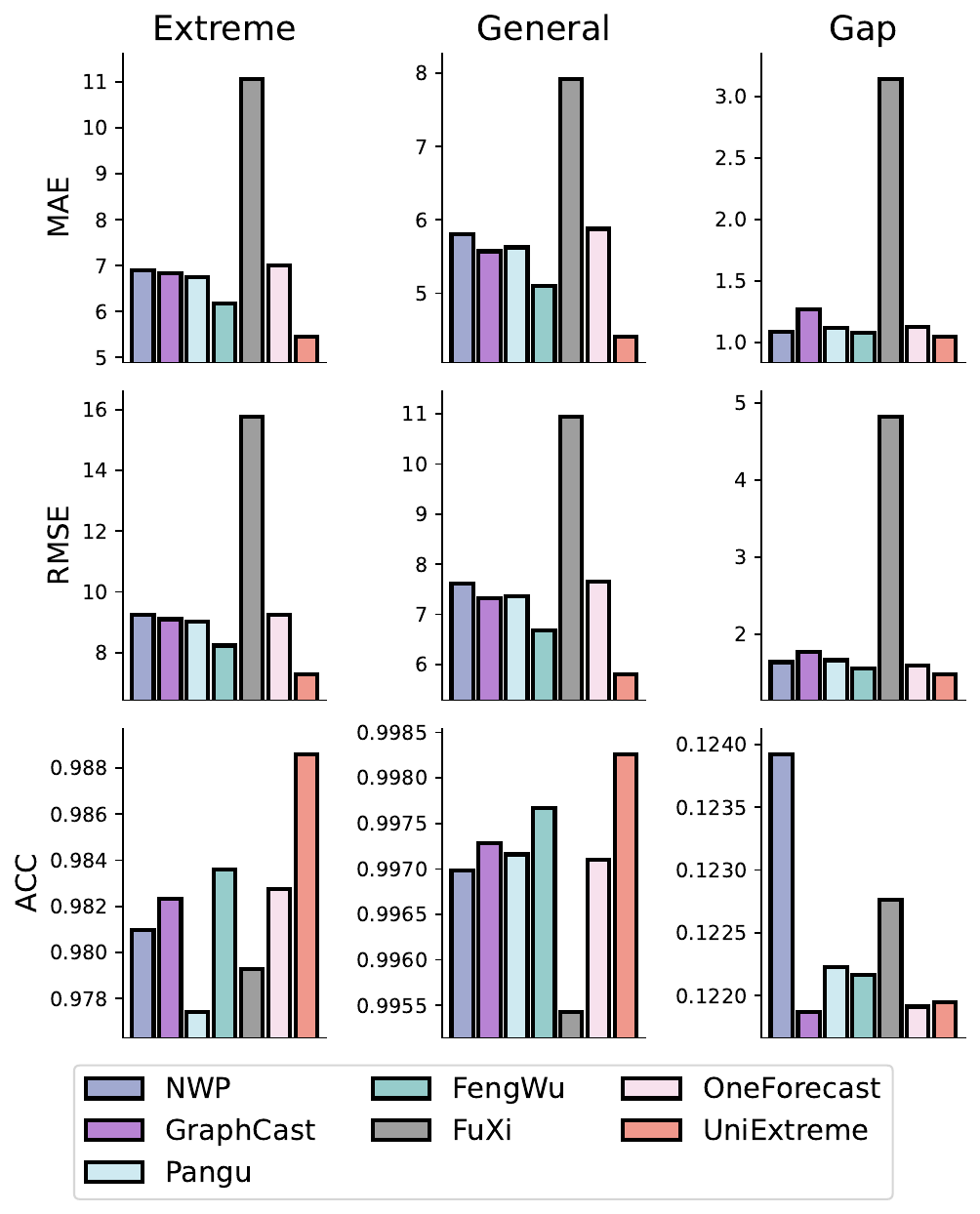}
    \vspace{-25pt}
    \caption{Raw forecasting results of variable Z150.}
    \vspace{-5pt}
    \label{fig:raw_z_150_left}
\end{minipage}
\hfill
\begin{minipage}[t]{0.48\textwidth}
    \centering
    \includegraphics[width=\linewidth]{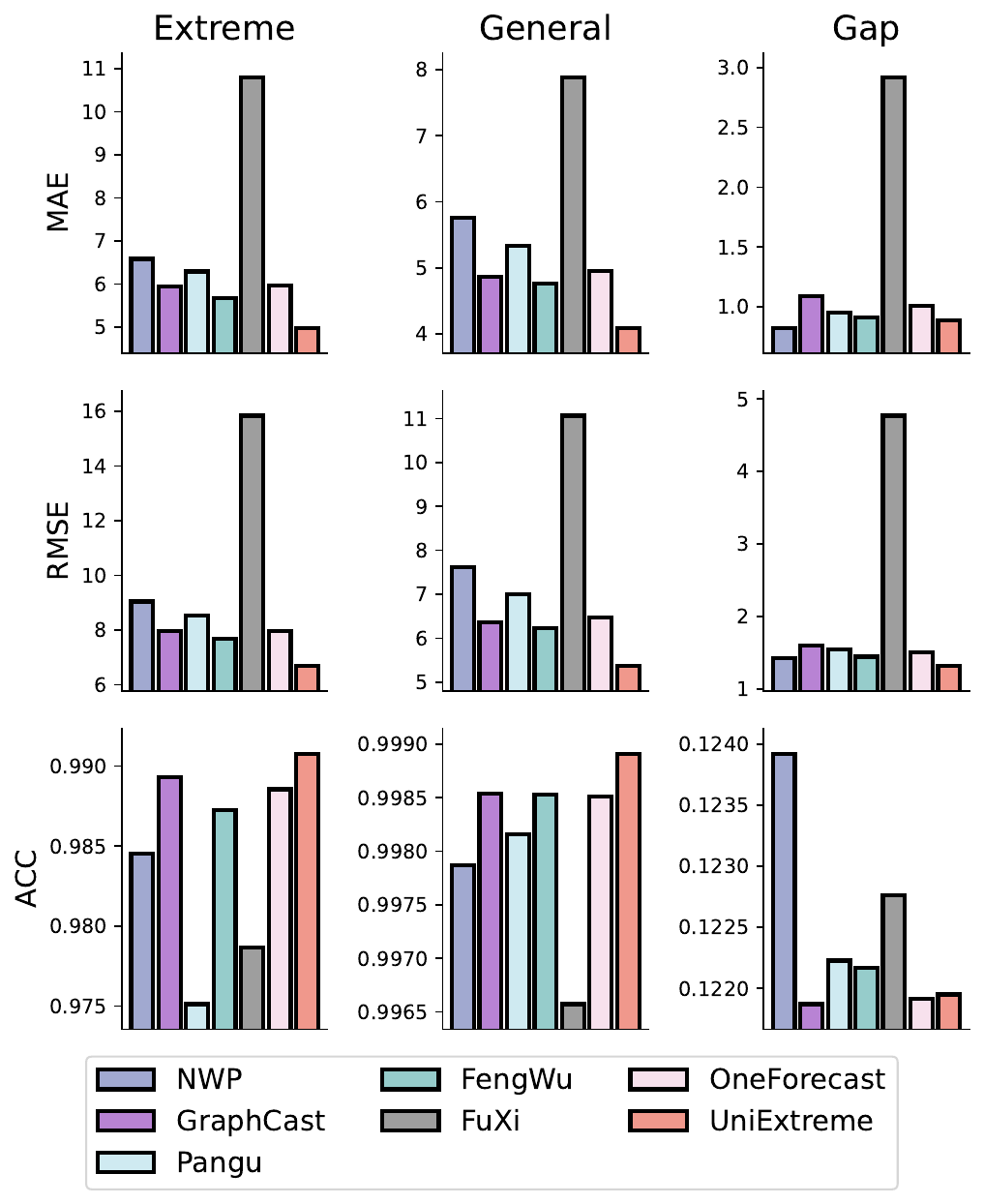}
    \vspace{-25pt}
    \caption{Raw forecasting results of variable Z200.}
    \vspace{-5pt}
    \label{fig:raw_z_200_right}
\end{minipage}
\end{figure*}

\begin{figure*}[!ht]
\begin{minipage}[t]{0.48\textwidth}
    \centering
    \includegraphics[width=\linewidth]{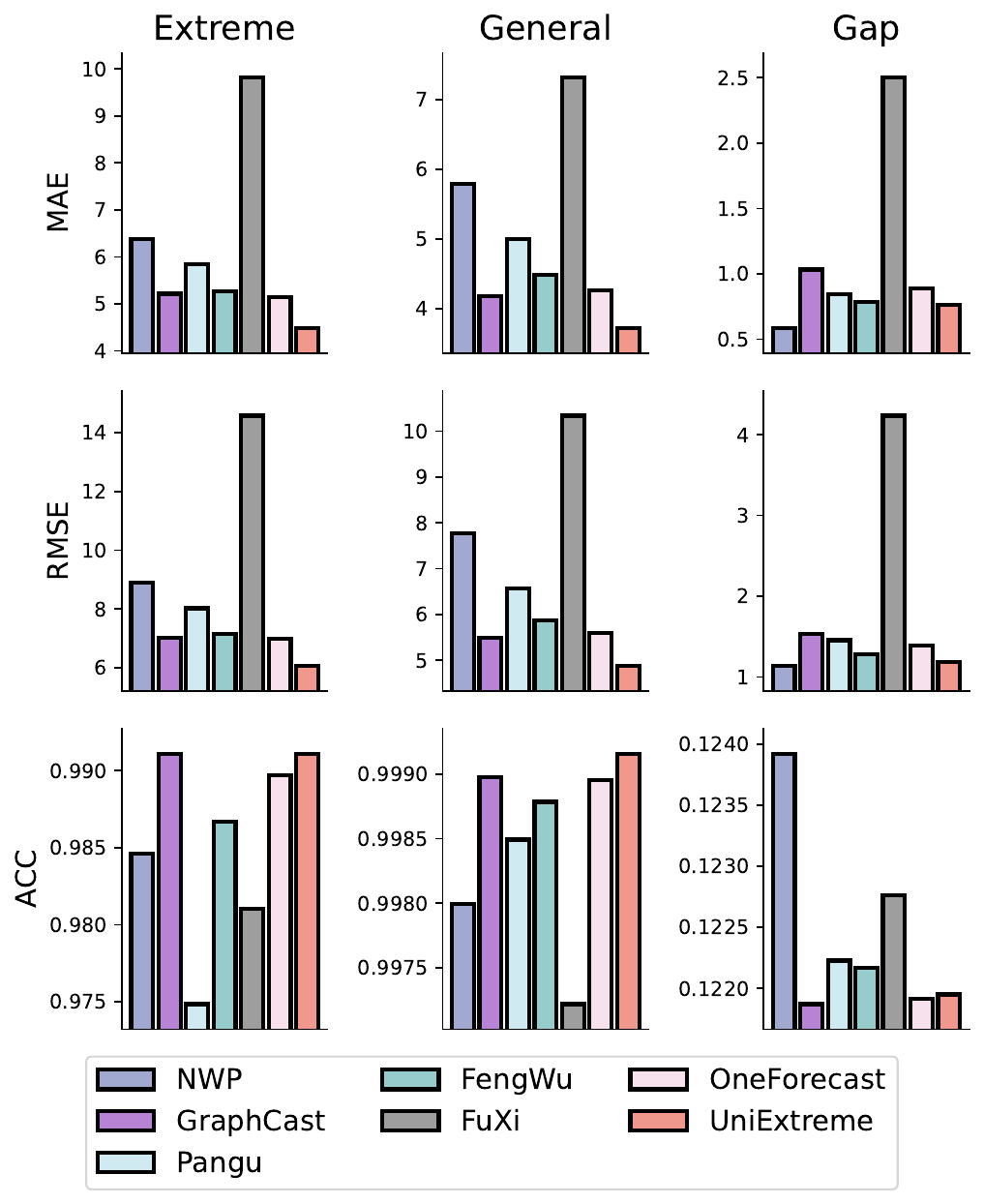}
    \vspace{-25pt}
    \caption{Raw forecasting results of variable Z250.}
    \vspace{-5pt}
    \label{fig:raw_z_250_left}
\end{minipage}
\hfill
\begin{minipage}[t]{0.48\textwidth}
    \centering
    \includegraphics[width=\linewidth]{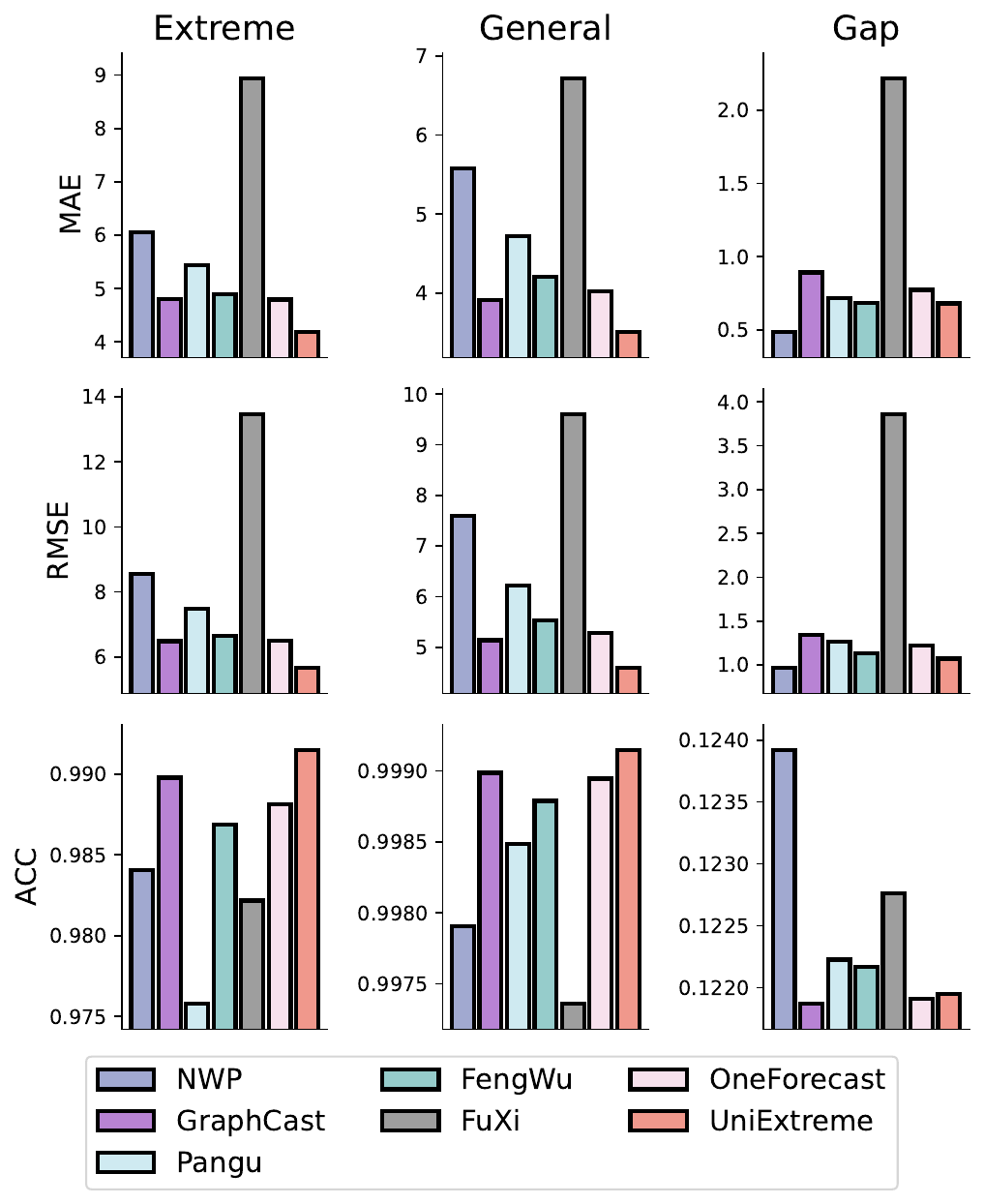}
    \vspace{-25pt}
    \caption{Raw forecasting results of variable Z300.}
    \vspace{-5pt}
    \label{fig:raw_z_300_right}
\end{minipage}
\\[10pt]
\begin{minipage}[t]{0.48\textwidth}
    \centering
    \includegraphics[width=\linewidth]{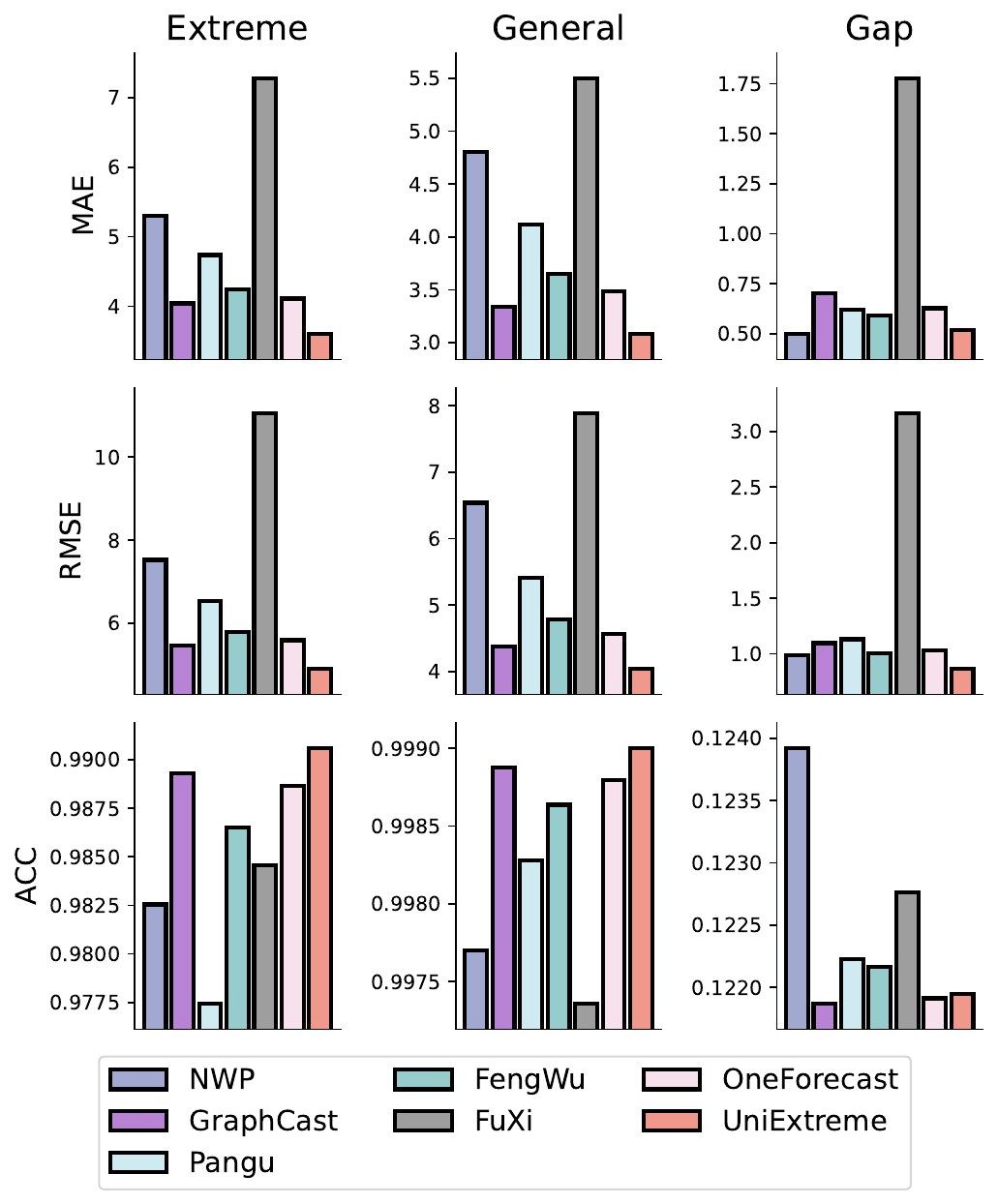}
    \vspace{-25pt}
    \caption{Raw forecasting results of variable Z400.}
    \vspace{-5pt}
    \label{fig:raw_z_400_left}
\end{minipage}
\hfill
\begin{minipage}[t]{0.48\textwidth}
    \centering
    \includegraphics[width=\linewidth]{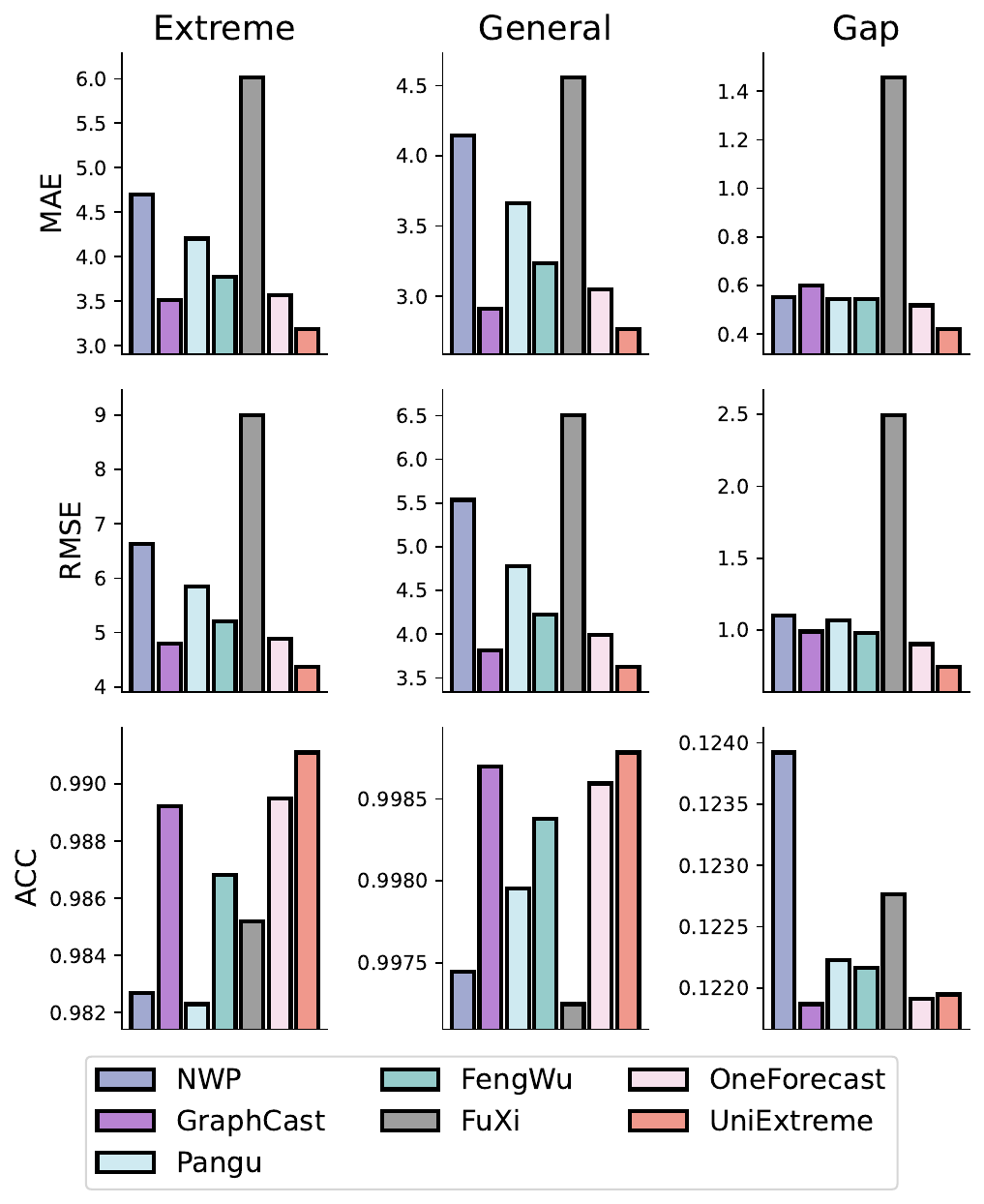}
    \vspace{-25pt}
    \caption{Raw forecasting results of variable Z500.}
    \vspace{-5pt}
    \label{fig:raw_z_500_right}
\end{minipage}
\end{figure*}

\begin{figure*}[!ht]
\begin{minipage}[t]{0.48\textwidth}
    \centering
    \includegraphics[width=\linewidth]{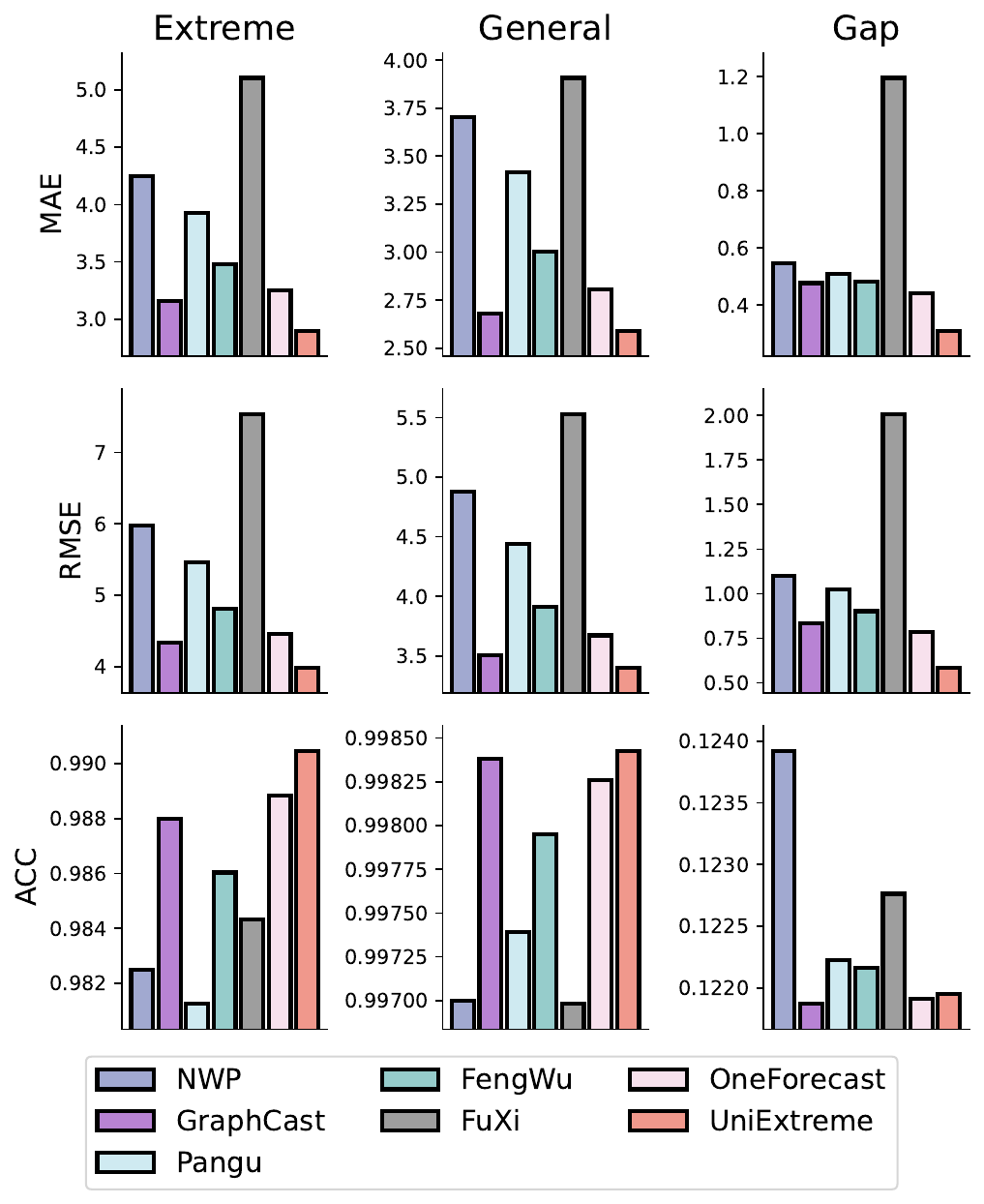}
    \vspace{-25pt}
    \caption{Raw forecasting results of variable Z600.}
    \vspace{-5pt}
    \label{fig:raw_z_600_left}
\end{minipage}
\hfill
\begin{minipage}[t]{0.48\textwidth}
    \centering
    \includegraphics[width=\linewidth]{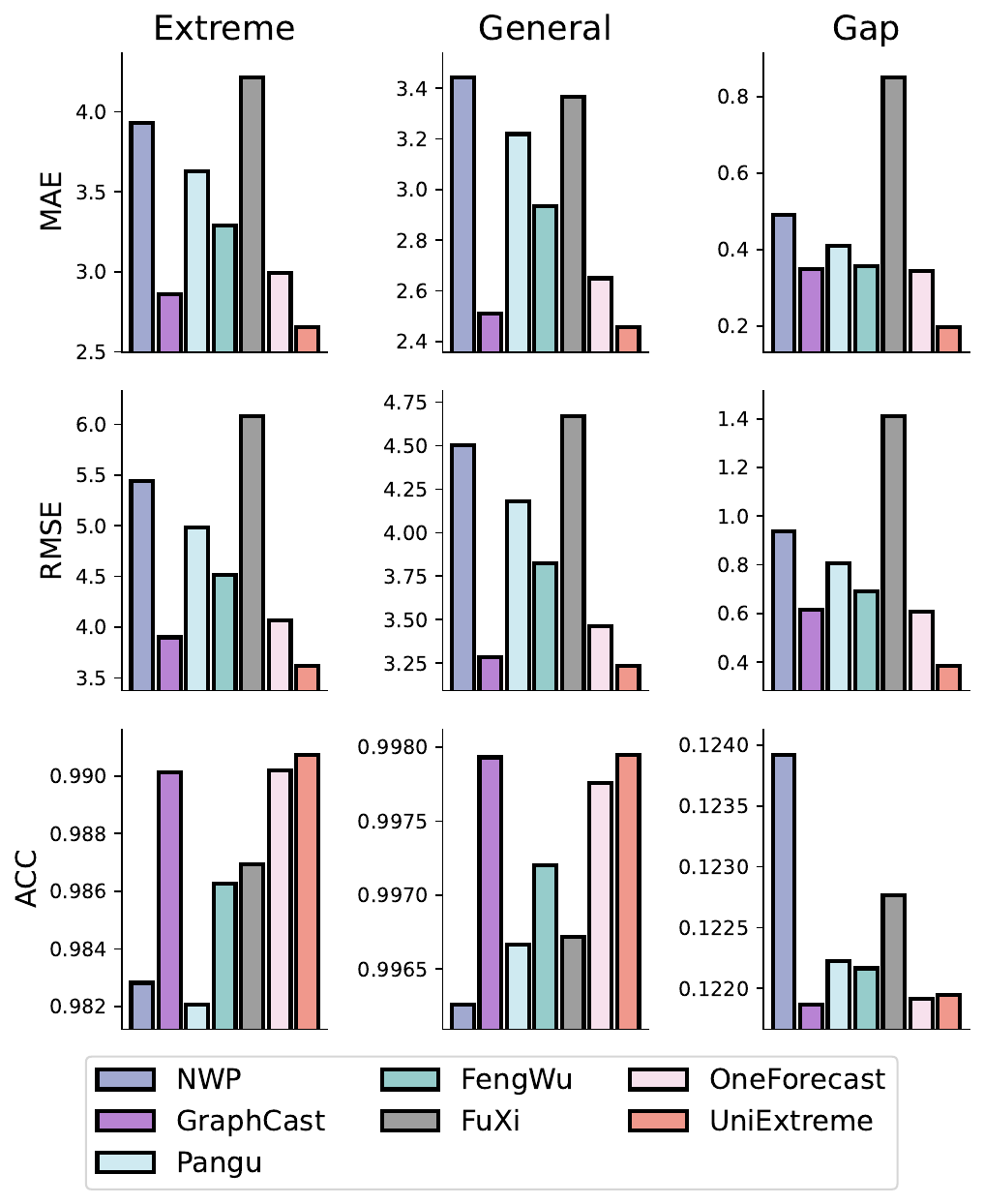}
    \vspace{-25pt}
    \caption{Raw forecasting results of variable Z700.}
    \vspace{-5pt}
    \label{fig:raw_z_700_right}
\end{minipage}
\\[10pt]
\begin{minipage}[t]{0.48\textwidth}
    \centering
    \includegraphics[width=\linewidth]{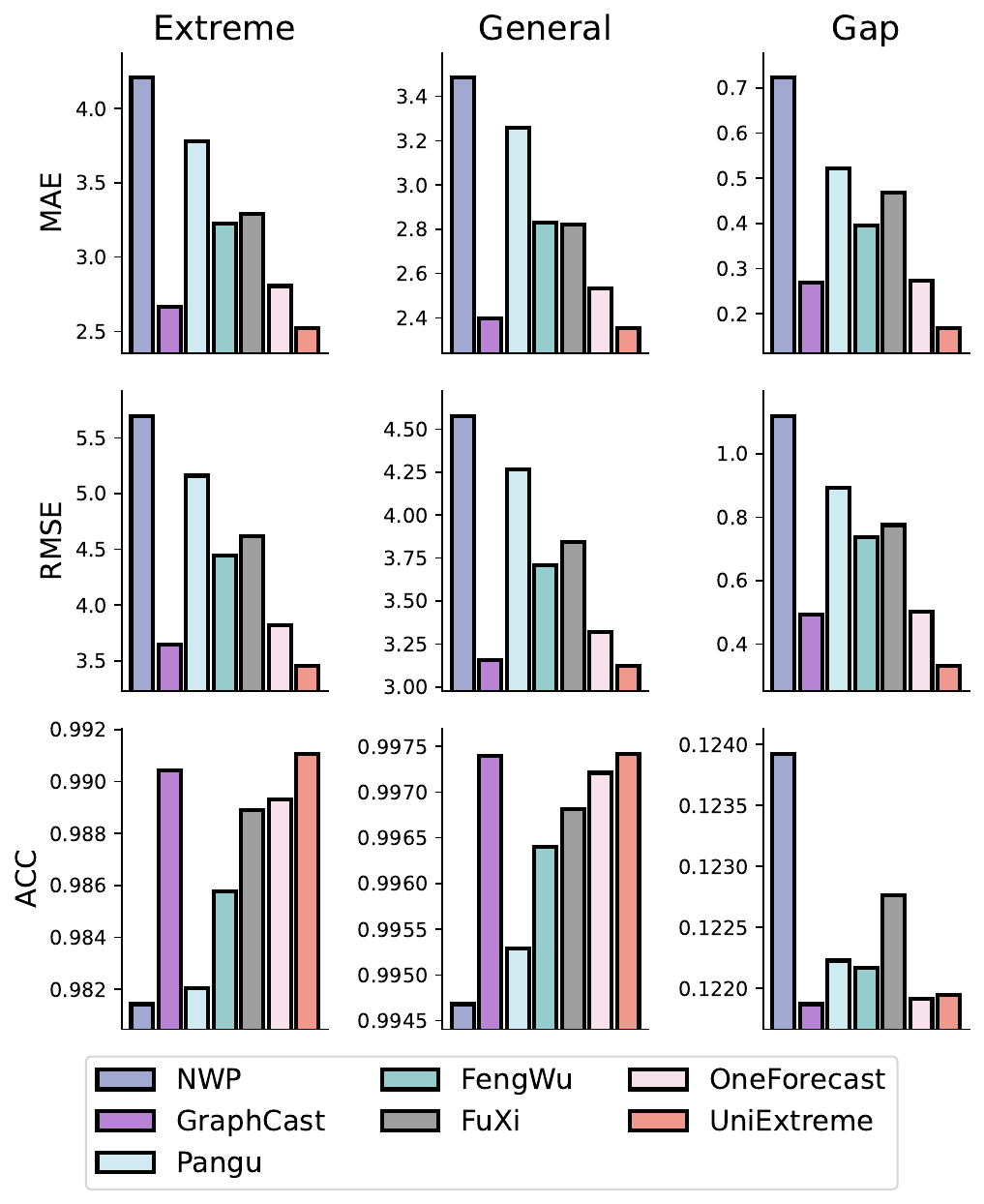}
    \vspace{-25pt}
    \caption{Raw forecasting results of variable Z850.}
    \vspace{-5pt}
    \label{fig:raw_z_850_left}
\end{minipage}
\hfill
\begin{minipage}[t]{0.48\textwidth}
    \centering
    \includegraphics[width=\linewidth]{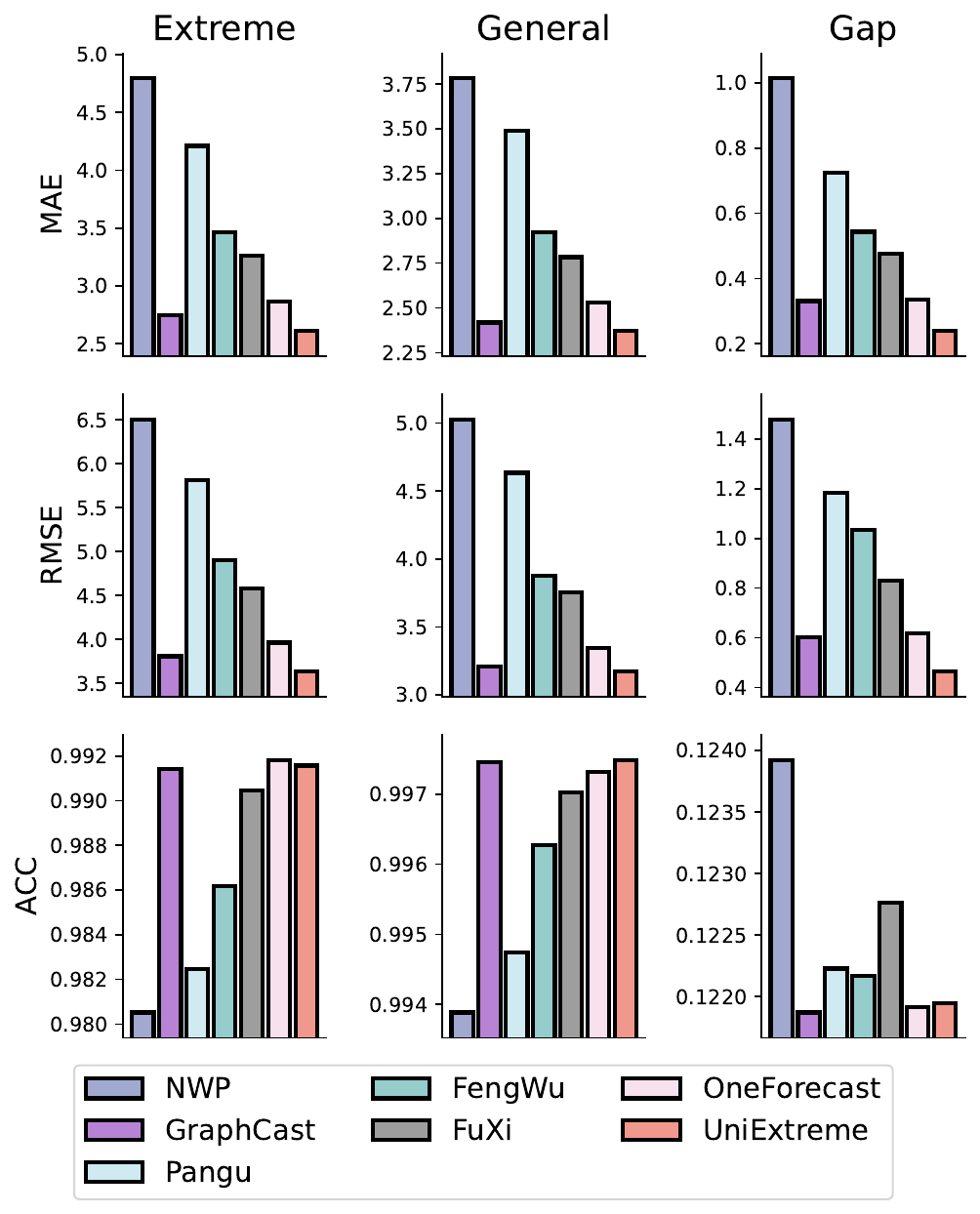}
    \vspace{-25pt}
    \caption{Raw forecasting results of variable Z925.}
    \vspace{-5pt}
    \label{fig:raw_z_925_right}
\end{minipage}
\end{figure*}

\begin{figure*}[!ht]
\begin{minipage}[t]{0.48\textwidth}
    \centering
    \includegraphics[width=\linewidth]{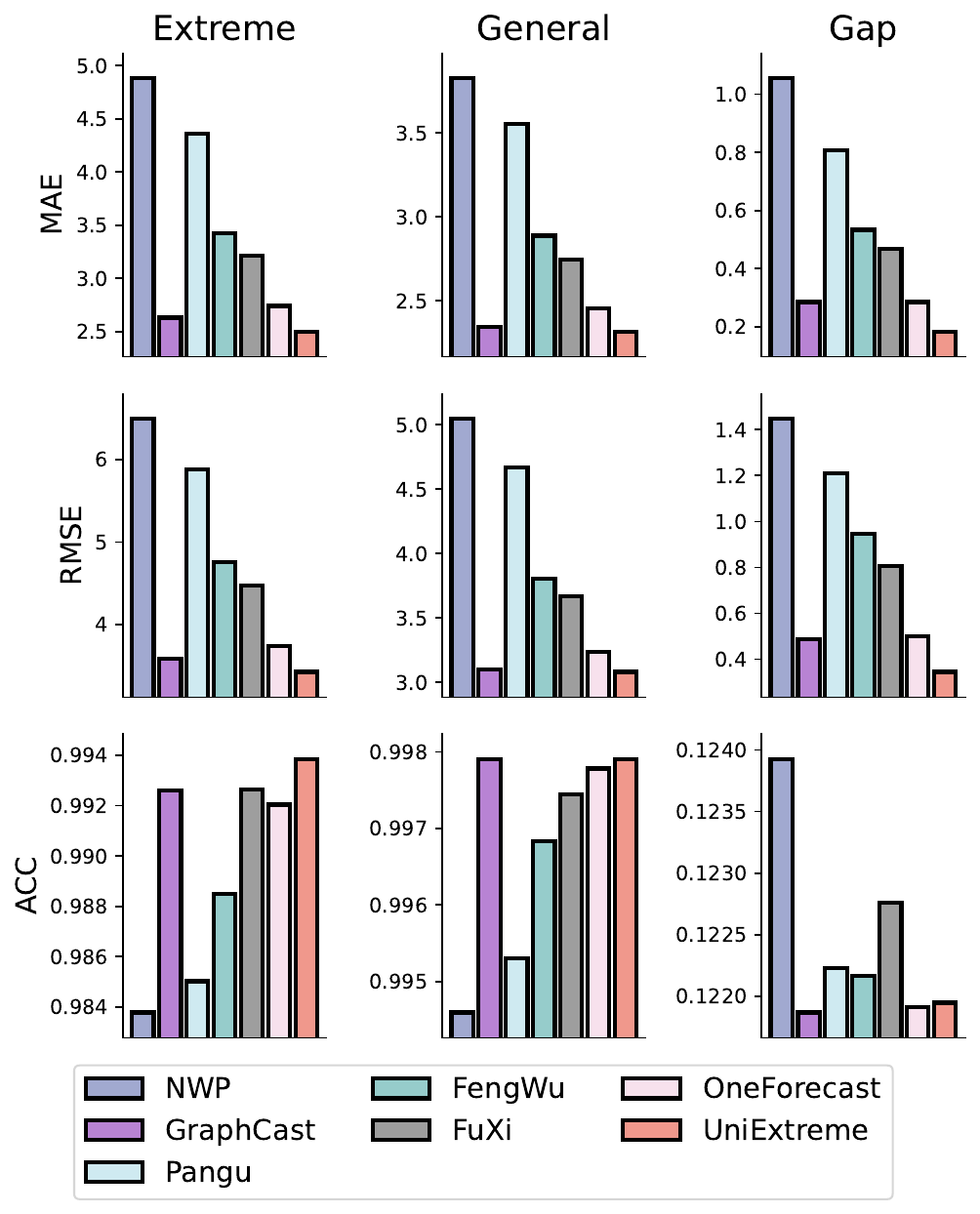}
    \vspace{-25pt}
    \caption{Raw forecasting results of variable Z1000.}
    \vspace{-5pt}
    \label{fig:raw_z_1000_left}
\end{minipage}
\hfill
\end{figure*}

\figref{fig:raw_msl_left}-\figref{fig:raw_z_1000_left} present the complete raw-scale performance of all atmospheric variables across all metrics. These results further highlight UniExtreme's effectiveness in forecasting diverse atmospheric phenomena at varying altitudes.

\section{Complete Categorical Results}
\label{appendix:categorical}

This section provides additional results for different methods in forecasting diverse extreme weather events.
Except for the MAE performance presented in \figref{figure:main_type}, we show the comparisons of RMSE and ACC performance in \figref{figure:type_rmse} and \figref{figure:type_acc}, respectively.
The RMSE results further reinforce our above conclusion that UniExtreme achieves the superior effectiveness with minimal forecasting errors. 
However, in terms of ACC, we also note that the baselines can surpass UniExtreme in some types of events, especially DstD, implying the challenge of accurately predicting the temporal deviations and anomaly patterns for dust devil events.

\section{Details of HFA for Band Weights}
\label{appendix:baa}
This section elaborates on the computation of the high-frequency area for the band weights. The learned weight tensor $\mathcal{W}^t_{r} \in \mathbb{R}^{N \times C}$ adheres to the frequency order, implying that $\mathcal{W}^t_{r;n}$ corresponds to the $n$-th band with the smallest frequency range.
We define the "energy ratio" for the $n$-th band and variable $c$ as:
\begin{align*}
\eta_{n,c}=\frac{\sum_{j=1}^n \mathcal{W}^t_{r;j,c}}{\sum_{j=1}^N \mathcal{W}^t_{r;j,c}}.
\end{align*}
Based on this, we compute the energy ratio curve and subsequently derive the high-frequency area using the same methodology outlined in \appref{appendix:empirical}. This proxy metric serves as an effective measure of the frequency concentration inherent in the band weights learned by the AFM module.

\clearpage

\clearpage

\clearpage

\clearpage

\clearpage

\clearpage

\clearpage

\clearpage

\clearpage

\clearpage

\clearpage

\clearpage

\clearpage

\clearpage

\clearpage

\clearpage

\clearpage

\clearpage

\end{document}